\documentclass[runningheads]{llncs}

\usepackage{eccv}

\usepackage{eccvabbrv}

\usepackage{graphicx}
\usepackage{booktabs}
\usepackage{multirow}
\usepackage{adjustbox}
\usepackage[accsupp]{axessibility}  

\usepackage{xcolor}
\usepackage{colortbl}

\definecolor{cat0}{HTML}{1F77B4} 
\definecolor{cat1}{HTML}{FF7F0E} 
\definecolor{cat2}{HTML}{2CA02C} 
\definecolor{cat3}{HTML}{D62728} 
\definecolor{cat4}{HTML}{9467BD} 
\definecolor{cat5}{HTML}{8C564B} 
\definecolor{cat6}{HTML}{E377C2} 
\definecolor{cat7}{HTML}{7F7F7F} 
\definecolor{cat8}{HTML}{BCBD22} 

\usepackage{hyperref}

\usepackage{orcidlink}
\begin{document}
\title{Metonymy in vision models undermines attention-based interpretability}

\titlerunning{Metonymy in vision models}

\author{Ananthu Aniraj\inst{1}\orcidlink{0009-0003-4521-036X} \and
Cassio F. Dantas\inst{1,2}\orcidlink{0000-0002-1934-0625} \and
Dino Ienco\inst{1,2}\orcidlink{0000-0002-8736-3132} \and
Massimiliano Mancini\inst{3}\orcidlink{0000-0001-8595-9955} \and  
Diego Marcos\inst{1}\orcidlink{0000-0001-5607-4445}}

\authorrunning{A. Aniraj et al.}

\institute{Inria, EVERGREEN, University of Montpellier, 34090 Montpellier, France
\and
INRAE, UMR TETIS, University of Montpellier, 34090 Montpellier, France
\and
University of Trento, Trento, Italy
}
\maketitle
\begin{abstract}
Part-based reasoning is a classical strategy to make a computer vision model directly focus on the object parts that are relevant to the downstream task.
In the context of deep learning, this also serves to improve by-design interpretability, often by using part-centric attention mechanisms on top of a latent image representation provided by a standard, black-box model. This approach is based on a locality assumption: that the latent representation of an object part encodes primarily information about the corresponding image region. In this work, we test this basic assumption, measuring intra-object leakage in vision models using part-based attribute annotations. Through a comprehensive experimental evaluation, we show that modern pretrained vision transformers violate the locality assumption and exhibit a strong intra-object leakage, in which each part encodes information from the whole object, a visual metonymy that compromises the faithfulness of attention-based interpretable-by-design methods for part-based reasoning, ultimately rendering them uninterpretable. In addition, we establish an upper bound using a two-stage approach that prevents leakage by design.
We then show that this inherently disentangled feature extraction improves attribute-driven part discovery on a variety of tasks, confirming the practical impact of intra-object leakage.
Our results uncover a neglected issue affecting the interpretability of part-based representations, such as those in CBMs relying on part-centric concepts, highlighting that two-stage approaches offer a promising way to mitigate it. 
\end{abstract}
\section{Introduction}

Part-based reasoning has been pursued as a strategy for visual recognition long before deep learning became commonplace~\cite{felzenszwalb2005pictorial,dalal2005histograms,felzenszwalb2009object}. Many modern concept-based interpretability methods, although not explicitly frame themselves as part-based, employ concepts that are inherently tied to object parts. By grounding predictions in such localized object parts, models offer users a natural entry point to inspect and verify the decision process. This principle underlies a range of recent approaches, from fine-grained recognition methods that learn part-centric attention~\cite{zheng2017learning,peng2017object,saha2023particle} to Concept Bottleneck Models (CBMs)~\cite{koh2020concept, chen2020concept, rao2024discover}, which predict human-readable attributes, such as ``has a black bill'' or ``has red legs'' in the context of bird species identification, as an intermediate step before mapping to class labels. The interpretability of all such methods rests on a shared, often implicit, assumption: that the latent representation of a given part encodes primarily information about the corresponding image region. However, a model trained on bird photos may learn a consistent co-occurrence between beak colour and leg colour, entangling their representations such that information about one is encoded in the features of the other, a form of visual \emph{metonymy}, in which some element of an object comes to stand for the whole. If this \emph{locality assumption} is violated in this way, then the reasoning exposed to the user is no longer faithful, regardless of how interpretable it appears. Indeed, there is growing evidence that concept representations, such as those in CBMs, leak information beyond their intended scope~\cite{mahinpei2021promises} and fail to respect locality~\cite{huang2024concept,raman2025do}: a prediction for ``has a black bill'' may be heavily influenced by features outside the bill region. This resists straightforward mitigation~\cite{mahinpei2021promises}, and higher-quality concept predictors that are causally influenced only by the relevant object part are a necessary part of any solution~\cite{havasi2022addressing, ruiz2024theoretical}.

\begin{figure}[t]
    \centering
    \includegraphics[width=\linewidth]{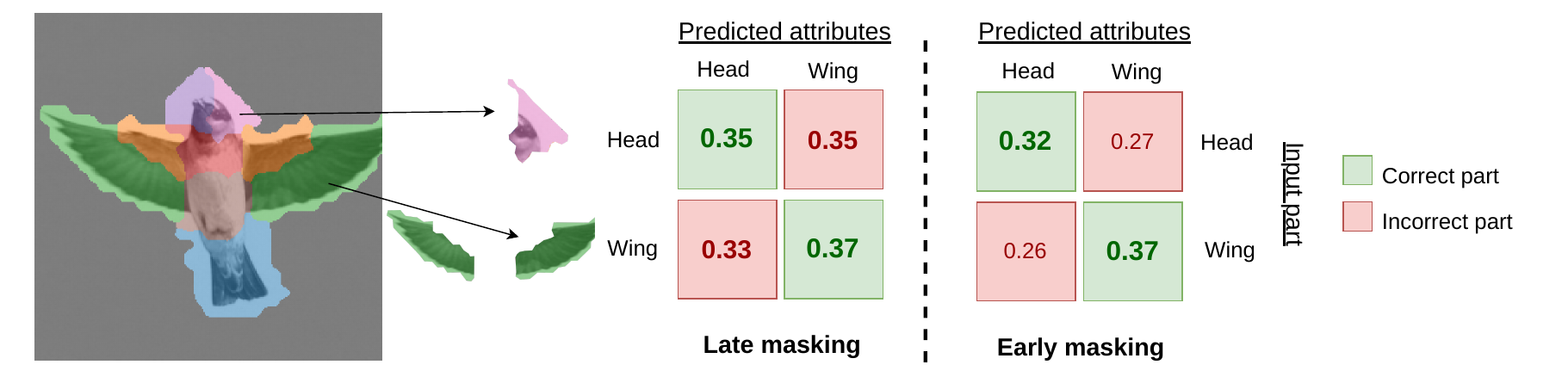}
    \caption{Intra-object leakage: Modern pretrained vision models violate locality. Early masking via two-stage approaches can mitigate this issue. Mean average precision (mAP) results for part attribute prediction using either the head or wing parts with a DINOv3 backbone. With late masking, the wrong part achieves almost as high mAP (anti-diagonal, red squares) as the correct part (diagonal, green squares).}
    \label{fig:splash}
\end{figure}

This problem represents a form of \emph{shortcut learning}~\cite{geirhos2020shortcut}, where models exploit readily available statistical regularities rather than the intended causal features. While prior work has extensively studied \emph{background bias}, where models spuriously associate objects with their contextual surroundings~\cite{beery2018recognition, geirhos2020shortcut, aniraj2023masking}, and developed attention-based methods to suppress it~\cite{locatello2020object,wang2021causal,dittadi2022generalization,cultrera2023leveraging}, there is comparatively little discussion of leakage \emph{within} the foreground object itself. Yet several families of methods, including CBMs, part prototype networks, and unsupervised part discovery, all rely on applying part-centric attention mechanisms to deep feature maps whose representations are already entangled by the large receptive fields of modern architectures. Even if such attention maps visually highlight the correct part region (e.g., the beak), the underlying features may still encode information leaked from other parts, undermining faithfulness. This phenomenon, illustrated in Fig.~\ref{fig:splash} using actual attribute prediction results from a DINOv3 model, is particularly critical in high-stakes domains such as medical imaging, where reliance on a correlated but unrelated object structure could lead to dangerous misdiagnoses. In this paper, we address this gap by proposing a benchmark to measure intra-object leakage and a way forward, based on two-stage transformer models, that yield better part-centric representations by design. To make this problem measurable, we propose a metric for intra-object leakage based on datasets with part-level attribute annotations. Using this metric, we evaluate several state-of-the-art self-supervised backbones, including DINO and DINOv2, across three diverse datasets. Despite evidence that such models preserve enough locality for tasks like semantic segmentation~\cite{vanyan2023analyzing}, we find that intra-object leakage is pervasive, putting into question the faithfulness of attention-based interpretability approaches built on top of them. We then study two-stage transformer models, in which attention maps are applied to the original image rather than to deep features, and show that this architectural change substantially reduces intra-object leakage, leading to significant improvements in part discovery.
In summary, our contributions are:
\begin{itemize}
\item A benchmark for intra-object leakage using part attribute annotations.
\item A comprehensive evaluation showing that intra-object leakage is widespread in modern pretrained backbones.
\item Evidence that two-stage feature extraction mitigates this leakage, improving attention-based part discovery.
\end{itemize}

\section{Related Work}

\noindent \textbf{Shortcut learning and feature coupling.} Deep neural networks are prone to shortcut learning~\cite{geirhos2020shortcut}, exploiting non-causal statistical regularities rather than causal structure. A well-studied manifestation is background bias, where an object's classification becomes spuriously tied to its context~\cite{beery2018recognition, aniraj2023masking, geirhos2020shortcut}: a model may learn to recognize ``cow'' by detecting ``pasture''~\cite{beery2018recognition}, and in extreme cases achieve high accuracy even when the foreground object is removed entirely, as long as the same biases are present in the test set~\cite{zhu2017object, xiao2021noise}. This problem persists even in large-scale Vision-Language Models~\cite{wang2024sober}, indicating that scaling alone is not enough to mitigate it. Attention-based approaches have been proposed to focus models on the foreground information, improving generalization~\cite{seitze2023bridging} and efficiency~\cite{kapl2025objectcentric}, and several works have aimed at encouraging locality in learned representations~\cite{yun2022patch,li2023localvit,ranasinghe2023perceptual}. 
However, these previous efforts mainly target information leakage between image background and foreground. In this work, we show that even models with improved locality still suffer from \emph{intra-object} leakage between part representations.

Object coupling is a variant of this problem in which the feature representations of two objects become entangled if they co-occur often, potentially leading models to rely on spurious features within the foreground or even within the object of interest~\cite{izmailov2022feature}. This issue affects many modern models trained with self supervision, although it can be partially mitigated by modifying the patch sampling strategy in augmentation-based self-supervised training to ensure there is enough overlap between patch pairs~\cite{qiu2024mind}. The authors of \cite{qiu2024mind}, although mainly focused on inter-object coupling, also provide empirical evidence of leakage at intra-object level, exploring how their approach helps mitigate it. Their aim, as well as that of the locality-encouraging methods cited above, is to provide better representations for spatial tasks such as semantic segmentation.
Unlike \cite{qiu2024mind}, we propose to quantify intra-object leakage by comparing it to an upper bound that uses a two-stage transformer where leakage is avoided by design. In addition, in contrast to the mentioned works, our aim is to study and improve the interpretability of part attention models.

\noindent \textbf{Part-based interpretability in deep learning.} 
The problem of intra-object leakage is especially consequential for methods that aim to ground predictions in localized parts. Several families of such methods exist.
CBMs~\cite{koh2020concept, chen2020concept} force predictions through an intermediate layer of human-understandable concepts that, in object-centric tasks, are often tied to specific parts. 
However, their learned concept representations can leak information beyond their intended scope~\cite{mahinpei2021promises}, fail to capture stable inter-concept relationships~\cite{raman2024understanding, bader2025sub}, and lack spatial grounding in the correct image region~\cite{raman2025do, bader2025sub}. 
While prior work has studied \emph{concept leakage} (label information bypassing the bottleneck)~\cite{mahinpei2021promises,havasi2022addressing,parisini2025leakage} and  whether concept predictions are grounded in the latent representation of the correct region~\cite{raman2025do,huang2024concept}, we focus on a more upstream problem, where the latent representation of one object part is itself contaminated by information from other parts.
Part Prototype Networks (PPNs)~\cite{chen2019looks} classify images by comparing local patches to learned prototypical parts, providing a more explicitly spatial form of reasoning. 
However, prototype activation maps reveal severe symptoms of leakage, and several studies have found a misalignment between similarity in the latent space and similarity as perceived by humans~\cite{hoffmann2021looks,huang2023evaluation,davoodi2023interpretability}, undermining the interpretability they are designed to provide.
Unsupervised part discovery methods~\cite{hung2019scops, huang2020interpretable, klis2023PDiscoNet, aniraj2024pdiscoformer, xia2024unsupervised, MPAE} offer a third strategy, learning to segment objects into semantically consistent parts using only image-level supervision via attention mechanisms analogous to slot attention~\cite{locatello2020object}. Recent approaches~\cite{aniraj2024pdiscoformer, xia2024unsupervised, MPAE} leveraging self-supervised ViT features~\cite{oquab2023dinov2, caron2021emerging} have relaxed the strict geometric priors required by previous approaches, enabling discovery of parts with diverse shapes. Despite their different designs, all three families share a common vulnerability: they apply attention or comparison operations to deep latent features in which information has already been entangled across spatial locations by the large receptive fields of modern architectures, what we refer to as \emph{late masking}~\cite{aniraj2023masking}. We posit that mitigating intra-object leakage at the representation level is a prerequisite for any of these approaches to achieve faithful, compositional part-based reasoning, and use part discovery as a benchmark to measure the impact of such mitigation.

\section{Measuring intra-object leakage in backbones} 

We propose two evaluation protocols to measure intra-object leakage. The first one, part specificity (PS), is adapted to tasks in which each local attribute is systematically linked to the same semantic part. For instance, in a dataset of bird photographs the beak-related attributes will always be localized to the beak.
The second protocol, the most predictive part overlap (MPPO), is adapted to settings in which each attribute, even if local, may be associated to different semantic parts. For instance, in medical imaging, a fracture can be located on any bone in an X-ray image.
In both cases we partition the foreground object into up to $K$ parts via part discovery, although ground truth part masks would also be suitable. For each part $k\in\{1,...,K\}$, we assume a meaningful feature vector $\mathbf{v}^k$ can be extracted using a pretrained model and the corresponding part mask.

\subsubsection{Part specificity (PS)} is computed as the difference between the attribute classification performance using masks that correspond to the associated part(s), $\mathcal{K}_g^+\subset\mathcal{K}$, \emph{versus} using masks corresponding to the other parts, $\mathcal{K}_g^-=\mathcal{K}-\mathcal{K}_g^+$. We assume the part-based attributes are each linked to a ground truth part $g\in\{1, ..., G\}$, where each $g$ would correspond to a semantic part (e.g. head, legs, etc.).
Since discovered parts do not come with semantic labels, we assign them to ground truth parts based on spatial overlap.
For every ground truth part $g$, we first need to assign a set of discovered parts $\mathcal{K}_g^+\subset\mathcal{K}$ to it according to spatial overlap. If ground truth part keypoints are available, we compute a contingency matrix with the co-ocurrence $c_g^k$ between each $k$ and $g$, and add $k$ to $\mathcal{K}_g^+$ if $c_g^k >\tau$, with $\tau$ a threshold.  
Alternatively, this can be performed manually via visual inspection of the discovered parts whenever no ground truth part information is available.
For every discovered part $k\in\{1, ..., K\}$, we then extract the corresponding feature vectors $\mathbf{v}^k_s\in\mathcal{V}^k$ for all samples $s=\{1,...,S\}$ in the dataset. We then train a linear layer $l^k: \mathbb{R}^D\mapsto \mathbb{R}^A$ for attribute classification using the attributes associated to all parts and evaluate its performance $\text{mAP}^k_g$ on each group of ground truth attributes using mean average precision.

For a ground truth part $g$, $\text{PS}_g$ is calculated as the difference between the average attribute prediction performance using the correct, matching discovered parts $\mathcal{K}_g^+$ and the average performance using the other discovered parts $\mathcal{K}_g^-$:
\begin{equation}
    \text{PS}_g=\frac{1}{|\mathcal{K}_g^+|}\sum_{k\in\mathcal{K}_g^+}\text{mAP}^k_g - \frac{1}{|\mathcal{K}_g^-|}\sum_{k\in\mathcal{K}_g^-}\text{mAP}^k_g
\end{equation}
The final PS score is simply the average $\text{PS}_g$ across all ground truth parts $\text{PS}=\sum_g \text{PS}_g / G$.
A low PS may indicate either intra-part leakage or merely reflect actual correlations between attributes across parts. In order to eliminate the latter, we compare PS using a regular late masking approach to obtain the part features against early masking, an approach that ensures that no leakage can happen and that is described in the following section.

\subsubsection{Most predictive part overlap (MPPO)} with ground truth masks is computed as the overlap between the most attribute-predictive discovered parts and the corresponding ground truth part masks, meaning that ground truth masks indicating the image region responsible for each attribute are available.
For every attribute $a \in \{1, \ldots, A\}$, we first identify the most predictive discovered part $k^*_a = \arg\max_{k} \, l^k(\mathbf{v}^k)_a$, where $l^k(\mathbf{v}^k)_a$ is the logit of the linear classifier $l^k$ corresponding to attribute $a$.
For every sample $s$ where attribute $a$ is annotated as present, we then evaluate whether the mask $\mathbf{M}_{k^*_a}$ of the most predictive part $k^*_a$ overlaps with the ground truth mask $\mathbf{M}_{g(a)}$, where $g(a)$ denotes the ground truth part associated with attribute $a$. We define overlap as a non-zero intersection between the two masks.
For an attribute $a$, $\text{MPPO}_a$ is calculated as the proportion of attribute-present samples for which the most predictive discovered part overlaps with the corresponding ground truth mask:
\begin{equation}
    \text{MPPO}_a = \frac{|\{s \in S_a : \mathbf{M}_{k^*_a}(s) \cap \mathbf{M}_{g(a)}(s) \neq \emptyset\}|}{|S_a|}
\end{equation}
where $S_a \subseteq \{1,\ldots,S\}$ denotes the set of samples in which attribute $a$ is annotated as present.
The final MPPO score is simply the average $\text{MPPO}_a$ across all attributes $\text{MPPO} = \sum_a \text{MPPO}_a / A$. 
The MPPO score measures how often the part with the highest attribute prediction confidence, measured as the magnitude of its logit, overlaps with the corresponding ground truth mask, indicating how often the correct part is the most predictive.


\section{Two-stage ViTs to mitigate intra-object leakage}
\label{sec:method_2}

\begin{figure}[t]
    \centering
    \includegraphics[width=\linewidth]{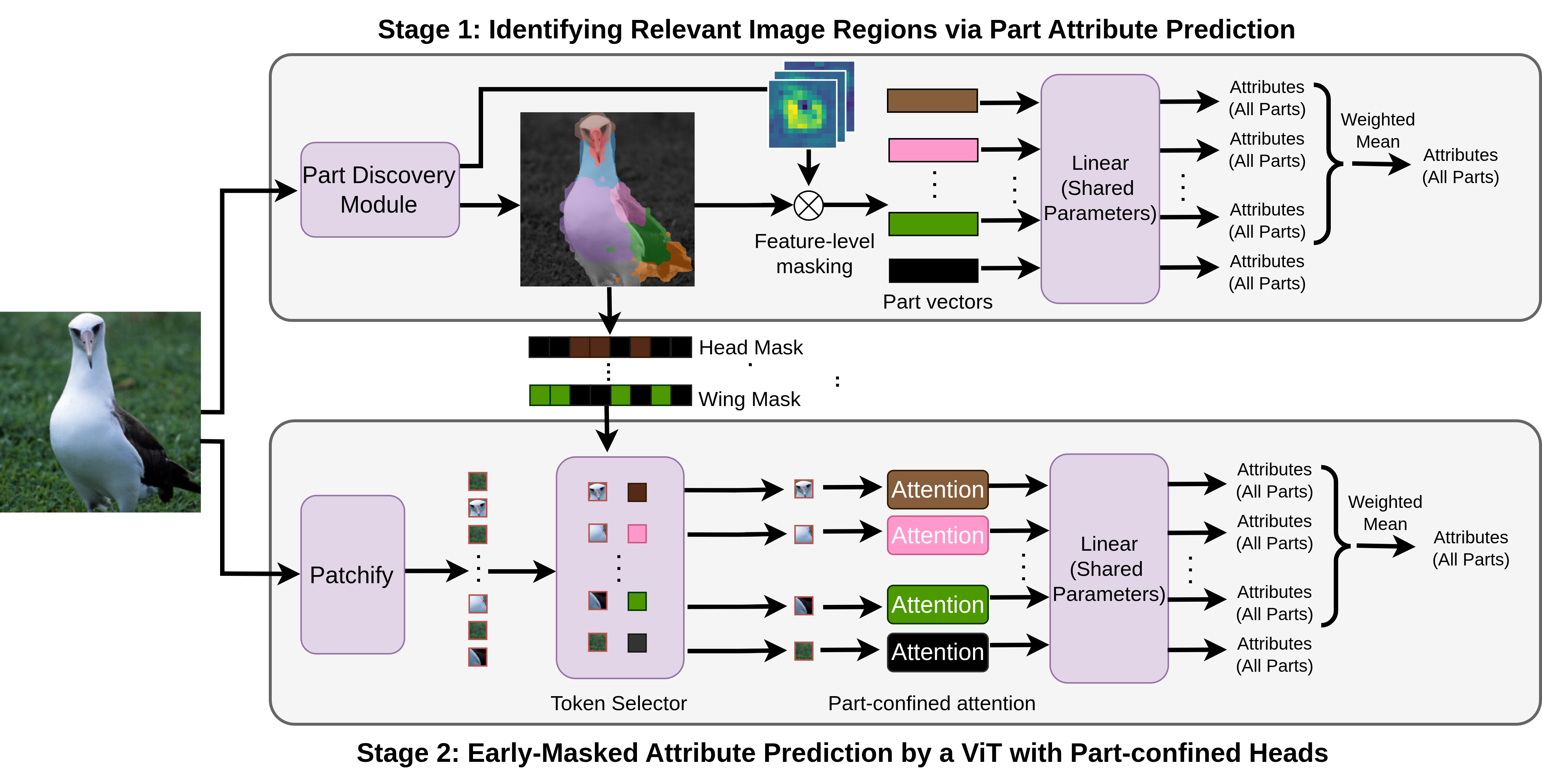}
    \caption{Overview of the proposed two-stage architecture.
In Stage 1, attribute-driven part discovery learns a set of attention masks that localize semantic parts. In Stage 2, these masks are applied directly to the input image tokens and different parts are processed in parallel through part-confined attention, preventing feature leakage between regions. This second stage provides an additional attribute prediction signal under strict spatial isolation, which helps guide and refine the part discovery process in Stage~1 during end-to-end training.}
    \label{fig:2-stage}
\end{figure}

To address intra-object leakage, where models provide part representations capturing information about the whole object, we propose a two-stage approach, in which part representations of the second stage are guaranteed not to be contaminated by features from the rest of the object. Figure~\ref{fig:2-stage} shows an overview of the proposed method. The first stage discovers semantic parts using a combination of part shaping losses and an attribute prediction task. The second stage uses attention masking to process each of the discovered parts independently, ensuring that the final part representation depends only on the image region that corresponds to the part mask.

\subsubsection{Stage 1: attribute-driven part discovery.}
We adapt the PDiscoFormer framework~\cite{aniraj2024pdiscoformer} to generate $K$ part masks. Given an image $\mathbf{x}$, we extract spatial features $\mathbf{z} \in \mathbb{R}^{D \times H \times W}$ using a frozen backbone $h_\theta$ (e.g., DINOv2). We compute attention maps $\mathbf{A} \in [0, 1]^{(K+1) \times H \times W}$ (where the $(K+1)$-th map represents the background) by comparing patch features to learnable prototypes $\mathbf{p}^k$, followed by a Gumbel-Softmax operation. These maps are used to pool the features into part embeddings $\mathbf{v}^k$.
\begin{equation}
    \mathbf{v}^{k} = \frac{1}{HW} \sum_{i,j} a_{ij}^{k} \mathbf{z}_{ij}
\end{equation}

\noindent \textbf{Shared layer normalization.} We apply a single shared Layer Normalization~\cite{ba2016layer} (LN) across all part embeddings to obtain $\mathbf{v}_{m}^{k} = \text{LN}_{\text{shared}}(\mathbf{v}^{k})$. This prevents the model from learning part-specific scaling artifacts and forces the distinctiveness to emerge from the features themselves rather than the normalization statistics.

\noindent \textbf{Attribute-specific routing.} 
Crucially, we do not aggregate parts into a global vector. Instead, we allow the model to dynamically select the most relevant part for each attribute.
Let $\mathbf{v}_{m}^{k}$ be the embedding for part $k$. We project this embedding to attribute scores using a linear layer (MLP):
\begin{equation}
    \mathbf{S}_{:,k} = \text{MLP}(\mathbf{v}_{m}^{k}) \in \mathbb{R}^{N_{attr}}
\end{equation}
where $S_{c,k}$ is the raw score for attribute $c$ predicted solely from part $k$. To obtain the final prediction, we compute routing weights $\mathbf{W}$ via a Softmax across the part dimension:
\begin{equation}
    W_{c,k} = \frac{\exp(S_{c,k})}{\sum_{j=1}^{K} \exp(S_{c,j})}, \quad \hat{y}_c = \sum_{k=1}^{K} W_{c,k} S_{c,k}
    \label{eq:smax_routing}
\end{equation}
This mechanism aligns parts with attributes: if part $k$ provides the highest confidence for attribute $c$, the gradient signal will encourage the attention mask for $k$ to focus on the image region containing evidence for $c$. The attribute prediction loss is the binary cross-entropy between these predictions and the ground truth: $\mathcal{L}_{\text{att}} = \text{BCE}(\mathbf{y}, \hat{\mathbf{y}})$.

\subsubsection{Stage 2: Parallel part processing via strict masking.}

To enforce strict compositionality, we process the discovered parts in parallel streams within a single ViT. We initialize Stage 2 with a standard ViT but replicate the prefix tokens (the \texttt{[CLS]} and register tokens) $K+1$ times, creating a dedicated set of query tokens for each discovered part.

We then enforce parallel processing by manipulating the self-attention mechanism. We construct a block-diagonal attention mask $\mathbf{M}$ where the $k$-th set of prefix tokens and the image patches assigned to part $k$ form an isolated clique. 

\noindent \textbf{Masking Variants and Gradient Estimation.}
A core challenge is balancing strict isolation (which requires binary masks) with the need to propagate gradients back to the part discovery module in Stage 1. To investigate this trade-off, we explore three masking strategies for the attention mechanism $\text{Attn}(\mathbf{Q}, \mathbf{K}, \mathbf{V}, \mathbf{M})$:

     \noindent \emph{Soft Masking:} We use the continuous probability maps from Stage 1 directly as attention biases:
        $\mathbf{H}_{\text{soft}} = \text{Attn}(\mathbf{Q}, \mathbf{K}, \mathbf{V}, \mathbf{M}_{\text{soft}})$. %
    This allows gradient flow but also leakage, as low-probability regions still contribute weak signals (leakage).

    \noindent \emph{Hard Masking:} We discretize the masks using an argmax operation, enforcing strict binary isolation.
    $\mathbf{H}_{\text{hard}} = \text{Attn}(\mathbf{Q}, \mathbf{K}, \mathbf{V}, \mathbf{M}_{\text{hard}})$.
    While this guarantees zero leakage, standard discrete operations block gradient flow, potentially hindering the end-to-end training of Stage 1 driven by Stage 2's attribute loss.

    \noindent \emph{Straight-Through Estimator (STE):} We propose a hybrid approach that combines the strict isolation of hard masking with the differentiability of soft masking. We compute the output using hard masks in the forward pass but utilize the gradients from the soft mask operation during backpropagation:
    $\mathbf{H}_{\text{ste}} = \text{SG}(\mathbf{H}_{\text{hard}}) + (\mathbf{H}_{\text{soft}} - \text{SG}(\mathbf{H}_{\text{soft}})),$
    where $\text{SG}(\cdot)$ is the stop-gradient operator. This ensures the model predicts based on strictly isolated parts while still allowing the attribute predictor to refine the part shapes.

\noindent \textbf{Attribute Prediction.}
Finally, we apply the attribute predictor head to each independent part feature and aggregate them using the same Softmax Routing mechanism described in Eq.~\ref{eq:smax_routing}.

\subsubsection{Part shaping objectives.}
To ensure the discovered masks correspond to coherent semantic regions, we employ a set of loss functions. We discard the concentration loss used in prior work~\cite{hung2019scops} to allow for larger, more natural part shapes.

\noindent \textbf{1. De-correlated orthogonality loss ($\mathcal{L}_{\perp}$).} 
The orthogonality loss used in prior works~\cite{hung2019scops, aniraj2024pdiscoformer} encourages orthogonality between all part vectors. This could be detrimental for intra-class part discovery: a bird's ``head'' and ``wing'' naturally share common texture and colour features. Forcing their raw vectors to be orthogonal suppresses this shared signal.

We propose a new formulation that strictly separates the background while allowing foreground parts to share a common mean, enforcing orthogonality only on their \textit{variations}.
Let $\mathbf{v}_{bg}$ be the background feature and $\mathcal{F} = \{\mathbf{v}^1, \dots, \mathbf{v}^K\}$ be the set of foreground part features.

First, we enforce separation between the background and all foreground parts:
\begin{equation}
    \mathcal{L}_{bg} = \frac{1}{K} \sum_{k=1}^{K} \left( \frac{\mathbf{v}_{bg} \cdot \mathbf{v}^k}{\|\mathbf{v}_{bg}\| \|\mathbf{v}^k\|} \right)^2
\end{equation}

Second, to decorrelate the foreground parts, we compute the centroid of the foreground features $\boldsymbol{\mu}_{\mathcal{F}} = \frac{1}{K} \sum_{k=1}^K \mathbf{v}^k$. We then center the part features to isolate their unique descriptive components:
\begin{equation}
    \mathbf{u}^k = \mathbf{v}^k - \boldsymbol{\mu}_{\mathcal{F}}
\end{equation}
We minimize the covariance between these centered features $\mathbf{u}^k$ to ensure that the ``leg'' and ``head'' capture distinct deviations from the global object appearance:
\begin{equation}
    \mathcal{L}_{decorr} = \frac{1}{K(K-1)} \sum_{i=1}^K \sum_{j \neq i} \left( \frac{\mathbf{u}^i \cdot \mathbf{u}^j}{\|\mathbf{u}^i\| \|\mathbf{u}^j\| + \epsilon} \right)^2
\end{equation}
The final loss is the sum: $\mathcal{L}_{\perp} = \mathcal{L}_{bg} + \mathcal{L}_{decorr}$.

\noindent \textbf{2. Geometric and semantic priors.} Following PDiscoFormer~\cite{aniraj2024pdiscoformer}, we regularize the attention maps with auxiliary losses encouraging spatial smoothness ($\mathcal{L}_{\text{tv}}$), equivariance to affine transformations ($\mathcal{L}_{\text{eq}}$), presence of all parts within each batch ($\mathcal{L}_{\text{p}}$), and low-entropy, binary-like pixel assignment ($\mathcal{L}_{\text{ent}}$). The final training objective combines the attribute prediction loss with these priors:
\begin{equation}
\mathcal{L} = \mathcal{L}_{\text{att, stage1}} + 
\mathcal{L}_{\text{att, stage2}}+
\lambda_{\perp}\mathcal{L}_{\perp} + \lambda_{\text{tv}}\mathcal{L}_{\text{tv}} + \lambda_{\text{eq}}\mathcal{L}_{\text{eq}} + \lambda_{\text{p}}\mathcal{L}_{\text{p}} + \lambda_{\text{ent}}\mathcal{L}_{\text{ent}}
\end{equation}

\section{Experimental Setup}

\subsection{Datasets}
\label{sec:datasets}
\noindent\textbf{CUB.}
We utilize the Caltech-UCSD Birds-200-2011~\cite{WelinderEtal2010} dataset, a benchmark for fine-grained classification containing 11,788 images of 200 bird species. Crucially for our work, CUB provides dense binary attribute annotations (312 binary attributes) mapped to specific body parts (e.g., \textit{has bill shape::hooked}, \textit{has wing color::blue}). We group these attributes by their corresponding semantic parts to evaluate the model's ability to localize attribute signals to their correct anatomical regions. Additionally, the CUB dataset provides per-image keypoint annotations for 15 ground truth parts. To evaluate the part discovery performance, we merge these 15 keypoints into 7 macro-regions that directly correspond to the attribute groupings: back, belly, breast, head, leg, tail, and wing.

\noindent\textbf{CelebA.}
We use the CelebFaces Attributes (CelebA) dataset~\cite{liu2015deep}, a large-scale facial attribute benchmark containing 202,599 images of 10,177 identities with 40 binary attribute annotations per image. In addition, the aligned version of CelebA provides five facial landmark annotations (left eye, right eye, nose, left mouth corner, right mouth corner), enabling coarse localization of facial parts. 
To evaluate intra-object leakage, we define semantic facial parts using two complementary sources. First, by visual inspection of the part segmentation maps produced by PDiscoFormer. Second, we leverage the landmark annotations to derive reference part locations. Specifically, we merge the left and right eye landmarks into a single \textit{eye} part and retain separate parts for the \textit{nose} and \textit{mouth}. This restricted set of 3 parts is used to compute the part discovery metrics (NMI and ARI) in Table~\ref{tab:single_vs_two_stage}.

\noindent\textbf{CheXpert.}
For medical images, we use the CheXpert chest radiograph dataset~\cite{irvin2019chexpert}, a large-scale benchmark for automated chest X-ray interpretation containing 224,316 radiographs from 65,240 patients. Each image is labeled for the presence of 14 observations derived from radiology reports, including common thoracic pathologies such as cardiomegaly, edema, and pleural effusion.
Since CheXpert itself provides only image-level labels, we rely on the CheXlocalize dataset~\cite{saporta2022benchmarking}, which augments a subset of CheXpert images with radiologist-annotated localization maps. CheXlocalize provides pixel-level segmentation masks 
created by board-certified radiologists on images from the CheXpert validation and test sets (of size 234 and 668, respectively). These segmentation maps were used for evaluation purpose only and enabled quantitative assessment of spatial localization via our proposed MPPO metric.

\begin{figure}[t]
\centering
\small
\setlength\tabcolsep{1.5pt} 
\centering
\begin{adjustbox}{width=0.95\textwidth}
 \begin{tabular}{ccc@{\hspace{0.3cm}}ccc@{\hspace{0.3cm}}ccc}
 \multicolumn{3}{c}{CUB} & \multicolumn{3}{c}{CelebA} & \multicolumn{3}{c}{CheXpert} \\
 \includegraphics[width=0.1\textwidth]{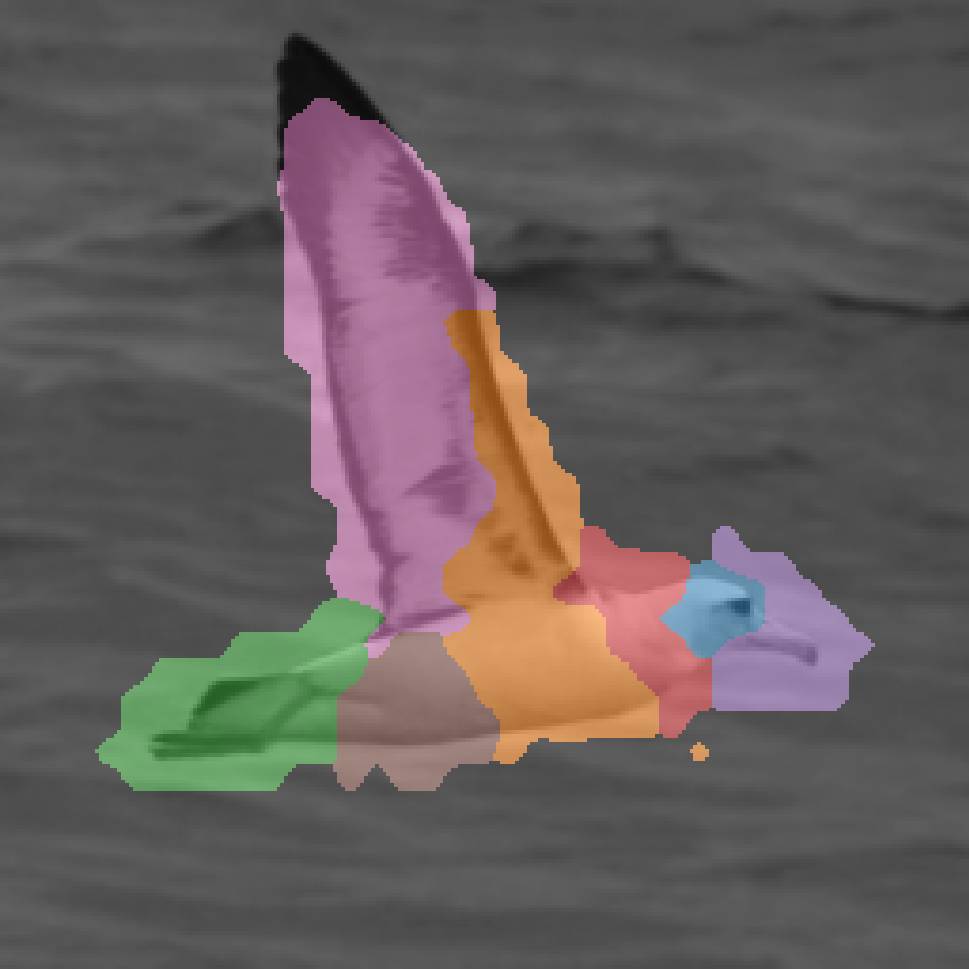}         &
 \includegraphics[width=0.1\textwidth]{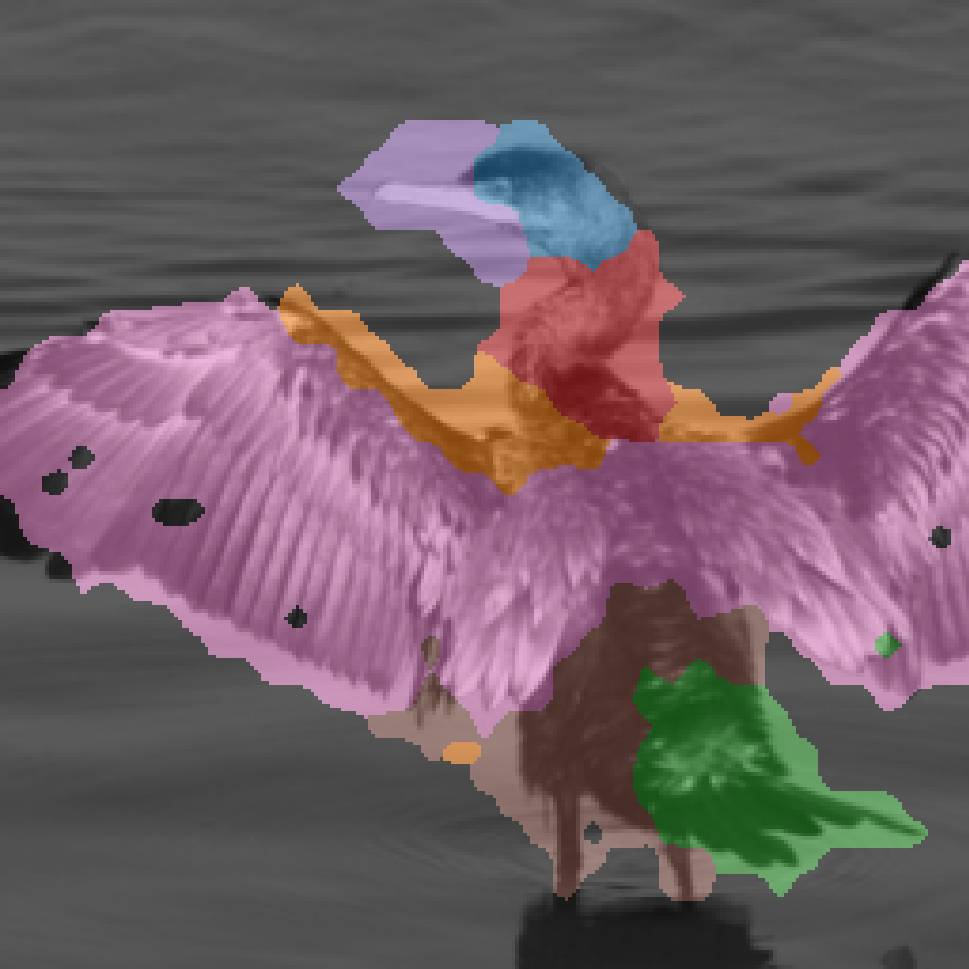}         &
 \includegraphics[width=0.1\textwidth]{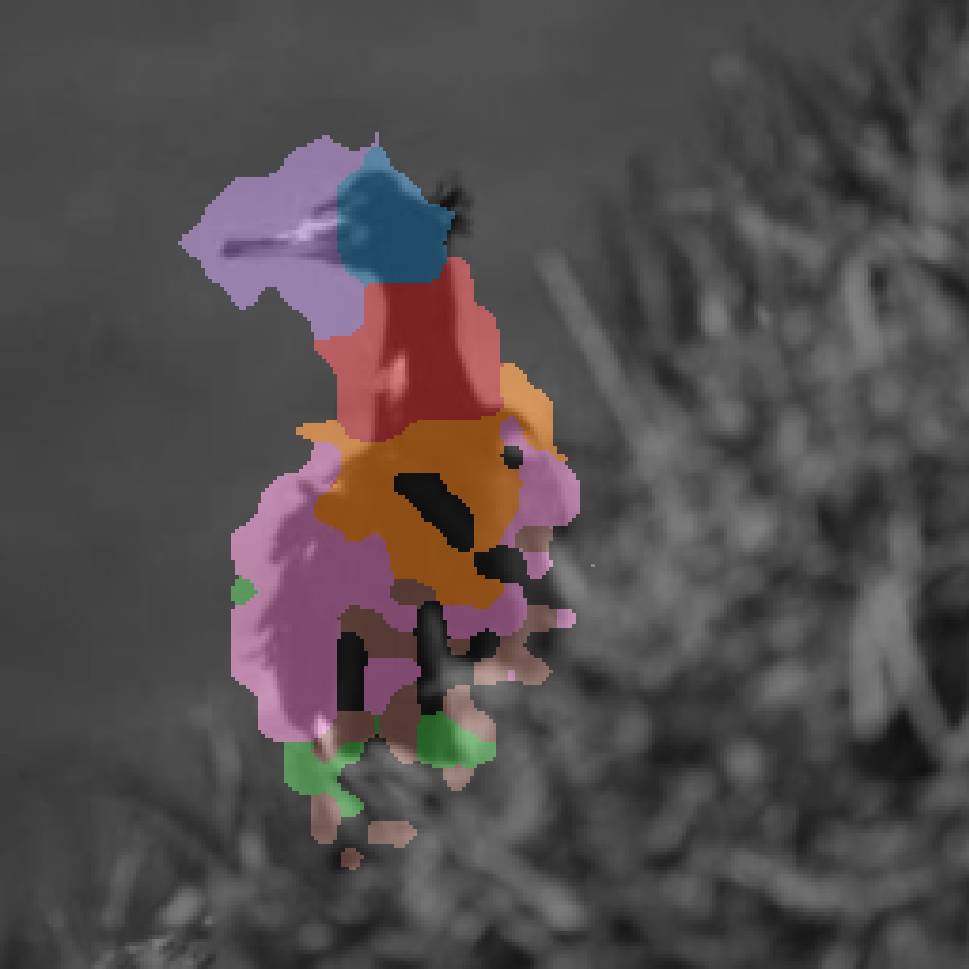}         &
 \includegraphics[width=0.1\textwidth]{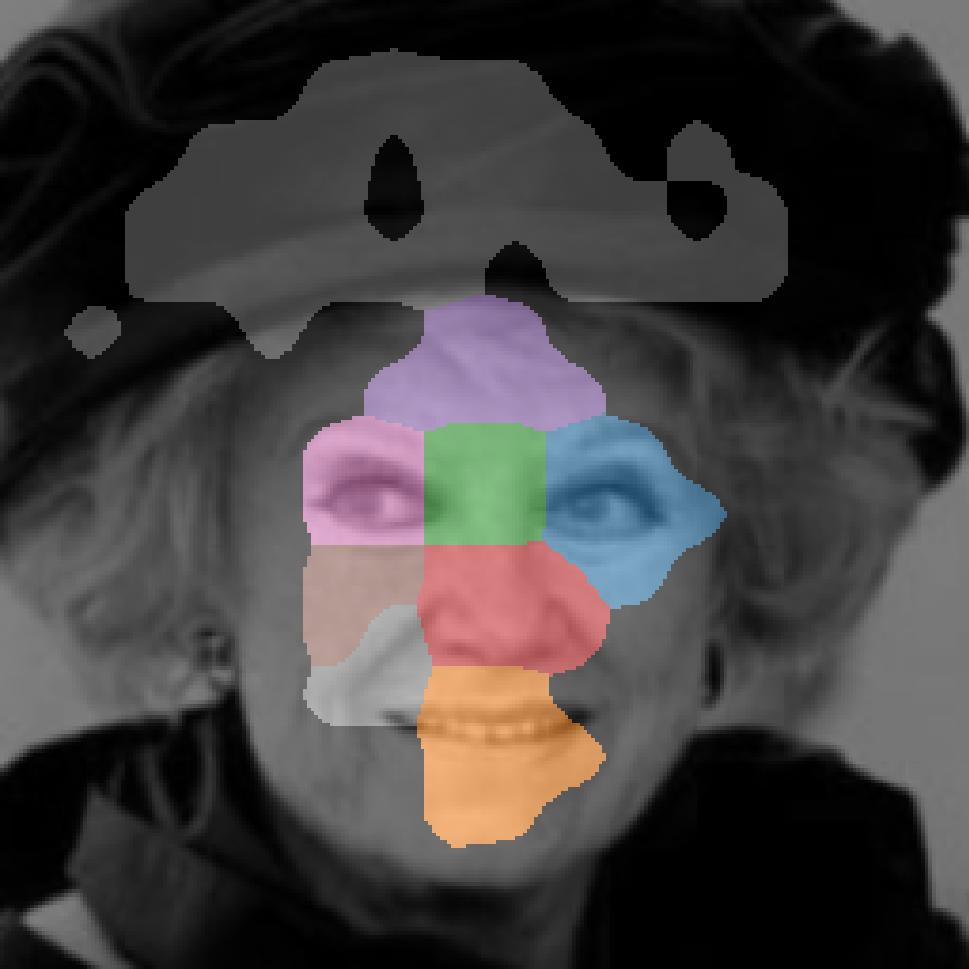}      &
 \includegraphics[width=0.1\textwidth]{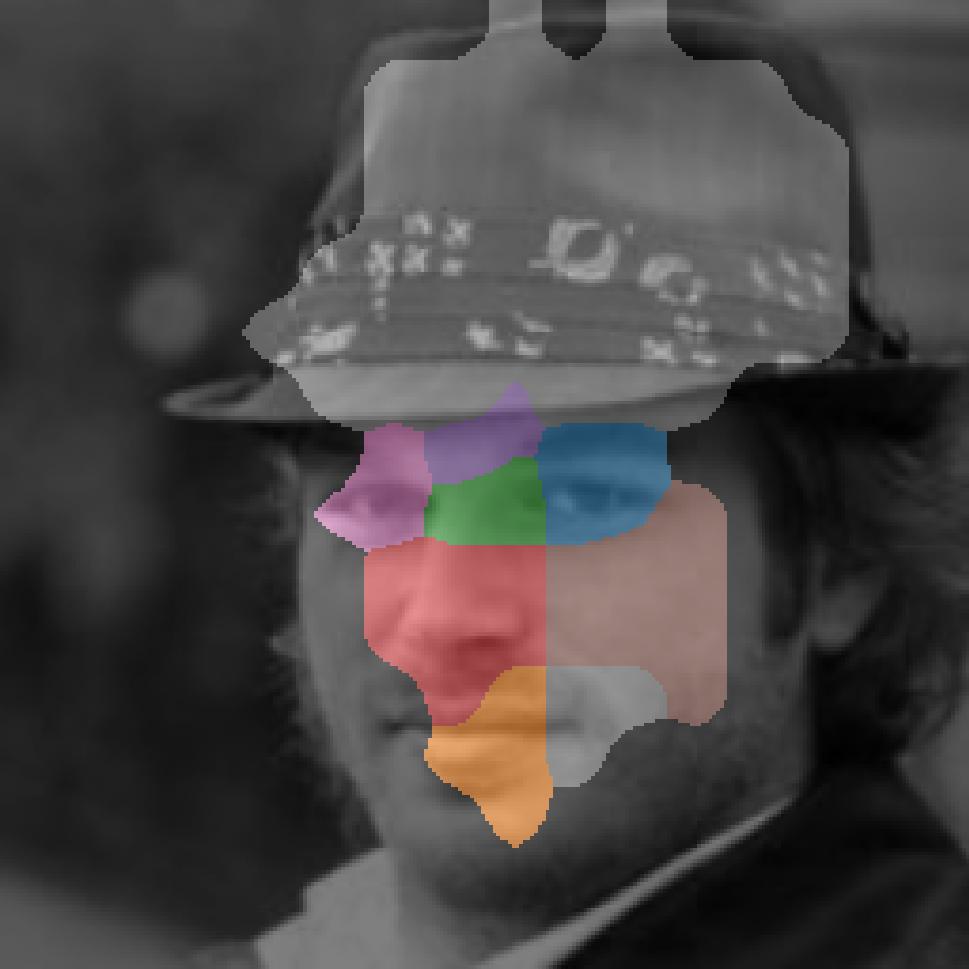}      &
  \includegraphics[width=0.1\textwidth]{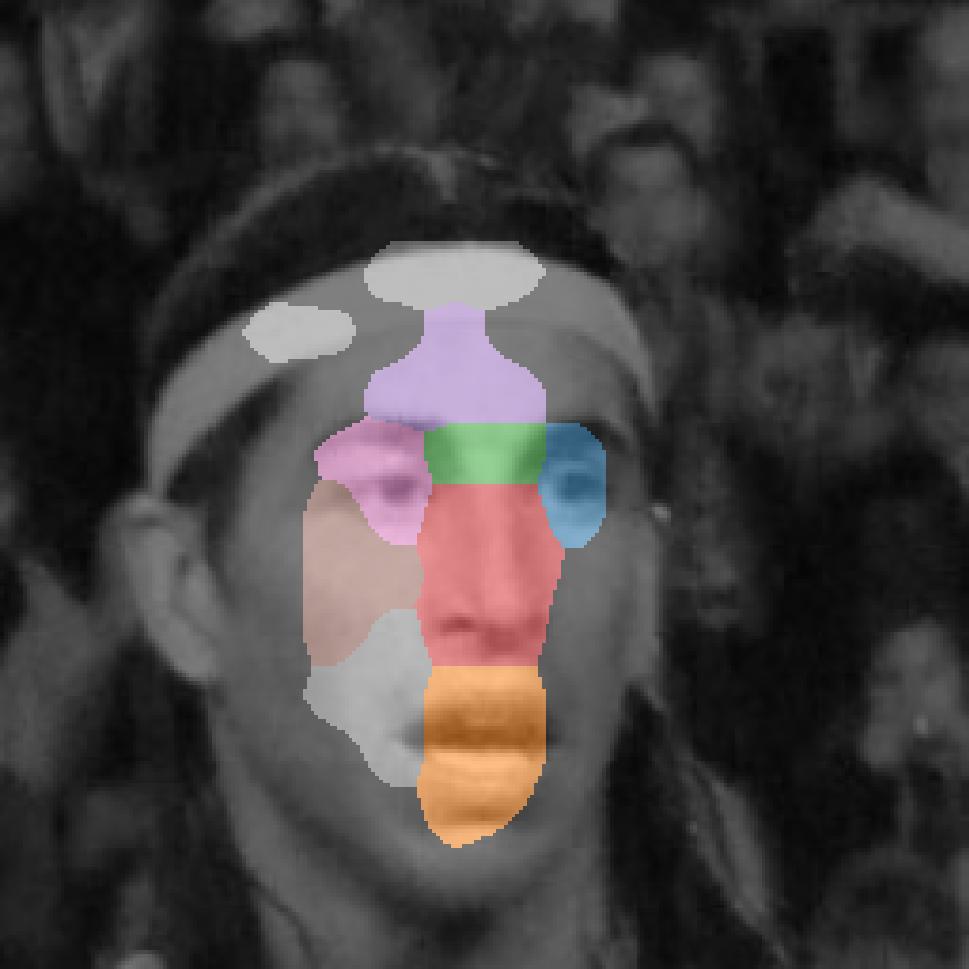}      &
 \includegraphics[width=0.1\textwidth]{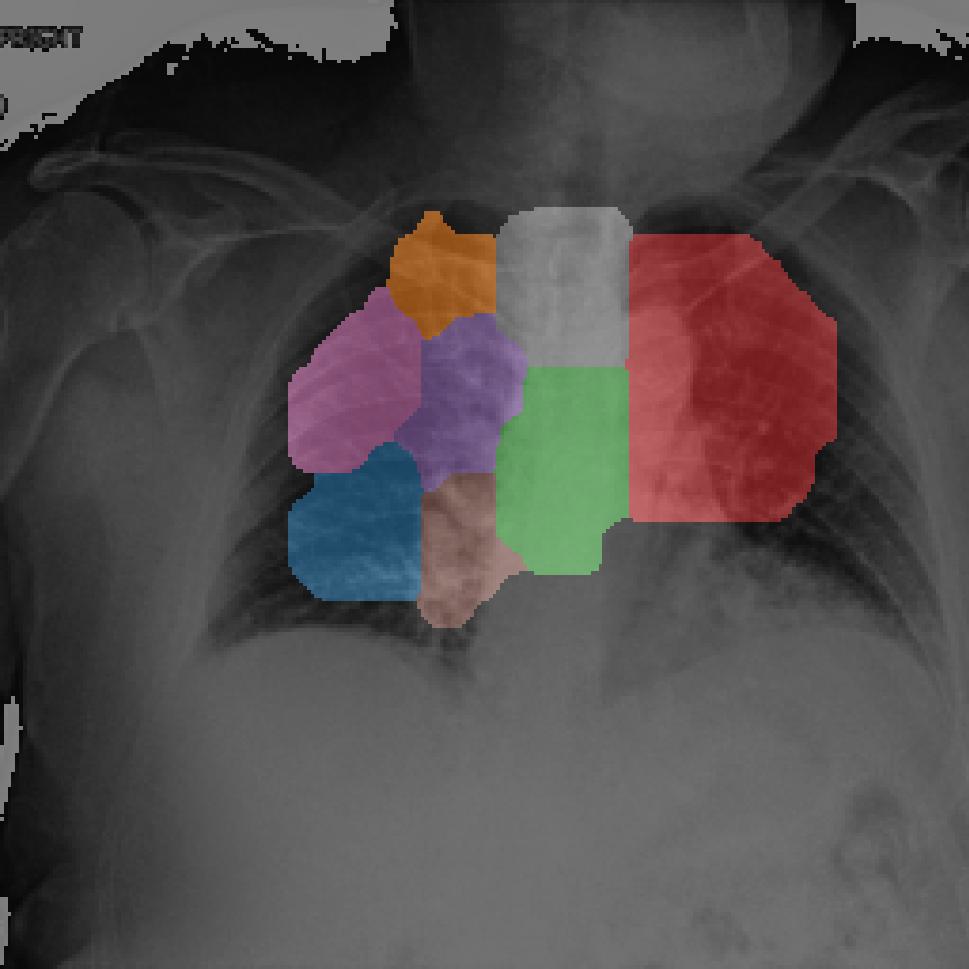}    &
 \includegraphics[width=0.1\textwidth]{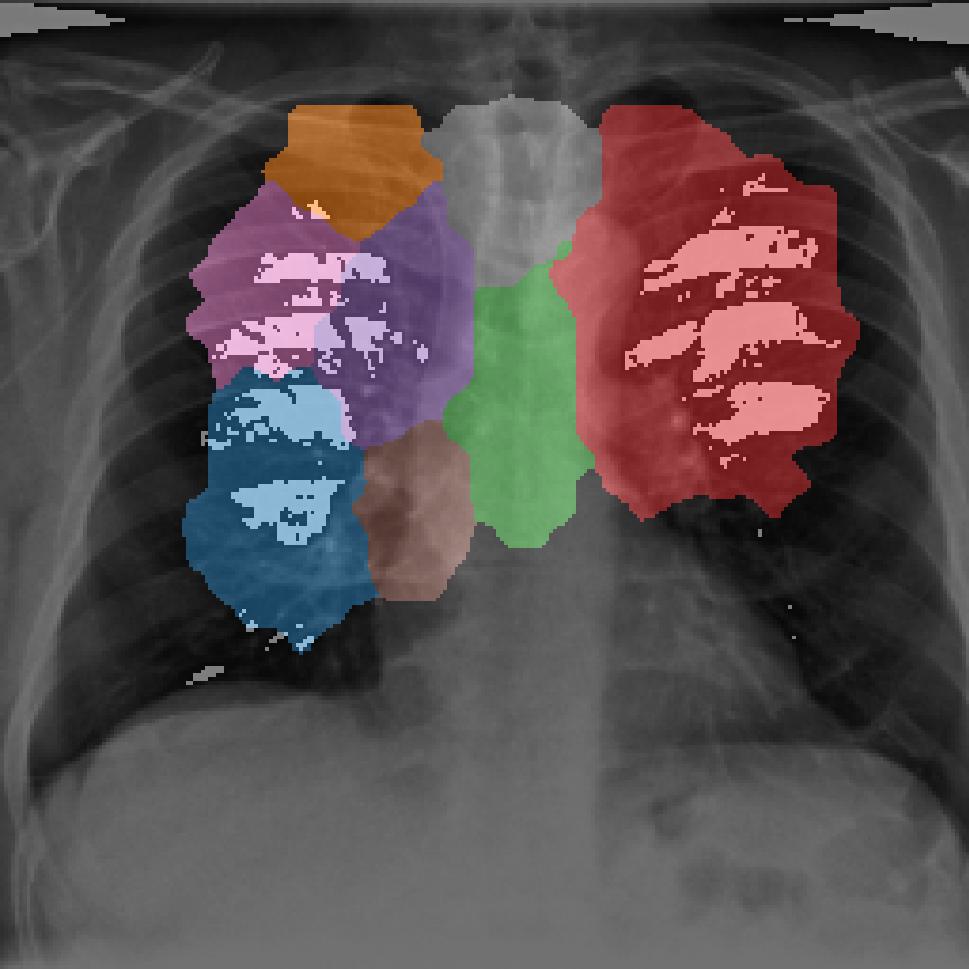}    &
 \includegraphics[width=0.1\textwidth]{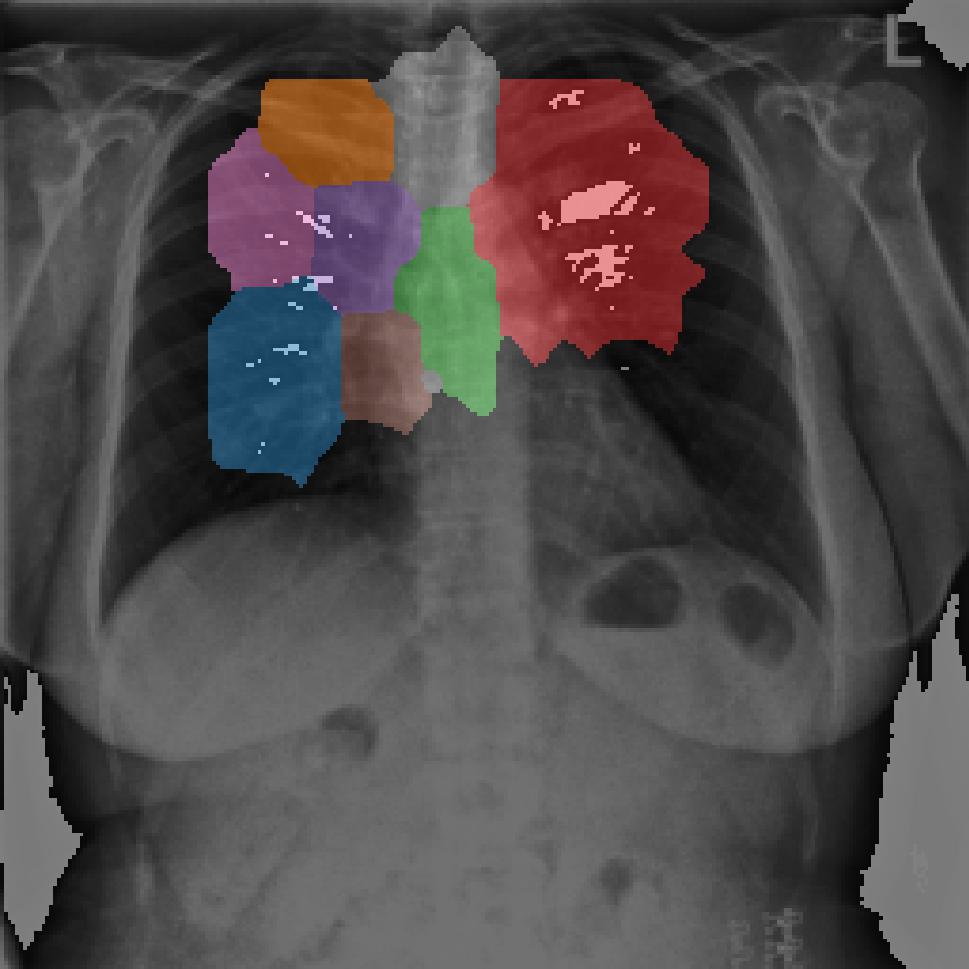}     
 \\
 \includegraphics[width=0.1\textwidth]{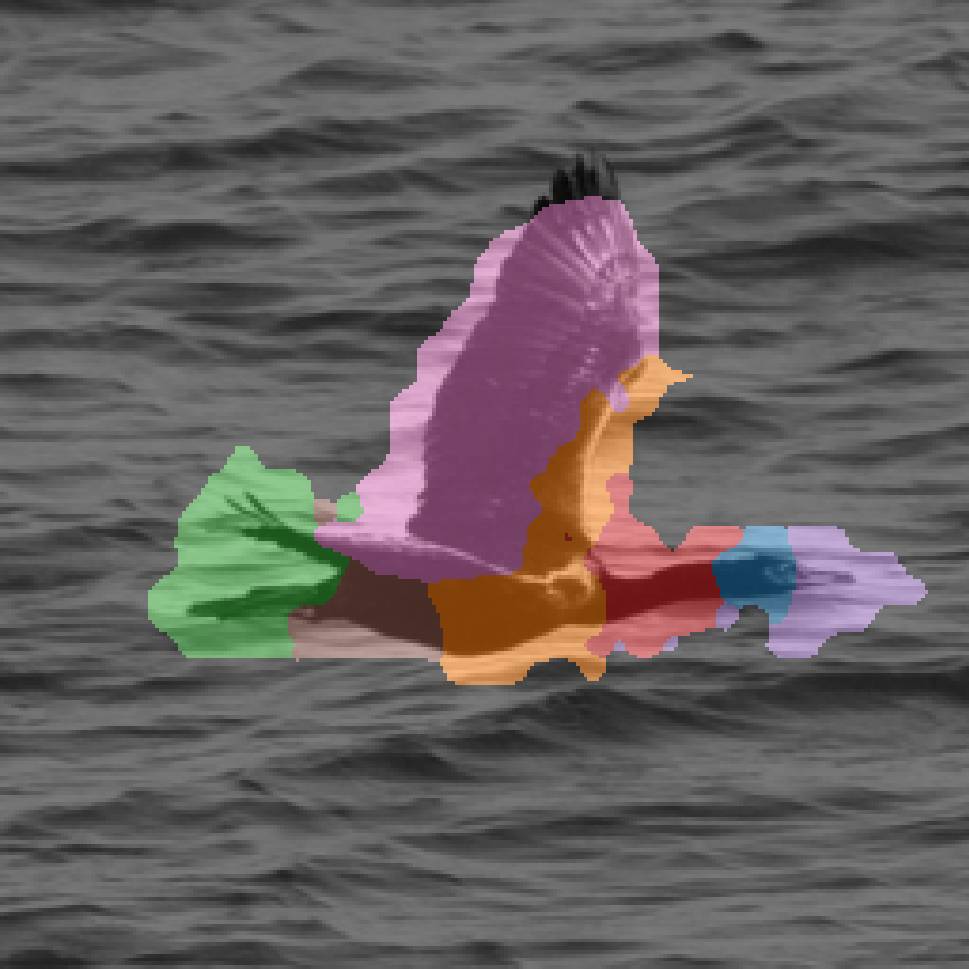}         &
 \includegraphics[width=0.1\textwidth]{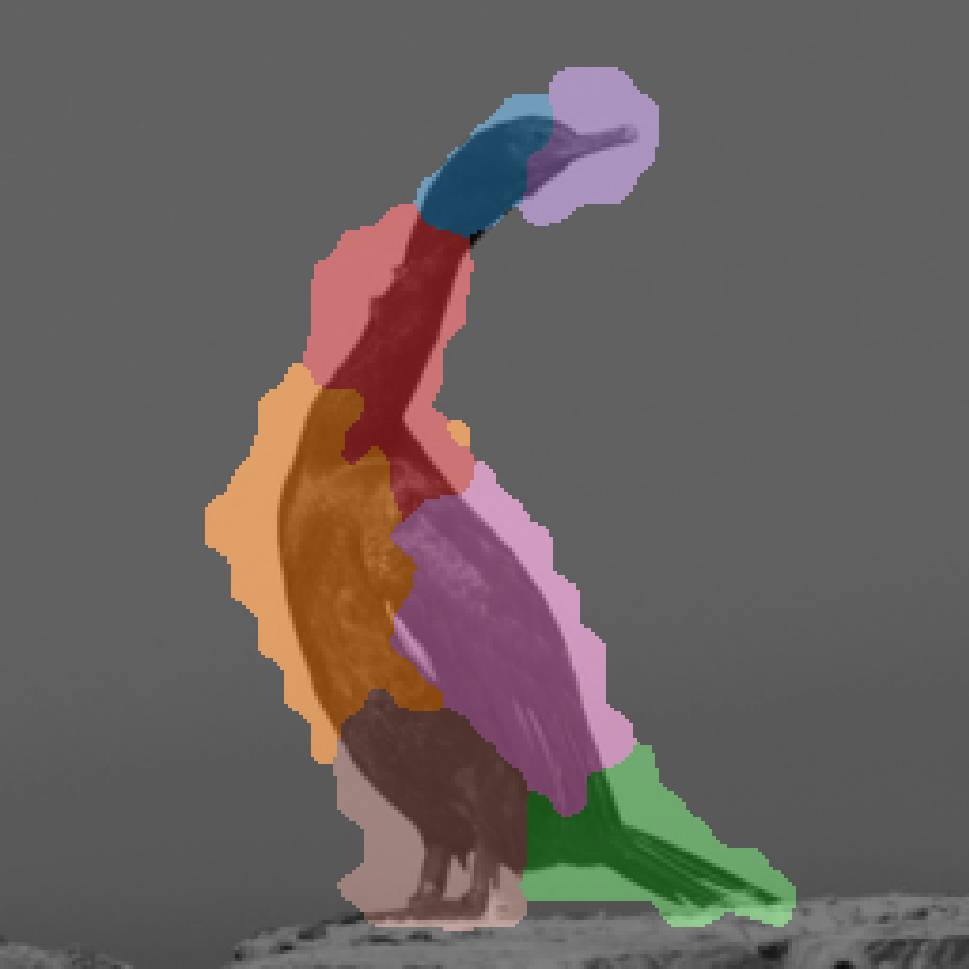}         &
 \includegraphics[width=0.1\textwidth]{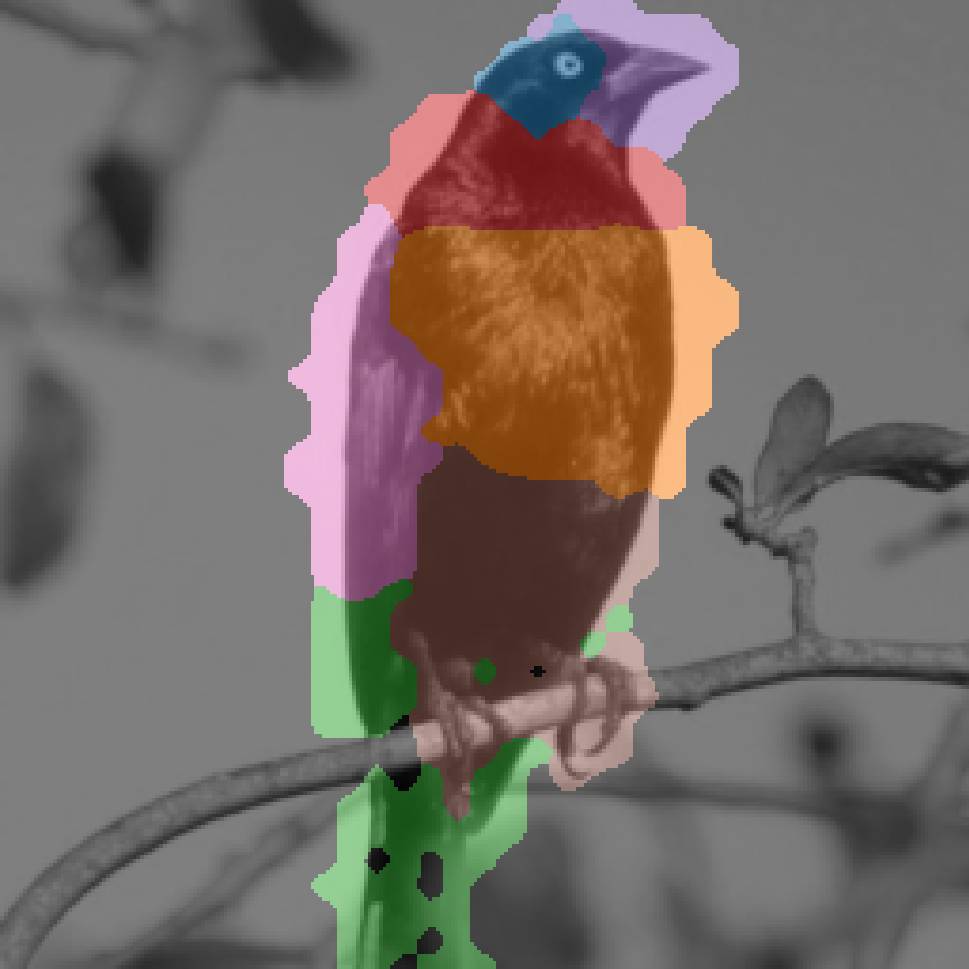}         &
 \includegraphics[width=0.1\textwidth]{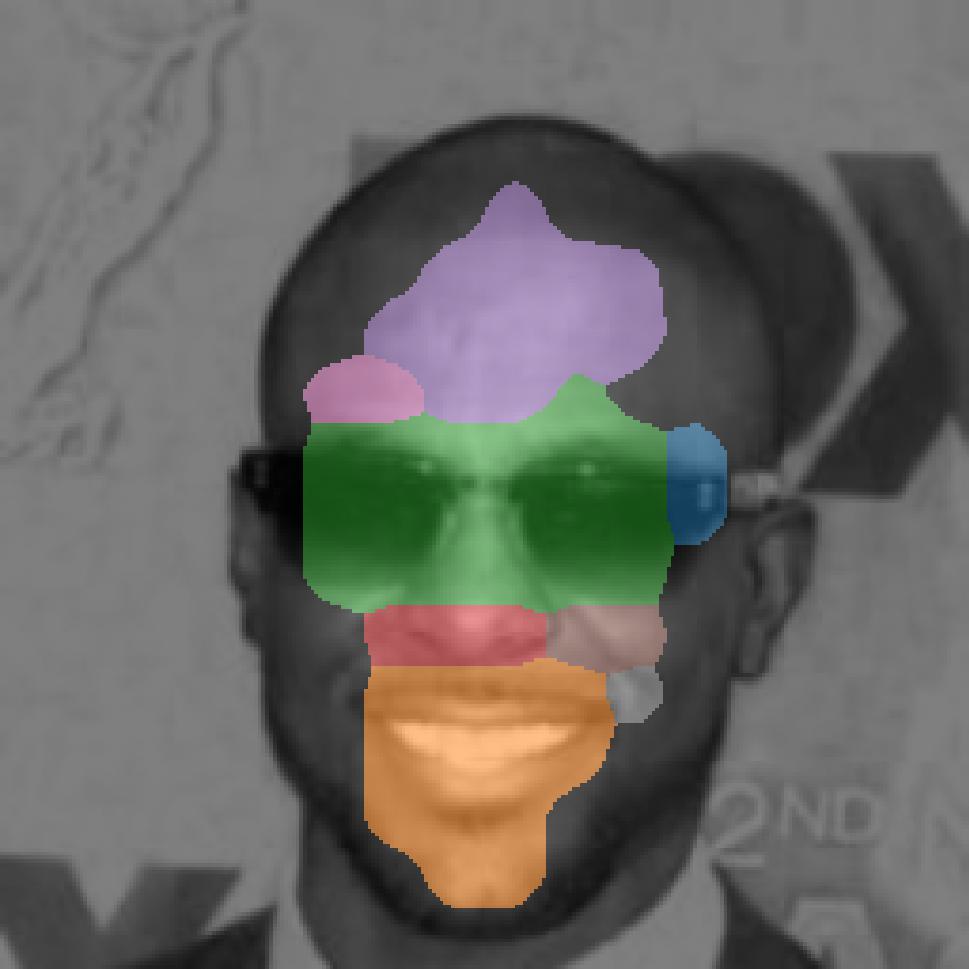}      &
 \includegraphics[width=0.1\textwidth]{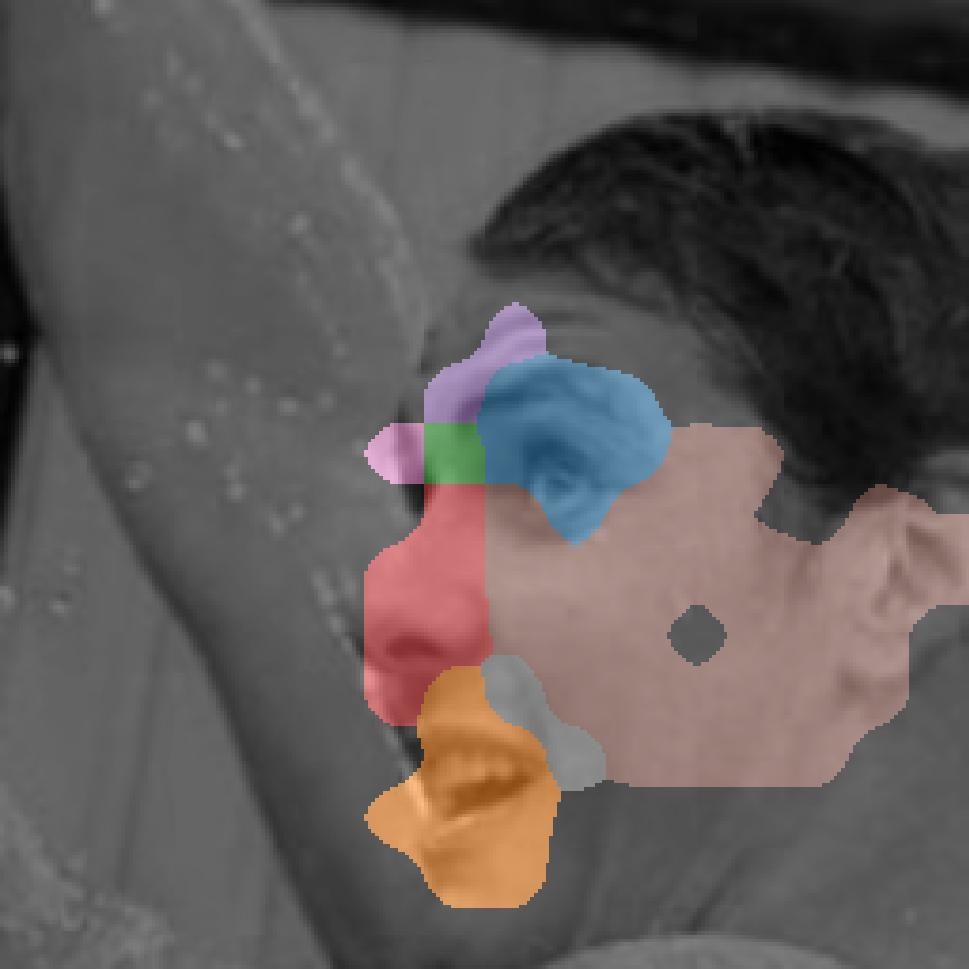}     &
 \includegraphics[width=0.1\textwidth]{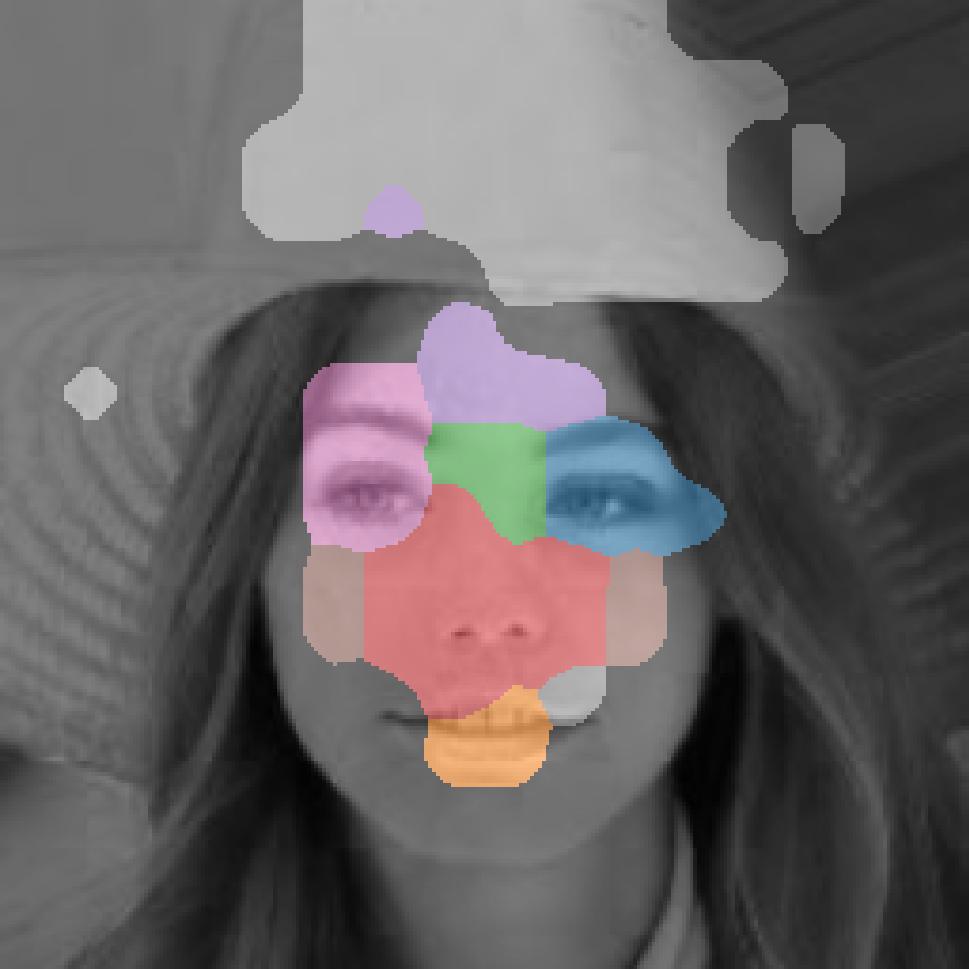}    &
 \includegraphics[width=0.1\textwidth]{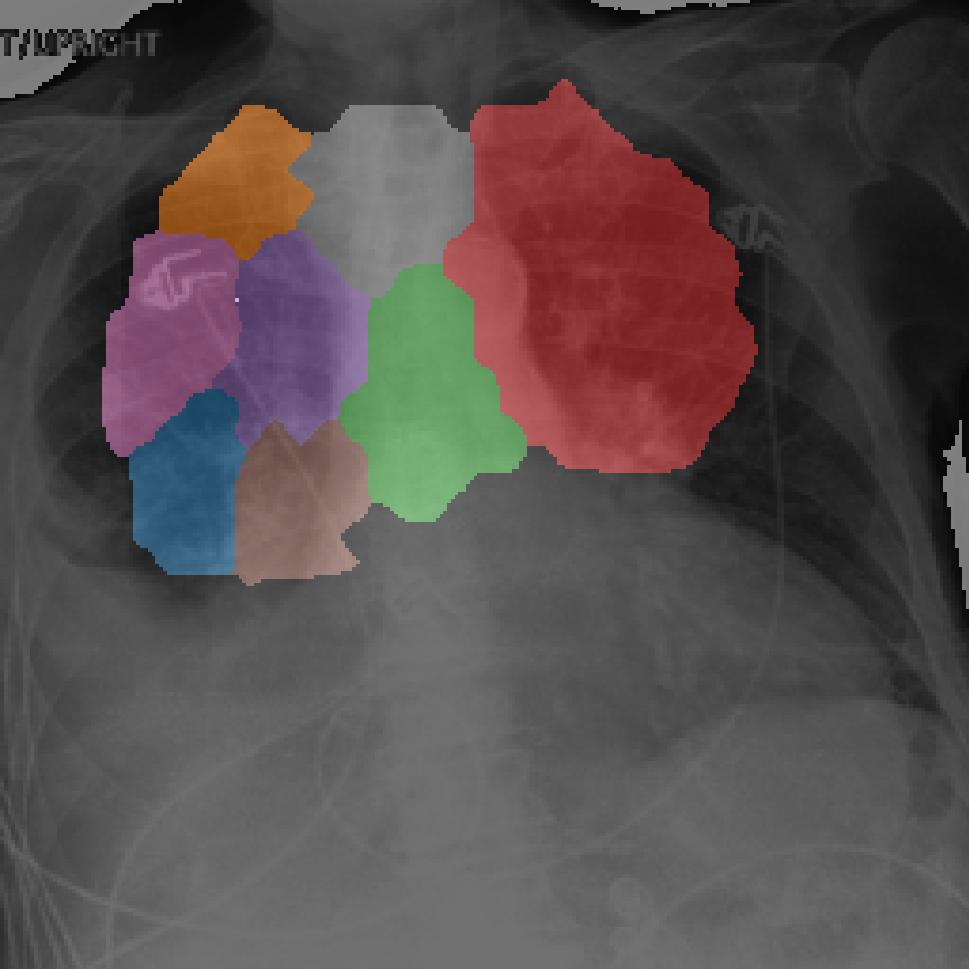}    &
 \includegraphics[width=0.1\textwidth]{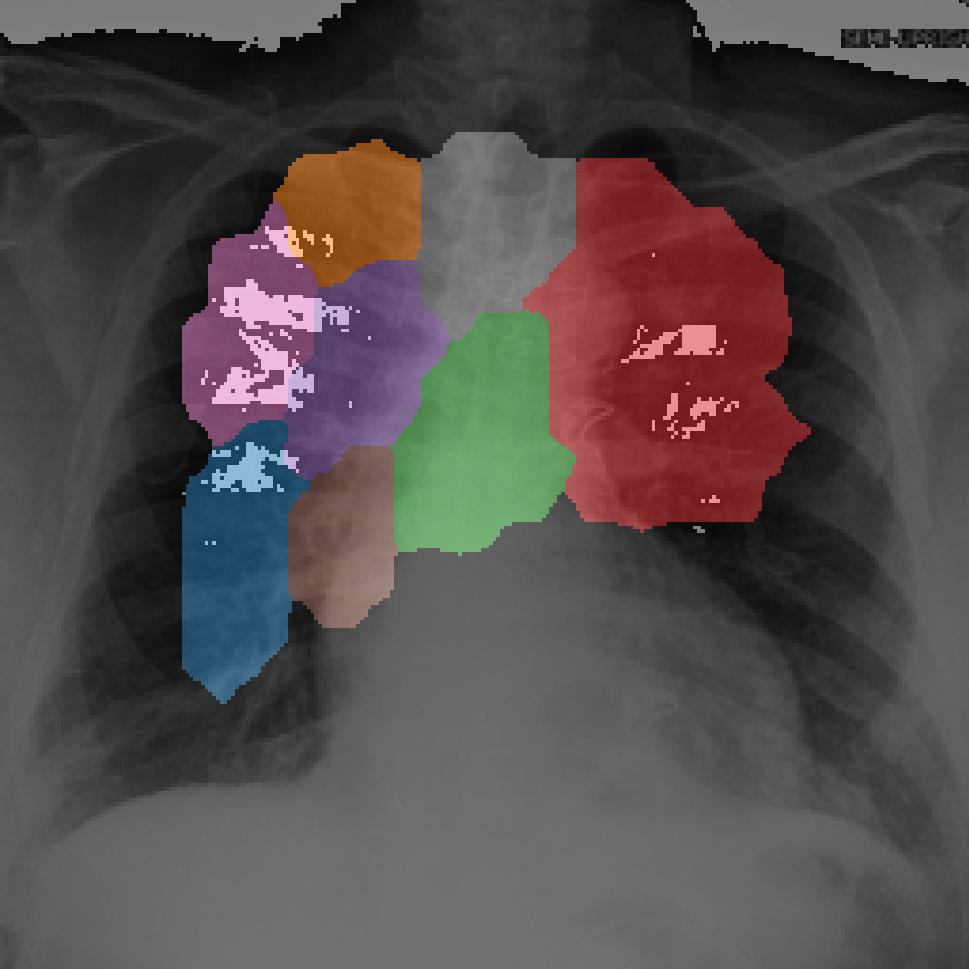}    &
 \includegraphics[width=0.1\textwidth]{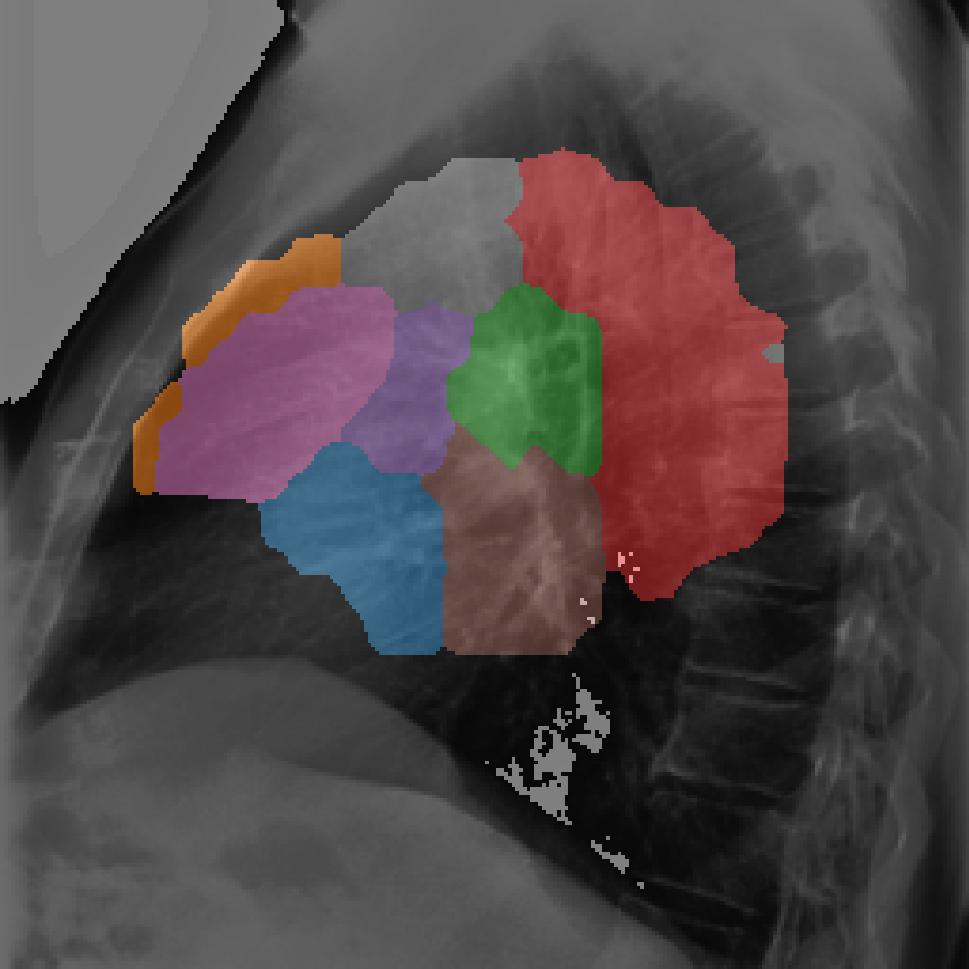}      
 \\
 \includegraphics[width=0.1\textwidth]{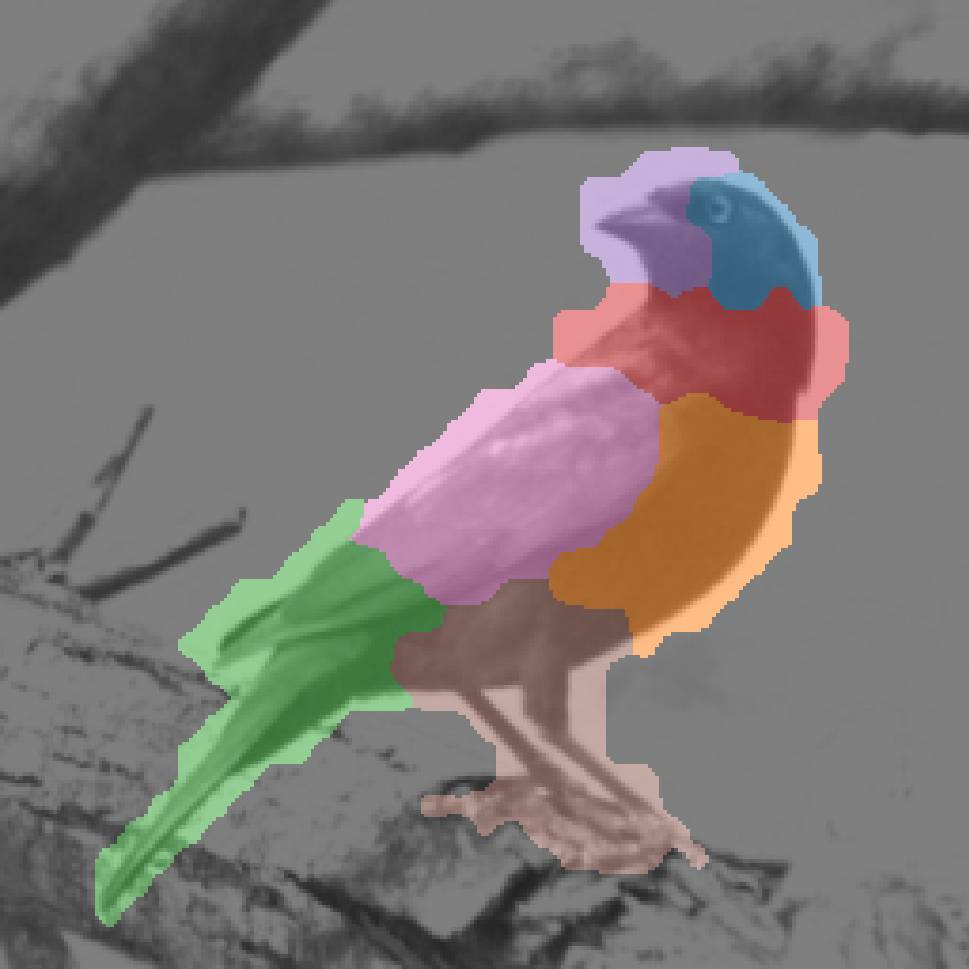}         &
 \includegraphics[width=0.1\textwidth]{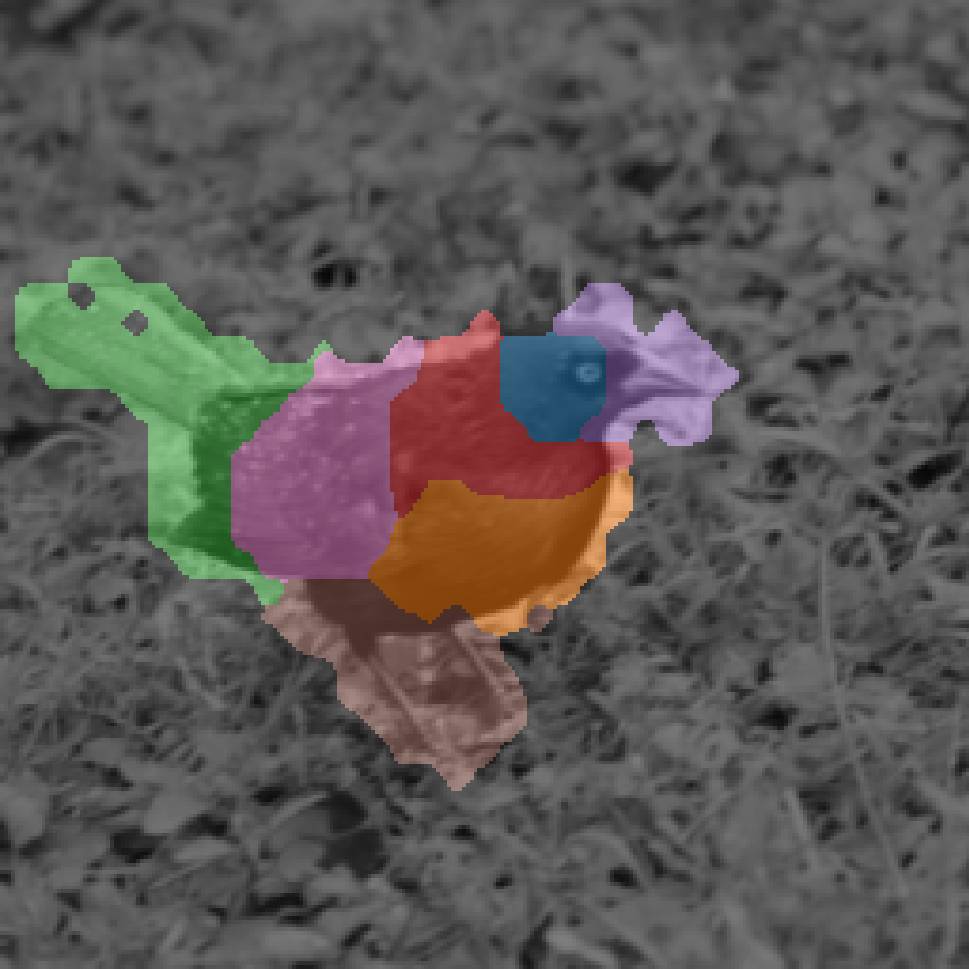}         &
 \includegraphics[width=0.1\textwidth]{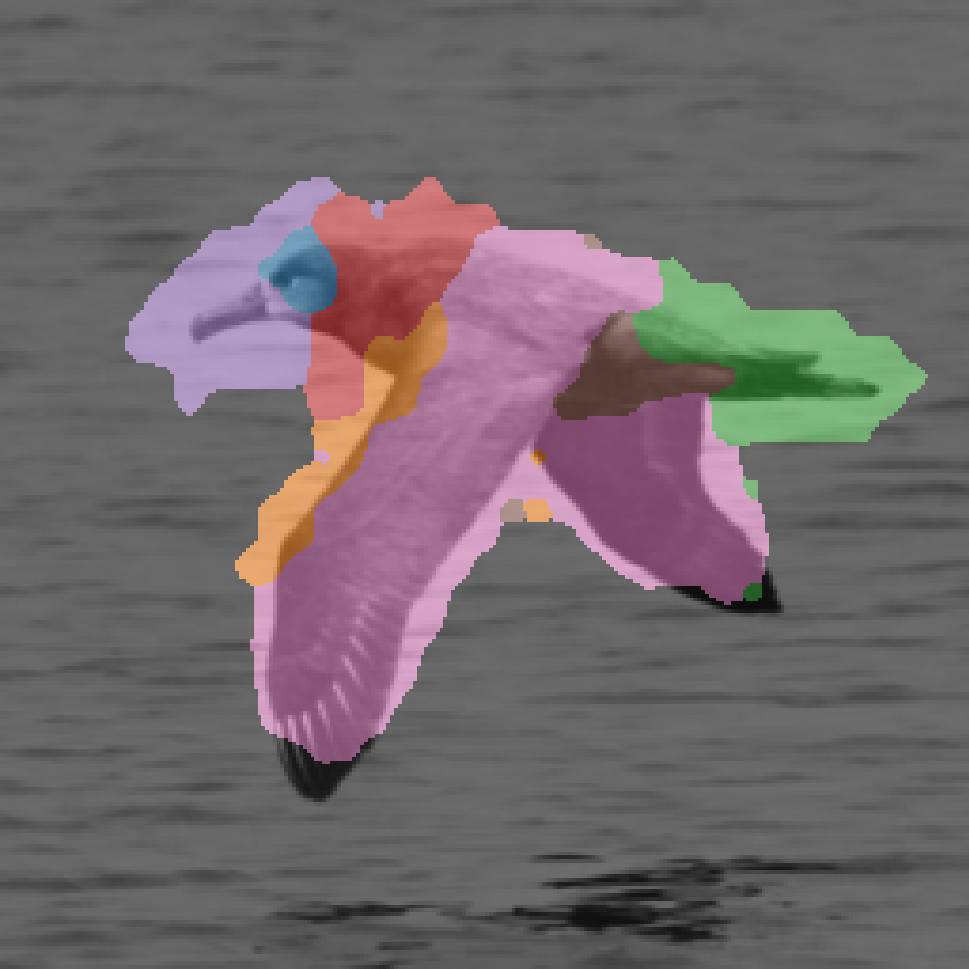}         &
 \includegraphics[width=0.1\textwidth]{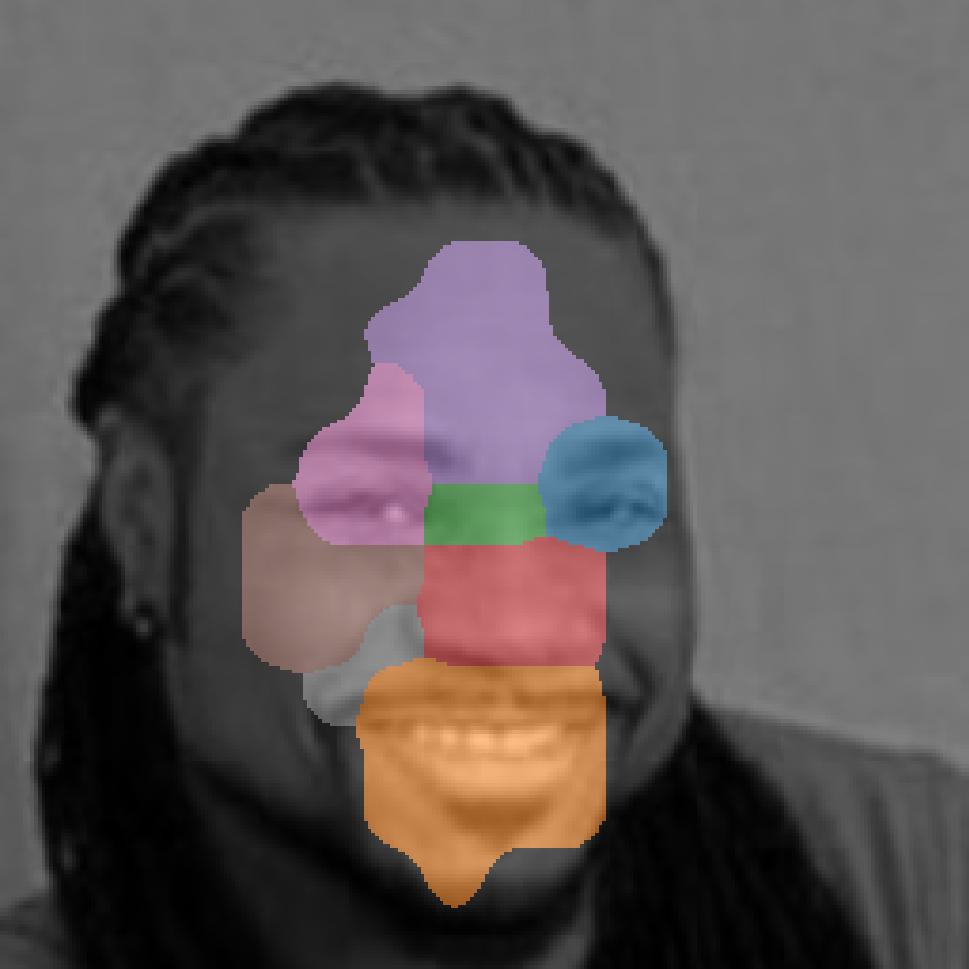}     &
 \includegraphics[width=0.1\textwidth]{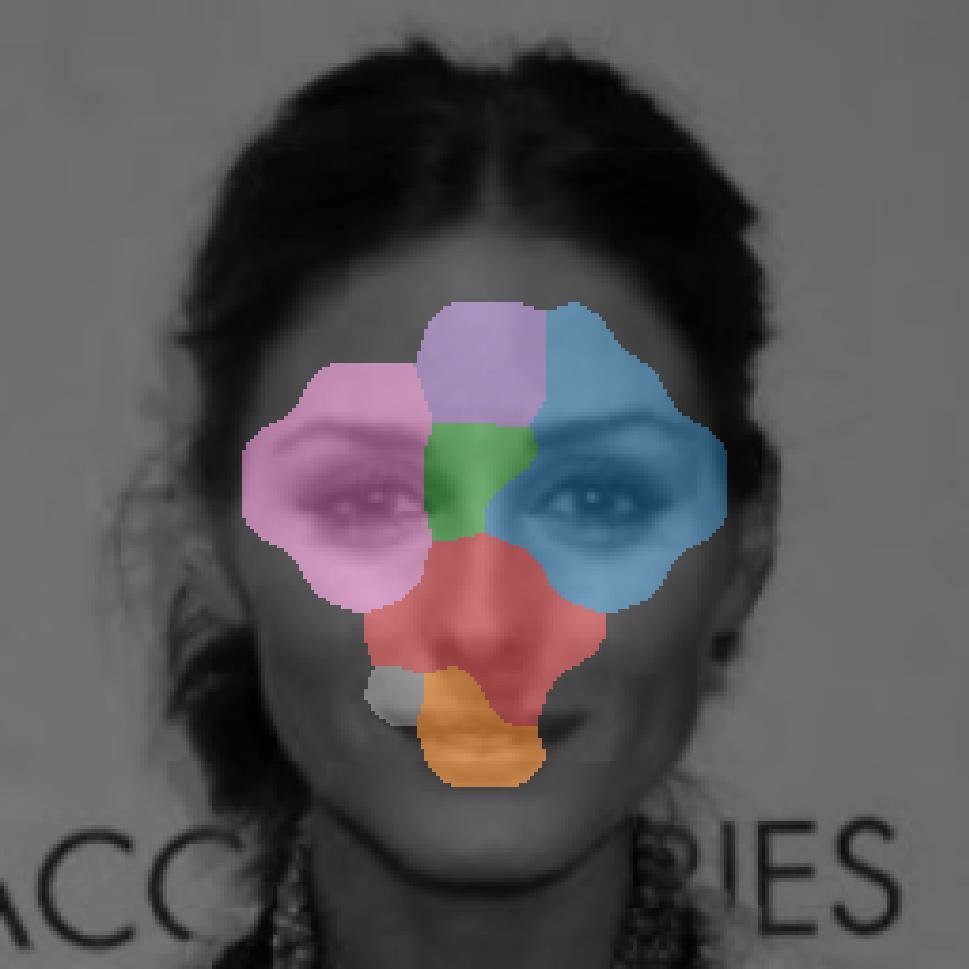}    &
 \includegraphics[width=0.1\textwidth]{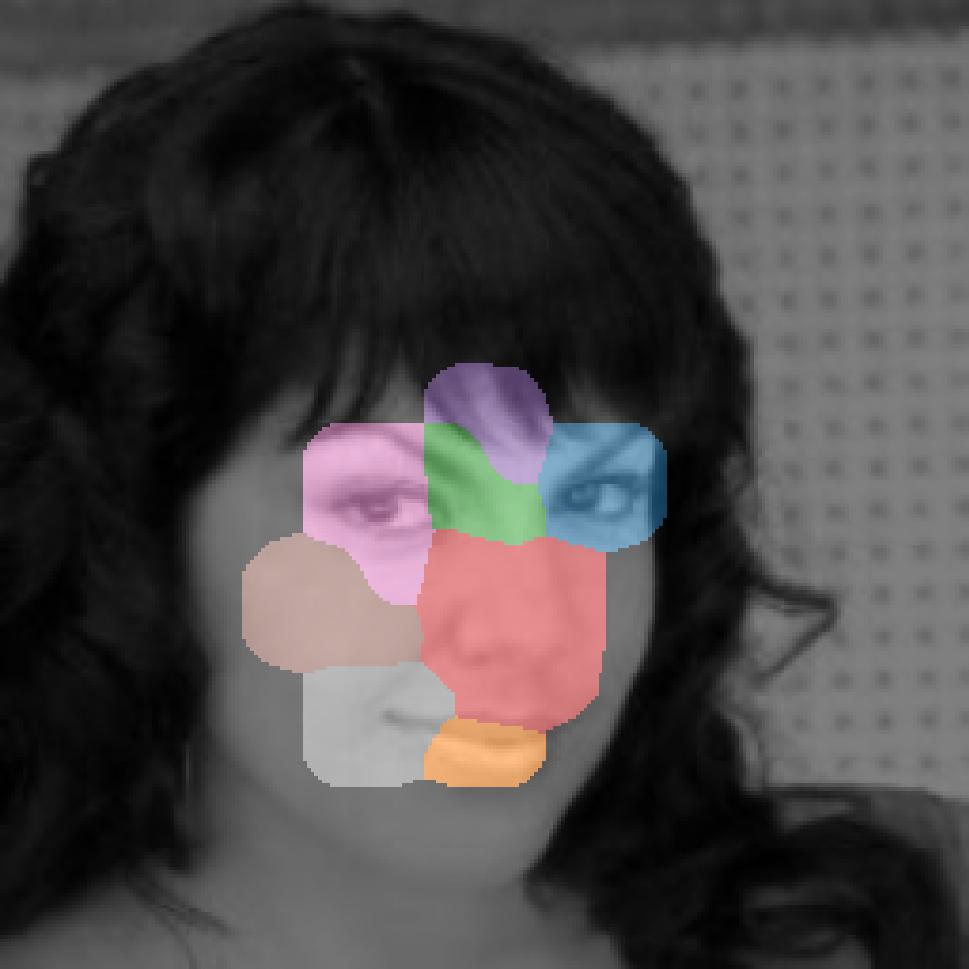}   &
 \includegraphics[width=0.1\textwidth]{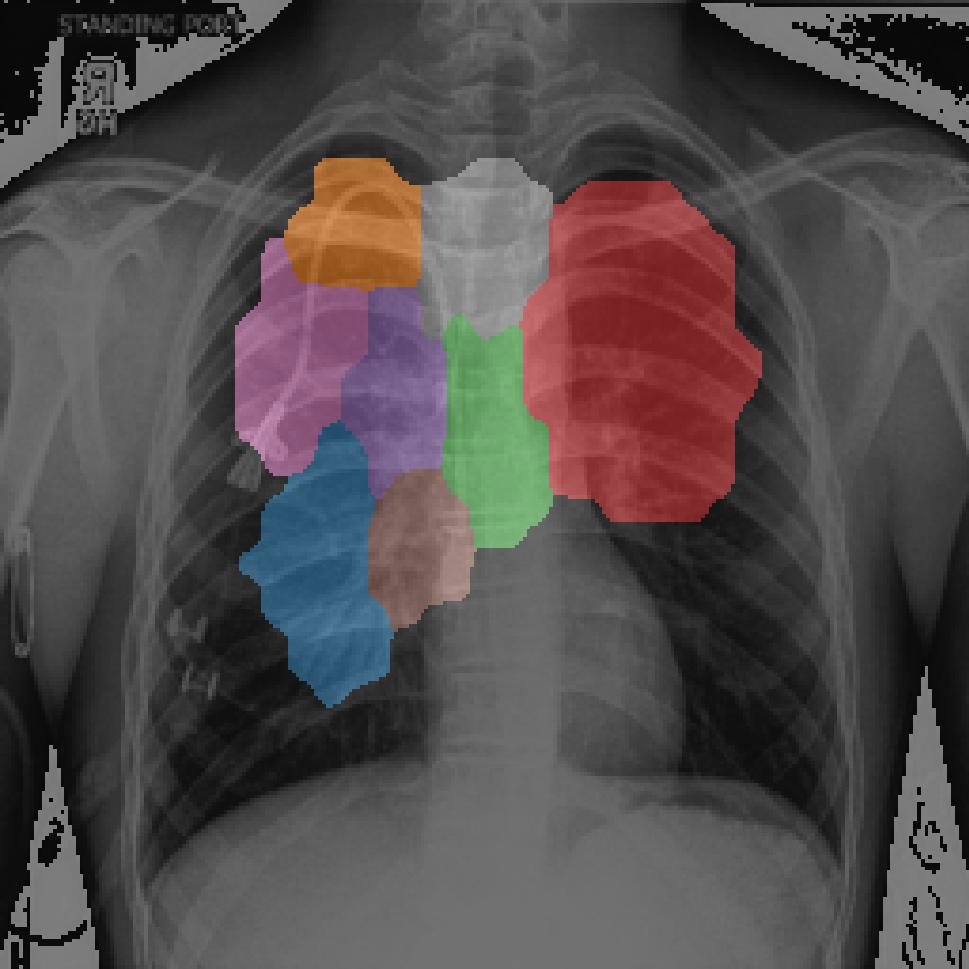}    &
 \includegraphics[width=0.1\textwidth]{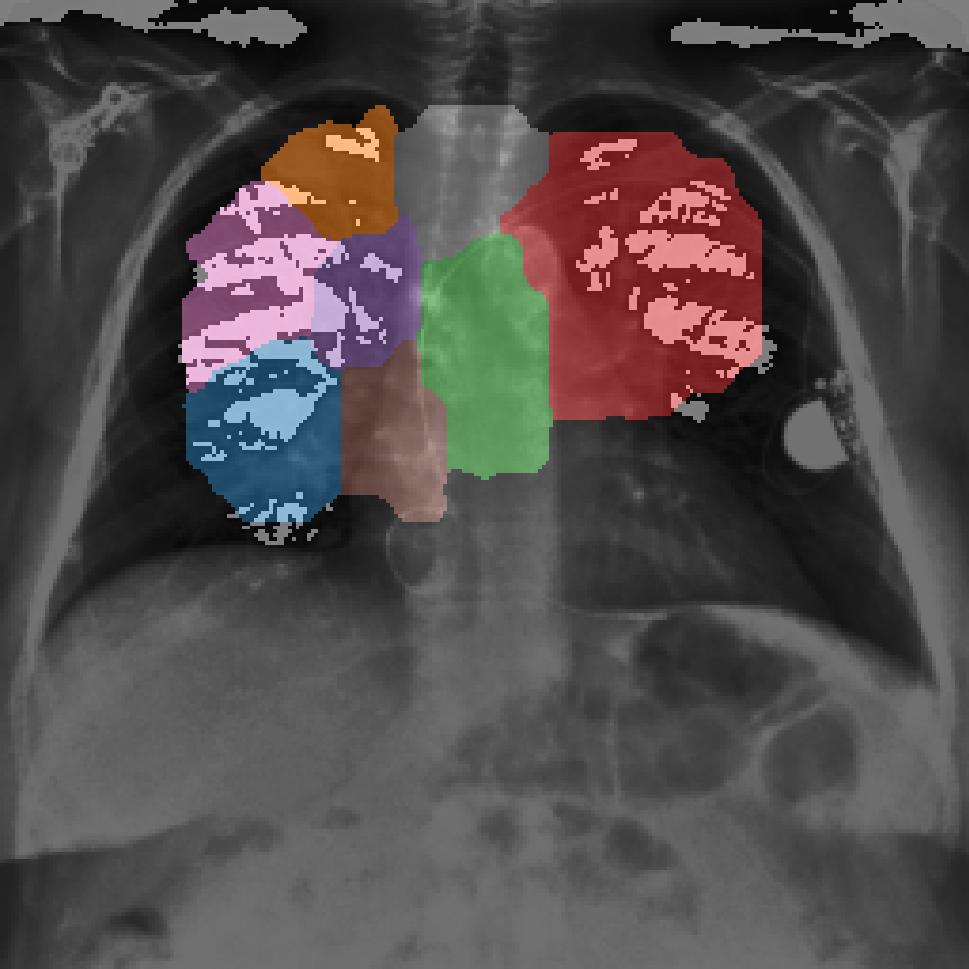}    &
 \includegraphics[width=0.1\textwidth]{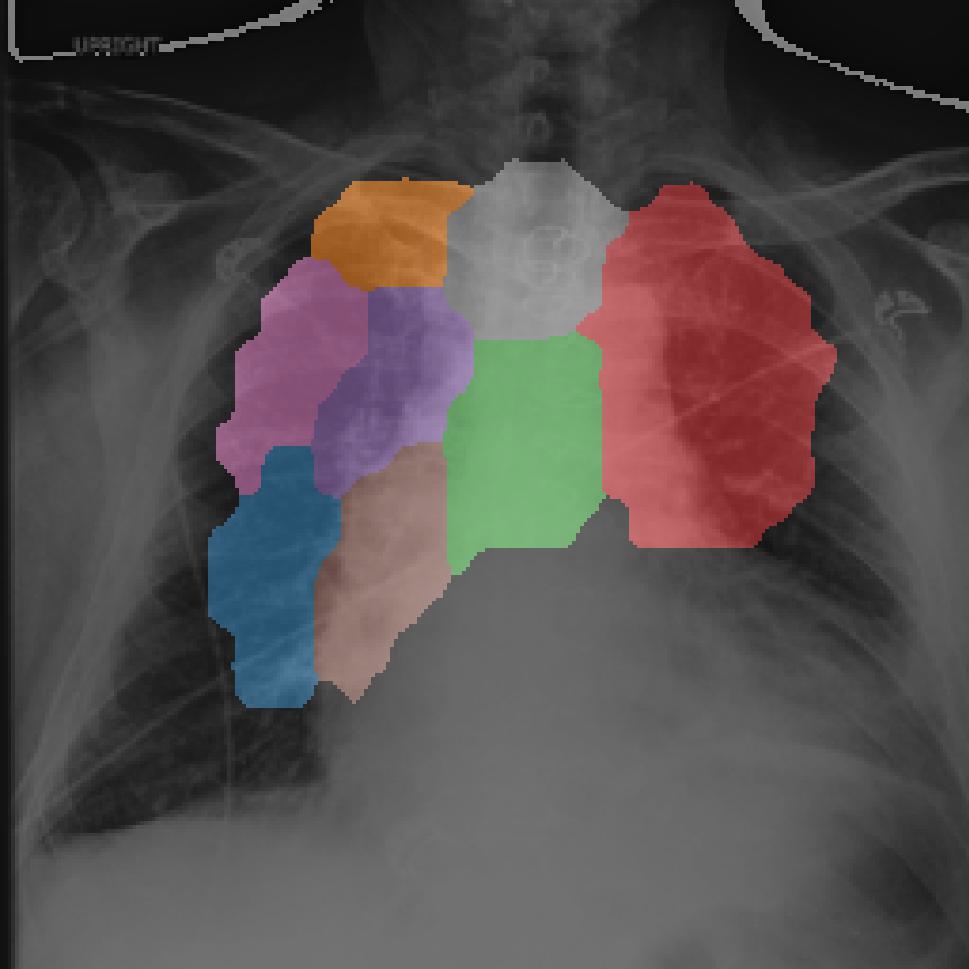}    
 \\
 \end{tabular}
\end{adjustbox}
\caption{Examples of part segmentation maps used in the linear probing experiments. }
\label{fig:qualitatives_part1}
\end{figure}

\subsection{Benchmarking intra-object leakage in vision backbones}

We quantify intra-object leakage across pretrained backbones using linear probing to predict image-level attributes from fixed part representations, extracted via either late or early masking. We generate the binary part masks using PDiscoFormer~\cite{aniraj2024pdiscoformer} (see \cref{fig:qualitatives_part1}).
For \textbf{late masking}, we pass the unmasked image through the frozen backbone and aggregate the resulting patch tokens via a mask-weighted average. Conversely, \textbf{early masking} enforces strict architectural isolation, mirroring our Stage 2 mechanism (\cref{sec:method_2}). We replicate the backbone's prefix tokens (e.g., \texttt{[CLS]}) for each discovered part and apply hard binary masks directly to the self-attention matrices across all layers. This restricts the $k$-th prefix token to attend exclusively to patches within the $k$-th mask. By severing attention links between distinct parts, early masking eliminates representation-level leakage and serves as a performance upper bound for our PS and MPPO metrics.
We evaluate PS on CUB and CelebA, and MPPO on CheXpert. For CUB and CelebA, we benchmark standard models (DINOv1~\cite{caron2021emerging}, DINOv2~\cite{oquab2023dinov2}, DINOv3~\cite{simeoni2025dinov3}, CLIP~\cite{radford2021learning}, MAE~\cite{he2022masked}) alongside locality-optimized backbones (CLIPpy \cite{ranasinghe2023perceptual}, SelfPatch~\cite{yun2022patch}, SAM~\cite{Kirillov2023SegmentAnything}, DenseSSL~\cite{qiu2024mind}). SelfPatch and SAM lack an aggregation token, precluding a straight-forward evaluation with early masking. For CheXpert, we evaluate the domain-specific RAD DINO~\cite{perez2024rad}, alongside DINOv2 and DINOv3 for reference.

\subsection{Impact of mitigating intra-object leakage on part discovery}

In this experimental section we investigate the impact that intra-object leakage has on part discovery driven by part attribute prediction.
To do this, we perform part-informed part discovery, as described in Section~\ref{sec:method_2}, using four different settings. The first is the stage 1 model that uses only late masking, serving as a baseline.
We then test three different settings in which a second stage is added that uses early masking with (1) the original soft masks stemming from part discovery, (2) hard masks obtained by applying an argmax on the soft masks and (3) a straight-through gradient approach that uses hard masks in the forward pass and soft masks for backpropagation.
This two-stage setting enables more informed part discovery, as the early-masked attribute prediction stage may provide explicit guidance to the part discovery module.
Unlike in previous works, the part quality evaluation on CUB and CelebA is performed on merged key point groups, as described in Section~\ref{sec:datasets}, favouring situations in which the discovered parts coincide with the defined semantic parts.
For reference, we provide part discovery metrics using this same setting with two state-of-the-art approaches: PdiscoFormer~\cite{aniraj2024pdiscoformer} and MPAE~\cite{MPAE} for CUB and CelebA.

\begin{table}[t]
\centering
\caption{
Part specificity (PS) with linear probing on CUB and CelebA using different backbones 
and either late or early masking. Early masking consistently yields substantially 
higher specificity. \label{tab:late_vs_early_cub_celeba}
}
\begin{adjustbox}{width=0.95\textwidth}
\setlength{\tabcolsep}{4pt}
\begin{tabular}{lc cc cc}
\toprule
 & & \multicolumn{2}{c}{CUB} 
 & \multicolumn{2}{c}{CelebA} \\
\cmidrule(lr){3-4}
\cmidrule(lr){5-6}
Method
& Arch
& Late masking & Early masking
& Late masking & Early masking \\
\midrule
DINOv3 \cite{simeoni2025dinov3} & ViT B-16        & 0.019 & \textbf{0.070} & 0.036 & 0.266 \\
DINOv2 \cite{oquab2023dinov2} & ViT B-14      & 0.032 & 0.062 & 0.018 & 0.265 \\
DINOv1 \cite{caron2021emerging} & ViT B-16       & 0.036 & 0.056 & 0.021 & 0.260 \\
CLIP  \cite{radford2021learning}   & ViT B-16      & 0.025 & 0.062 & 0.040 & \textbf{0.272} \\
MAE \cite{he2022masked} & ViT B-16  & 0.033 & 0.040 & 0.061 & 0.241 \\
\midrule
CLIPpy\cite{ranasinghe2023perceptual} & ViT B-16 & 0.027	& 0.057	& 0.037	& 0.261 \\
SelfPatch\cite{yun2022patch} & ViT S-16 & \textbf{0.047}	& -- & 0.056 & --  \\
SAM\cite{Kirillov2023SegmentAnything} & ViT B-16 & 0.017	& -- & \textbf{0.094} & --  \\
Dense SSL\cite{qiu2024mind} & ViT S-16 & 0.017	& 0.043 & 0.036 & 0.257  \\
\bottomrule
\end{tabular}
\end{adjustbox}

\end{table}
\begin{table}[t]
\centering
\small 
\setlength{\tabcolsep}{3pt} 
\caption{Results on the most predictive part overlap (MPPO) with ground truth masks on the CheXpert dataset. All the methods use the ViT-Base architecture. }
\label{tab:late_vs_early_chext}
\begin{adjustbox}{width=0.95\textwidth}
\begin{tabular}{l *{10}{c} |c}
\toprule
& \rotatebox{80}{Enlarged Cardio.} & \rotatebox{80}{Cardiomegaly} & \rotatebox{80}{Lung Opacity} & \rotatebox{80}{Lung Lesion} & \rotatebox{80}{Edema} & \rotatebox{80}{Consolidation} & \rotatebox{80}{Atelectasis} & \rotatebox{80}{Pneumothorax} & \rotatebox{80}{Pleural Effusion} & \rotatebox{80}{Support Devices} & \rotatebox{80}{\textbf{Average}} \\
\midrule
DINOv3 \cite{simeoni2025dinov3} late & 0.68	& 0.57	& 0.43 & 0.36	& 0.78	& 0.46	& 0.37	& 0.30	& 0.24	& 0.68	& 0.49 \\
DINOv3 \cite{simeoni2025dinov3} early & 0.77	& 0.69  & 0.44	& \textbf{0.43}	& 0.76	& 0.66	& 0.37	& 0.50	& 0.31	& 0.75	& 0.57 \\
\midrule
DINOv2 \cite{oquab2023dinov2} late & 0.71	& 0.59	& 0.44	& \textbf{0.43}	& 0.82	& 0.49	& 0.30	& 0.40	& 0.26	& 0.75	&  0.52 \\
DINOv2 \cite{oquab2023dinov2} early & 0.78	& 0.65	& 0.41	& 0.36	& 0.87	& \textbf{0.60}	& 0.33	& 0.60 & 0.20	& 0.78	& 0.56 \\
\midrule
RAD DINO \cite{perez2024rad} late & 0.80 & 0.64 & 0.38 & \textbf{0.43} & 0.79 & 0.51 & 0.35 & 0.40 & 0.20 & 0.77 & 0.53 \\
RAD DINO \cite{perez2024rad} early & \textbf{0.81} & \textbf{0.69} & \textbf{0.48} & 0.36 & \textbf{0.84} & \textbf{0.60} & \textbf{0.36} & \textbf{0.90} & \textbf{0.28} & \textbf{0.89} & \textbf{0.62} \\
\bottomrule
\end{tabular}
\end{adjustbox}
\end{table}
\begin{table}[t]
\centering
\setlength{\tabcolsep}{5pt}
\caption{Single vs. Two-stage variants for attribute-prediction-driven part discovery. Models trained on CUB and CelebA use the DinoV3 ViT-Base backbone and part-merged keypoints for evaluation, while CheXpert models use RAD-DINO ViT-Base.}
\label{tab:single_vs_two_stage}
\begin{adjustbox}{width=0.95\textwidth}
\begin{tabular}{l l ccc ccc c}
\toprule
 &  & \multicolumn{3}{c}{CUB} 
 & \multicolumn{3}{c}{CelebA}
 & \multicolumn{1}{c}{CheXpert} 
 \\
\cmidrule(lr){3-5} 
\cmidrule(lr){6-8}
\cmidrule(lr){9-9} 
Setting & Method 
& ARI & NMI & PS 
& ARI & NMI & PS 
& MPPO 
\\
\midrule
Single-stage 
& -- 
& 50.27 & 62.03 & 0.035 
& 49.46 & 57.50 & 0.017 
& 0.58 
\\
\midrule
\multirow{3}{*}{Two-stage}
& Hard masks 
& 71.01 & 67.32 & 0.052 
& 80.34	& 79.44 & 0.163 
& 0.56 
\\
& STE masks
& \textbf{73.93} & \textbf{70.85} & \textbf{0.056} 
& \textbf{97.51}	& \textbf{94.49} & \textbf{0.167} 
& \textbf{0.76} 
\\
& Soft masks 
& 41.95 & 59.72 & 0.046  
& 69.38	& 81.26	& \textbf{0.167} 
& 0.28 
\\
\midrule
\multirow{2}{*}{}
& PDiscoFormer~\cite{aniraj2024pdiscoformer}  
& 47.76 & 69.37	 & - 
& 57.94 & 69.44 & - 
& - 
\\
& MPAE \cite{MPAE}
& 44.90 & 69.54 & -
& 91.77 &86.91	& - 
& - 
\\
\bottomrule
\end{tabular}
\end{adjustbox}
\end{table}

\section{Results and Discussion}

\noindent \textbf{Benchmarking intra-object leakage in vision backbones.}
The backbone benchmarking results in Table \ref{tab:late_vs_early_cub_celeba} show a very consistent behaviour of PS for both datasets, with much higher values of PS when using early masking (around $3\times$ in CUB and $10\times$ in CelebA) compared to the same backbone using late masking. This confirms that there is a big intra-object leakage problem in all tested backbones. The differences between both datasets are also remarkable, with both showing similar PS with late masking, but CelebA resulting in a much higher score than CUB with early masking. This is likely due to the nature of the attribute sets, since CUB attributes will naturally incur in stronger correlations: we can imagine that the colour of a bird's head and its wing's colour may display a more consistent correlation than the attributes ``has hat'' and ``is smiling'' in CelebA. Among the tested locality-inducing methods, we also observe varying behavior for the different datasets: SelfPatch~\cite{yun2022patch} results in a consistent, although small, improvement in PS on both CUB (from the 0.032 of DINOv2 to 0.047) and CelebA (from 0.018 to 0.056), SAM~\cite{Kirillov2023SegmentAnything} only provides an improvement (to 0.094) in CelebA, while DenseSSL~\cite{qiu2024mind} provides results in line with DINOv3 in CelebA but the lowest early masking results in CUB. This is likely down to the way and the dataset on which these models were trained, with SelfPatch trained in a task agnostic manner, while SAM was trained using segmentation masks, in which human faces are likely to be overrepresented, encouraging the model to learn a more disentangled representation of human face parts than bird body parts. In any case, there remains a large gap to fill in order to reach the disentanglement of the two-stage upper bound.

\noindent \textbf{Intra-object leakage in a safety-critical domain (CheXpert).} 
The consequences of this leakage are especially severe in medical imaging, where an interpretable model must predict a pathology by looking at affected tissue, not correlated clinical artifacts like a chest drain tube. As shown in \cref{tab:late_vs_early_chext}, standard late masking severely compromises spatial grounding: even RAD DINO~\cite{perez2024rad}, a chest X-ray foundation model, achieves a Most Predictive Part Overlap (MPPO) of only 0.53, indicating predictions frequently stem from outside the ground-truth pathology region. Enforcing early masking on RAD DINO jumps the MPPO to 0.62, proving the underlying model inherently entangles spatial features. Our proposed two-stage architecture effectively mitigates this. As reported in \cref{tab:single_vs_two_stage}, replacing the single-stage baseline (MPPO 0.58) with our two-stage Straight-Through (STE) strategy dramatically increases MPPO to 0.76. This strict architectural isolation forces the model to base predictions on correct anatomical regions, whereas the failure of the soft masking variant (MPPO 0.28) proves that allowing even minimal continuous gradient flow across boundaries causes the model to revert to exploiting leakage.

\noindent \textbf{Impact of mitigating intra-object leakage on part discovery.} 
Our two-stage approach significantly improves both faithfulness and semantic quality. Architectural isolation translates directly into improved PS: the STE variant raises CUB PS from 0.035 to 0.056, and achieves a massive ten-fold increase (0.017 to 0.167) on CelebA compared to the single-stage baseline (Table~\ref{tab:single_vs_two_stage}). This leap confirms our representations are far less contaminated by leakage. Consequently, this reduced leakage yields superior part discovery. Two-stage settings with hard or STE masks systematically outperform single-stage counterparts in clustering (e.g., ARI jumps from 50.27 to 73.93 on CUB, and 49.46 to 97.51 on CelebA). STE consistently performs best: its hard forward pass prevents leakage, while its soft backward pass signals the part discovery module to expand masks over the full region necessary for attribute prediction.

\begin{figure}[t]
\centering
\small
\setlength\tabcolsep{6pt} 
\renewcommand{\arraystretch}{0.3}
\centering
 \begin{tabular}{cccccc}
& Input & One-stage & Hard & STE & Soft  \\
 \multirow{2}{*}{\rotatebox[origin=tl]{90}{{\parbox{1.1cm}{\centering CUB}}}} & 
 \includegraphics[width=0.118\textwidth]{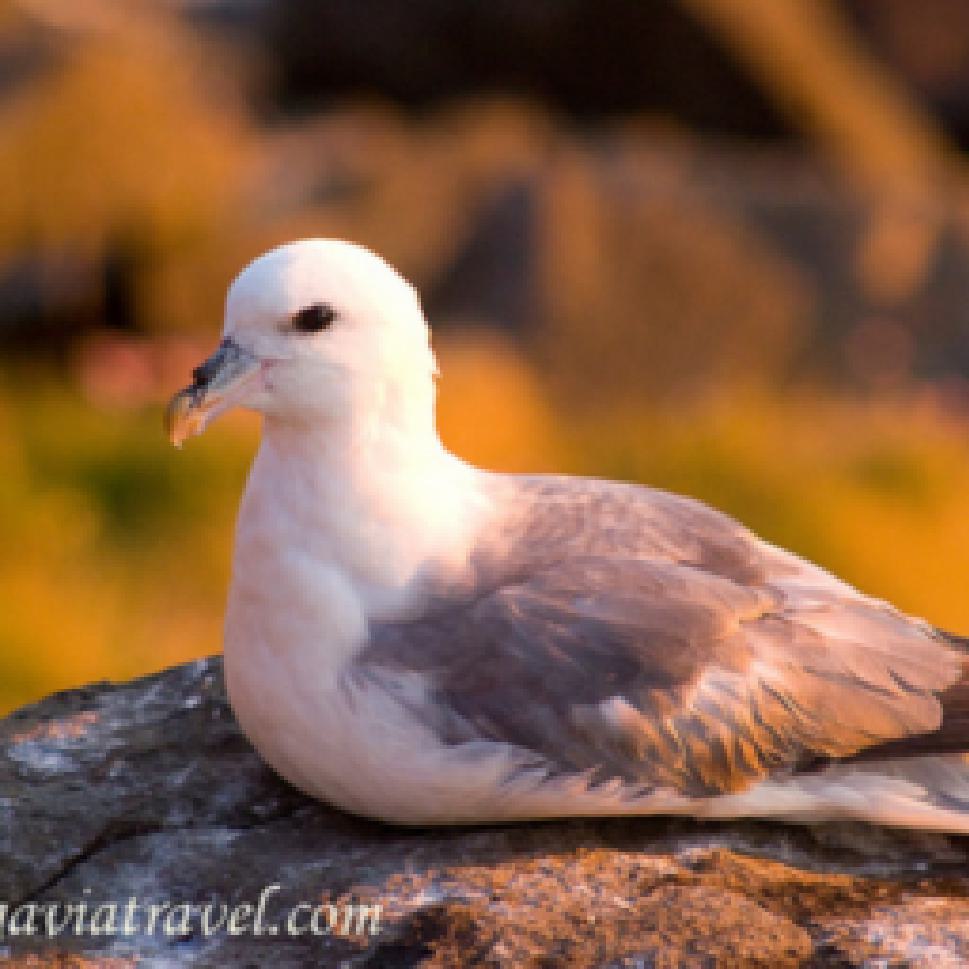} & 
 \includegraphics[width=0.118\textwidth]{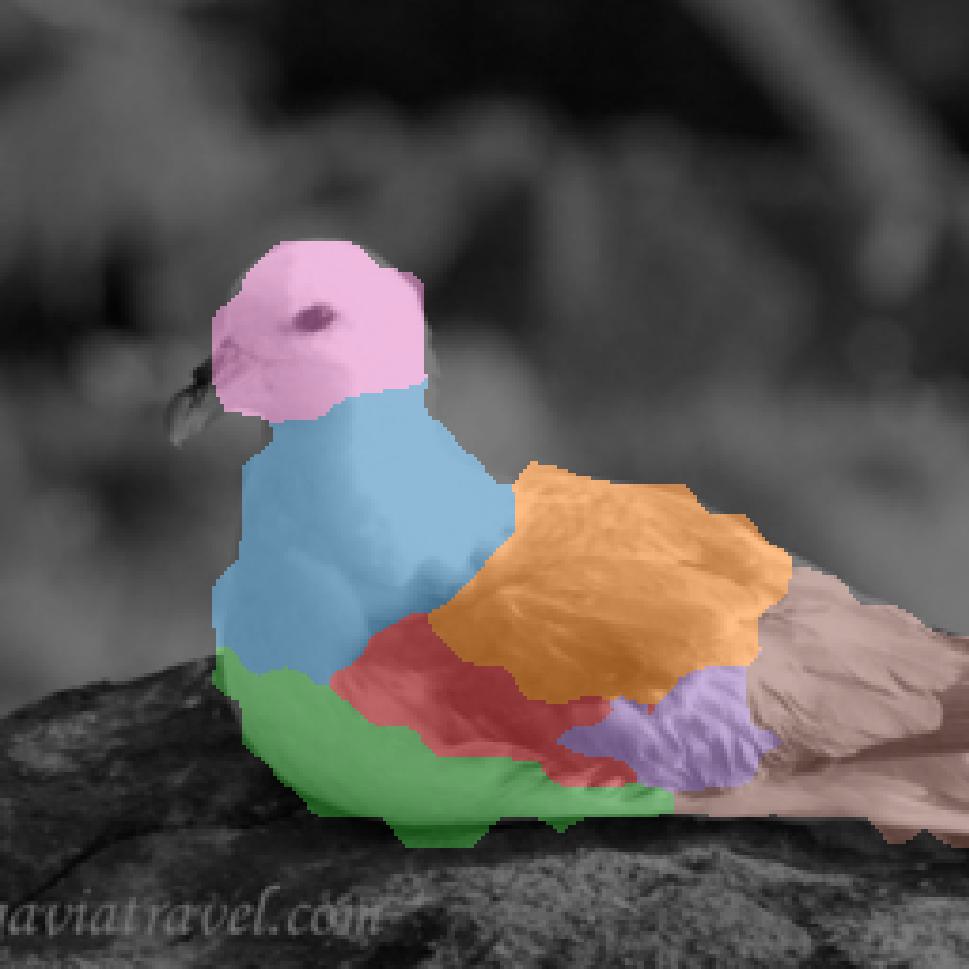} &

 \includegraphics[width=0.118\textwidth]{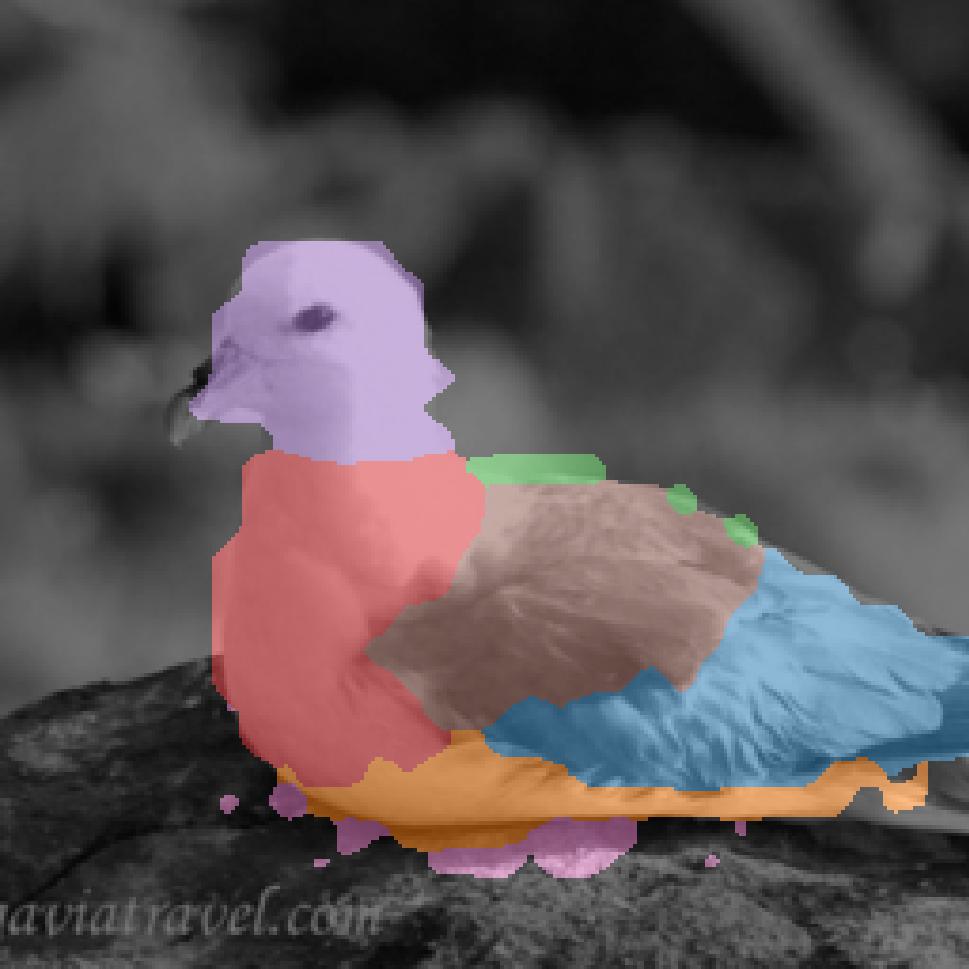}         &
 \includegraphics[width=0.118\textwidth]{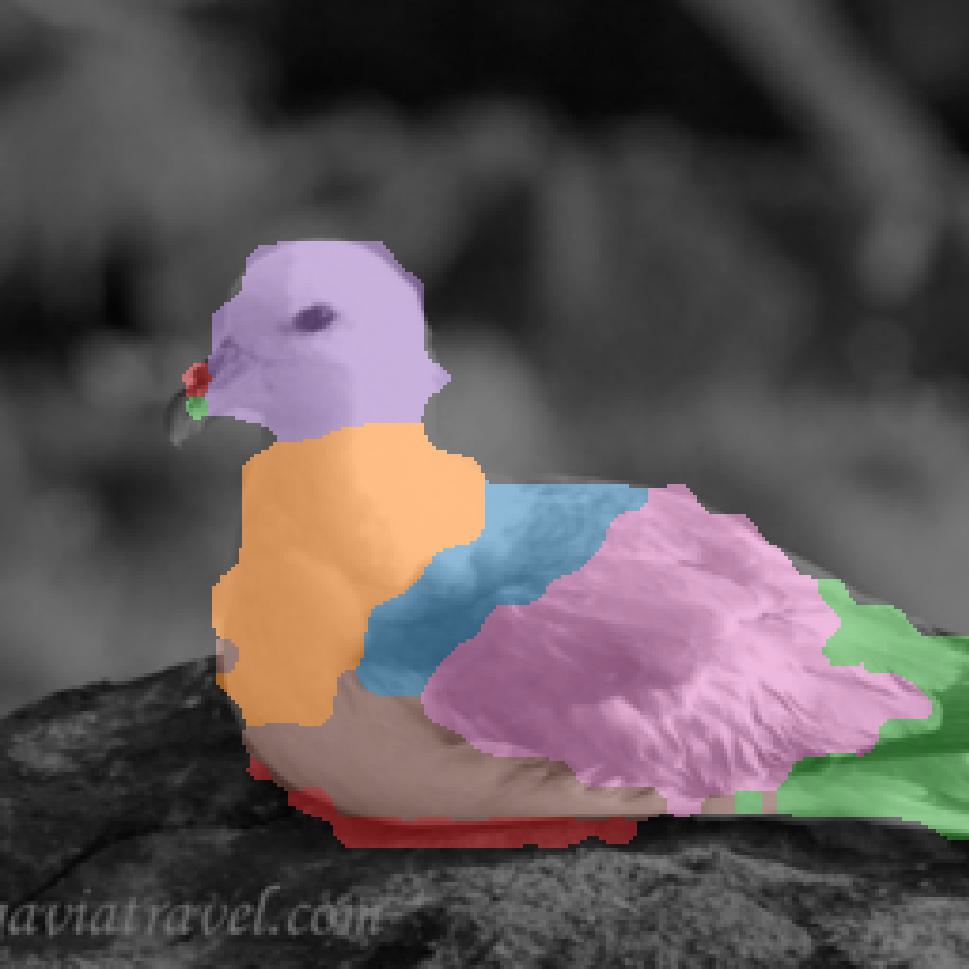}           &
 \includegraphics[width=0.118\textwidth]{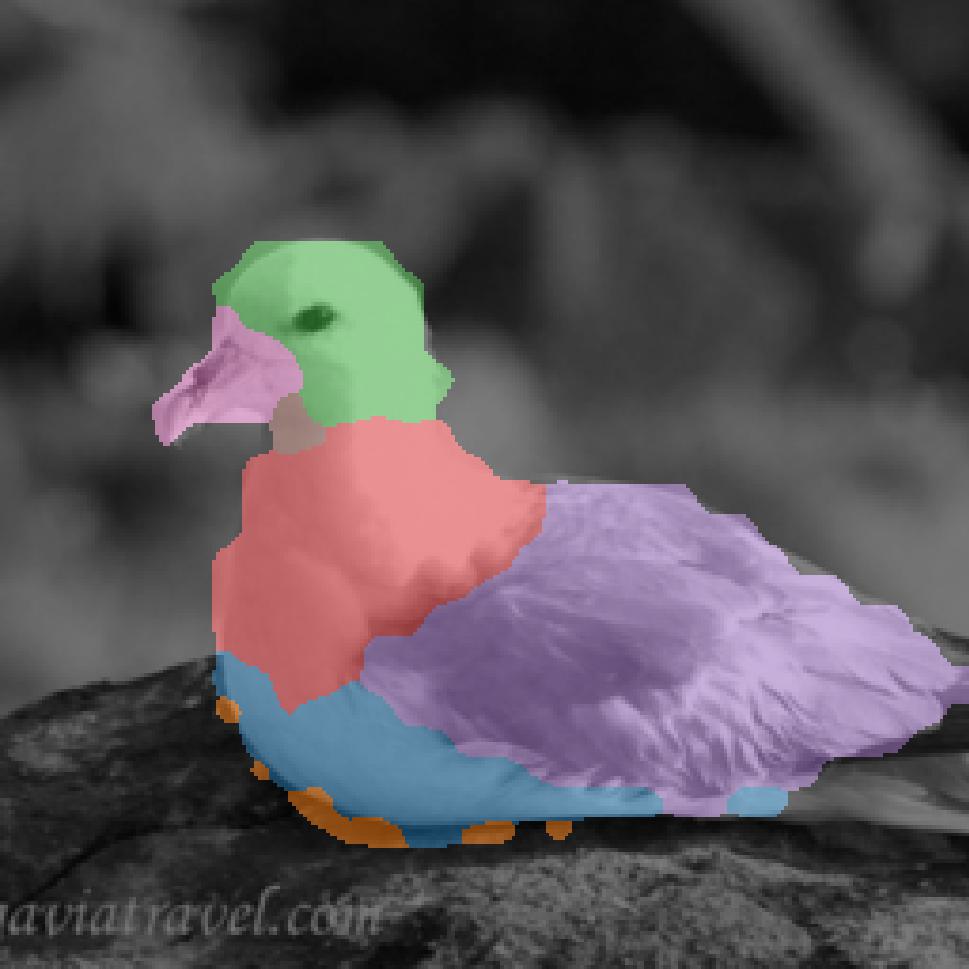}         \\
 &
\includegraphics[width=0.118\textwidth]{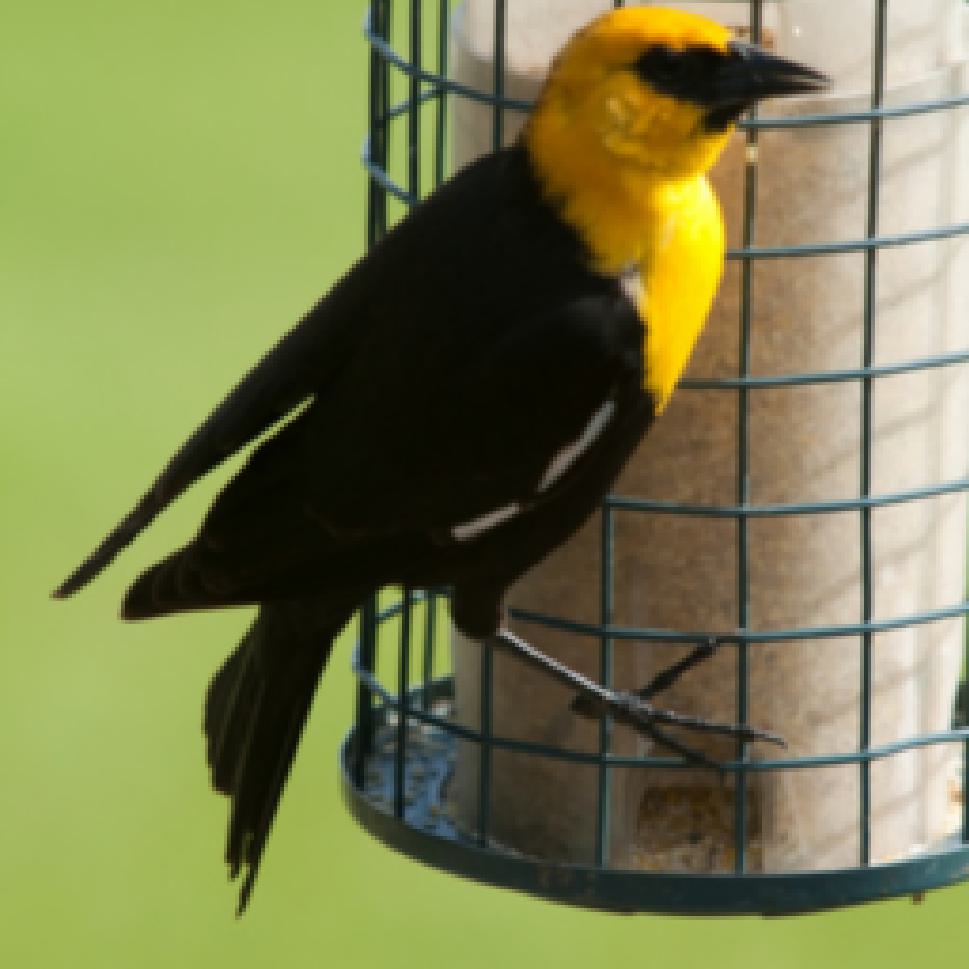} & 
 \includegraphics[width=0.118\textwidth]{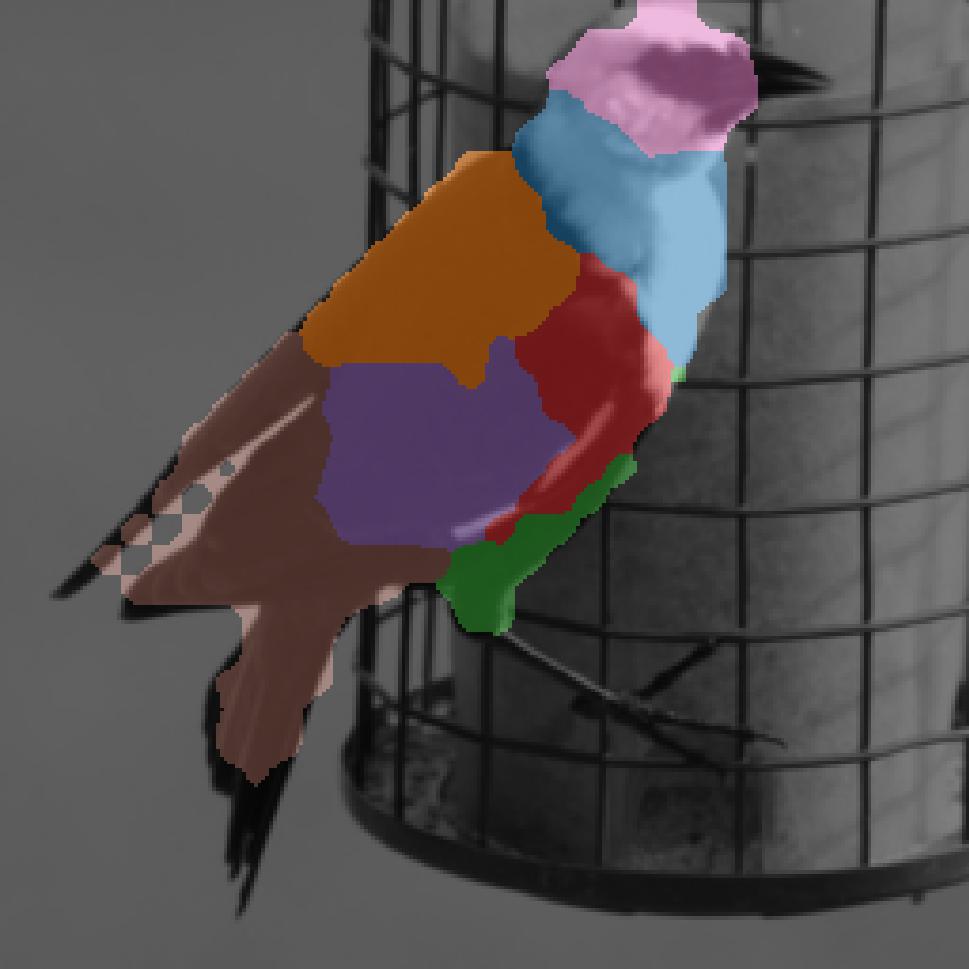} &
 \includegraphics[width=0.118\textwidth]{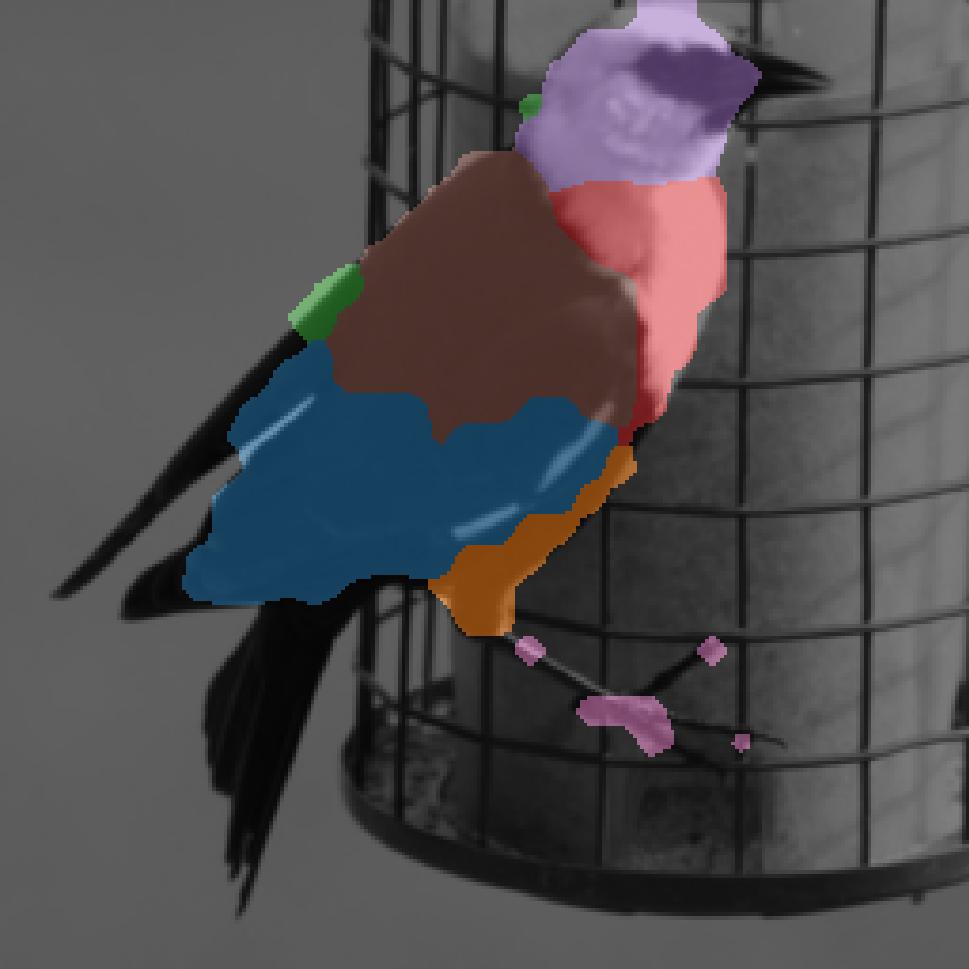}         &
 \includegraphics[width=0.118\textwidth]{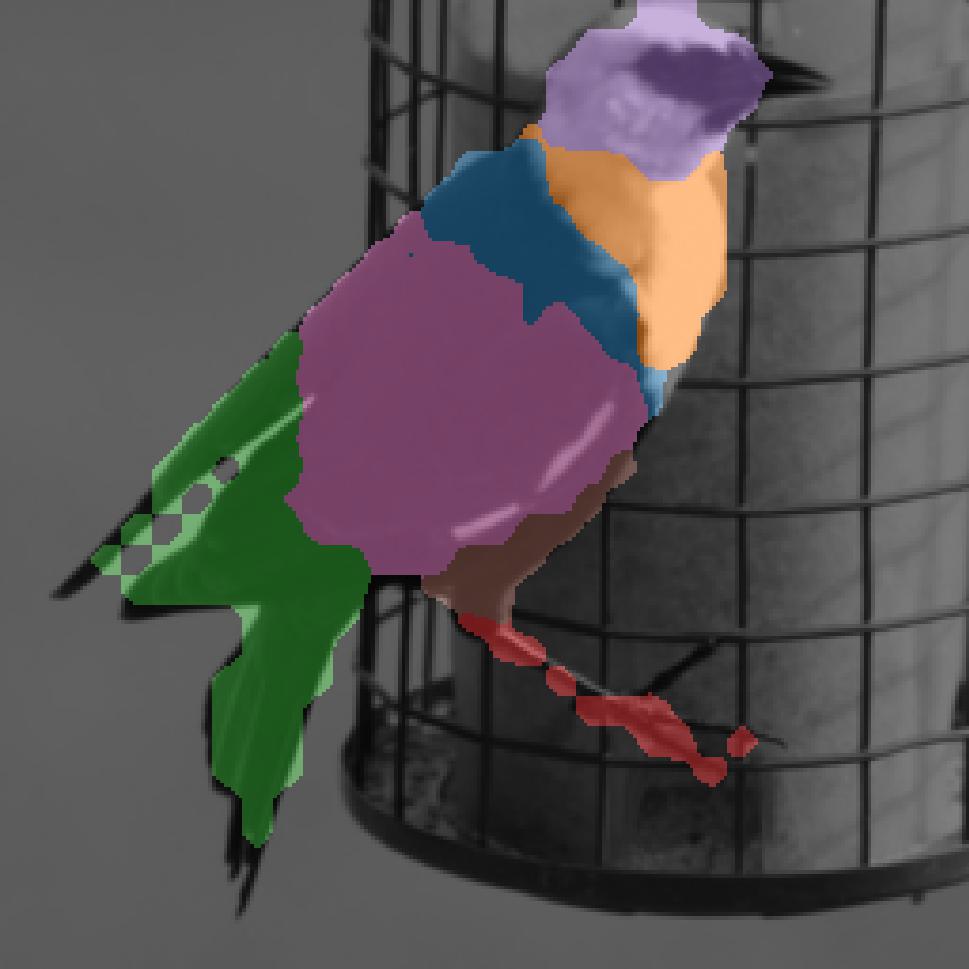}           &
 \includegraphics[width=0.118\textwidth]{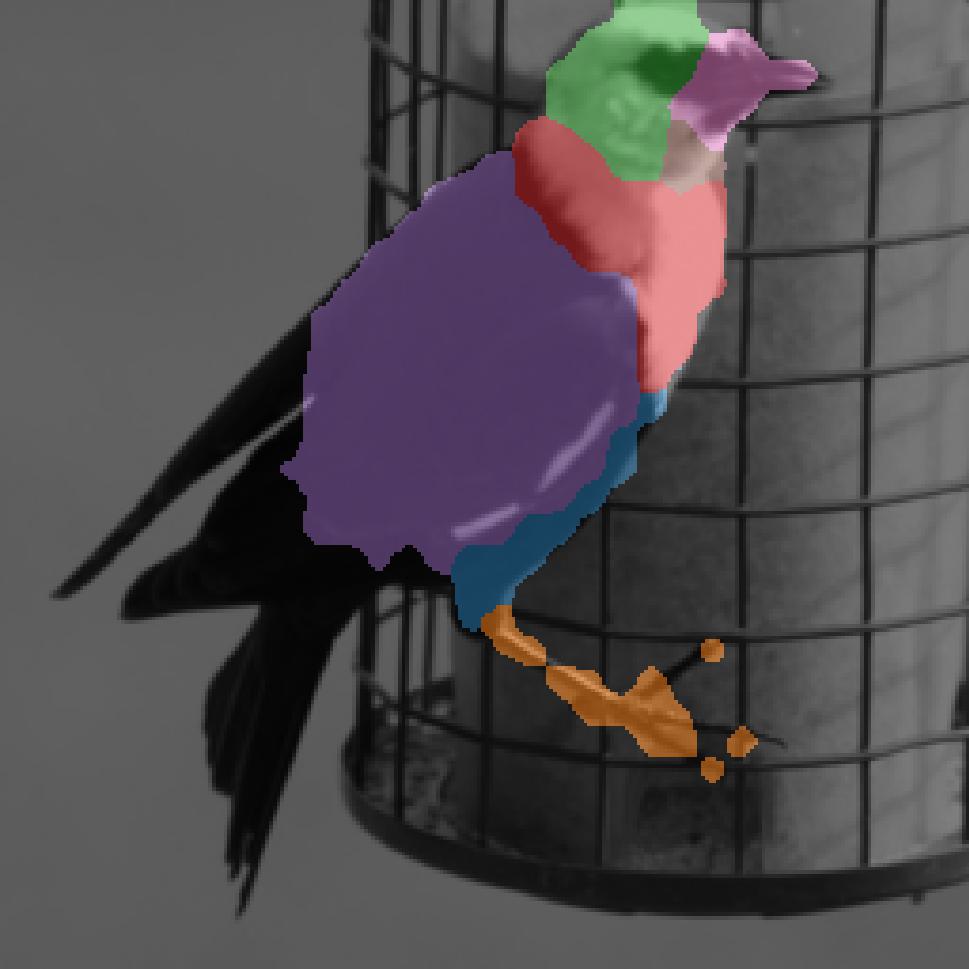}         \\ 
 \multirow{2}{*}{\rotatebox[origin=tl]{90}{{\parbox{1.1cm}{\centering CelebA}}}}  & 
 \includegraphics[width=0.118\textwidth]{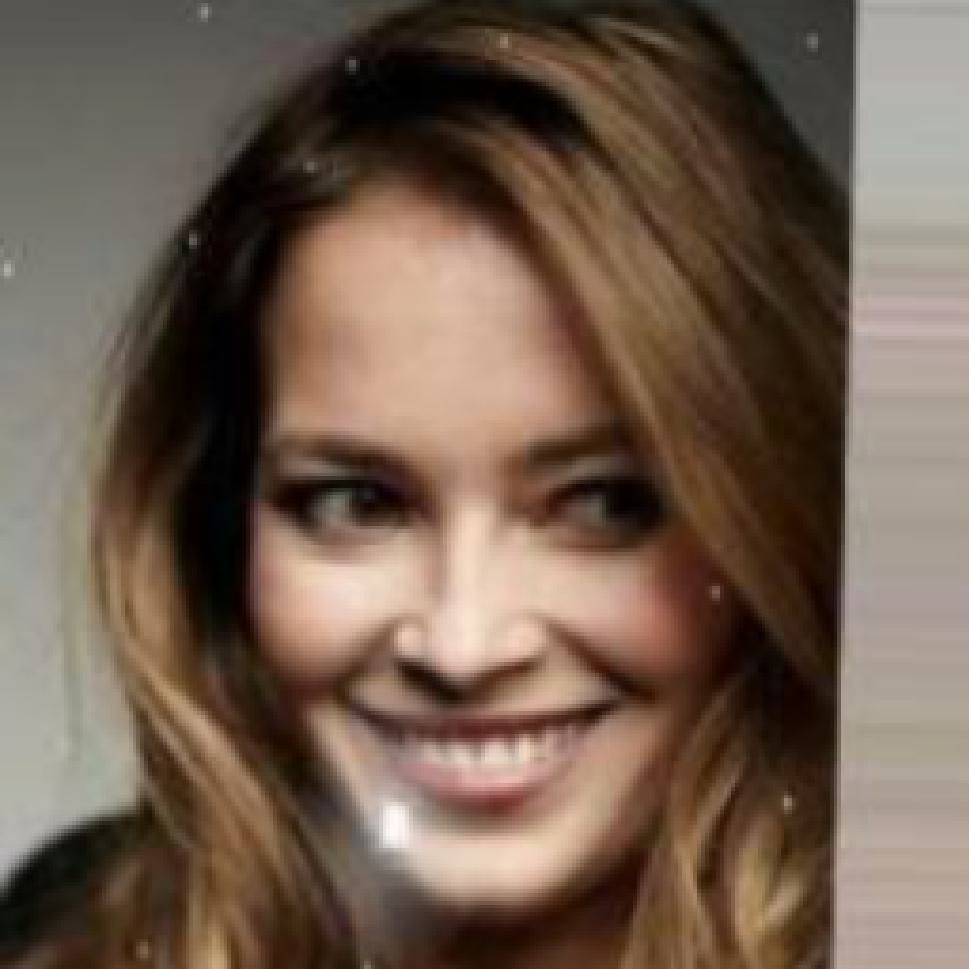} & 
 \includegraphics[width=0.118\textwidth]{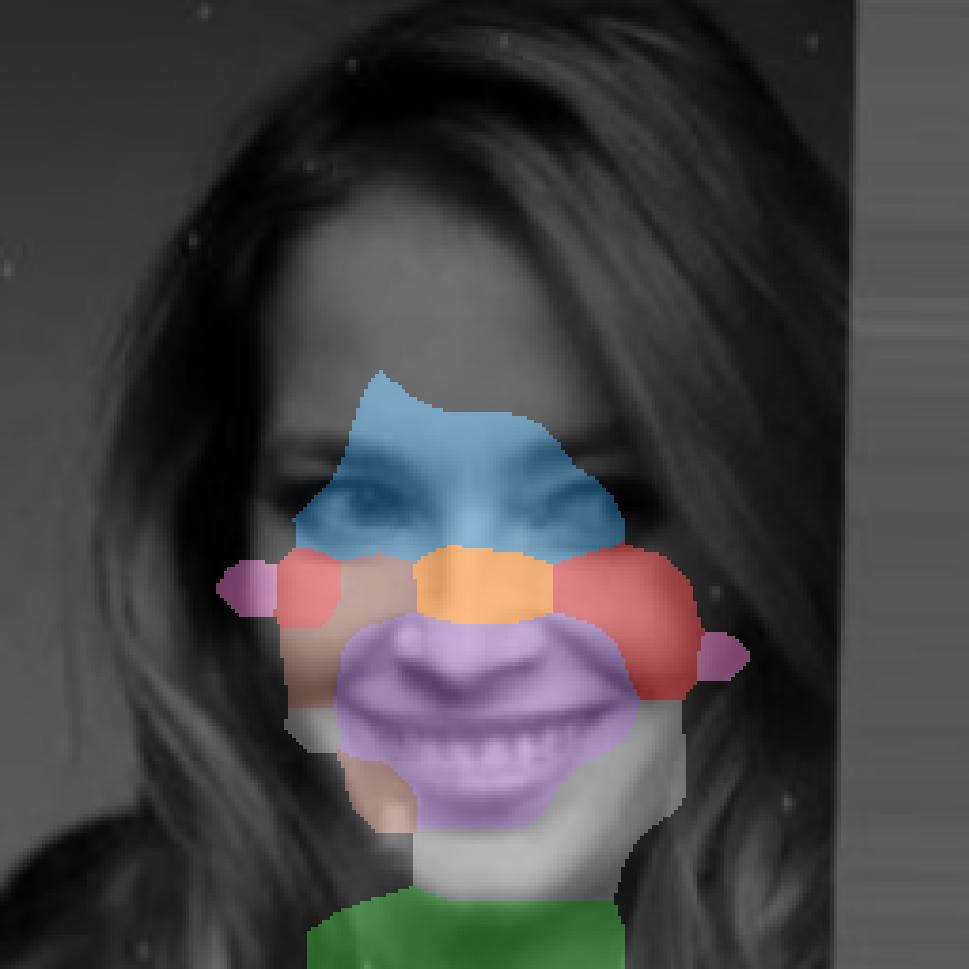} &
 \includegraphics[width=0.118\textwidth]{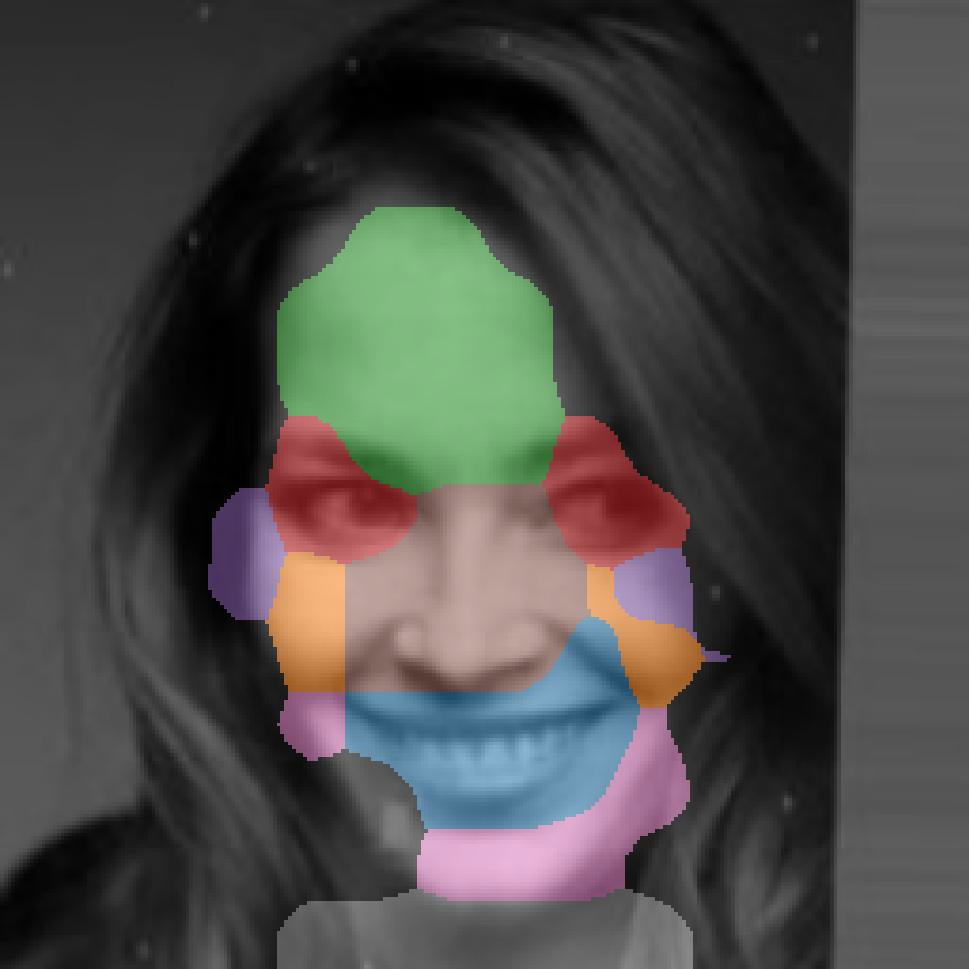}         &
 \includegraphics[width=0.118\textwidth]{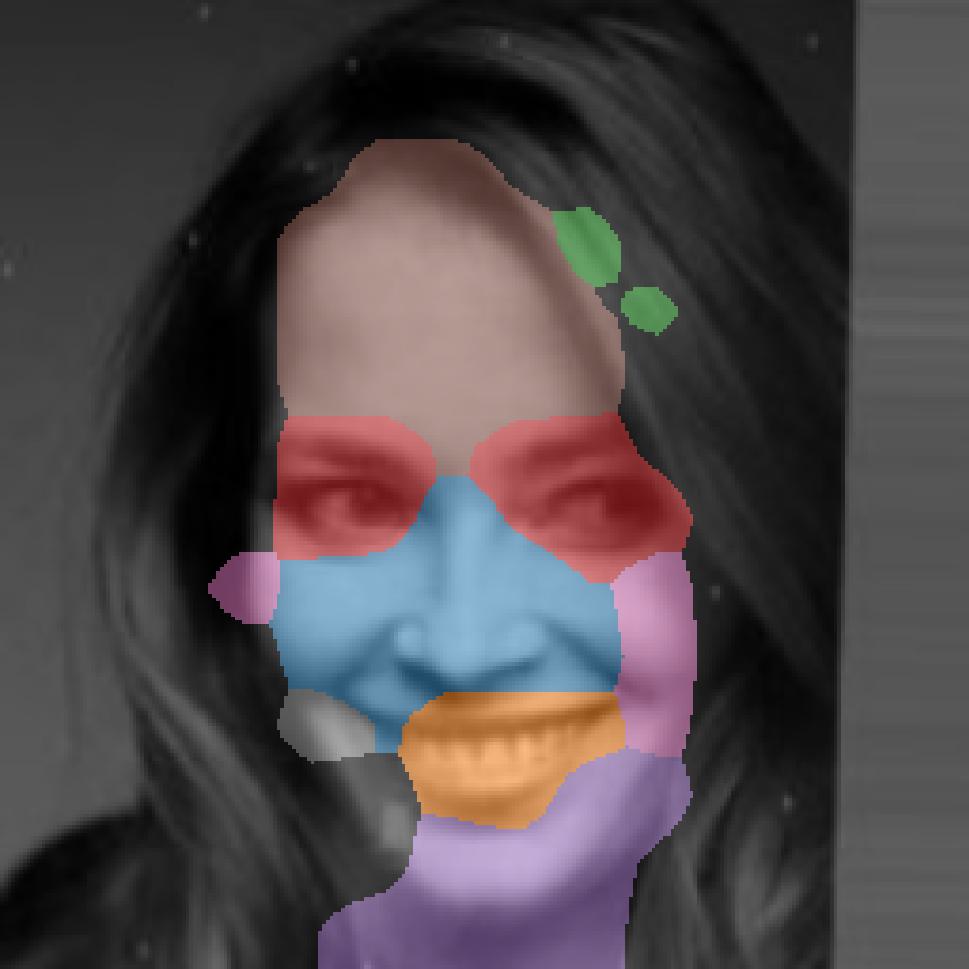}           &
 \includegraphics[width=0.118\textwidth]{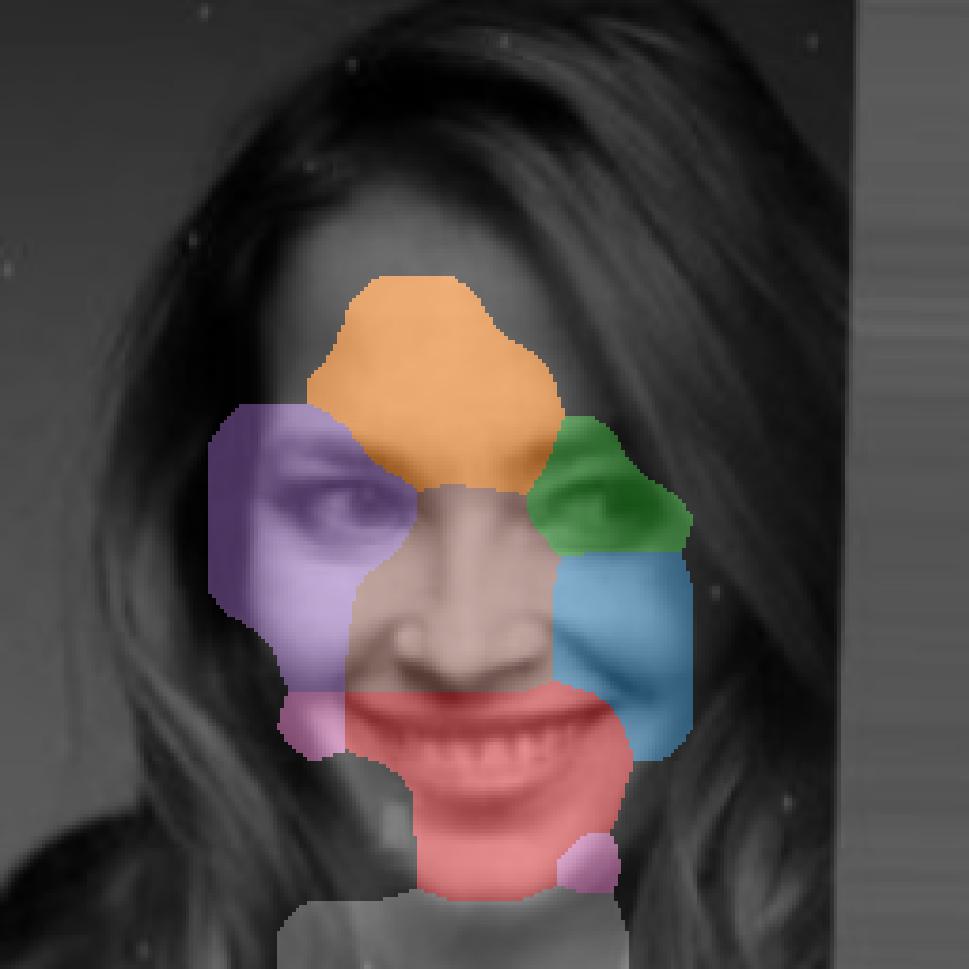}         \\     
 &
 \includegraphics[width=0.118\textwidth]{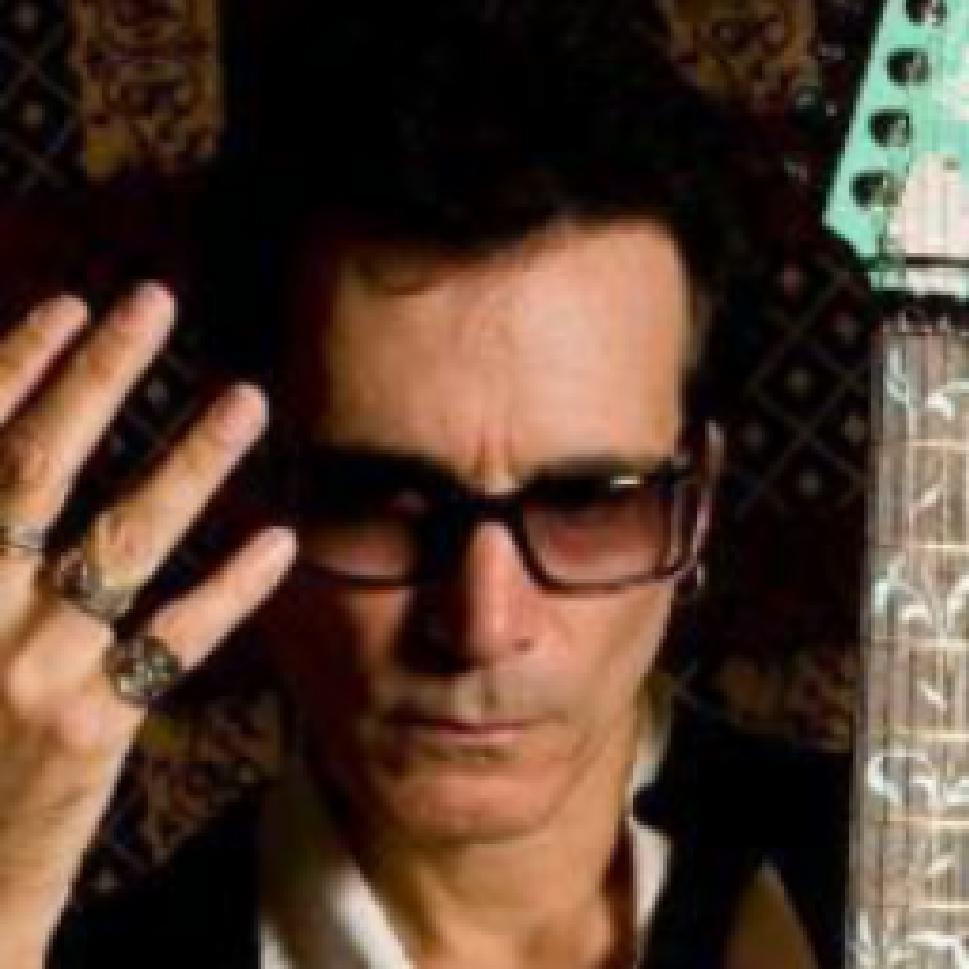} & 
 \includegraphics[width=0.118\textwidth]{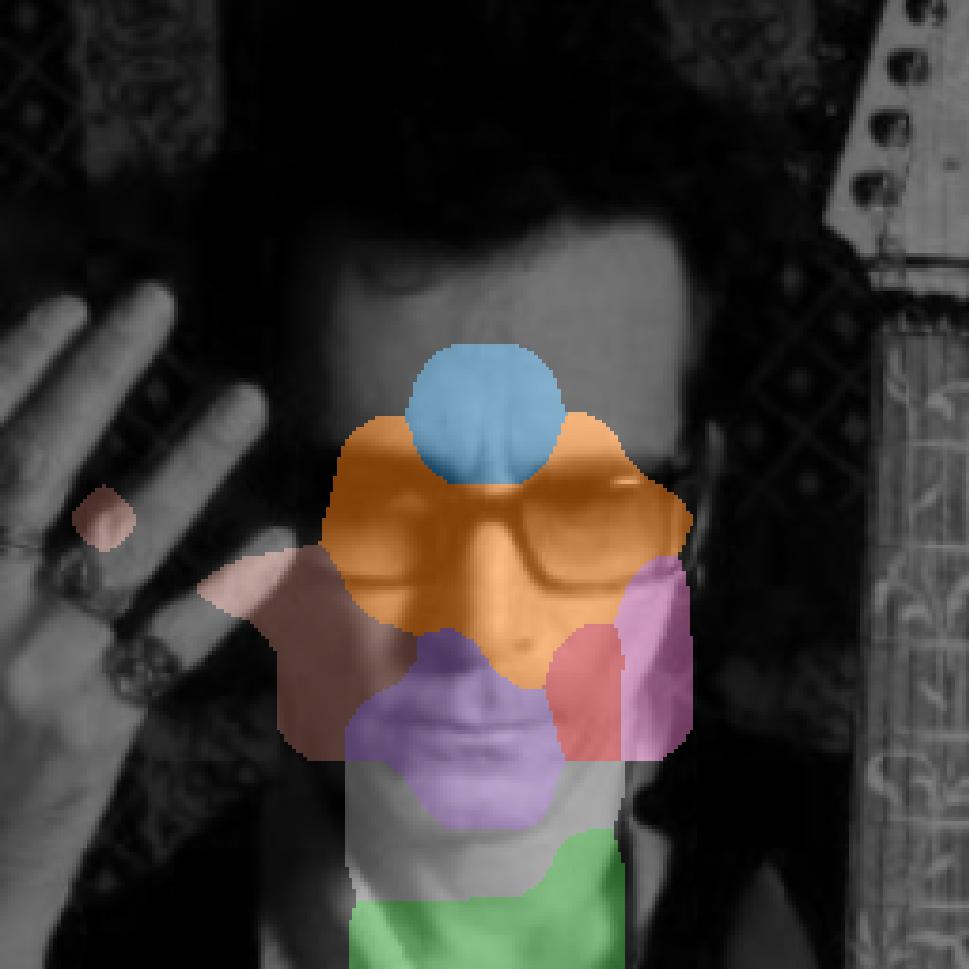} &
 \includegraphics[width=0.118\textwidth]{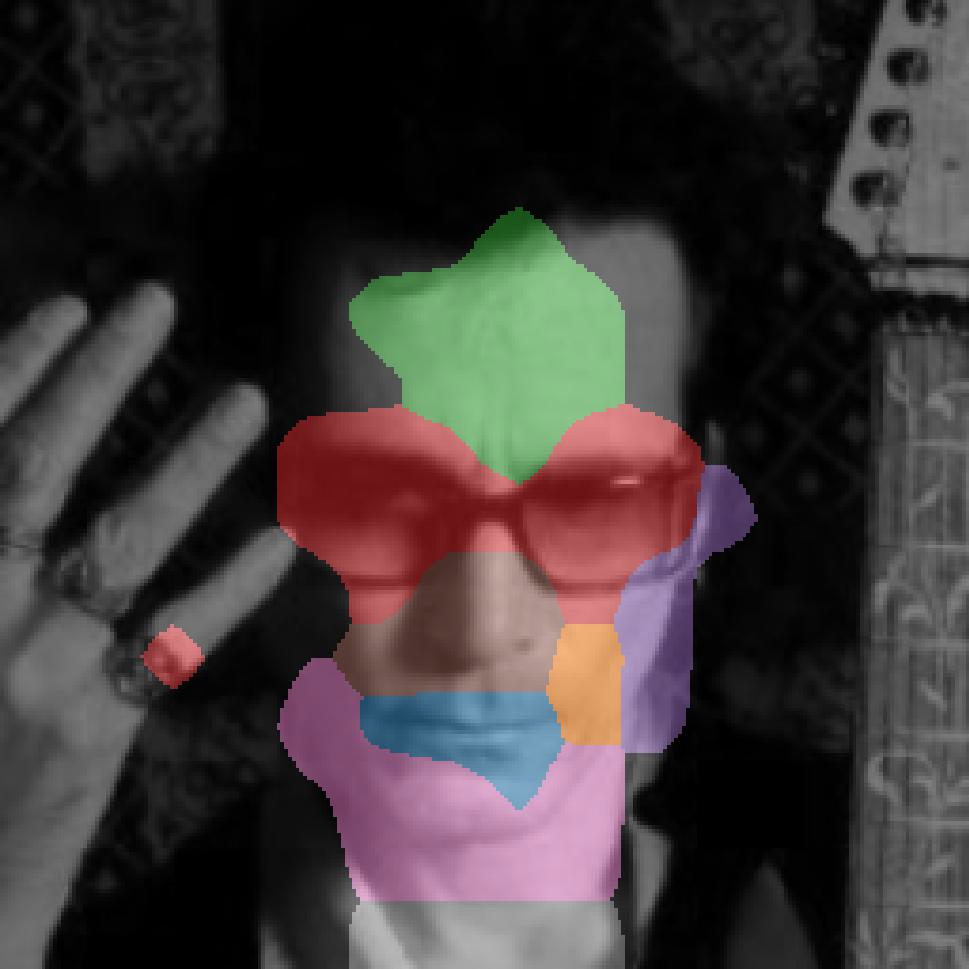}         &
 \includegraphics[width=0.118\textwidth]{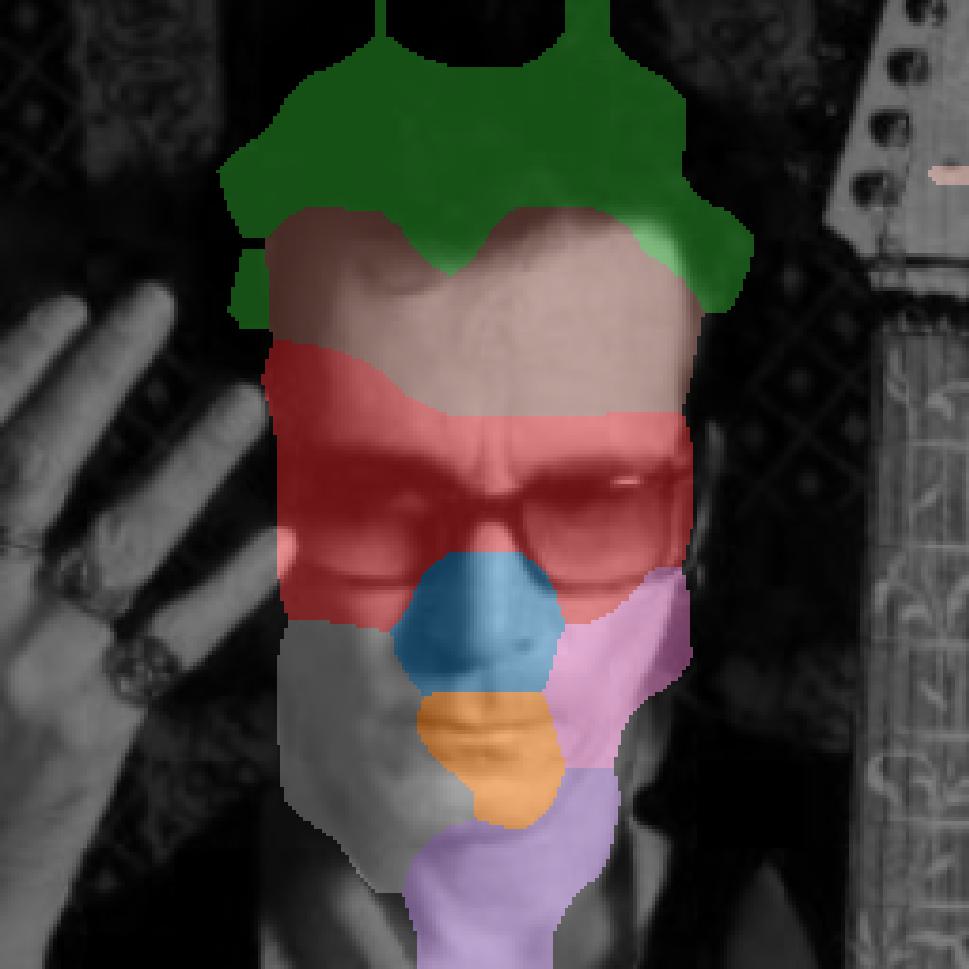}           &
 \includegraphics[width=0.118\textwidth]{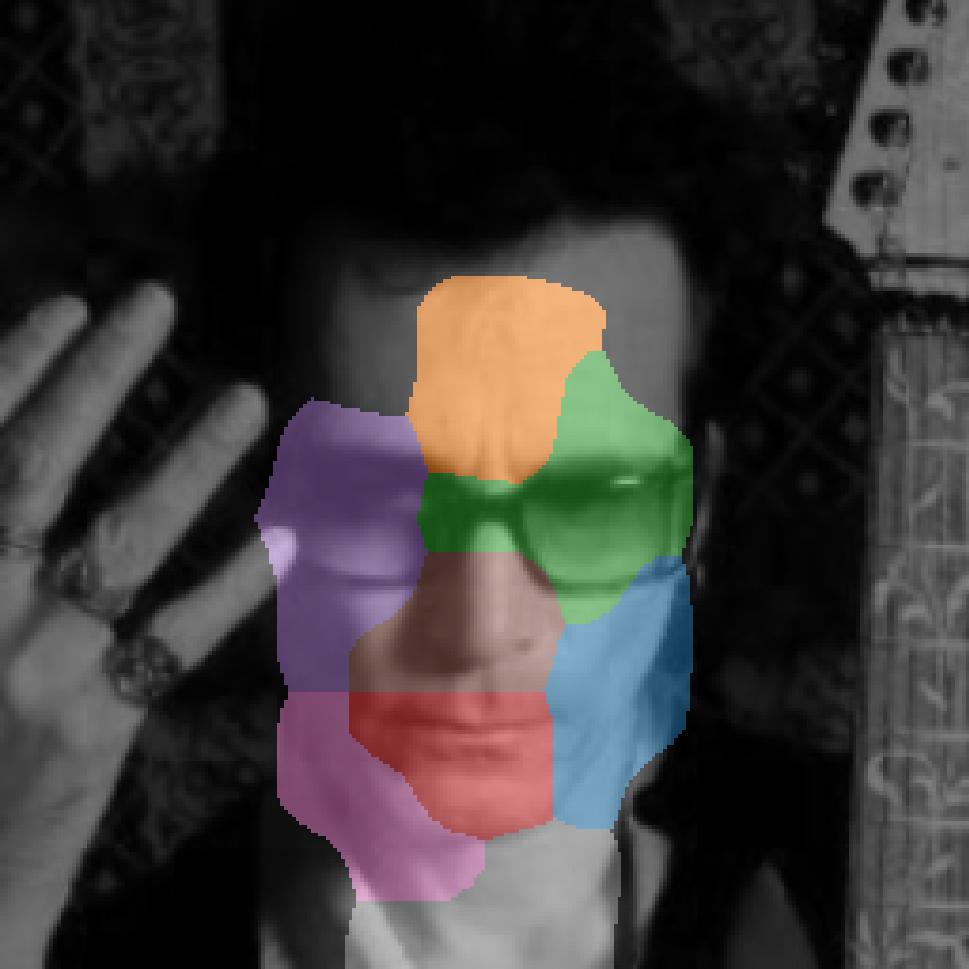}         \\     
 \multirow{2}{*}{\rotatebox[origin=tl]{90}{{\parbox{1.1cm}{\centering CheXpert}}}}  & 
 \includegraphics[width=0.118\textwidth]{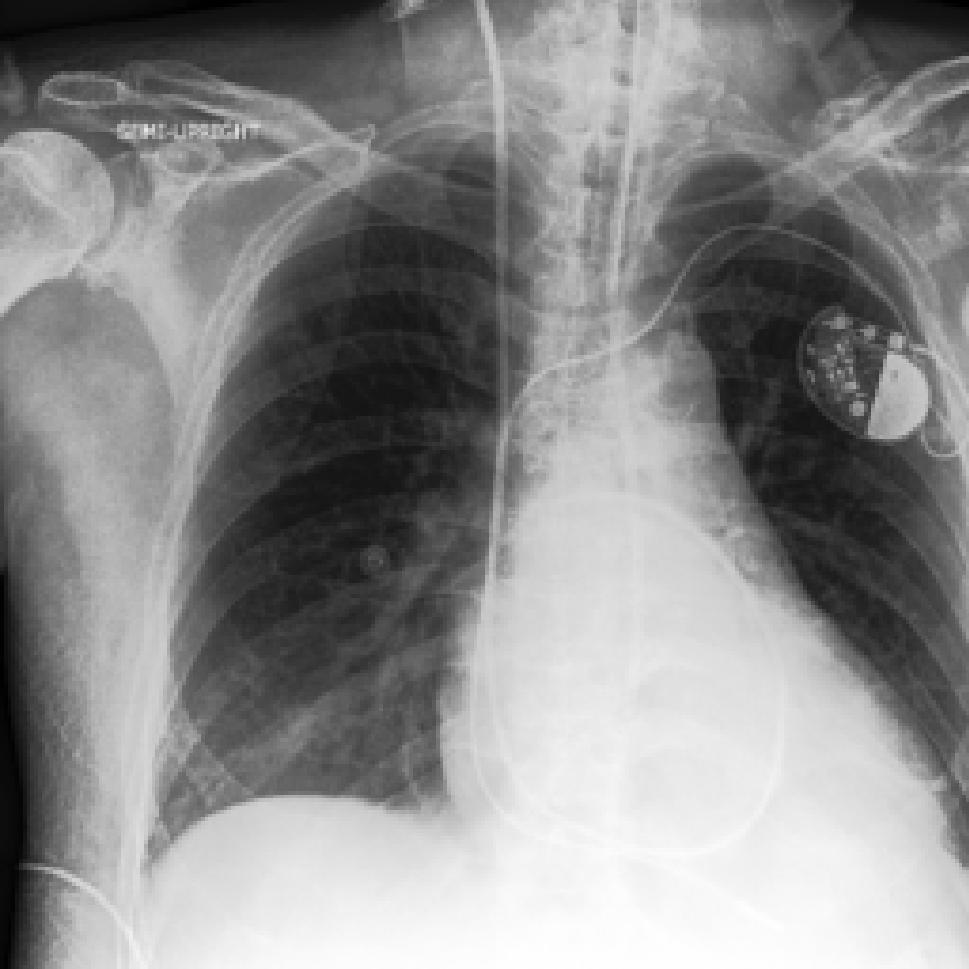} & 
 \includegraphics[width=0.118\textwidth]{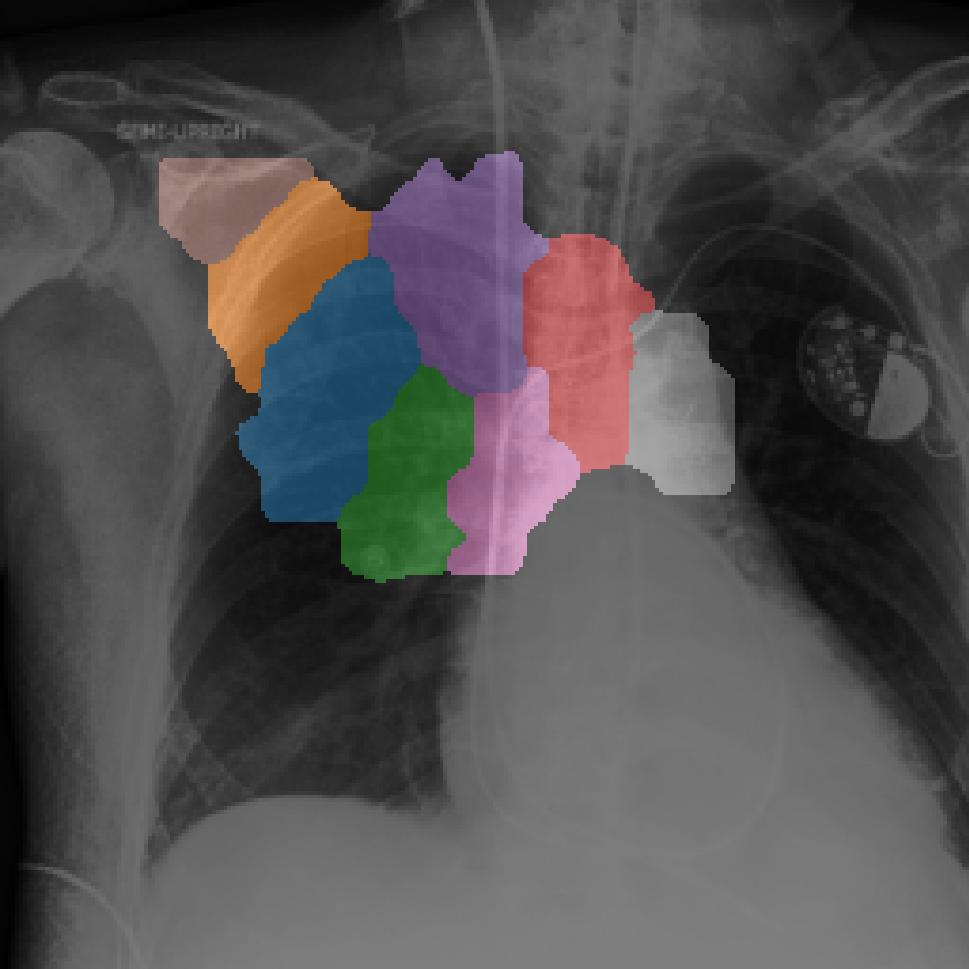} &
 \includegraphics[width=0.118\textwidth]{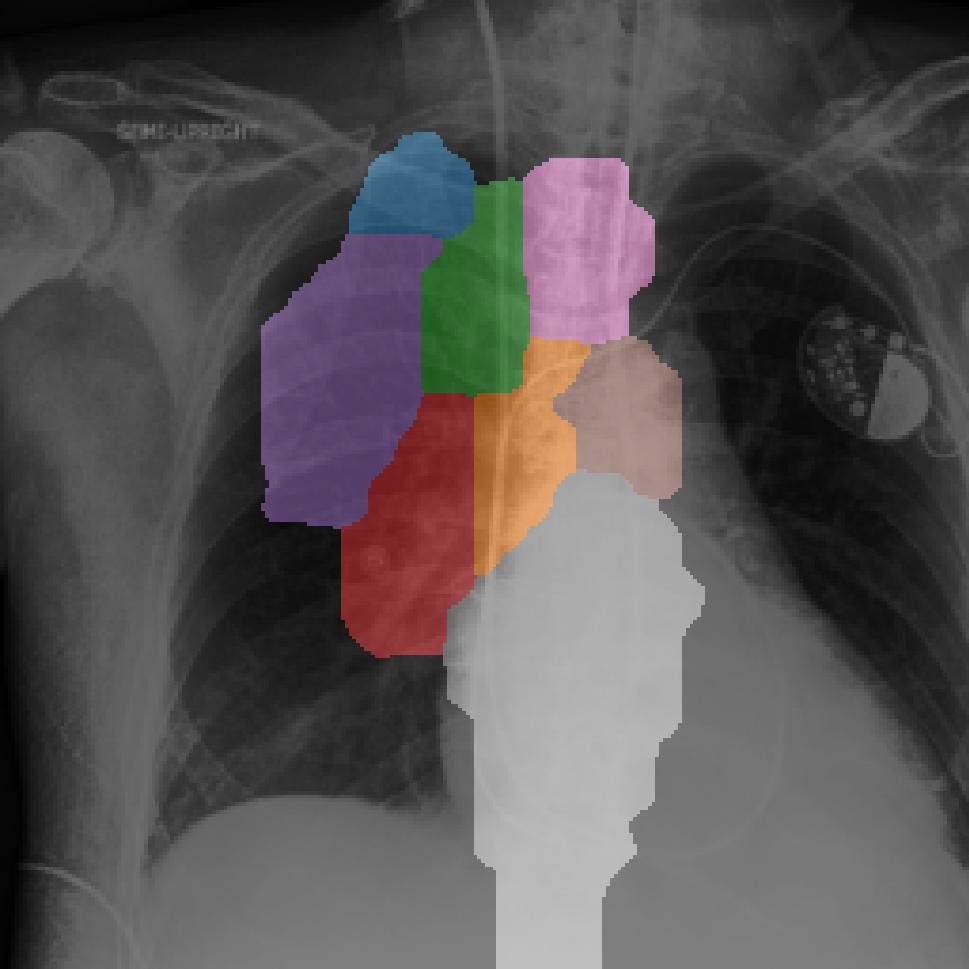}         &
 \includegraphics[width=0.118\textwidth]{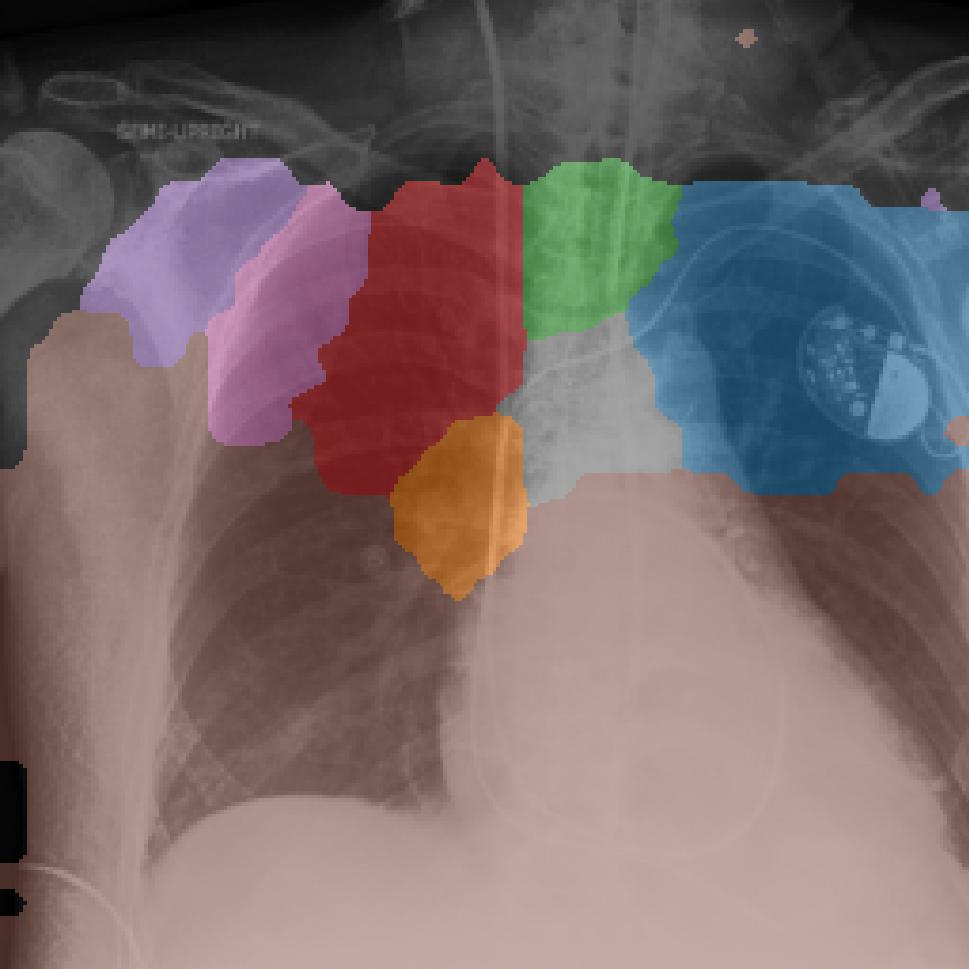}           &
 \includegraphics[width=0.118\textwidth]{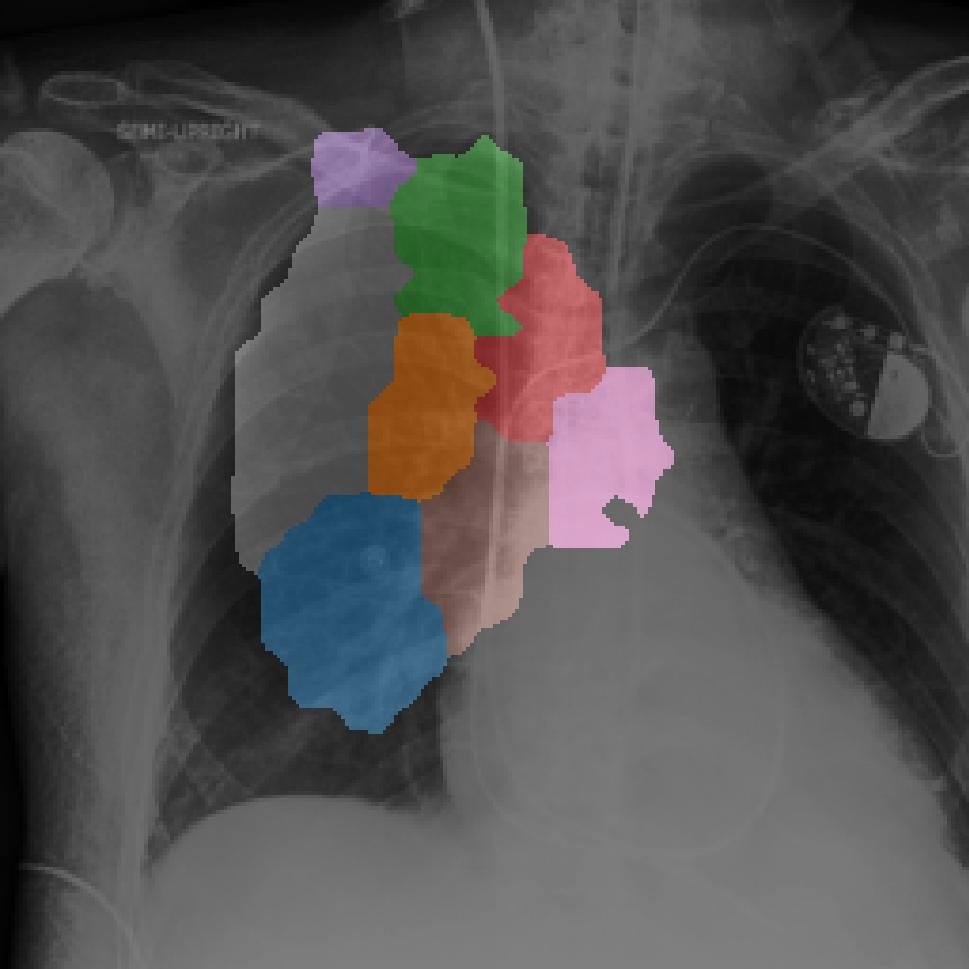}         \\ 
 & 
 \includegraphics[width=0.118\textwidth]{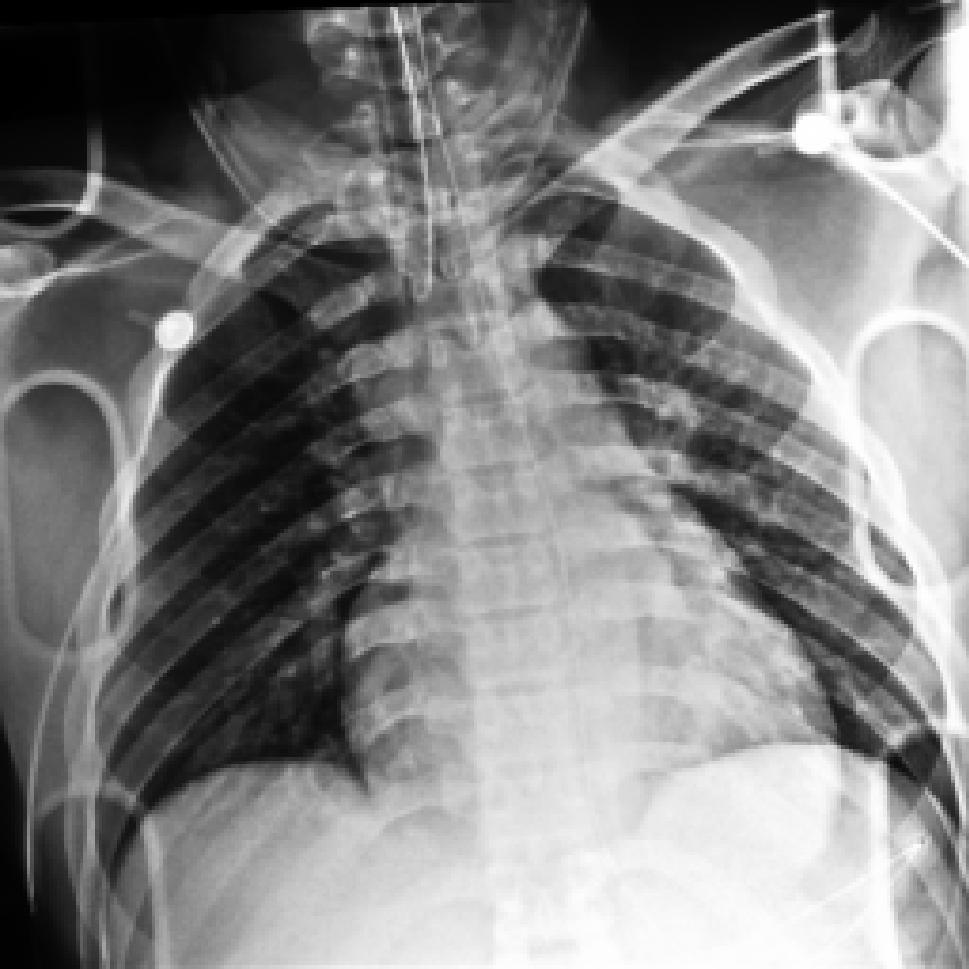} & 
 \includegraphics[width=0.118\textwidth]{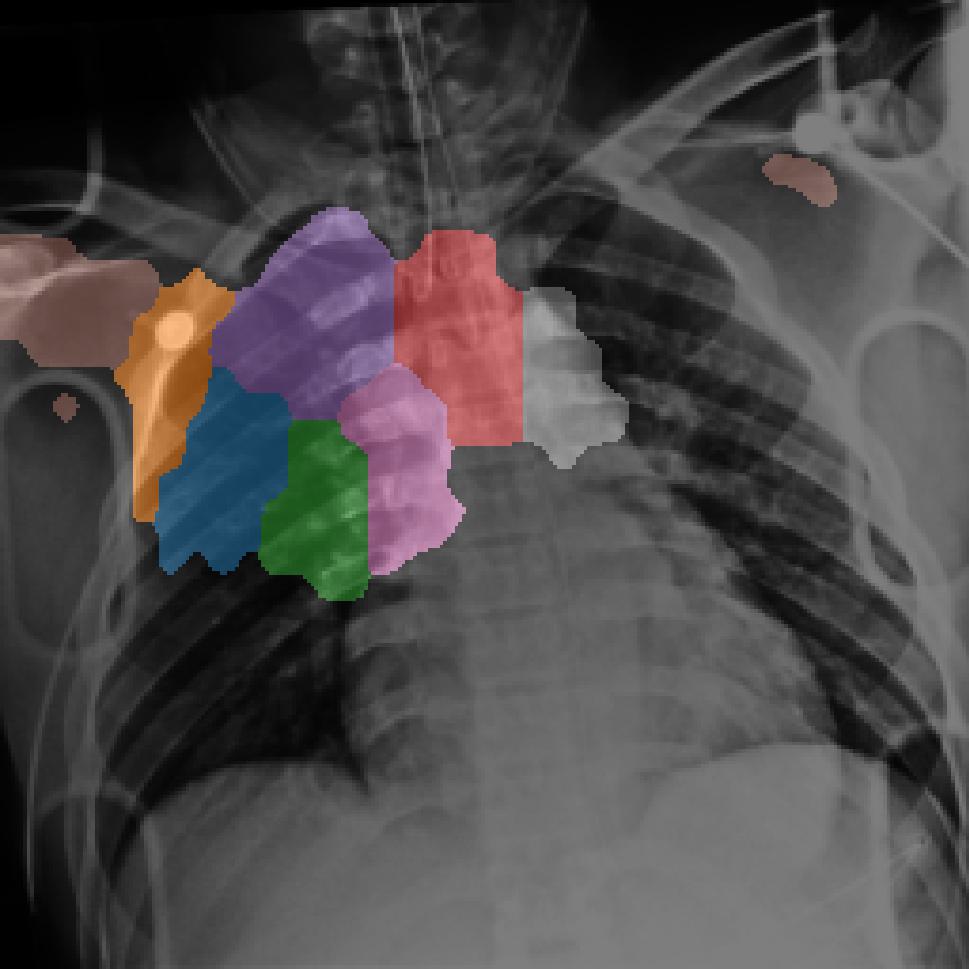} &
 \includegraphics[width=0.118\textwidth]{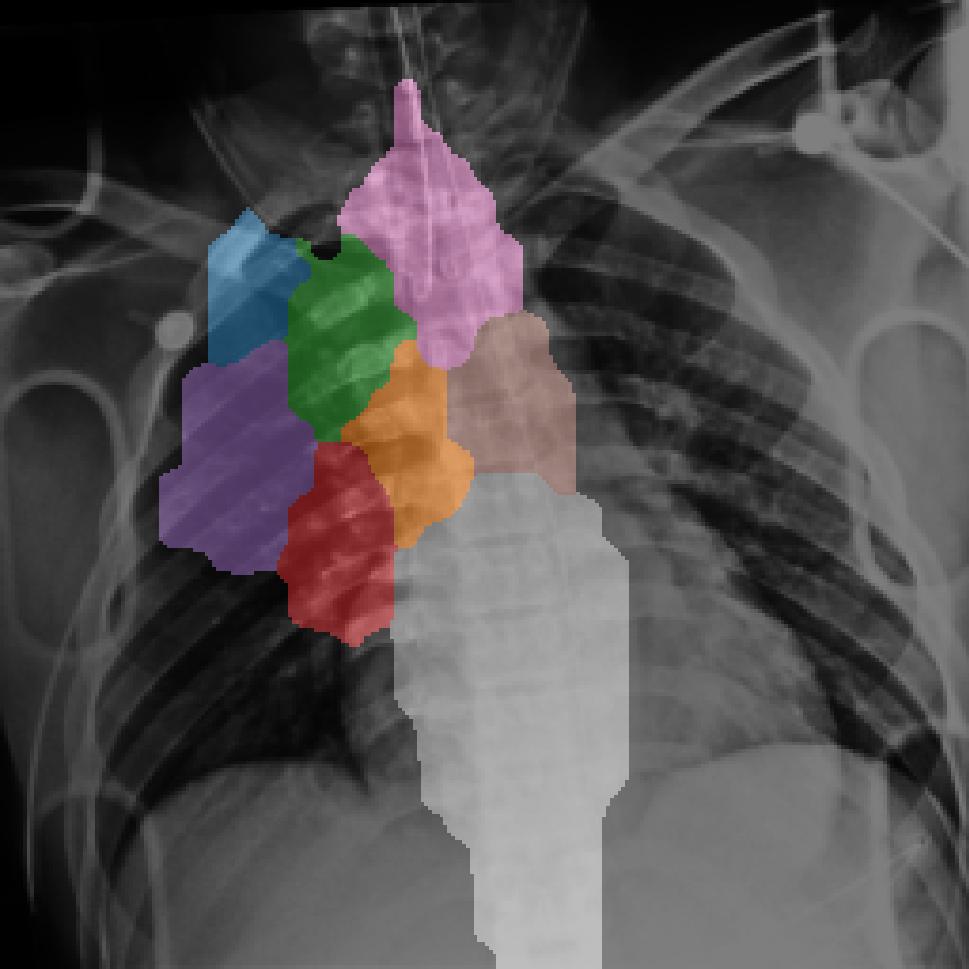}         &
 \includegraphics[width=0.118\textwidth]{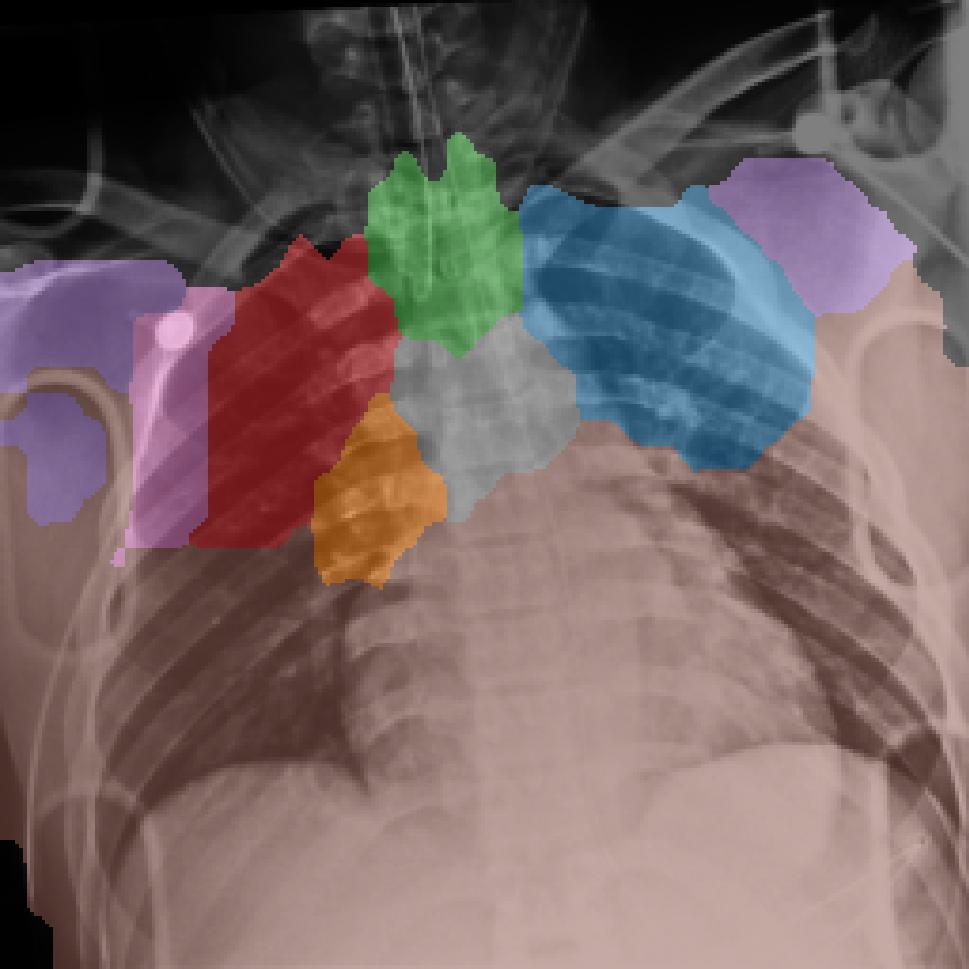}           &
 \includegraphics[width=0.118\textwidth]{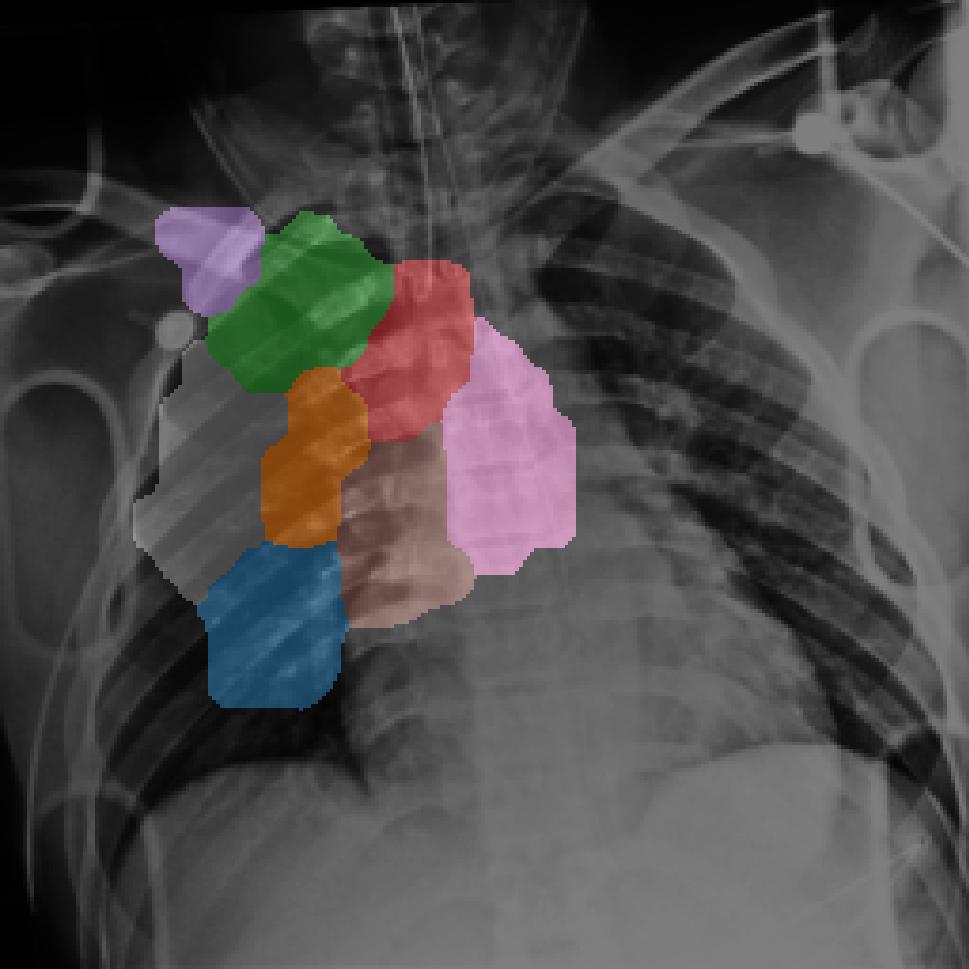}           \\
 \end{tabular}
\caption{Part discovery qualitative assessment: one-stage vs. two-stage (hard, straight-through and soft variants)}
\label{fig:qualitatives_part2}
\end{figure}

Qualitatively (Fig.~\ref{fig:qualitatives_part2}), two-stage models produce more spatially expanded masks. Because the early-masked second stage only accesses pixels within the identified part, the mask must cover the full relevant region for accurate attribute prediction. STE produces the most expanded masks because its soft backward pass signals that including additional pixels improves predictions, a signal pure hard masks cannot provide as effectively. Conversely, single-stage and soft-mask settings exploit leakage to predict attributes from partial masks, removing this pressure. Consequently, STE discovers parts that align much more closely with semantic ground-truth.

Compared to the state-of-the-art on part discovery, the STE two-stage method outperforms them in our setting, that promotes the discovery of parts aligned with the attributes. On CUB, PDiscoFormer~\cite{aniraj2024pdiscoformer} and MPAE~\cite{MPAE} report NMI around 69 but substantially lower ARI (47.76 and 44.90, respectively). In contrast, our two-stage STE variant reaches higher ARI (73.93) and NMI (70.85), indicating more consistent clustering. On CelebA, MPAE achieves strong results (ARI 91.77, NMI 86.91), but STE further improves these to 97.51 ARI and 94.49 NMI. While differences in training warrant cautious comparison, these results suggest our two-stage learning scheme substantially strengthens part discovery, closing the gap with, and often surpassing, state-of-the-art approaches.

\section{Conclusion}

This work examined a key assumption underlying many interpretable-by-design vision models: that the representation of an object part primarily encodes information from its corresponding image region. Through a benchmark based on part-level attribute annotations with dataset- and model-specific upper bounds, we showed that modern pretrained vision transformers violate this assumption, exhibiting strong intra-object leakage that undermines the faithfulness of attention-based interpretability approaches, such part-centric models and CBMs. We demonstrated that a two-stage architecture, in which attention masks are applied directly to the input image, significantly reduces this leakage and improves attribute-driven part discovery, with a straight-through masking strategy providing the best balance between strict isolation and effective gradient flow. More broadly, if the underlying representations already entangle information across spatial regions, attention-based interpretability mechanisms cannot recover faithful explanations. Addressing this requires representation-level solutions that enforce locality, which in our two-stage approach comes at roughly double the computational cost. Future work should explore reducing this overhead, for instance by training only the last layer of the second stage or by designing backbones with parallel local and global streams.

\bibliographystyle{splncs04}
\bibliography{main_2}
\newpage
\appendix

\renewcommand\thefigure{S\arabic{figure}}
\renewcommand\thetable{S\arabic{table}}

\section{Implementation details}
\label{sec_supp:imp_details}  
All models are implemented in PyTorch, using a ViT-B backbone initialized with DINOv3 weights~\cite{simeoni2025dinov3} (or RAD-DINO~\cite{perez2024rad} for CheXpert). Training was conducted using 8 NVIDIA H100 GPUs. To ensure statistical robustness, all reported quantitative results for our trained models are averaged over three independent runs using different random seeds. Code will be released upon paper acceptance.

\subsection{Training settings}  
\label{sec_supp:training_settings}  

We trained all models, both the single-stage baselines and the proposed two-stage architectures, for 30 epochs using the AdamW optimizer~\cite{loshchilov2018decoupled}. 

For the \textbf{single-stage baseline}, the pretrained ViT backbone was kept strictly frozen. We only updated the randomly initialized layers, which consist of the linear projection for part prototypes, the shared layer normalization, and the final attribute prediction layers. 

For the \textbf{two-stage architecture}, which was trained end-to-end in a single phase, the pretrained ViT backbone used in the first stage (for part discovery) was similarly kept strictly frozen. We updated the randomly initialized layers of the first stage alongside all parameters of the second stage, which includes the fully unfrozen second-stage ViT backbone and its corresponding shared layer normalization and final attribute prediction layers.

To adjust the learning rate dynamically, we employed a cosine annealing schedule~\cite{loshchilov2022sgdr}. With a defined base batch size of 64, the initial base learning rates applied to both architectures were $10^{-3}$ for the linear projection layer forming the part prototypes, and $10^{-2}$ for the shared layer norm and final linear layers. Additionally, for the two-stage models, the unfrozen ViT backbone in the second stage used a base learning rate of $2 \times 10^{-6}$.

We used a variable batch size ranging from a minimum of 16 to a maximum of 256, depending on available computational resources. To scale the learning rates appropriately across different batch sizes, we applied the square root scaling rule~\cite{krizhevsky2014one}. For example, at the maximum batch size of 256, the second-stage backbone learning rate was scaled to $4 \times 10^{-6}$. Regularization was performed using gradient norm clipping~\cite{Pascanu2013OnNetworks} with a constant maximum norm value of 2, and normalized weight decay~\cite{loshchilov2018decoupled} set to 0.05. 

The PDiscoFormer part shaping losses ($\mathcal{L}_{tv}$, $\mathcal{L}_{eq}$, $\mathcal{L}_{p}$, $\mathcal{L}_{ent}$) and their respective hyperparameters were configured identically to the original paper~\cite{aniraj2024pdiscoformer}.

\begin{figure}[t]
\centering

\begin{subfigure}[b]{0.49\textwidth}
\centering
\begin{adjustbox}{max width=\linewidth}
\begin{tabular}{@{}cccccccc@{}}
\toprule
\multicolumn{1}{l}{} & 
  \textbf{Back} & \textbf{Belly} & \textbf{Breast} & \textbf{Head} & \textbf{Leg} & \textbf{Tail} & \textbf{Wing} \\ \midrule
\textcolor{white}{\textbf{P0}} \cellcolor{cat0}  & 0.00          & 0.00          & 0.00          & \textbf{0.31}          & 0.00          & 0.00          & 0.00          \\
\rowcolor[HTML]{FFFFFF} \textcolor{white}{\textbf{P1}} 
\cellcolor{cat1}  & 0.01          & \textbf{0.81} & \textbf{0.83} & 0.00          & 0.00          & 0.00          & 0.04          \\
\rowcolor[HTML]{FFFFFF} \textcolor{white}{\textbf{P2}} 
\cellcolor{cat2} & 0.00          & 0.00          & 0.00          & 0.00          & 0.01          & \textbf{0.99} & 0.01          \\
\rowcolor[HTML]{FFFFFF} \textcolor{white}{\textbf{P3}} 
\cellcolor{cat3} & \textbf{0.37}          & 0.00          & 0.15          & 0.18          & 0.00          & 0.00          & 0.00          \\
\rowcolor[HTML]{FFFFFF} \textcolor{white}{\textbf{P4}} 
\cellcolor{cat4} & 0.00          & 0.00          & 0.00          & \textbf{0.50} & 0.00          & 0.00          & 0.00          \\
\rowcolor[HTML]{FFFFFF} \textcolor{white}{\textbf{P5}} 
\cellcolor{cat5} & 0.00          & 0.17          & 0.00          & 0.00          & \textbf{0.98} & 0.00          & 0.00          \\
\rowcolor[HTML]{FFFFFF} \textcolor{white}{\textbf{P6}} 
\cellcolor{cat6} & \textbf{0.60} & 0.01          & 0.00          & 0.00          & 0.00          & 0.00          & \textbf{0.93} \\ 
\rowcolor[HTML]{FFFFFF} 
\midrule         
\end{tabular}%

\end{adjustbox}
\caption{Spatial Contingency Matrix}
		\end{subfigure}
		\begin{subfigure}[b]{0.3\textwidth}
			\centering
			\includegraphics[width=\linewidth]{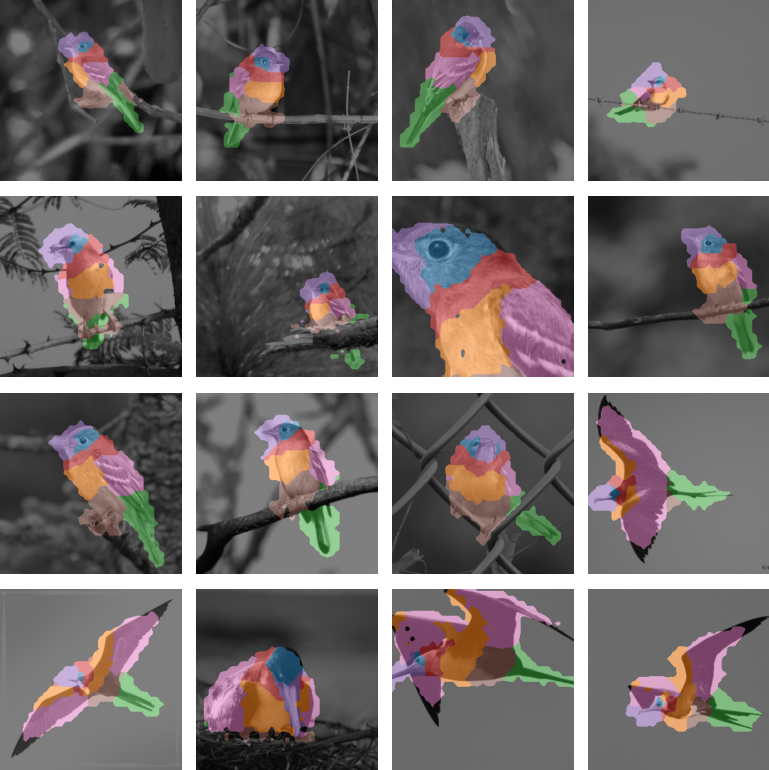}
            \caption{Qualitative Results}
		\end{subfigure}
\begin{subfigure}[b]{0.49\textwidth}
\centering
\label{tab_supp:map_dinov2_late}
\begin{adjustbox}{max width=\linewidth}
\begin{tabular}{@{}l ccccccc@{}}
\toprule
\rowcolor[HTML]{FFFFFF}\multicolumn{1}{l}{} & 
  \textbf{Back} & \textbf{Belly} & \textbf{Breast} & \textbf{Head} & \textbf{Leg} & \textbf{Tail} & \textbf{Wing} \\ \midrule
\cellcolor{cat0}{\textcolor{white}{\textbf{P0}}} & \cellcolor{red!55}\textcolor{white}{0.29} & \cellcolor{red!52}\textcolor{white}{0.30} & \cellcolor{red!58}\textcolor{white}{0.32} & \cellcolor{red!86}\textcolor{white}{\textbf{0.30}} & \cellcolor{red!25}0.13 & \cellcolor{red!44}0.20 & \cellcolor{red!44}0.30 \\
\cellcolor{cat1}{\textcolor{white}{\textbf{P1}}} & \cellcolor{red!46}0.28 & \cellcolor{red!84}\textcolor{white}{\textbf{0.35}} & \cellcolor{red!82}\textcolor{white}{\textbf{0.35}} & \cellcolor{red!57}\textcolor{white}{0.27} & \cellcolor{red!30}0.14 & \cellcolor{red!48}0.21 & \cellcolor{red!49}0.30 \\
\cellcolor{cat2}{\textcolor{white}{\textbf{P2}}} & \cellcolor{red!62}\textcolor{white}{\textbf{0.29}} & \cellcolor{red!60}\textcolor{white}{0.31} & \cellcolor{red!60}\textcolor{white}{0.32} & \cellcolor{red!44}0.26 & \cellcolor{red!32}0.14 & \cellcolor{red!85}\textcolor{white}{\textbf{0.24}} & \cellcolor{red!55}\textcolor{white}{\textbf{0.31}} \\
\cellcolor{cat3}{\textcolor{white}{\textbf{P3}}} & \cellcolor{red!60}\textcolor{white}{0.29} & \cellcolor{red!60}\textcolor{white}{0.31} & \cellcolor{red!71}\textcolor{white}{0.34} & \cellcolor{red!70}\textcolor{white}{0.28} & \cellcolor{red!19}0.13 & \cellcolor{red!46}0.21 & \cellcolor{red!50}0.30 \\
\cellcolor{cat4}{\textcolor{white}{\textbf{P4}}} & \cellcolor{red!10}0.23 & \cellcolor{red!19}0.25 & \cellcolor{red!26}0.27 & \cellcolor{red!44}0.26 & \cellcolor{red!12}0.12 & \cellcolor{red!10}0.17 & \cellcolor{red!10}0.26 \\
\cellcolor{cat5}{\textcolor{white}{\textbf{P5}}} & \cellcolor{red!27}0.25 & \cellcolor{red!53}\textcolor{white}{0.30} & \cellcolor{red!47}0.30 & \cellcolor{red!21}0.23 & \cellcolor{red!53}\textcolor{white}{\textbf{0.16}} & \cellcolor{red!25}0.18 & \cellcolor{red!23}0.28 \\
\cellcolor{cat6}{\textcolor{white}{\textbf{P6}}} & \cellcolor{red!36}0.26 & \cellcolor{red!10}0.23 & \cellcolor{red!10}0.25 & \cellcolor{red!10}0.22 & \cellcolor{red!11}0.12 & \cellcolor{red!27}0.19 & \cellcolor{red!39}0.29 \\
\bottomrule
\end{tabular}
\end{adjustbox}
\caption{Late Masking DinoV2 (Standard Probing)}
\end{subfigure}%
\hfill
\begin{subfigure}[b]{0.49\textwidth}
\centering
\label{tab_supp:map_dinov2_early}
\begin{adjustbox}{max width=\linewidth}
\begin{tabular}{@{}l ccccccc@{}}
\toprule
& \textbf{Back} & \textbf{Belly} & \textbf{Breast} & \textbf{Head} & \textbf{Leg} & \textbf{Tail} & \textbf{Wing} \\
\midrule
\cellcolor{cat0}{\textcolor{white}{\textbf{P0}}} & \cellcolor{red!36}0.26 & \cellcolor{red!42}0.28 & \cellcolor{red!54}\textcolor{white}{0.31} & \cellcolor{red!91}\textcolor{white}{0.31} & \cellcolor{red!14}0.12 & \cellcolor{red!32}0.19 & \cellcolor{red!17}0.27 \\
\cellcolor{cat1}{\textcolor{white}{\textbf{P1}}} & \cellcolor{red!45}0.27 & \cellcolor{red!100}\textcolor{white}{\textbf{0.37}} & \cellcolor{red!100}\textcolor{white}{\textbf{0.38}} & \cellcolor{red!37}0.25 & \cellcolor{red!10}0.12 & \cellcolor{red!50}0.21 & \cellcolor{red!39}0.29 \\
\cellcolor{cat2}{\textcolor{white}{\textbf{P2}}}  & \cellcolor{red!62}\textcolor{white}{0.29} & \cellcolor{red!44}0.28 & \cellcolor{red!34}0.28 & \cellcolor{red!20}0.23 & \cellcolor{red!13}0.12 & \cellcolor{red!100}\textcolor{white}{\textbf{0.26}} & \cellcolor{red!57}\textcolor{white}{0.31} \\
\cellcolor{cat3}{\textcolor{white}{\textbf{P3}}} & \cellcolor{red!70}\textcolor{white}{0.30} & \cellcolor{red!71}\textcolor{white}{0.33} & \cellcolor{red!87}\textcolor{white}{0.36} & \cellcolor{red!71}\textcolor{white}{0.29} & \cellcolor{red!12}0.12 & \cellcolor{red!50}0.21 & \cellcolor{red!45}0.30 \\
\cellcolor{cat4}{\textcolor{white}{\textbf{P4}}} & \cellcolor{red!31}0.26 & \cellcolor{red!50}0.30 & \cellcolor{red!53}\textcolor{white}{0.31} & \cellcolor{red!100}\textcolor{white}{\textbf{0.31}} & \cellcolor{red!37}0.14 & \cellcolor{red!38}0.20 & \cellcolor{red!20}0.27 \\
\cellcolor{cat5}{\textcolor{white}{\textbf{P5}}} & \cellcolor{red!23}0.25 & \cellcolor{red!81}\textcolor{white}{0.34} & \cellcolor{red!56}\textcolor{white}{0.31} & \cellcolor{red!15}0.23 & \cellcolor{red!100}\textcolor{white}{\textbf{0.21}} & \cellcolor{red!43}0.20 & \cellcolor{red!15}0.27 \\
\cellcolor{cat6}{\textcolor{white}{\textbf{P6}}}  & \cellcolor{red!100}\textcolor{white}{\textbf{0.34}} & \cellcolor{red!56}\textcolor{white}{0.30} & \cellcolor{red!49}0.30 & \cellcolor{red!47}0.26 & \cellcolor{red!24}0.13 & \cellcolor{red!83}\textcolor{white}{0.24} & \cellcolor{red!100}\textcolor{white}{\textbf{0.35}} \\
\bottomrule
\end{tabular}
\end{adjustbox}
\caption{Early Masking DinoV2 (Strict Isolation)}
\end{subfigure}

\begin{subfigure}[b]{0.49\textwidth}
\centering
\begin{adjustbox}{max width=\linewidth}
\begin{tabular}{@{}l ccccccc@{}}
\toprule
& \textbf{Back} & \textbf{Belly} & \textbf{Breast} & \textbf{Head} & \textbf{Leg} & \textbf{Tail} & \textbf{Wing} \\
\midrule
\cellcolor{cat0}{\textcolor{white}{\textbf{P0}}} & \cellcolor{red!87}\textcolor{white}{0.35} & \cellcolor{red!78}\textcolor{white}{0.38} & \cellcolor{red!83}\textcolor{white}{0.40} & \cellcolor{red!99}\textcolor{white}{0.35} & \cellcolor{red!56}\textcolor{white}{0.17} & \cellcolor{red!77}\textcolor{white}{0.26} & \cellcolor{red!85}\textcolor{white}{0.35} \\
\cellcolor{cat1}{\textcolor{white}{\textbf{P1}}} & \cellcolor{red!95}\textcolor{white}{0.36} & \cellcolor{red!100}\textcolor{white}{\textbf{0.41}} & \cellcolor{red!100}\textcolor{white}{\textbf{0.42}} & \cellcolor{red!86}\textcolor{white}{0.34} & \cellcolor{red!70}\textcolor{white}{0.18} & \cellcolor{red!90}\textcolor{white}{0.28} & \cellcolor{red!94}\textcolor{white}{0.37} \\
\cellcolor{cat2}{\textcolor{white}{\textbf{P2}}} & \cellcolor{red!93}\textcolor{white}{0.36} & \cellcolor{red!83}\textcolor{white}{0.39} & \cellcolor{red!79}\textcolor{white}{0.39} & \cellcolor{red!71}\textcolor{white}{0.32} & \cellcolor{red!74}\textcolor{white}{0.19} & \cellcolor{red!100}\textcolor{white}{\textbf{0.29}} & \cellcolor{red!92}\textcolor{white}{0.36} \\
\cellcolor{cat3}{\textcolor{white}{\textbf{P3}}} & \cellcolor{red!100}\textcolor{white}{\textbf{0.37}} & \cellcolor{red!92}\textcolor{white}{0.40} & \cellcolor{red!97}\textcolor{white}{0.42} & \cellcolor{red!100}\textcolor{white}{\textbf{0.35}} & \cellcolor{red!67}\textcolor{white}{0.18} & \cellcolor{red!90}\textcolor{white}{0.28} & \cellcolor{red!96}\textcolor{white}{0.37} \\
\cellcolor{cat4}{\textcolor{white}{\textbf{P4}}} & \cellcolor{red!86}\textcolor{white}{0.35} & \cellcolor{red!77}\textcolor{white}{0.38} & \cellcolor{red!82}\textcolor{white}{0.39} & \cellcolor{red!92}\textcolor{white}{0.35} & \cellcolor{red!61}\textcolor{white}{0.17} & \cellcolor{red!80}\textcolor{white}{0.27} & \cellcolor{red!84}\textcolor{white}{0.35} \\
\cellcolor{cat5}{\textcolor{white}{\textbf{P5}}} & \cellcolor{red!88}\textcolor{white}{0.35} & \cellcolor{red!89}\textcolor{white}{0.40} & \cellcolor{red!86}\textcolor{white}{0.40} & \cellcolor{red!71}\textcolor{white}{0.32} & \cellcolor{red!85}\textcolor{white}{\textbf{0.20}} & \cellcolor{red!89}\textcolor{white}{0.28} & \cellcolor{red!88}\textcolor{white}{0.36} \\
\cellcolor{cat6}{\textcolor{white}{\textbf{P6}}} & \cellcolor{red!97}\textcolor{white}{0.37} & \cellcolor{red!83}\textcolor{white}{0.39} & \cellcolor{red!86}\textcolor{white}{0.40} & \cellcolor{red!83}\textcolor{white}{0.33} & \cellcolor{red!72}\textcolor{white}{0.18} & \cellcolor{red!93}\textcolor{white}{0.28} & \cellcolor{red!100}\textcolor{white}{\textbf{0.37}} \\
\bottomrule
\end{tabular}
\end{adjustbox}
\caption{Late Masking DinoV3 (Standard Probing)}
\label{tab_supp:map_dinov3_late}
\end{subfigure}%
\hfill
\begin{subfigure}[b]{0.49\textwidth}
\centering

\begin{adjustbox}{max width=\linewidth}
\begin{tabular}{@{}l ccccccc@{}}
\toprule
& \textbf{Back} & \textbf{Belly} & \textbf{Breast} & \textbf{Head} & \textbf{Leg} & \textbf{Tail} & \textbf{Wing} \\
\midrule
\cellcolor{cat0}{\textcolor{white}{\textbf{P0}}} & \cellcolor{red!10}0.25 & \cellcolor{red!10}0.26 & \cellcolor{red!10}0.29 & \cellcolor{red!44}0.28 & \cellcolor{red!10}0.11 & \cellcolor{red!10}0.18 & \cellcolor{red!10}0.25 \\
\cellcolor{cat1}{\textcolor{white}{\textbf{P1}}} & \cellcolor{red!49}0.30 & \cellcolor{red!89}\textcolor{white}{\textbf{0.40}} & \cellcolor{red!85}\textcolor{white}{\textbf{0.40}} & \cellcolor{red!26}0.26 & \cellcolor{red!20}0.12 & \cellcolor{red!48}0.23 & \cellcolor{red!59}\textcolor{white}{0.32} \\
\cellcolor{cat2}{\textcolor{white}{\textbf{P2}}} & \cellcolor{red!58}\textcolor{white}{0.31} & \cellcolor{red!36}0.31 & \cellcolor{red!17}0.30 & \cellcolor{red!10}0.24 & \cellcolor{red!27}0.13 & \cellcolor{red!89}\textcolor{white}{\textbf{0.28}} & \cellcolor{red!63}\textcolor{white}{0.32} \\
\cellcolor{cat3}{\textcolor{white}{\textbf{P3}}} & \cellcolor{red!69}\textcolor{white}{0.33} & \cellcolor{red!62}\textcolor{white}{0.35} & \cellcolor{red!76}\textcolor{white}{0.38} & \cellcolor{red!58}\textcolor{white}{0.30} & \cellcolor{red!22}0.13 & \cellcolor{red!46}0.22 & \cellcolor{red!54}\textcolor{white}{0.31} \\
\cellcolor{cat4}{\textcolor{white}{\textbf{P4}}} & \cellcolor{red!19}0.26 & \cellcolor{red!32}0.30 & \cellcolor{red!37}0.33 & \cellcolor{red!73}\textcolor{white}{\textbf{0.32}} & \cellcolor{red!36}0.14 & \cellcolor{red!26}0.20 & \cellcolor{red!25}0.27 \\
\cellcolor{cat5}{\textcolor{white}{\textbf{P5}}} & \cellcolor{red!31}0.28 & \cellcolor{red!80}\textcolor{white}{0.38} & \cellcolor{red!51}\textcolor{white}{0.35} & \cellcolor{red!11}0.24 & \cellcolor{red!100}\textcolor{white}{\textbf{0.22}} & \cellcolor{red!41}0.22 & \cellcolor{red!40}0.29 \\
\cellcolor{cat6}{\textcolor{white}{\textbf{P6}}} & \cellcolor{red!81}\textcolor{white}{\textbf{0.34}} & \cellcolor{red!46}0.32 & \cellcolor{red!33}0.32 & \cellcolor{red!26}0.26 & \cellcolor{red!30}0.13 & \cellcolor{red!70}\textcolor{white}{0.25} & \cellcolor{red!95}\textcolor{white}{\textbf{0.37}} \\
\bottomrule
\end{tabular}
\end{adjustbox}
\caption{Early Masking DinoV3 (Strict Isolation)}
\label{tab_supp:map_dinov3_early}
\end{subfigure}
\caption{\textbf{Part Specificity (PS) analysis on CUB.} (a) Contingency matrix (matching between discovered and ground truth parts). (b) Qualitative part discovery. (c--f) Attribute mAP matrices (column-normalized) for DINOv2 and DINOv3. Late Masking (c, e) shows severe intra-object leakage. In contrast, our Early Masking (d, f) strictly isolates parts, concentrating predictive power on the correct parts.}
\label{fig:contingency-analysis} 
\end{figure}

\begin{figure}[t]
    \centering
    \includegraphics[width=0.9\linewidth]{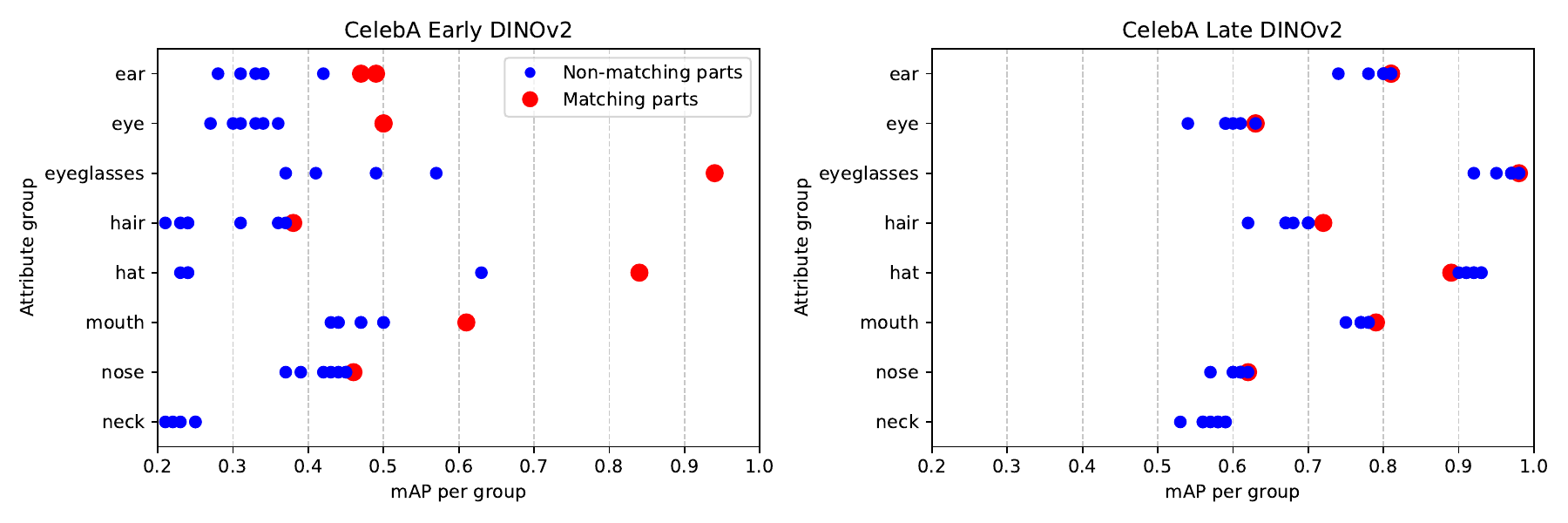}
    \includegraphics[width=0.9\linewidth]{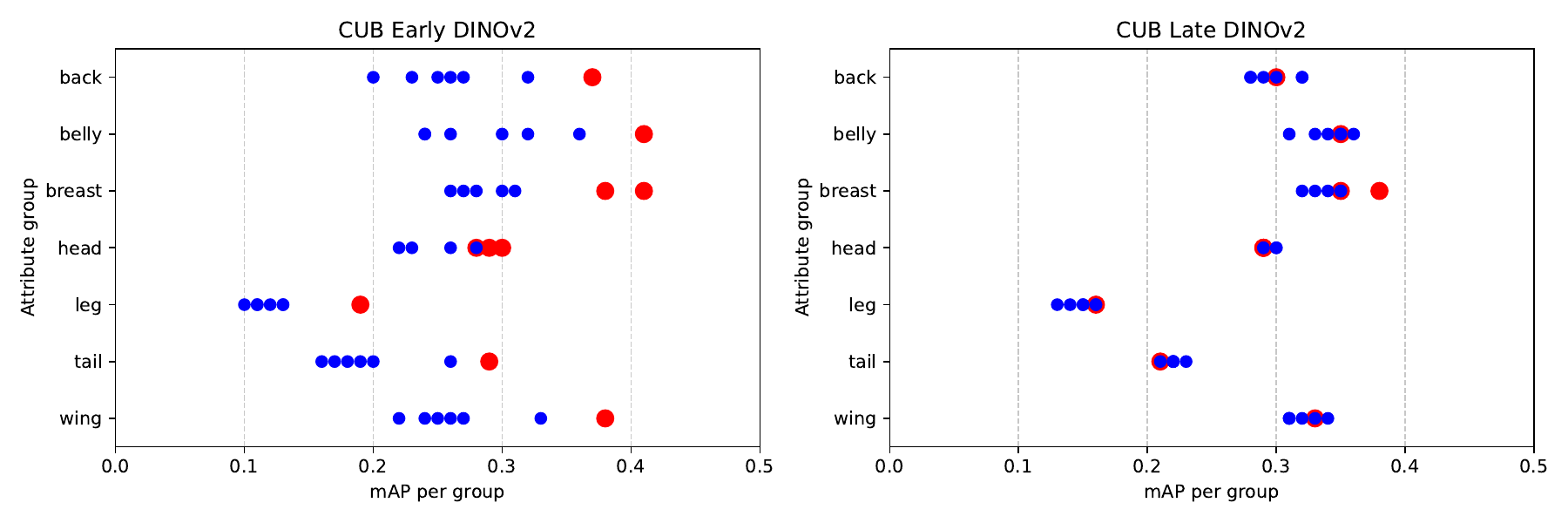}
    \caption{Plots of attribute prediction accuracy for CelebA and CUB using the representation of individual discovered parts, as described in the main paper. It is presented as mean average precision (mAP) averaged for each corresponding ground truth part. Every dot represents the accuracy using one discovered parts, with red dots corresponding to parts that are deemed good matches for the ground truth part.}
    \label{fig:matching_vs_nonmatching}
\end{figure}

\section{Detailed Calculation of the Part Specificity Metric}
\label{sec_supp:ps_metric}

In the main paper, we introduce Part Specificity (PS), a metric that measures intra-object leakage by evaluating the attribute prediction capabilities of individual object parts. To provide complete transparency into how this metric exposes metonymy, Figure~\ref{fig:contingency-analysis} illustrates the end-to-end calculation pipeline for representative models (DINOv2 and DINOv3) on the CUB dataset.

\textbf{Step 1: Spatial Alignment.} We first calculate a spatial contingency matrix (Fig.~\ref{fig:contingency-analysis}a) to mathematically map the unsupervised discovered parts (visually represented in Fig.~\ref{fig:contingency-analysis}b) to semantic regions based on spatial overlap with ground-truth keypoints (see \cref{sec_supp:attribute_groupings} for more info).

\textbf{Step 2: Attribute Probing (Exposing Leakage).} Using these spatial assignments, we evaluate whether a specific semantic region (e.g., the physical head) is actually the best predictor of its corresponding attributes (for more info see \cref{sec_supp:attribute_groupings}), or if the model is ``cheating'' by using features leaked from other parts. The contrast between the two masking strategies is stark. In the \textbf{Late Masking} matrices (Fig.~\ref{fig:contingency-analysis}c, e), predictive power is smeared across the rows and columns. For instance, attributes belonging to the ``Tail'' or ``Leg'' can be predicted with relatively high accuracy using features extracted from the ``Head'' part mask. This is direct empirical evidence of intra-object leakage.

\textbf{Step 3: Restoring Locality.} In the \textbf{Early Masking} matrices (Fig.~\ref{fig:contingency-analysis}d, f), our strict architectural isolation successfully restricts this unreasonable predictive power. The performance of non-matching parts drop significantly, and the predictive peaks (dark red) align tightly with the spatial assignments derived from the contingency matrix. The resulting margin between the matching and the non-matching parts is exactly what the Part Specificity (PS) score captures. 

Finally, Figure~\ref{fig:matching_vs_nonmatching} visualizes this dynamic as an aggregate trend across all discovered parts for both the CUB and CelebA datasets. Under early masking, spatially aligned ``correct'' parts (red dots) substantially outperform non-matching parts, whereas under late masking, all parts perform equivalently well due to the unconstrained leakage.

\section{Effect of the new orthogonality loss}

We also investigate the effect of the new orthogonality loss proposed in our paper compared to the one used in previous works on part discovery~\cite{hung2019scops, aniraj2024pdiscoformer}. The results on CUB and CelebA, in Table~\ref{tab:orth_ablation}, show that the new loss consistently results in better part discovery.

\begin{table}[t]
\centering
\caption{Part discovery results on CUB and CelebA with the orthogonality loss proposed in previous works~\cite{hung2019scops, aniraj2024pdiscoformer} compared with the new formulation used in this paper.}
\label{tab:orth_ablation}
\setlength{\tabcolsep}{5pt}
\begin{tabular}{llcc}
\toprule
\textbf{Dataset} & \textbf{Setting} & \textbf{NMI} & \textbf{ARI} \\
\midrule
\multirow{2}{*}{CUB}
  & $\mathcal{L}_{\perp}$~\cite{hung2019scops, aniraj2024pdiscoformer} & 65.22 & 62.34 \\
  & New $\mathcal{L}_{\perp}$    & \textbf{70.85} & \textbf{73.93} \\
\midrule
\multirow{2}{*}{CelebA}
  & $\mathcal{L}_{\perp}$~\cite{hung2019scops, aniraj2024pdiscoformer} & 80.88 & 76.85 \\
  & New $\mathcal{L}_{\perp}$    & \textbf{94.49} & \textbf{97.51} \\
\bottomrule
\end{tabular}
\end{table}

\section{The Role of Downstream Tasks in Controlling Semantic Part Discovery}
\label{sec:task_driven_bias}

In Table 3 of the main paper, we provide comparisons with state-of-the-art part discovery methods, namely PDiscoFormer~\cite{aniraj2024pdiscoformer} and MPAE~\cite{MPAE}. The primary objective of this comparison is not merely to establish a new state-of-the-art on a static benchmark, but to demonstrate how the choice of downstream task inherently controls the semantic alignment of the discovered parts. 

\noindent \textbf{Unconstrained Task Objectives.} Existing methods optimize for downstream tasks such as global image classification (PDiscoFormer) or masked image reconstruction (MPAE). Consequently, these models partition the image into regions that best serve those specific mathematical objectives. A model optimizing for reconstruction may discover parts based on high-frequency textures or lighting, while a model optimizing for classification may isolate small, arbitrary discriminative patches. In these paradigms, controlling the semantic meaning of the discovered parts is not inherently possible.

\noindent \textbf{Inducing semantic alignment via attributes.} By contrast, our framework utilizes fine-grained attribute prediction as the driving downstream task. Because the attributes we predict are inherently tied to specific semantic macro-regions (e.g., grouping beak, eye, and crown attributes under a unified ``Head'' concept), the model is forced to discover parts that best satisfy this specific task. 

\noindent \textbf{Discussion.} In order to minimize the attribute prediction loss, the part discovery module is induced to discover parts in a way that closely mirrors the human-defined ground-truth attribute annotations. The substantial improvements in clustering metrics (ARI and NMI) observed in Table 3 reflect this phenomenon. The attribute prediction task serves as a semantic anchor, providing a level of control over the part discovery process that enables the model to find parts significantly closer to ground-truth groupings than tasks lacking this semantic constraint. This is why we merge the ground truth keypoints to match the groupings in the attribute annotations. For more info, please refer to \cref{sec_supp:attribute_groupings}.

\section{Additional qualitative part discovery results}

\begin{figure}[t]
  \centering
  \setlength{\tabcolsep}{0pt}
  \renewcommand{\arraystretch}{0.5}
  \begin{tabular}{@{}c@{\hspace{2pt}}c@{\hspace{1pt}}c@{\hspace{1pt}}c@{\hspace{1pt}}c@{\hspace{1pt}}c@{\hspace{1pt}}c@{\hspace{1pt}}c@{\hspace{1pt}}c@{\hspace{1pt}}c@{\hspace{1pt}}c@{\hspace{1pt}}@{}}
  \toprule
    \rotatebox{90}{\makebox[1cm][c]{\scriptsize Images}} & \includegraphics[width=0.088\linewidth]{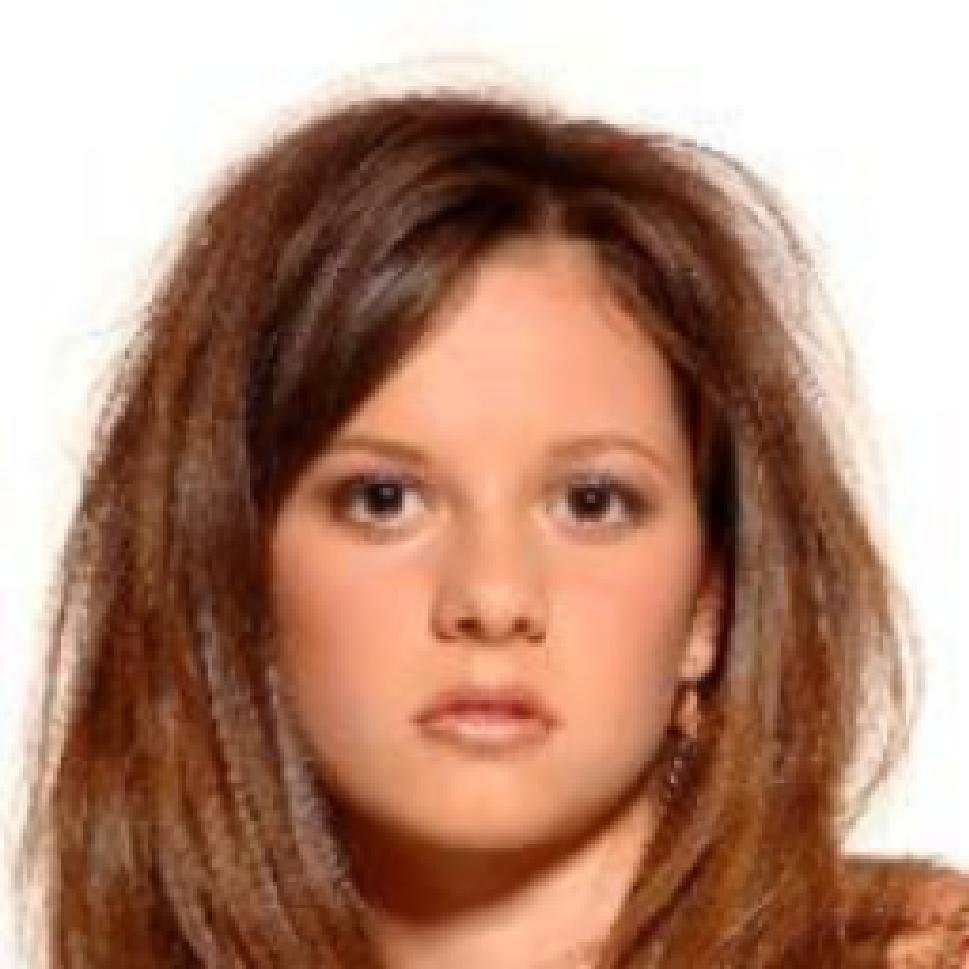} & \includegraphics[width=0.088\linewidth]{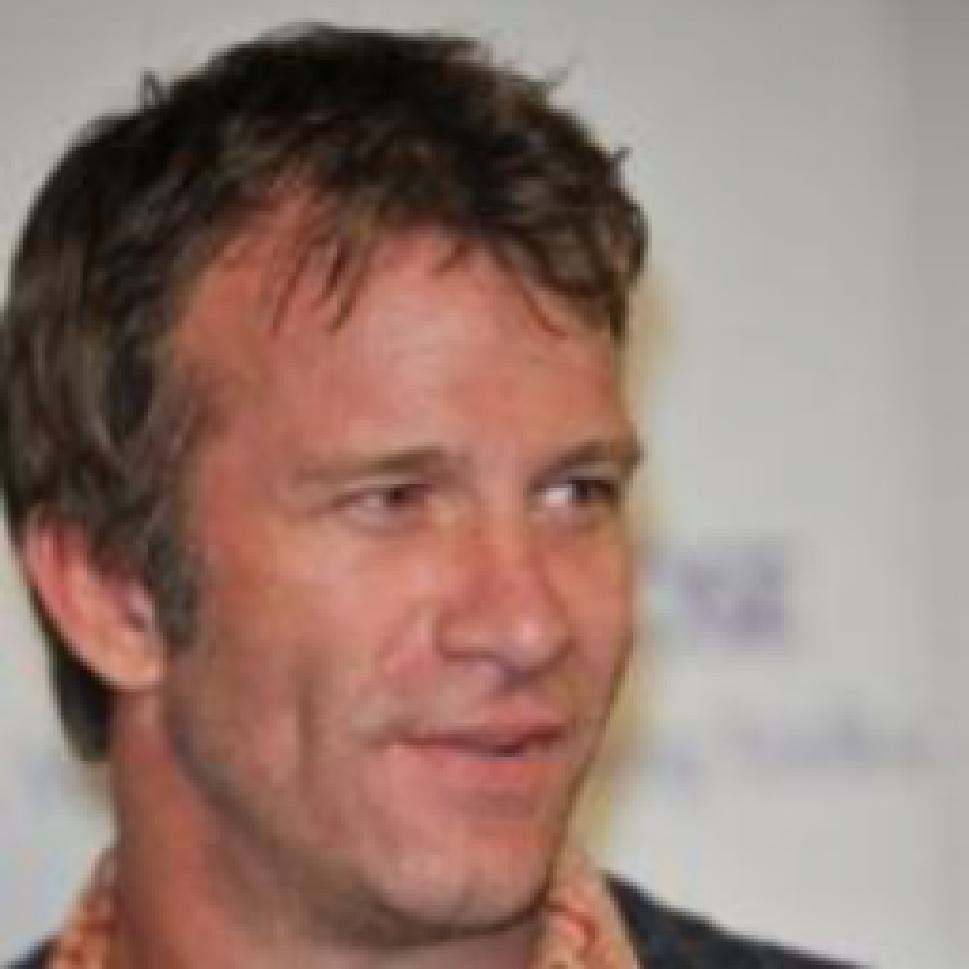} & \includegraphics[width=0.088\linewidth]{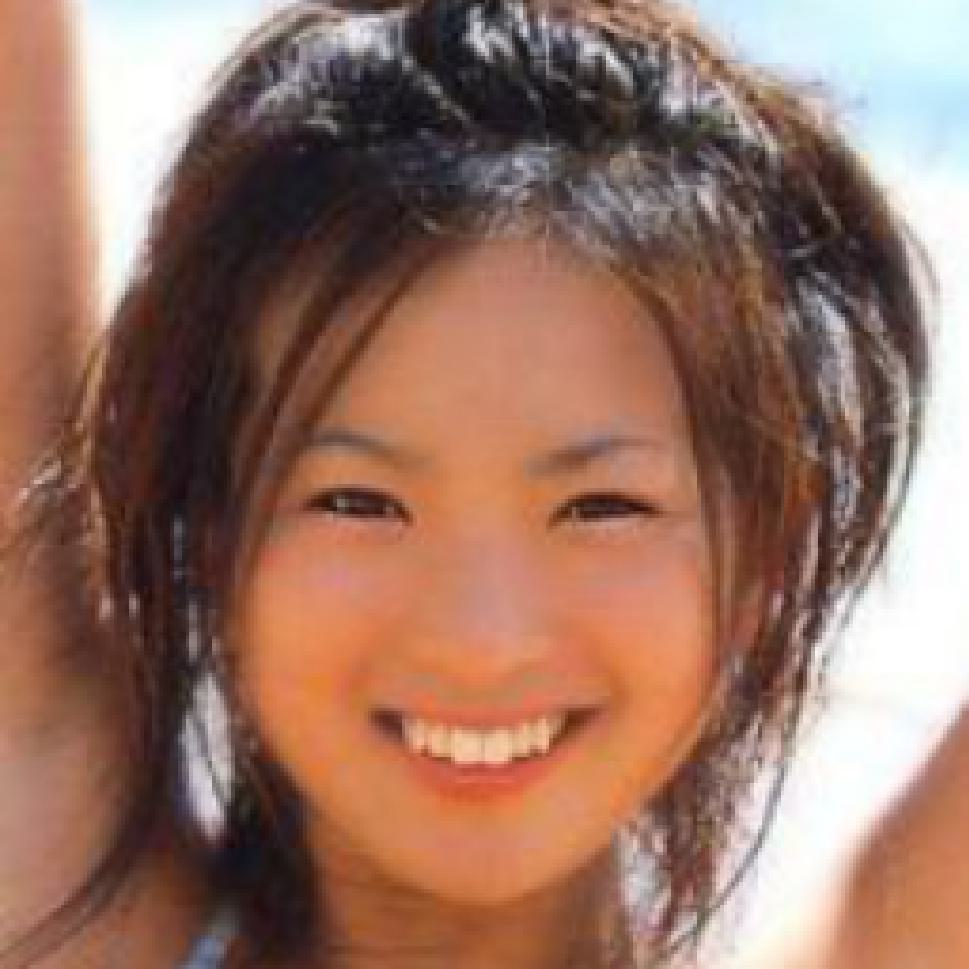} & \includegraphics[width=0.088\linewidth]{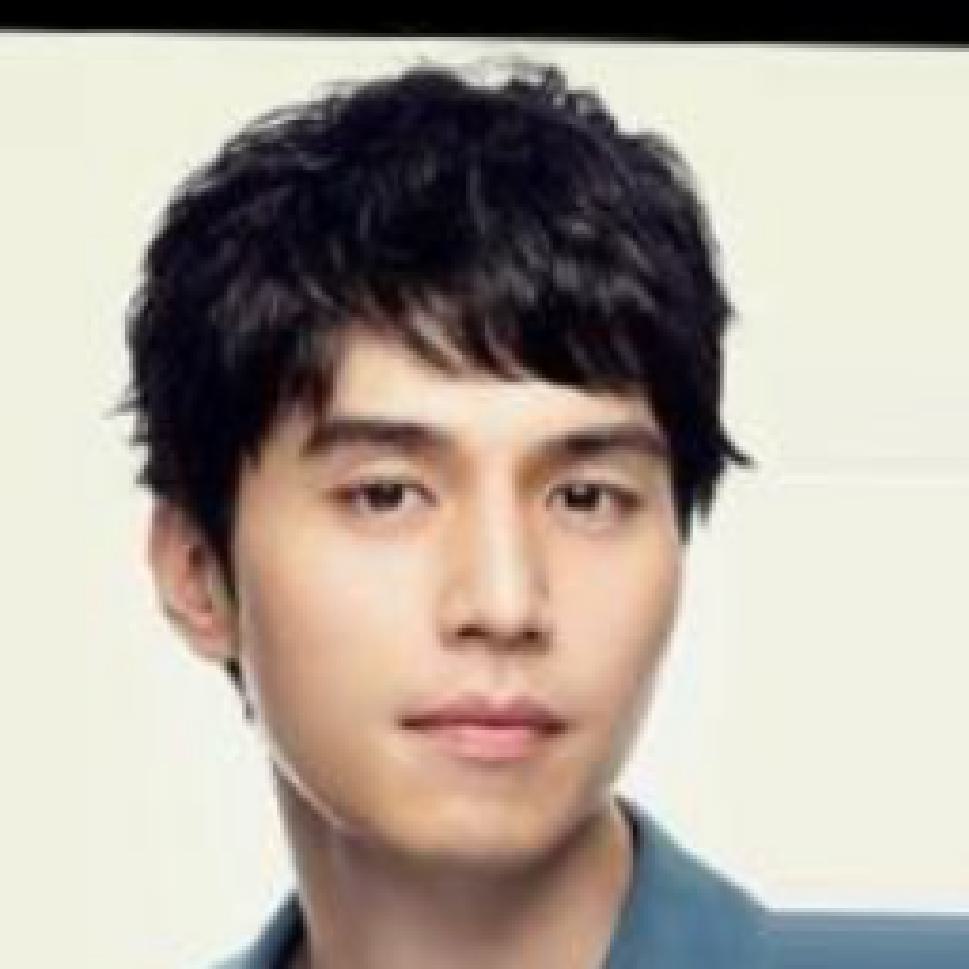} & \includegraphics[width=0.088\linewidth]{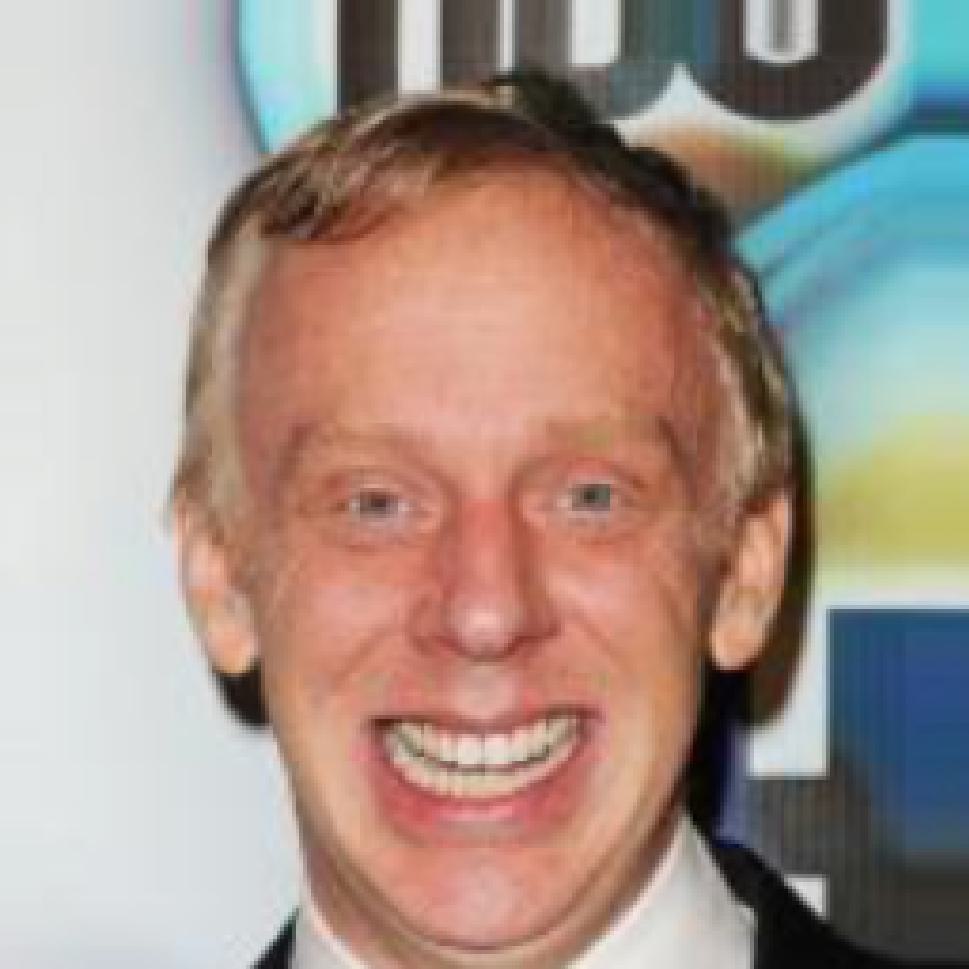} & \includegraphics[width=0.088\linewidth]{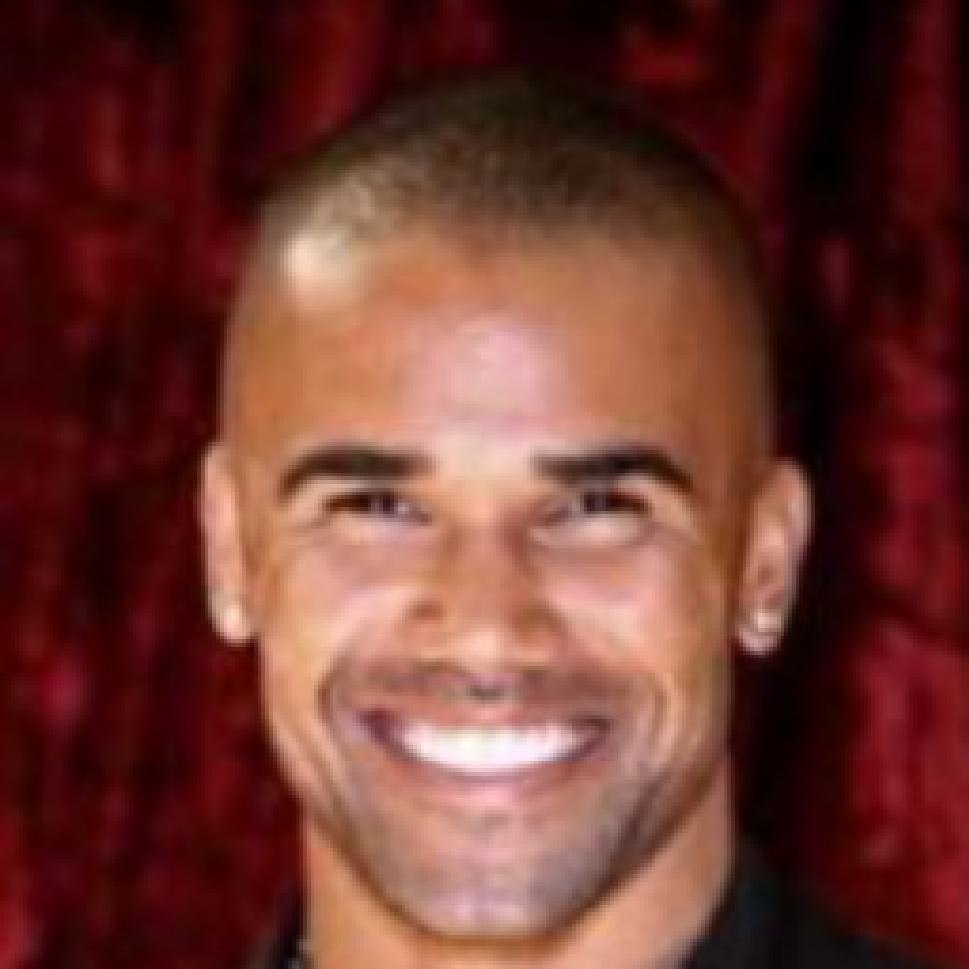} & \includegraphics[width=0.088\linewidth]{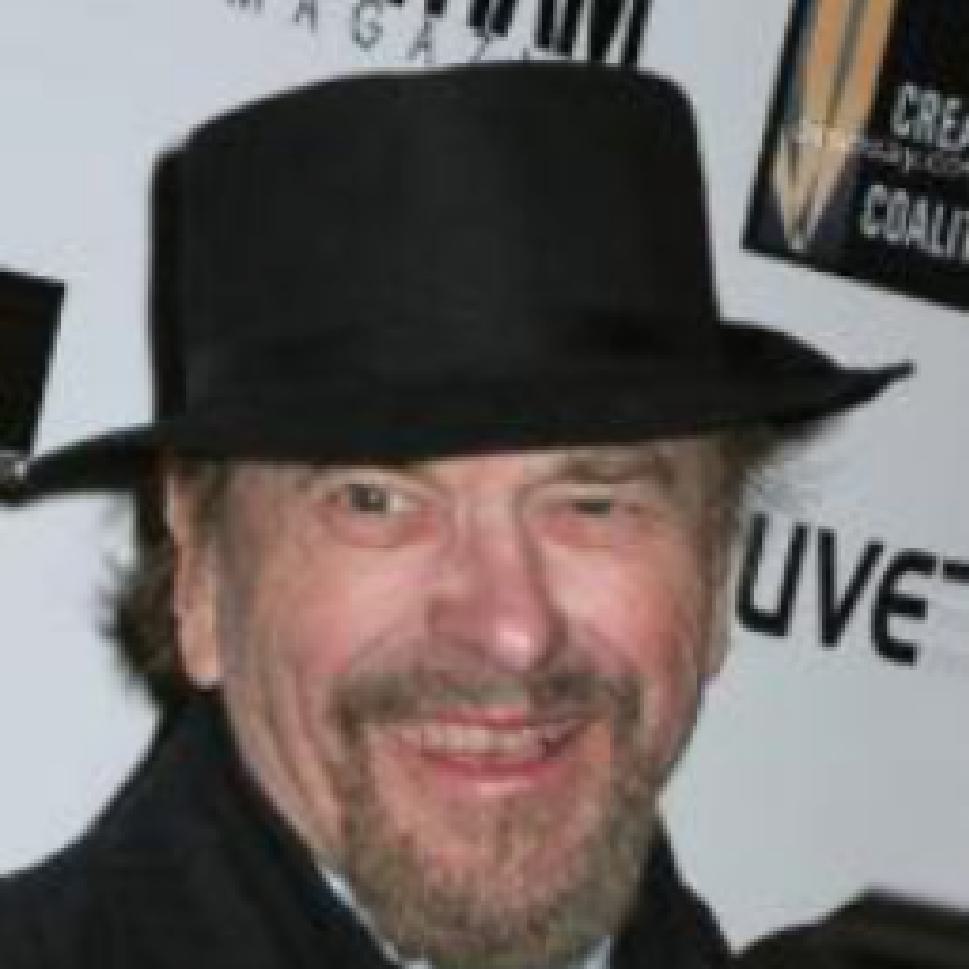} & \includegraphics[width=0.088\linewidth]{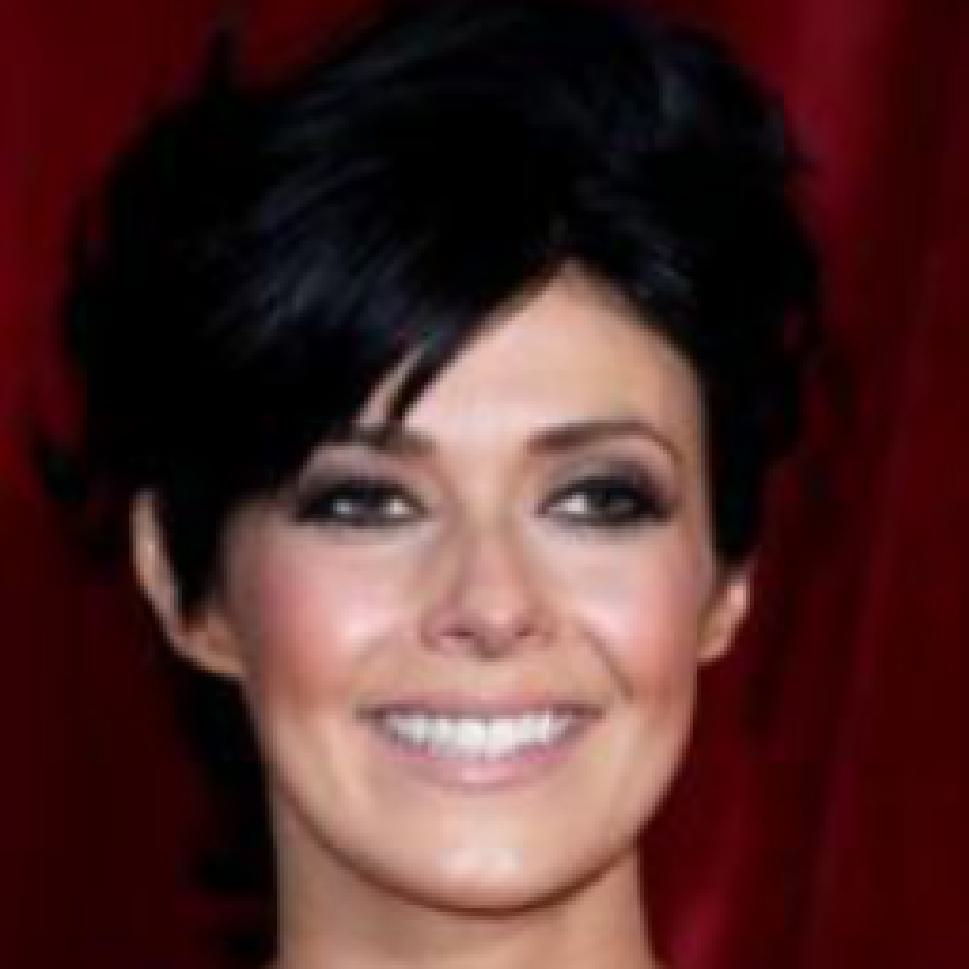} & \includegraphics[width=0.088\linewidth]{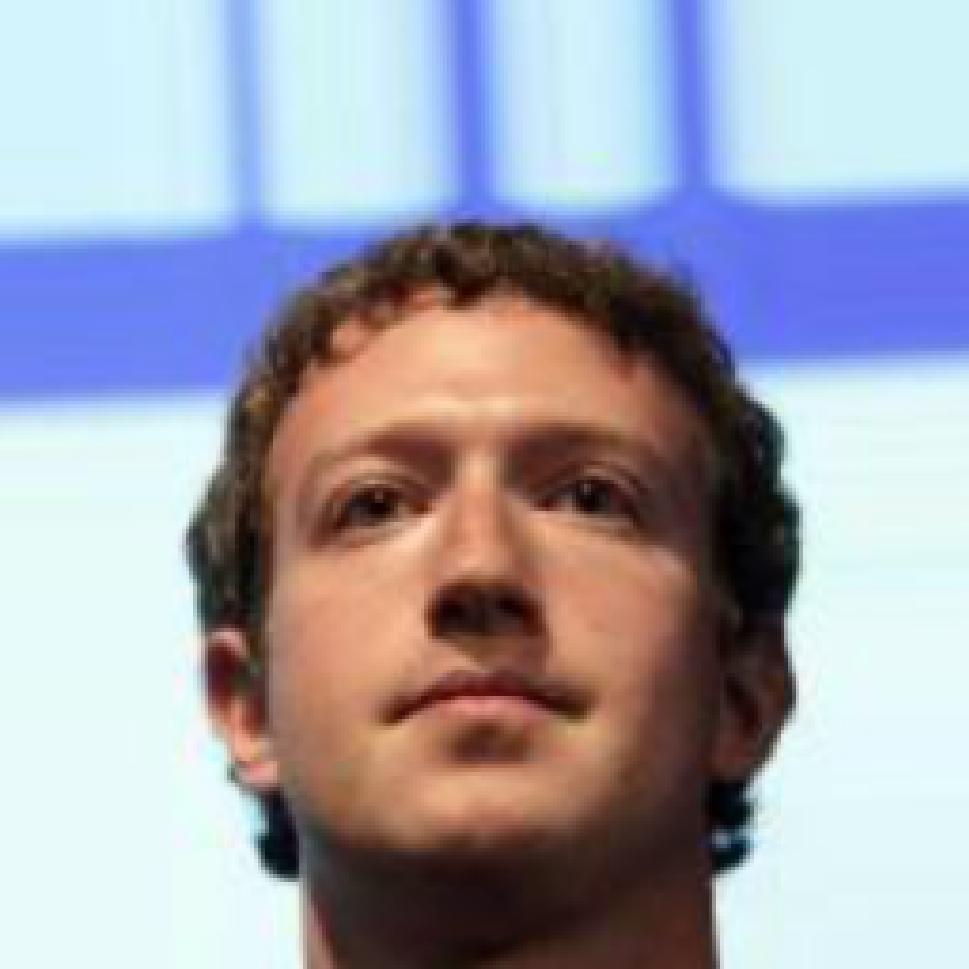} & \includegraphics[width=0.088\linewidth]{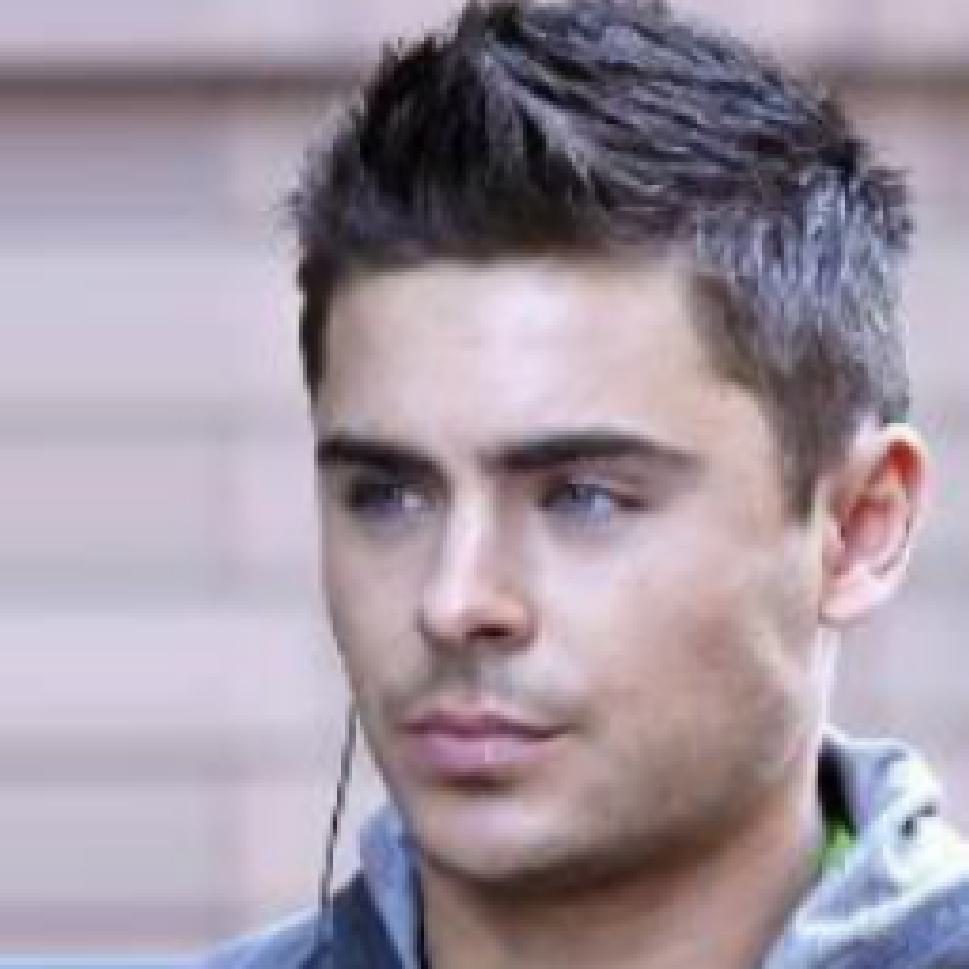} \\
    \rotatebox{90}{\makebox[1cm][c]{\scriptsize 1-stage}} & \includegraphics[width=0.088\linewidth]{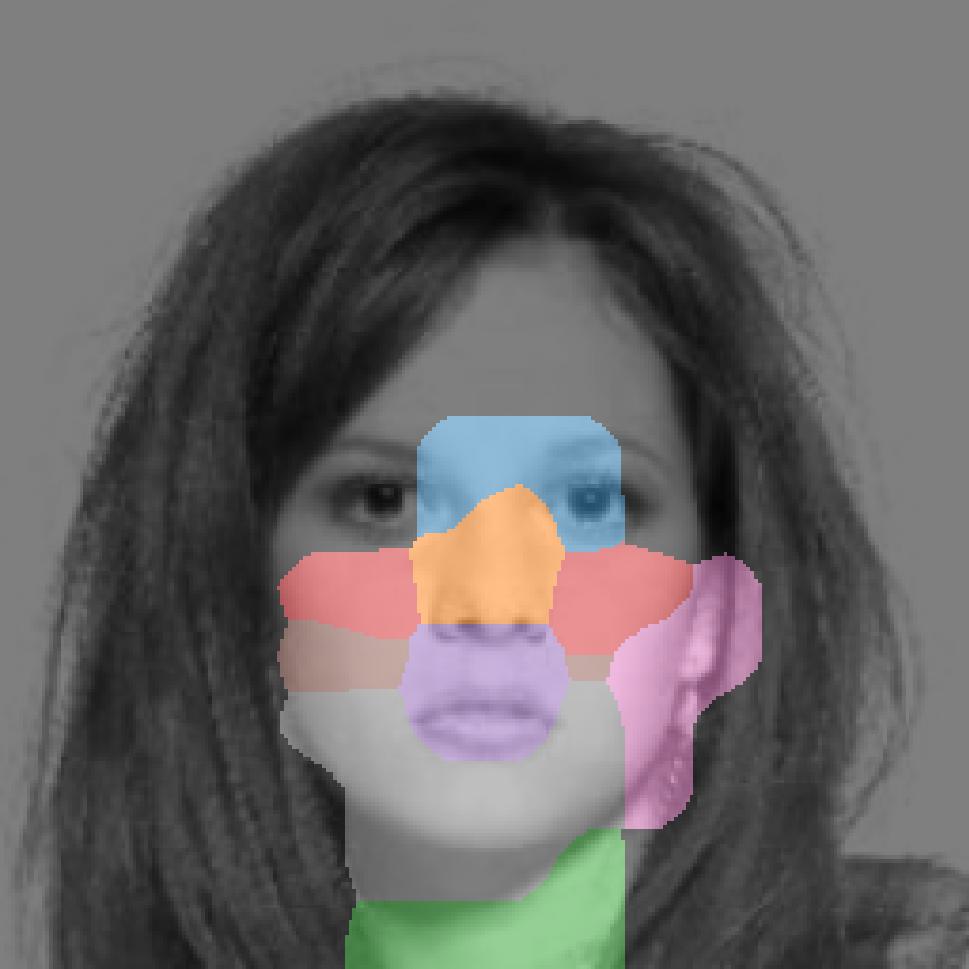} & \includegraphics[width=0.088\linewidth]{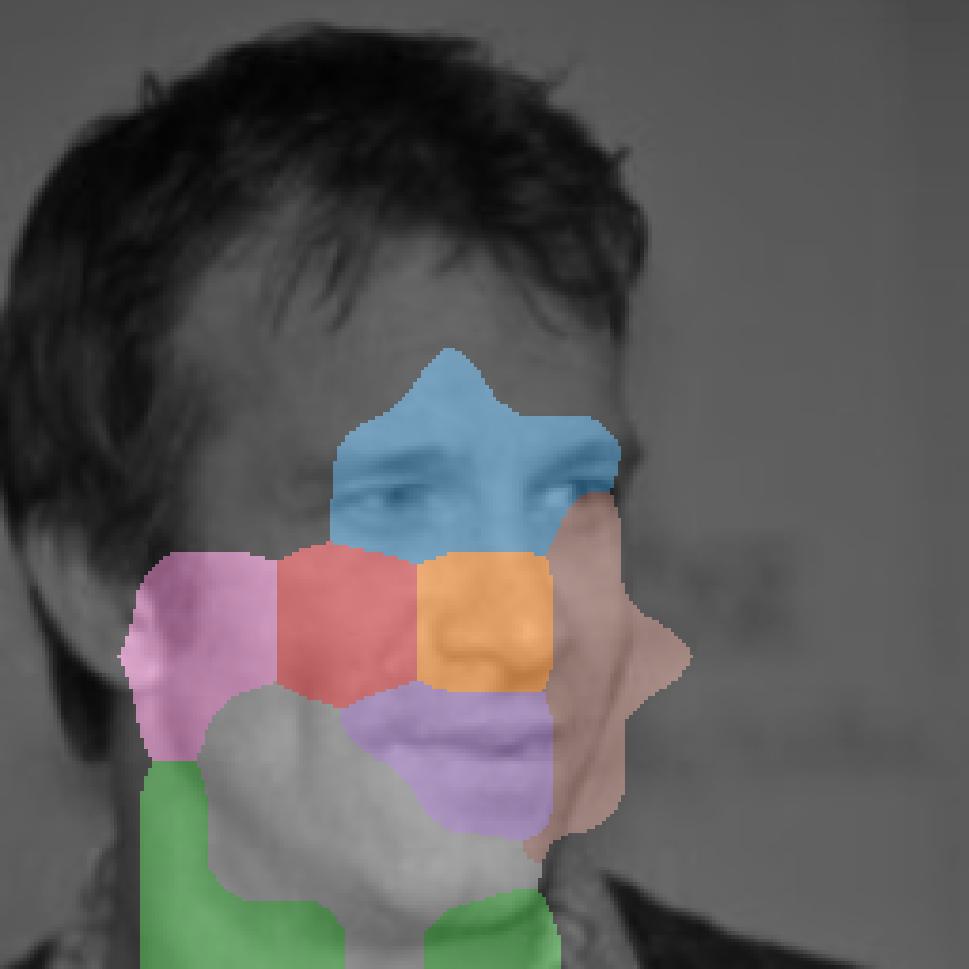} & \includegraphics[width=0.088\linewidth]{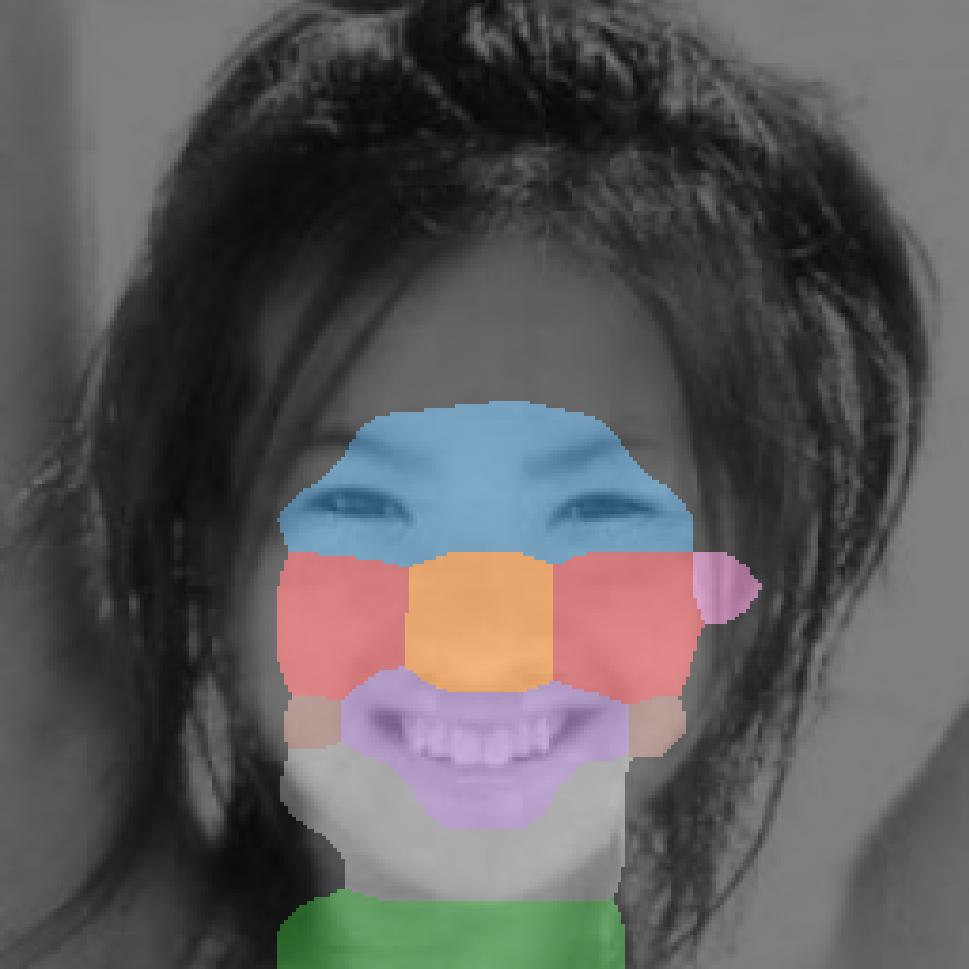} & \includegraphics[width=0.088\linewidth]{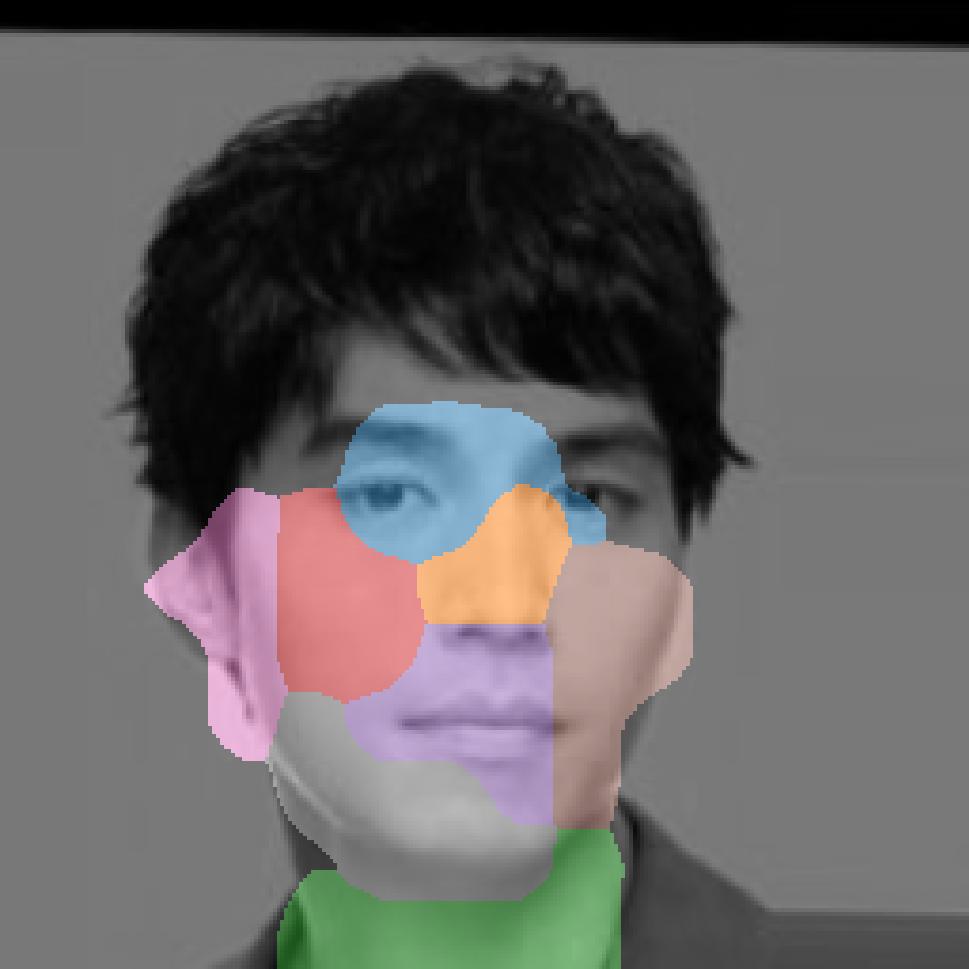} & \includegraphics[width=0.088\linewidth]{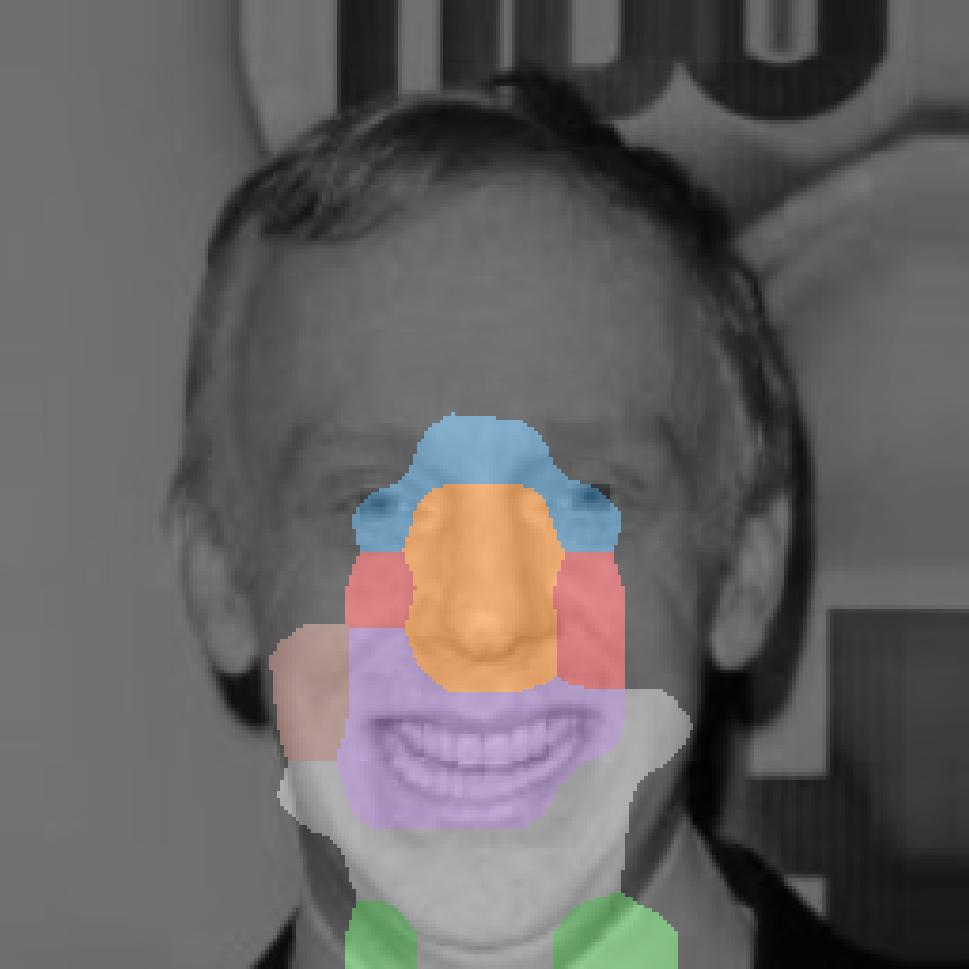} & \includegraphics[width=0.088\linewidth]{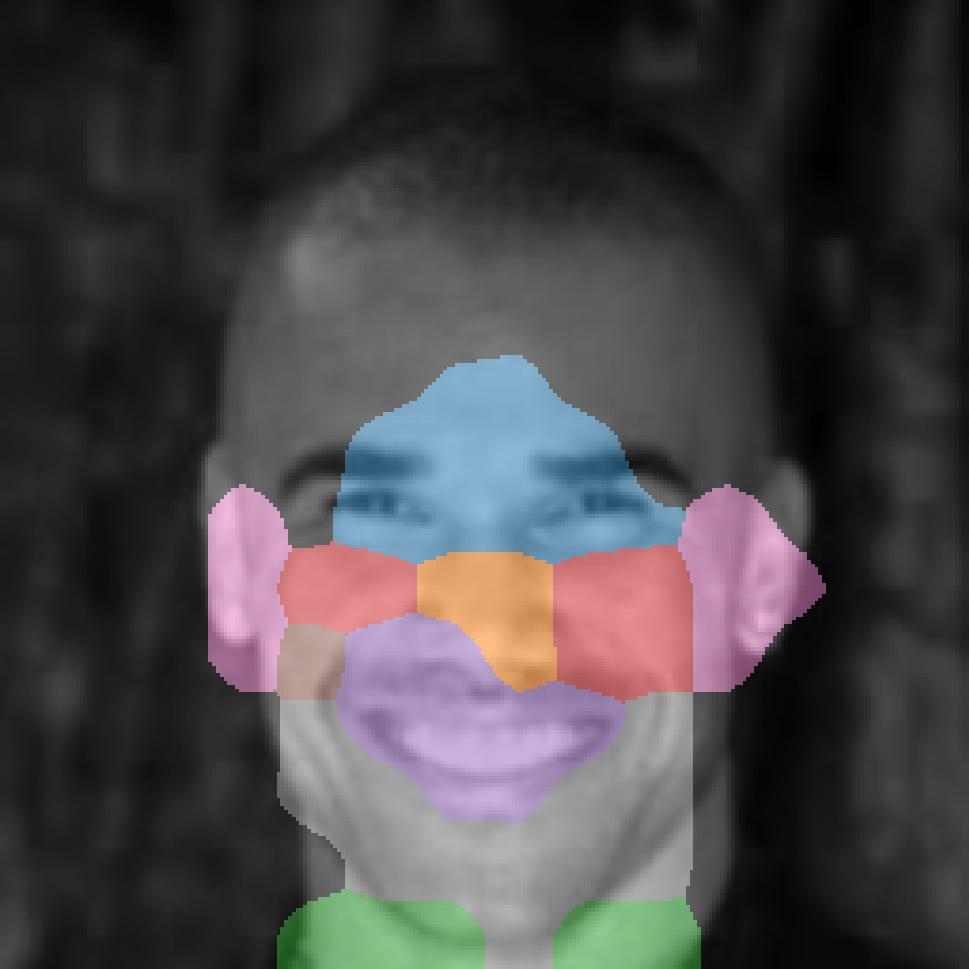} & \includegraphics[width=0.088\linewidth]{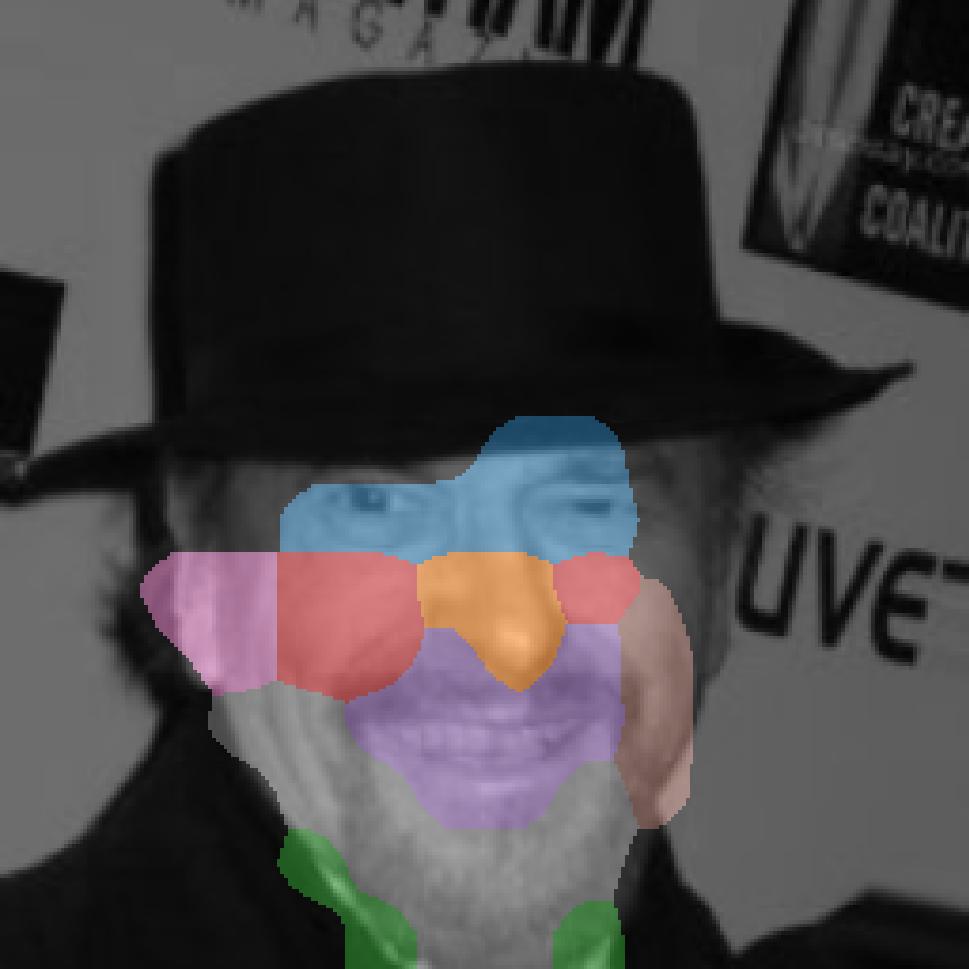} & \includegraphics[width=0.088\linewidth]{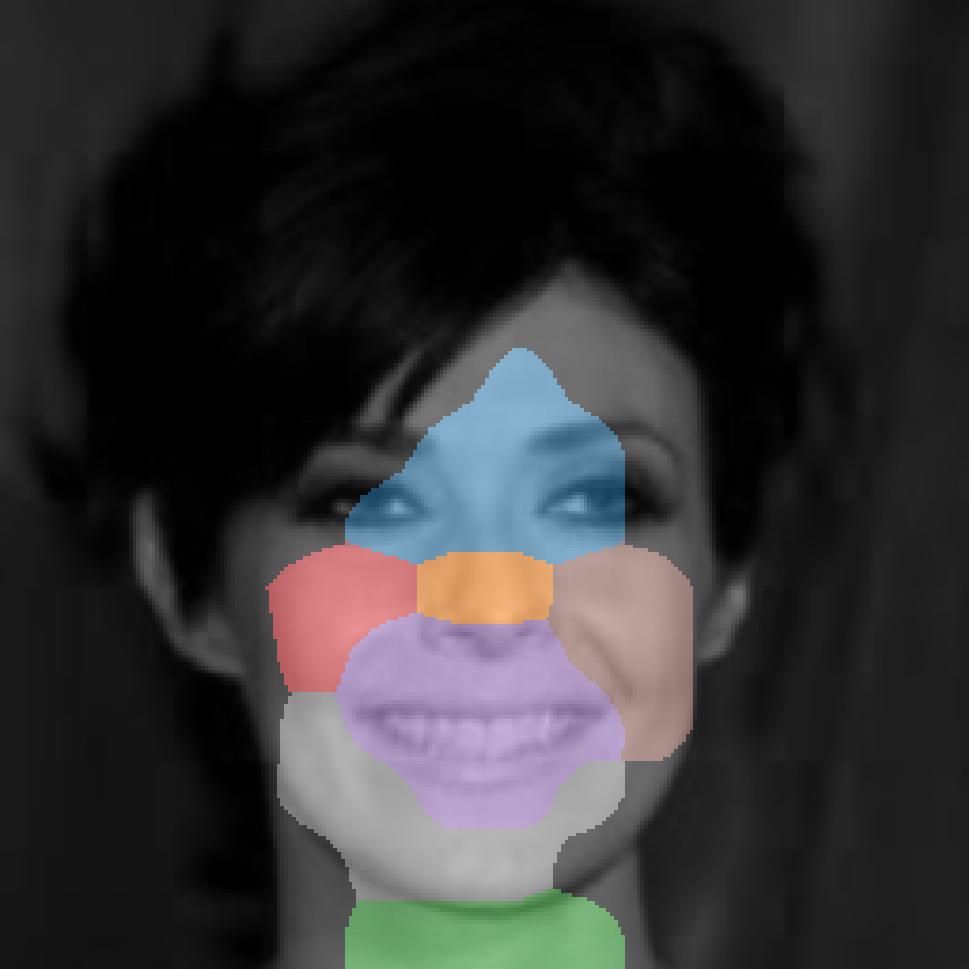} & \includegraphics[width=0.088\linewidth]{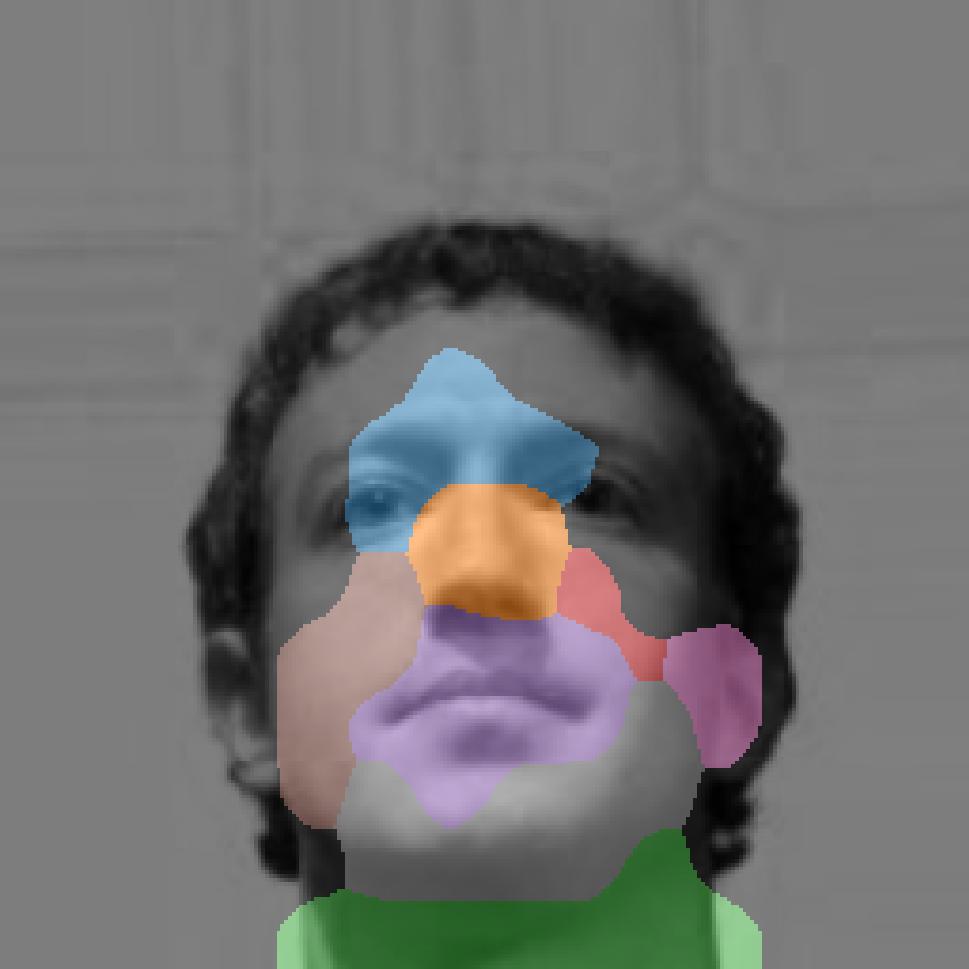} & \includegraphics[width=0.088\linewidth]{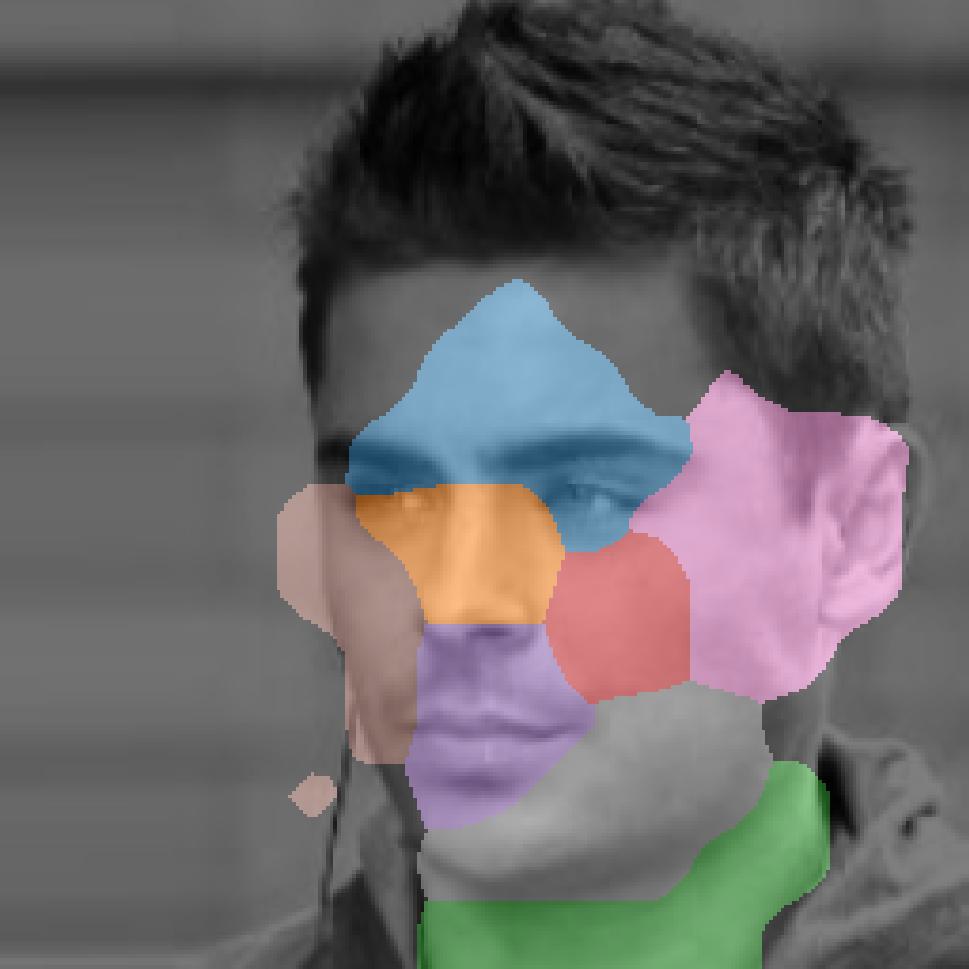} \\
    \rotatebox{90}{\makebox[1cm][c]{\scriptsize Soft}} & \includegraphics[width=0.088\linewidth]{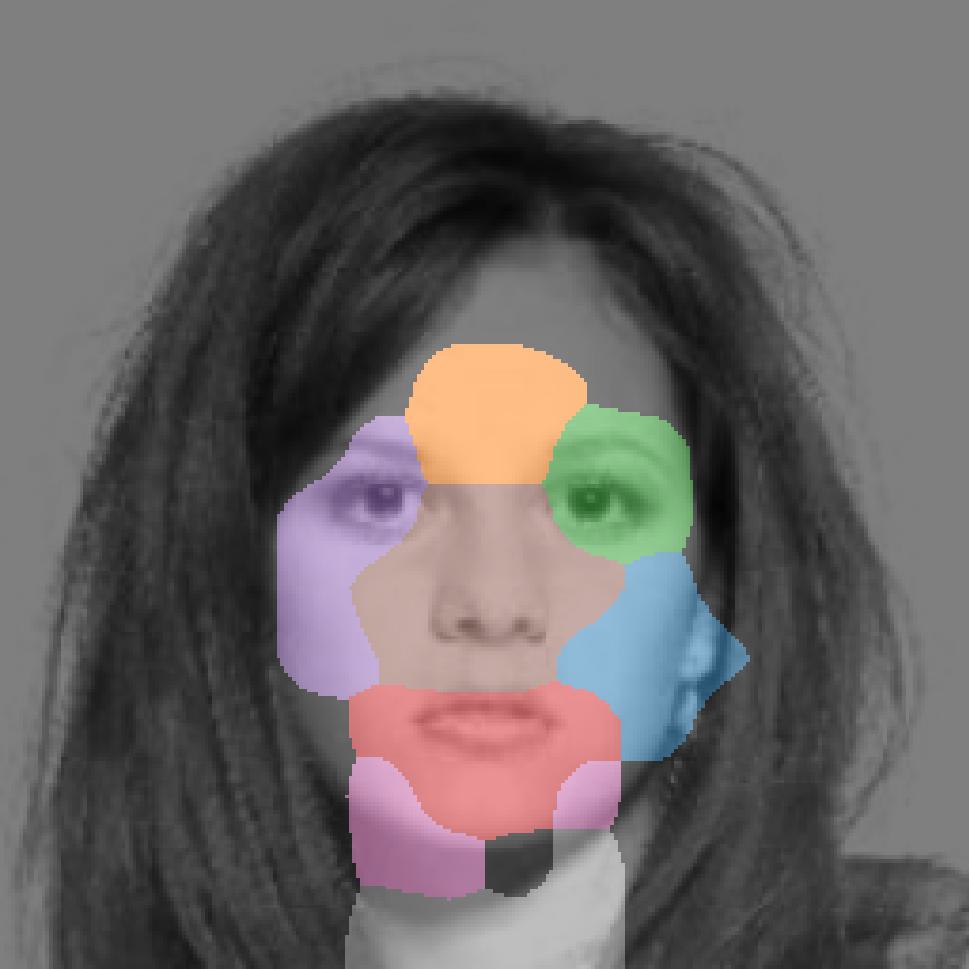} & \includegraphics[width=0.088\linewidth]{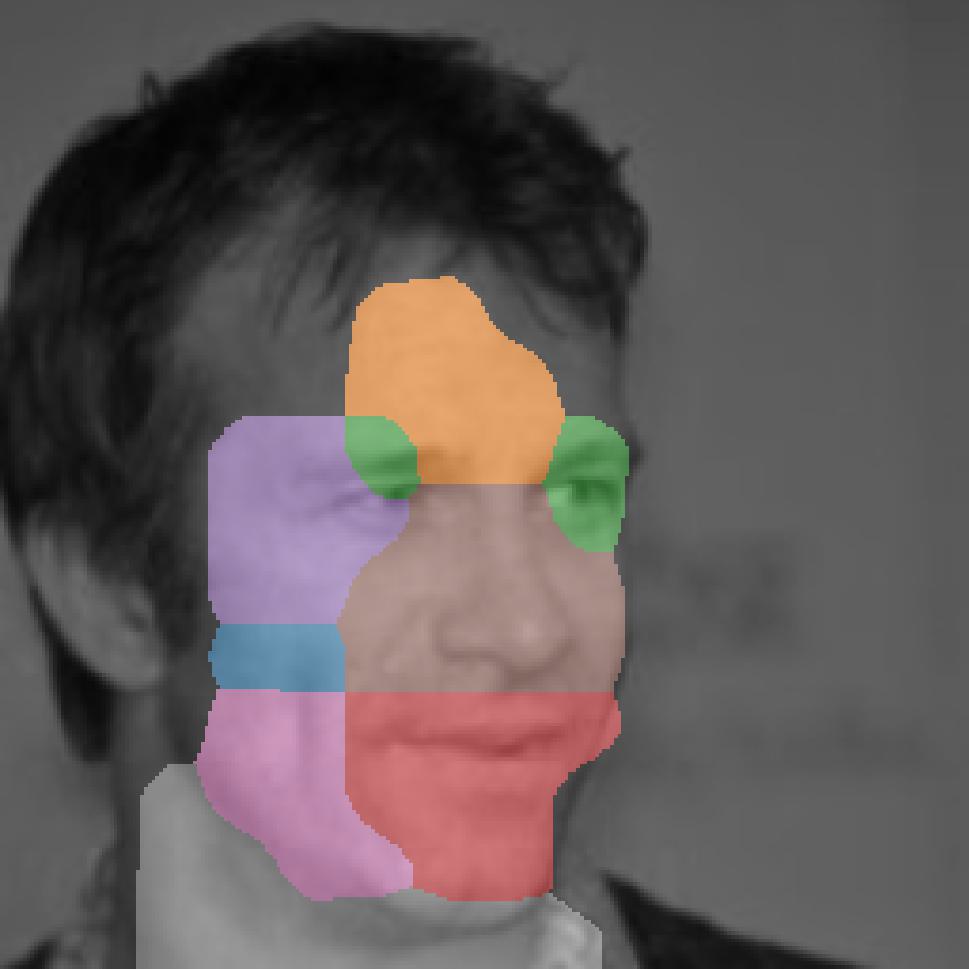} & \includegraphics[width=0.088\linewidth]{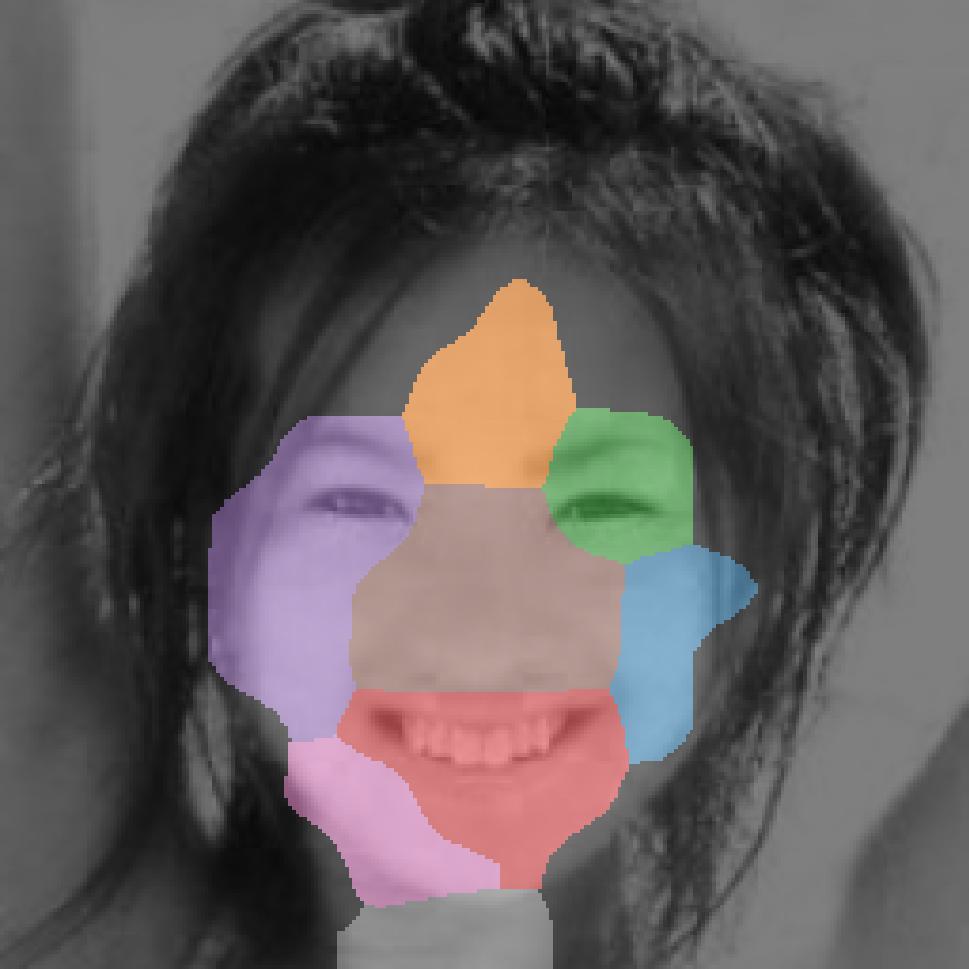} & \includegraphics[width=0.088\linewidth]{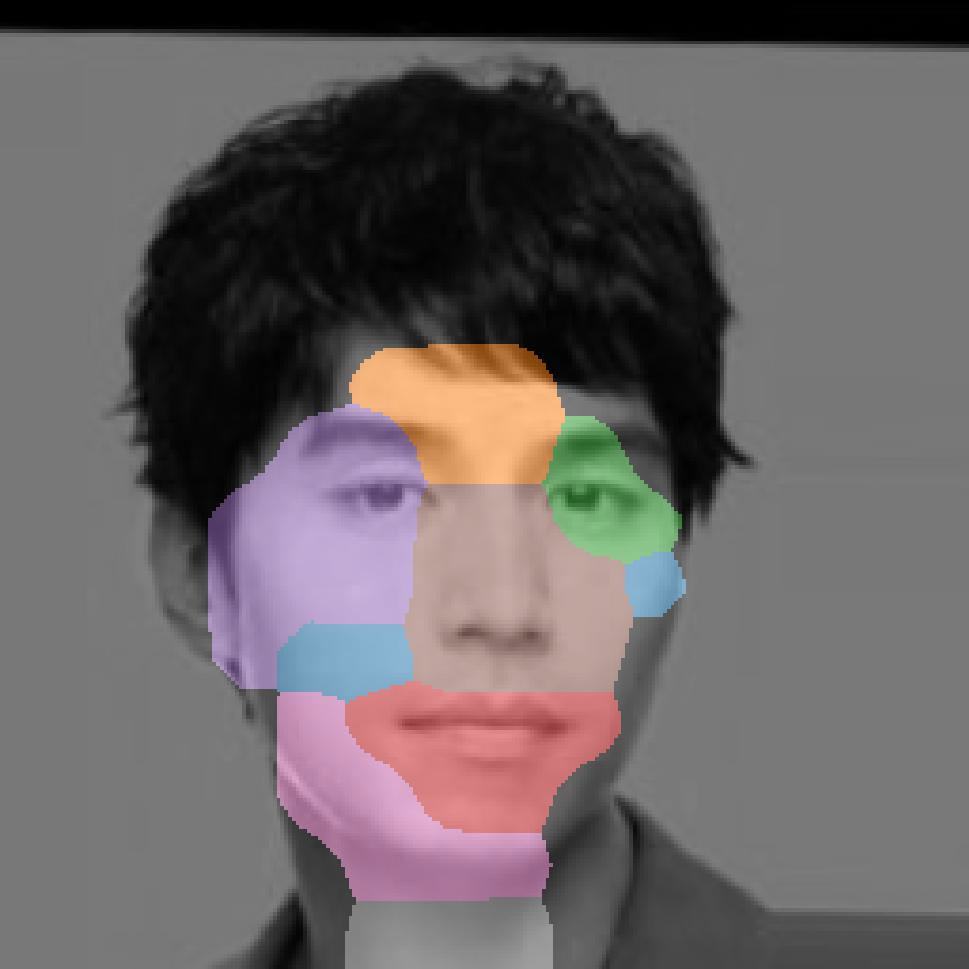} & \includegraphics[width=0.088\linewidth]{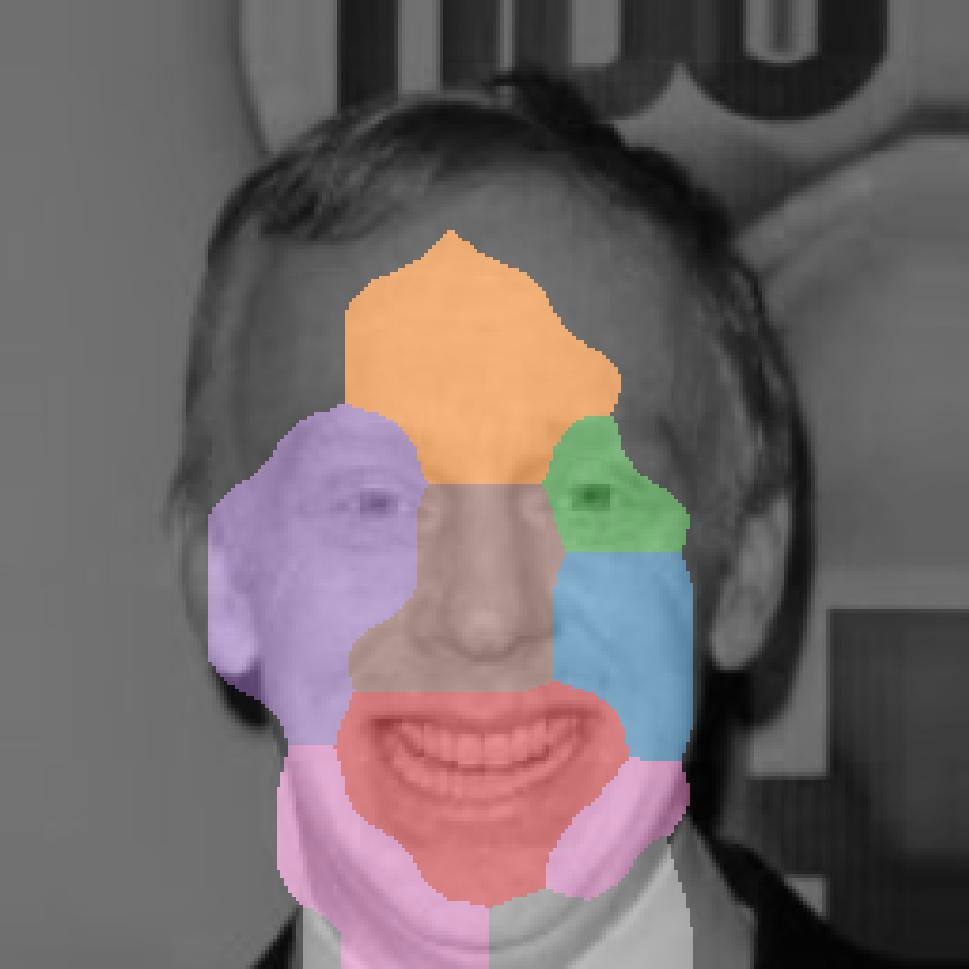} & \includegraphics[width=0.088\linewidth]{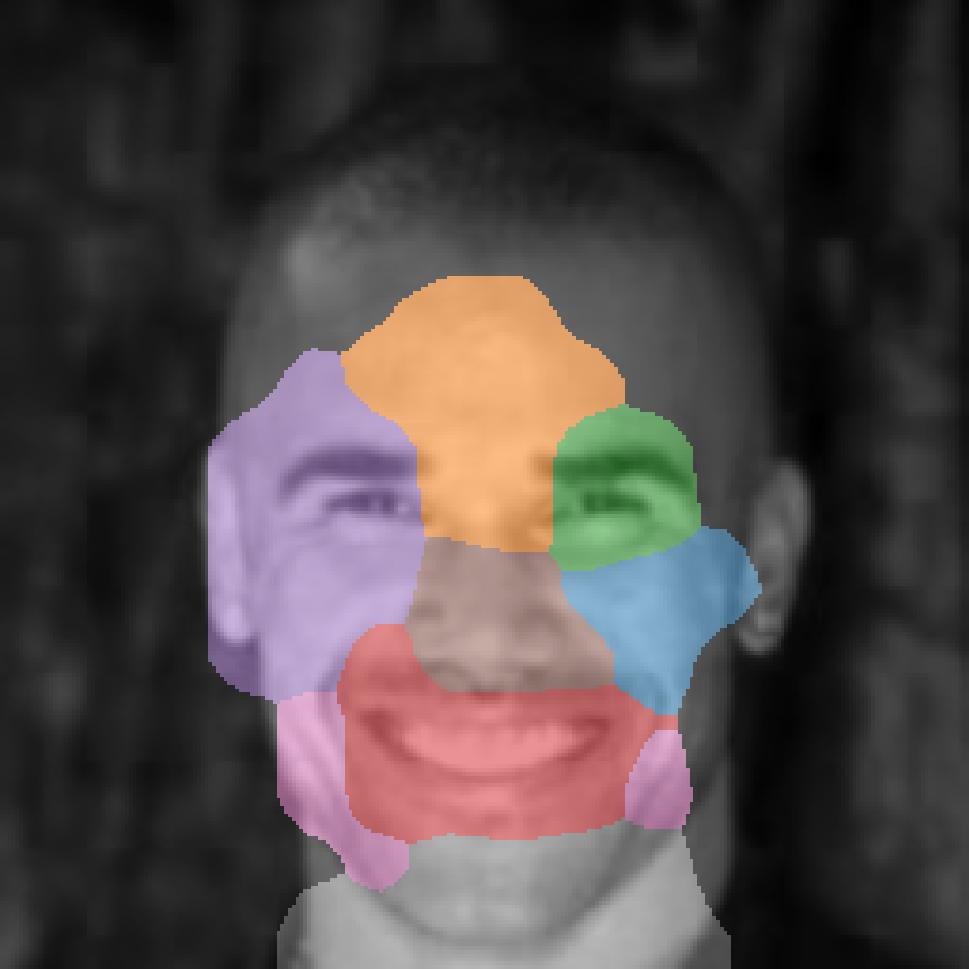} & \includegraphics[width=0.088\linewidth]{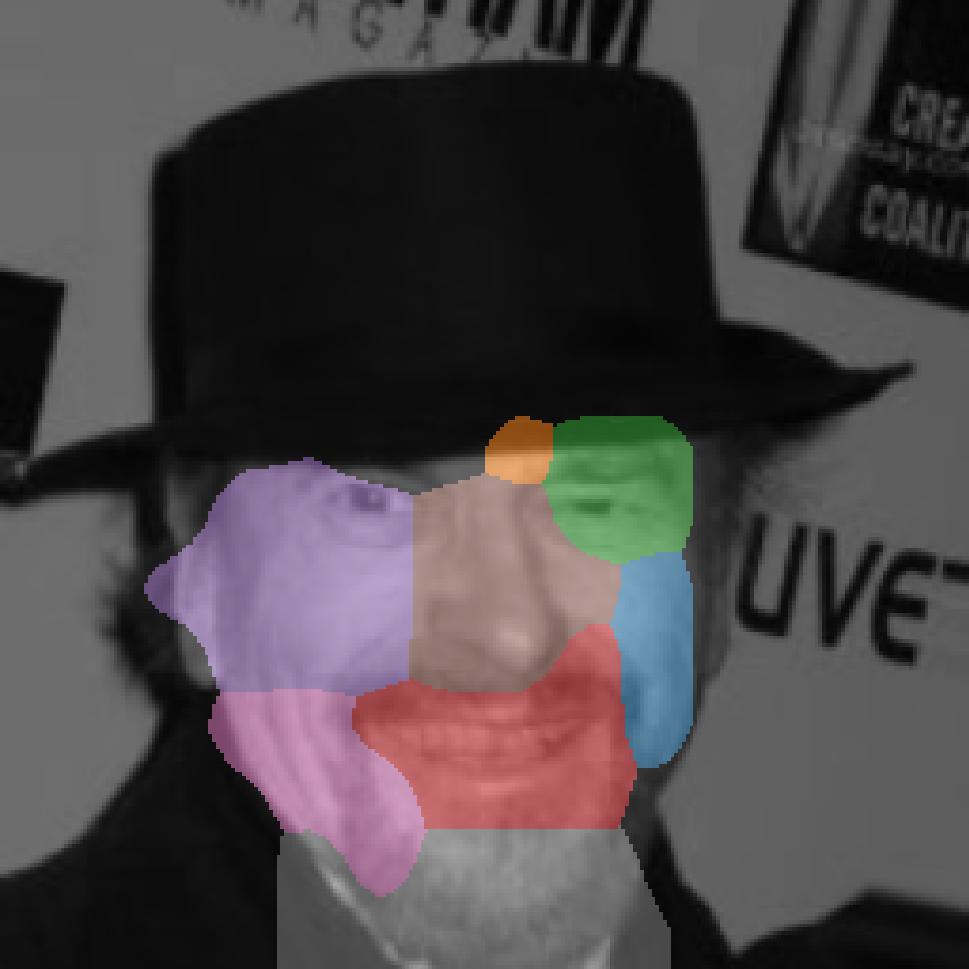} & \includegraphics[width=0.088\linewidth]{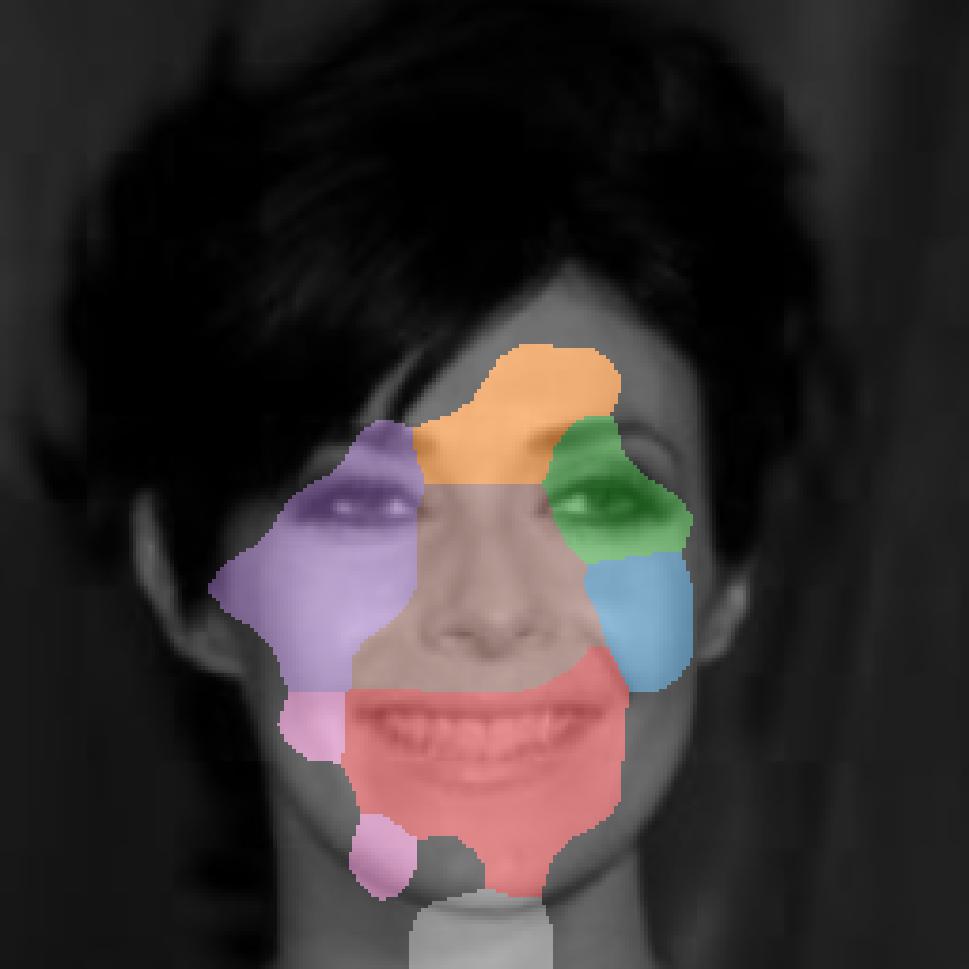} & \includegraphics[width=0.088\linewidth]{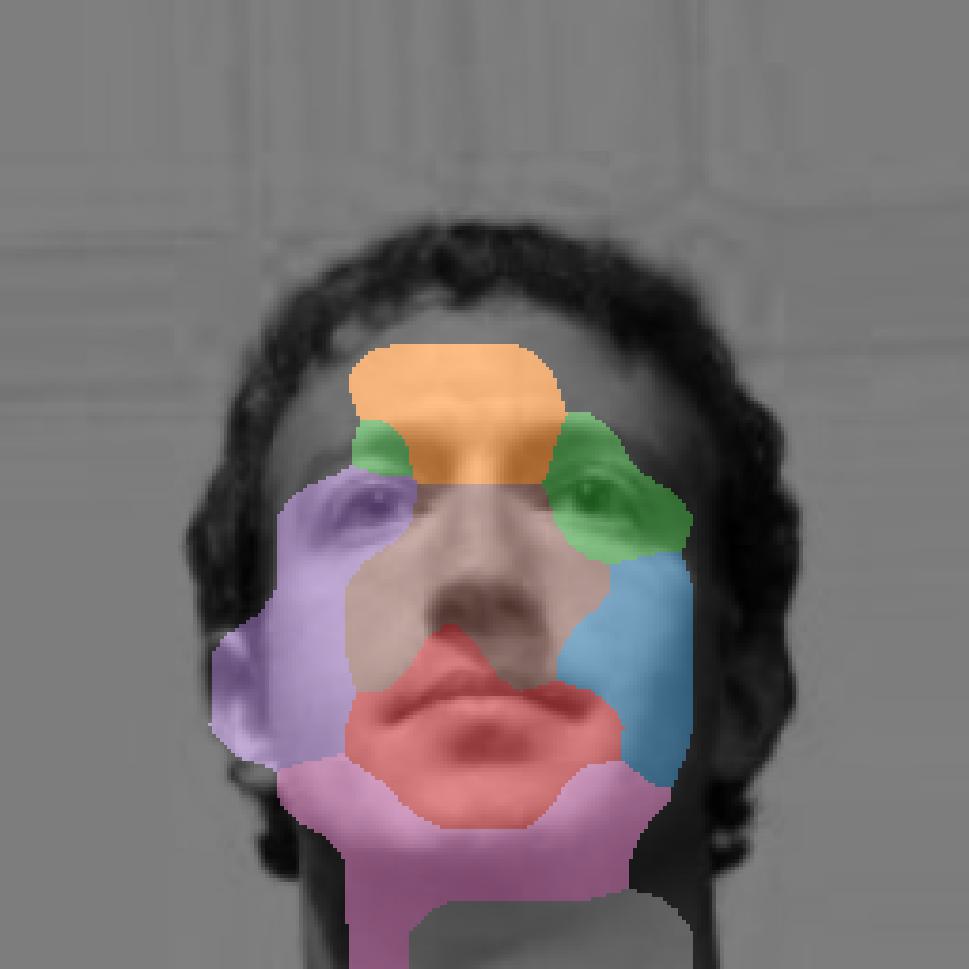} & \includegraphics[width=0.088\linewidth]{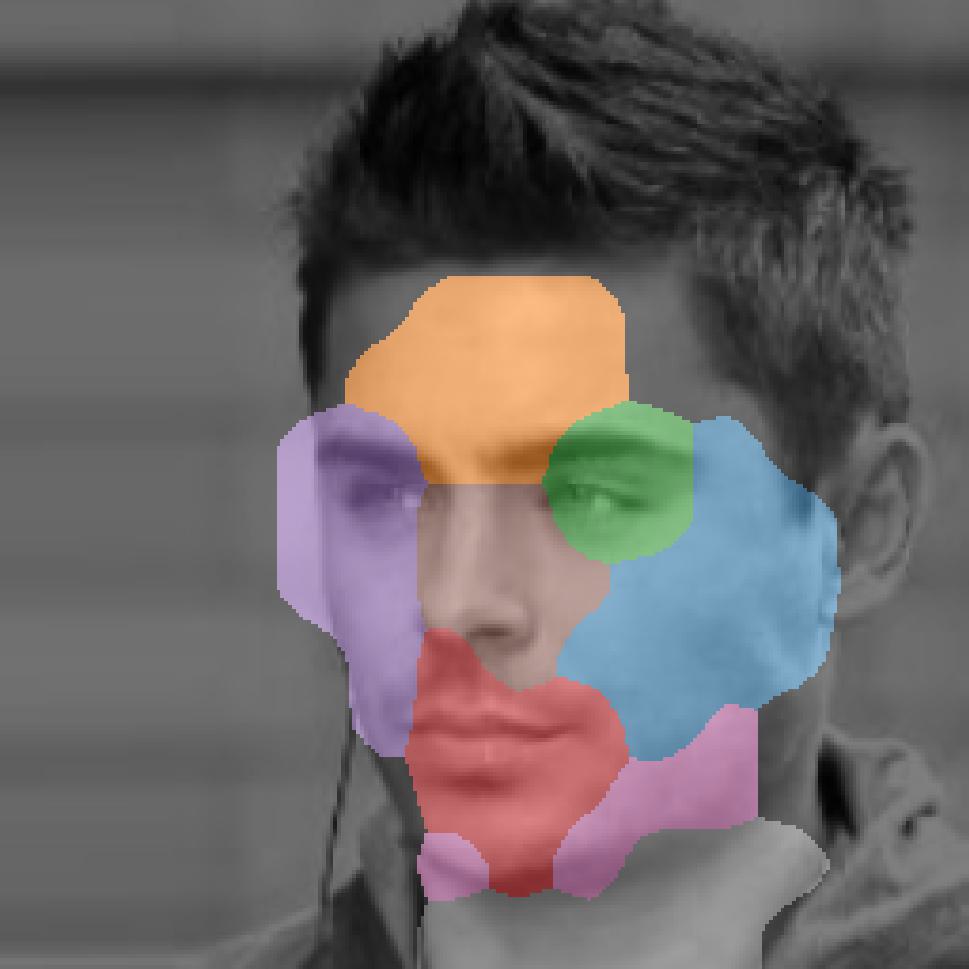} \\
    \rotatebox{90}{\makebox[1cm][c]{\scriptsize Hard}} & \includegraphics[width=0.088\linewidth]{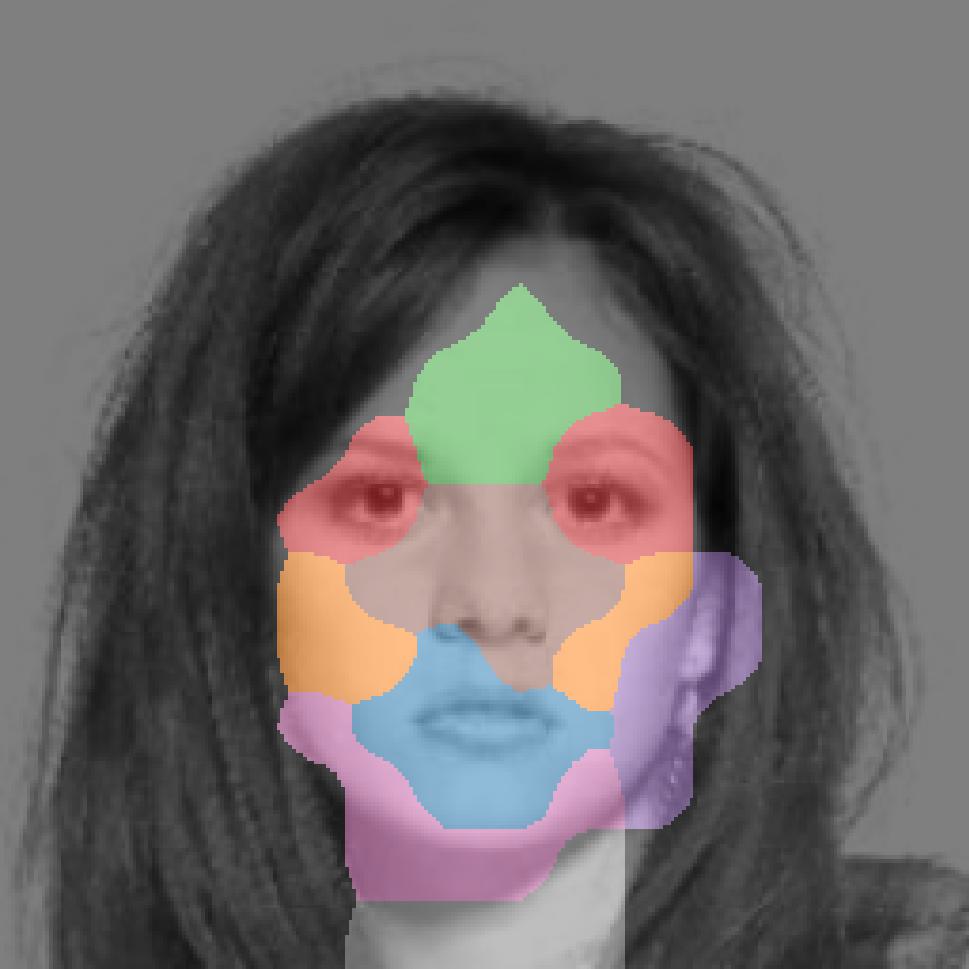} & \includegraphics[width=0.088\linewidth]{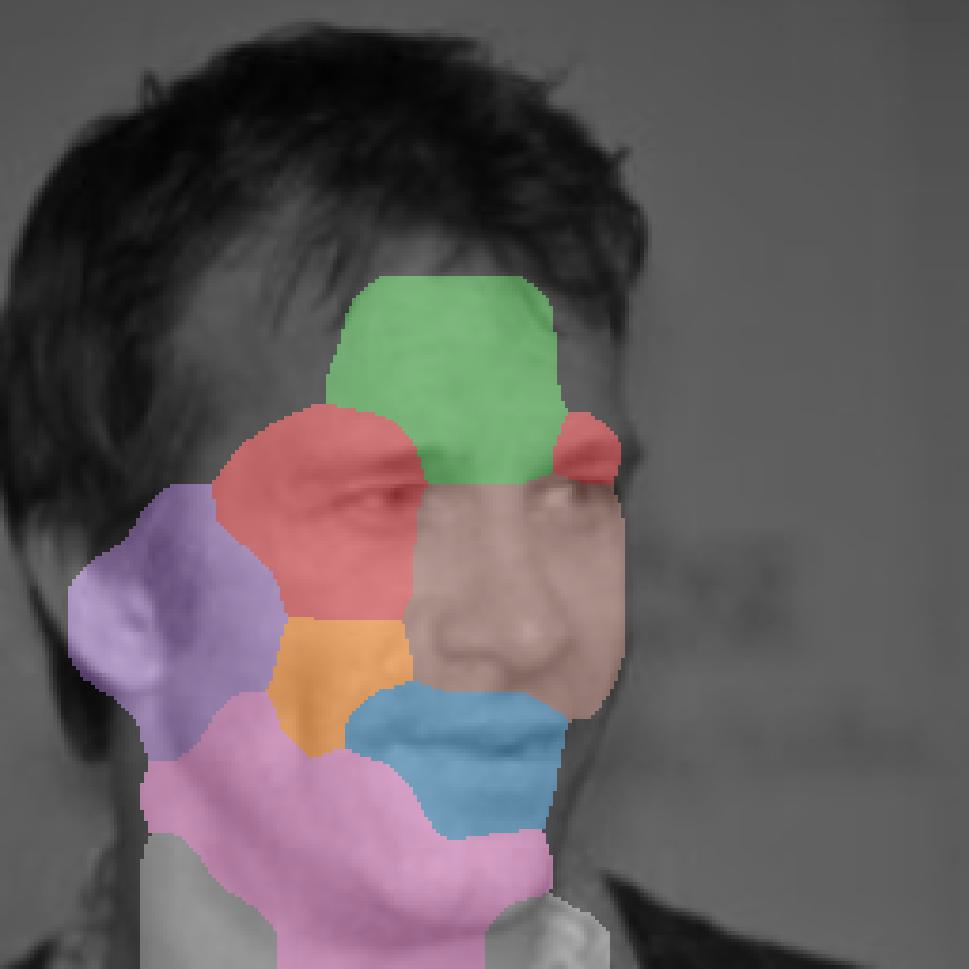} & \includegraphics[width=0.088\linewidth]{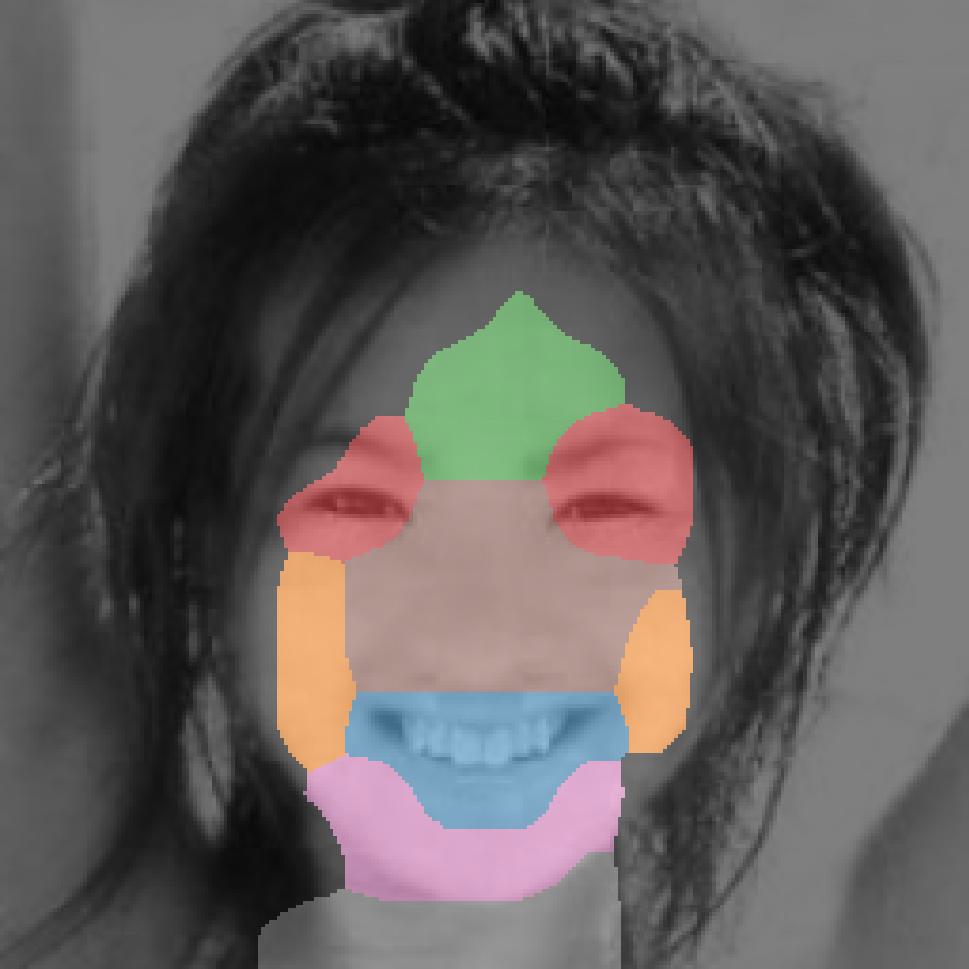} & \includegraphics[width=0.088\linewidth]{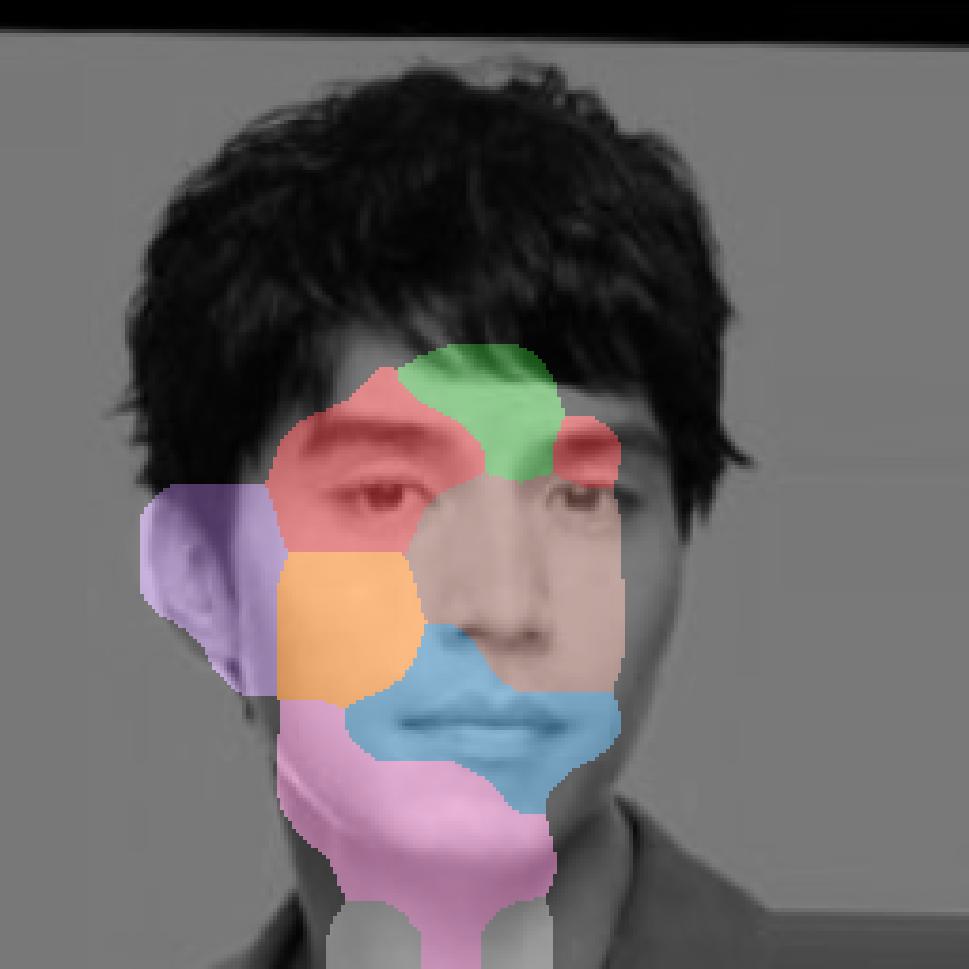} & \includegraphics[width=0.088\linewidth]{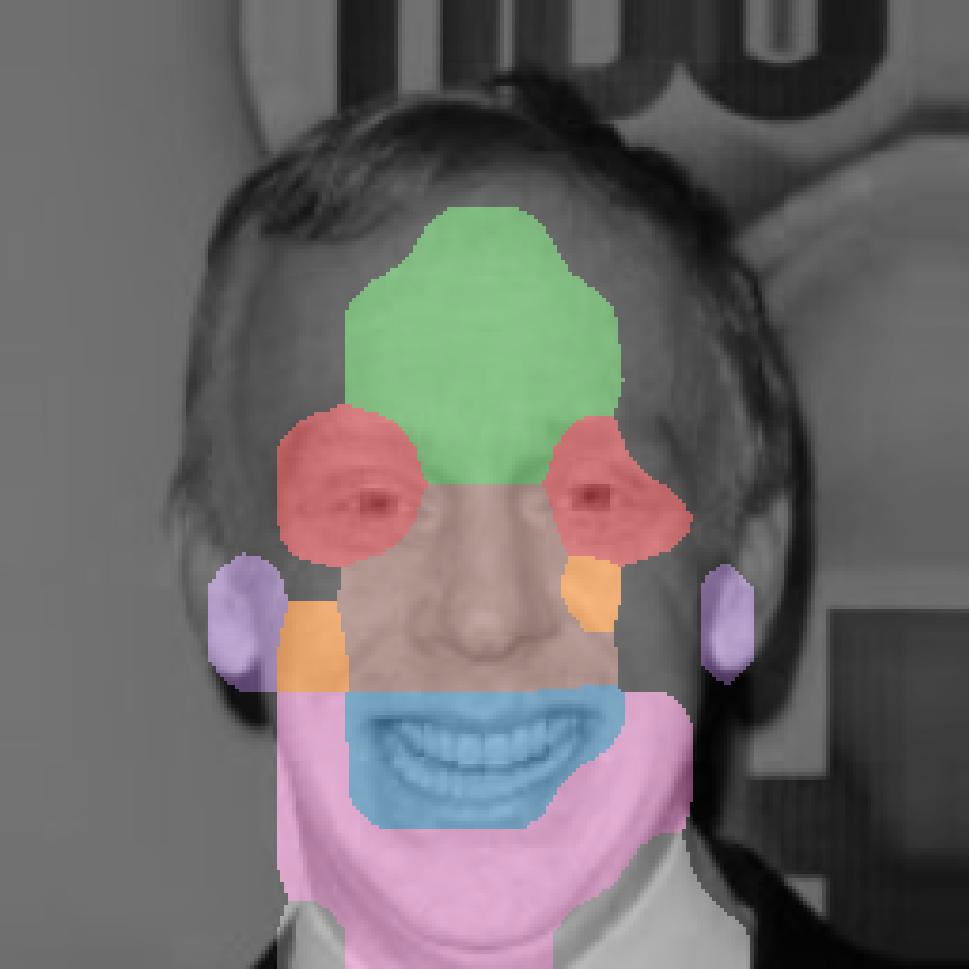} & \includegraphics[width=0.088\linewidth]{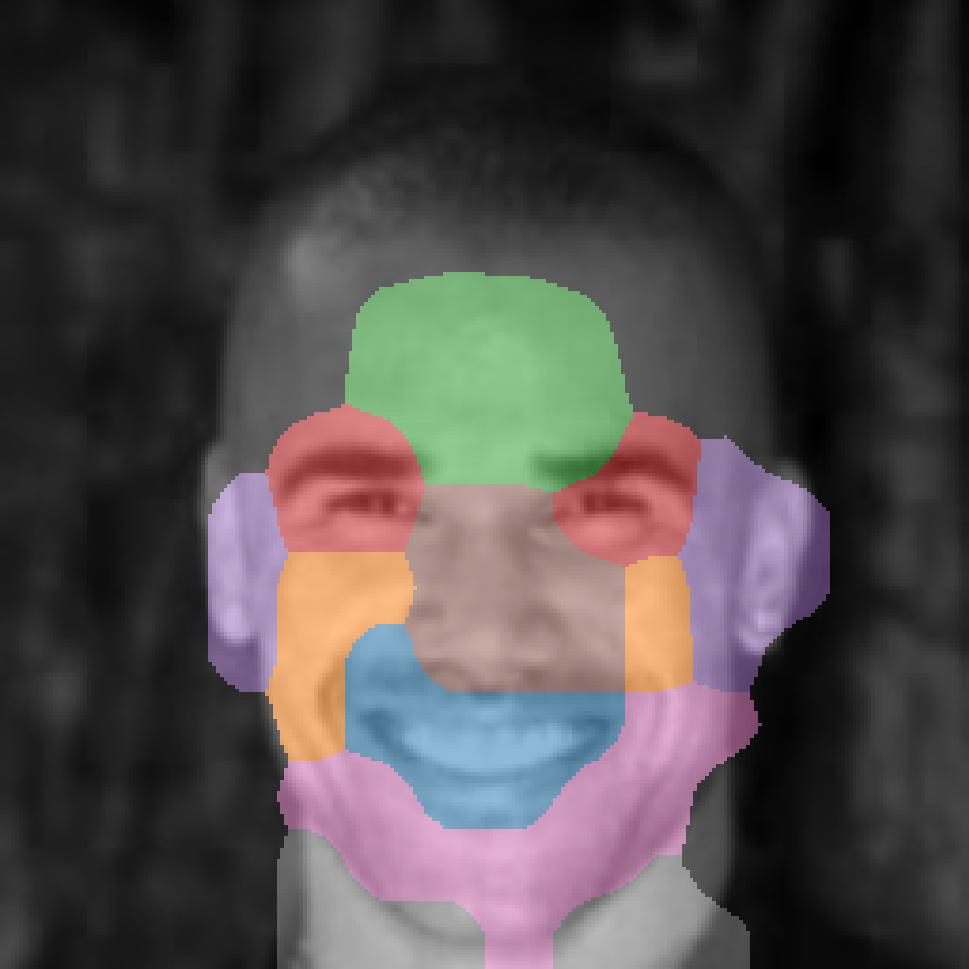} & \includegraphics[width=0.088\linewidth]{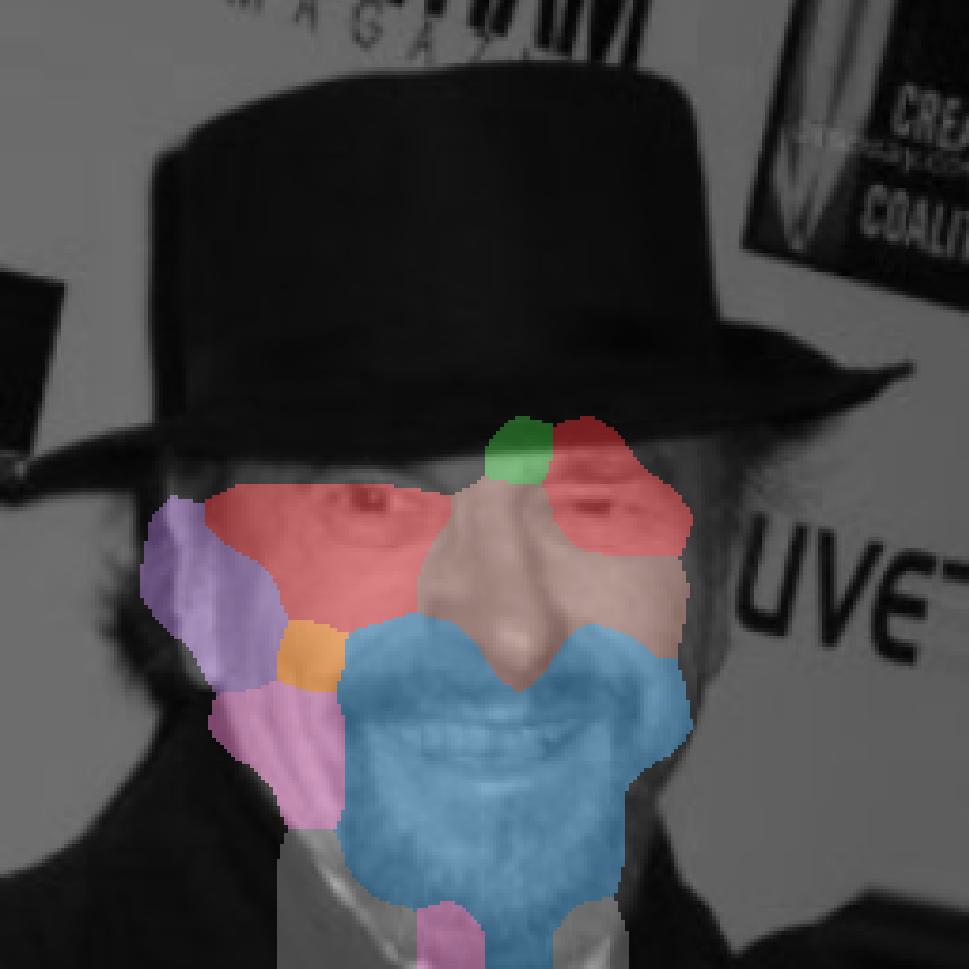} & \includegraphics[width=0.088\linewidth]{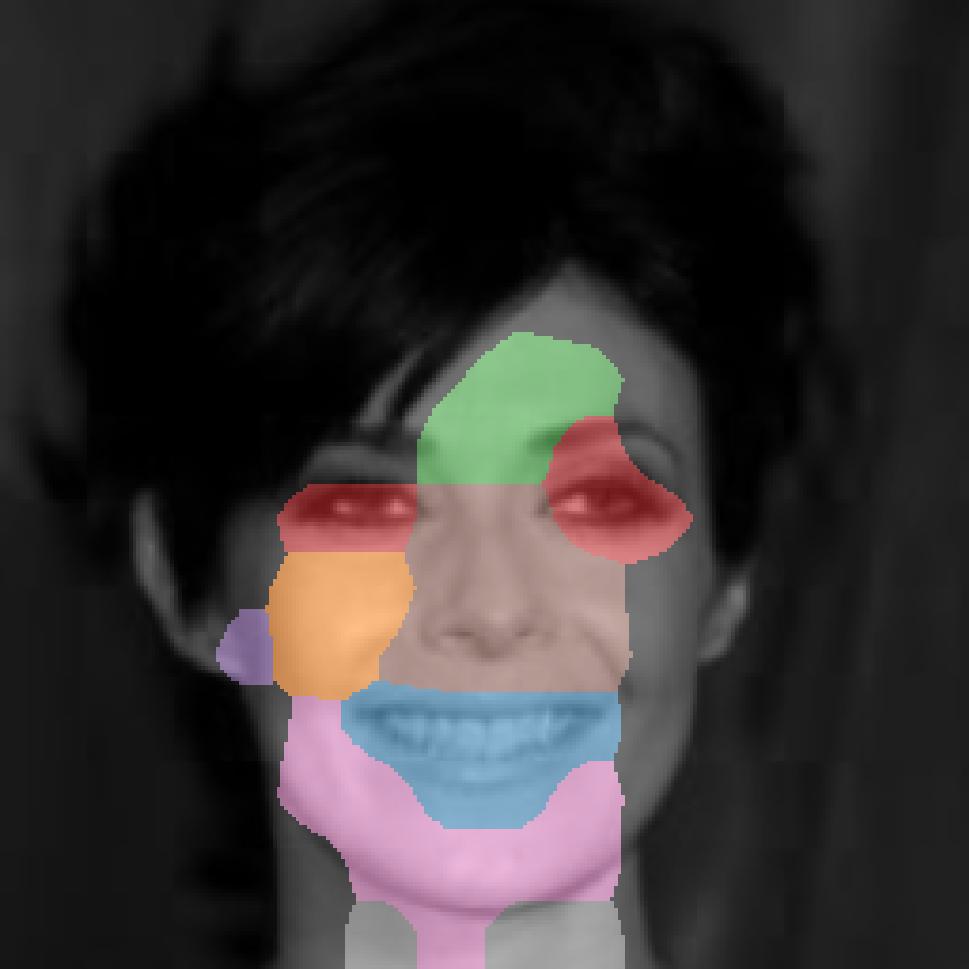} & \includegraphics[width=0.088\linewidth]{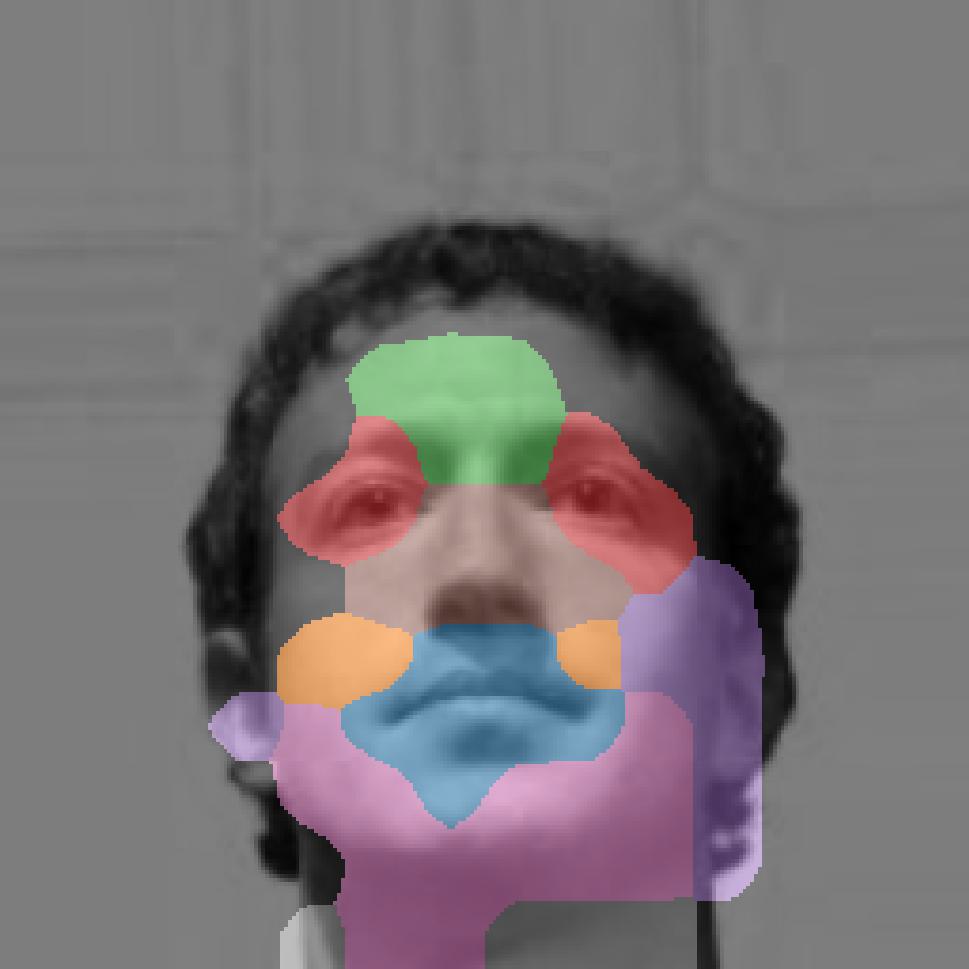} & \includegraphics[width=0.088\linewidth]{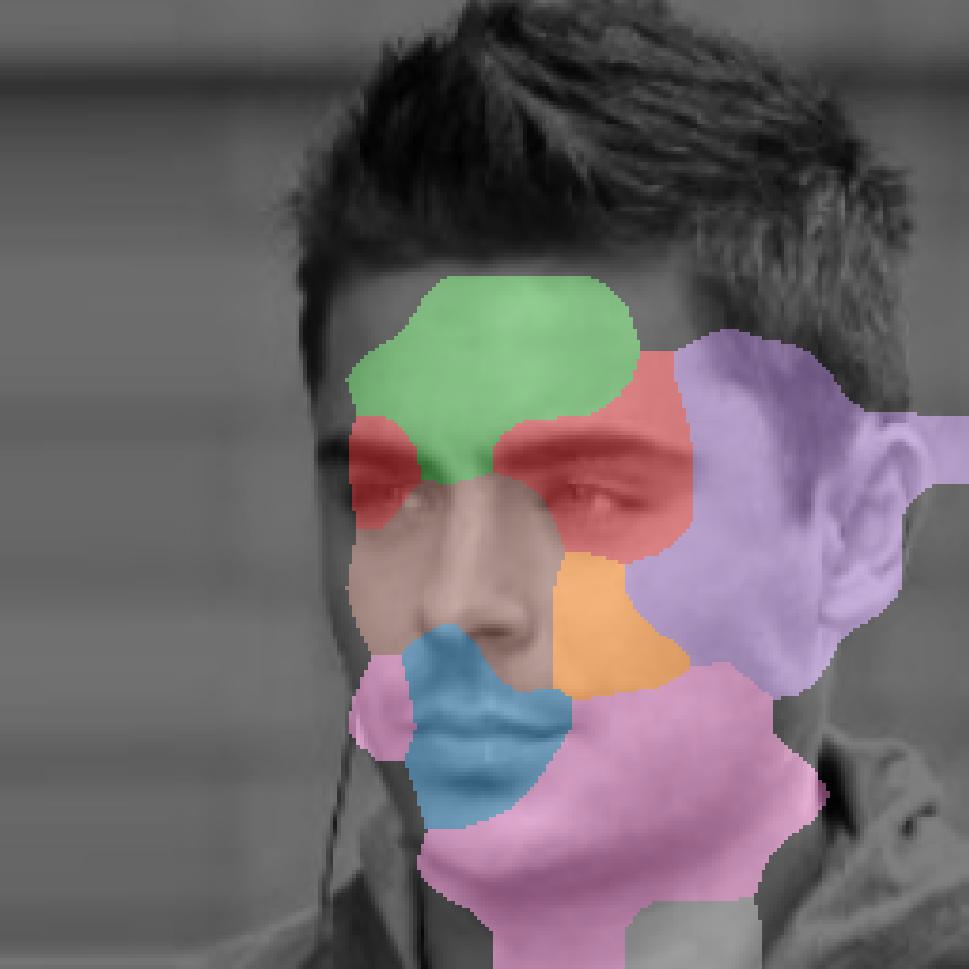} \\
    
    \rotatebox{90}{\makebox[1cm][c]{\scriptsize ST}} & \includegraphics[width=0.088\linewidth]{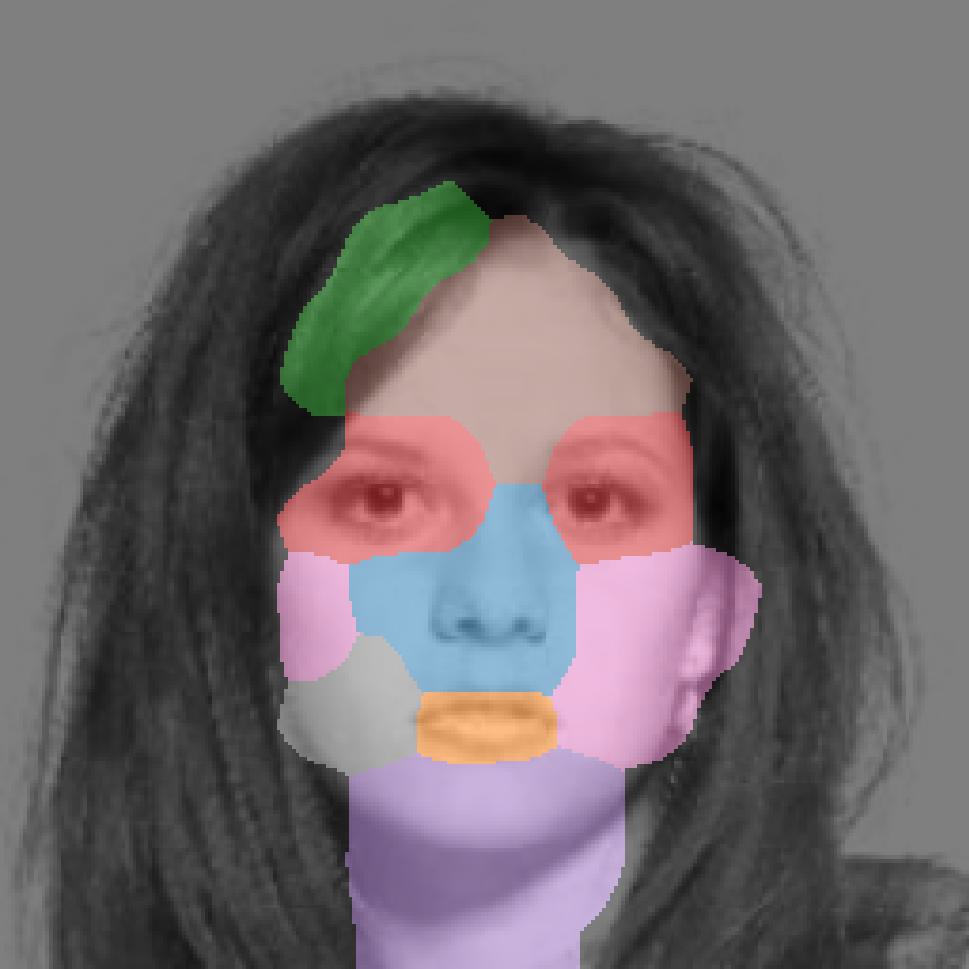} & \includegraphics[width=0.088\linewidth]{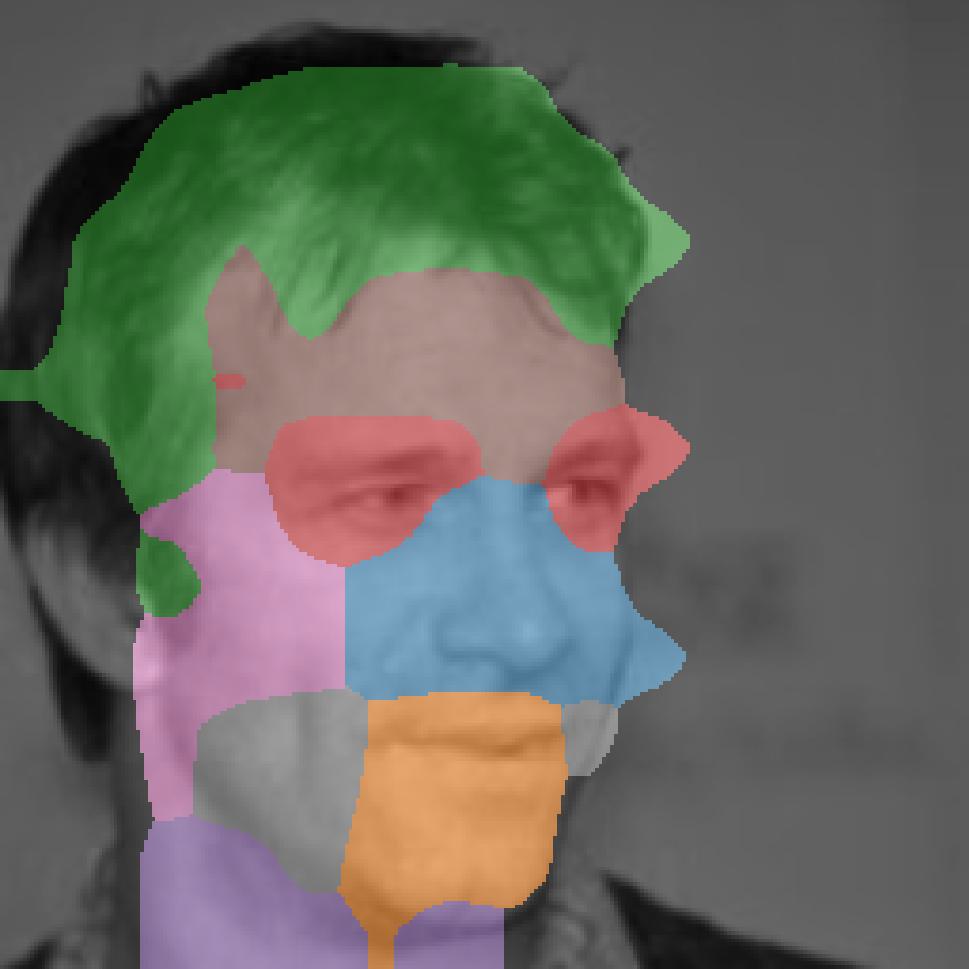} & \includegraphics[width=0.088\linewidth]{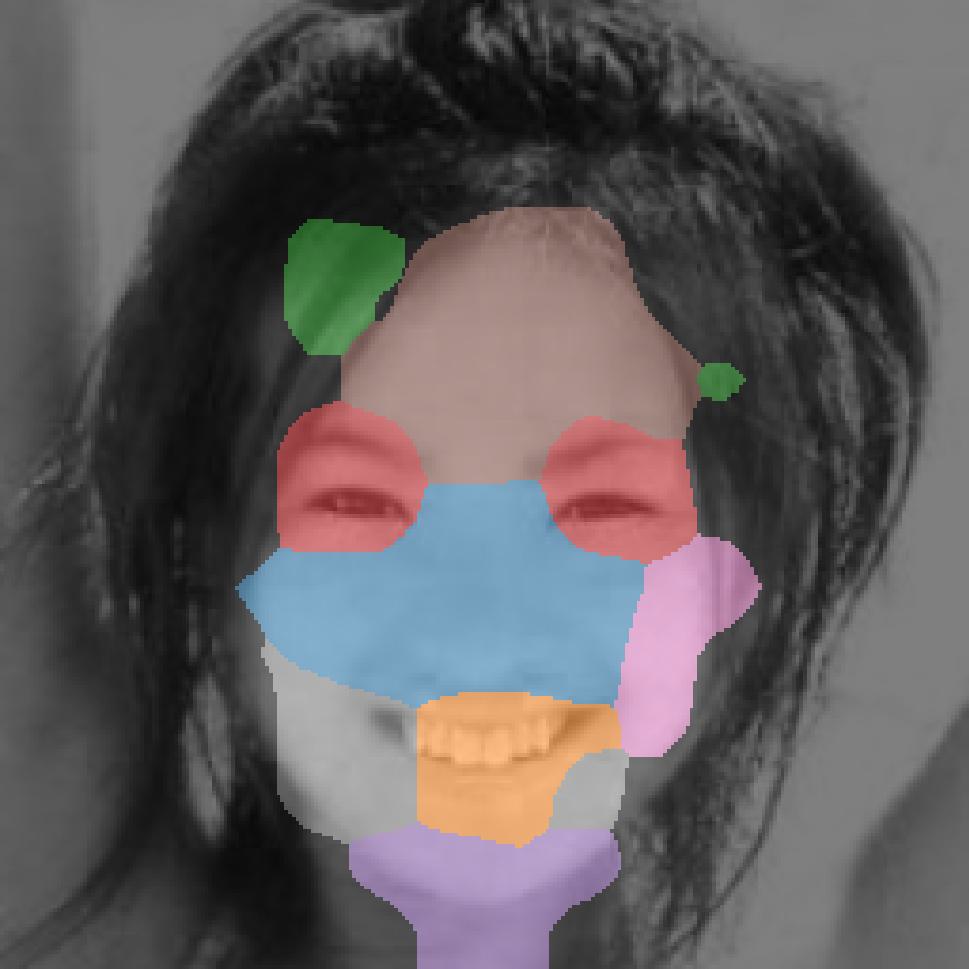} & \includegraphics[width=0.088\linewidth]{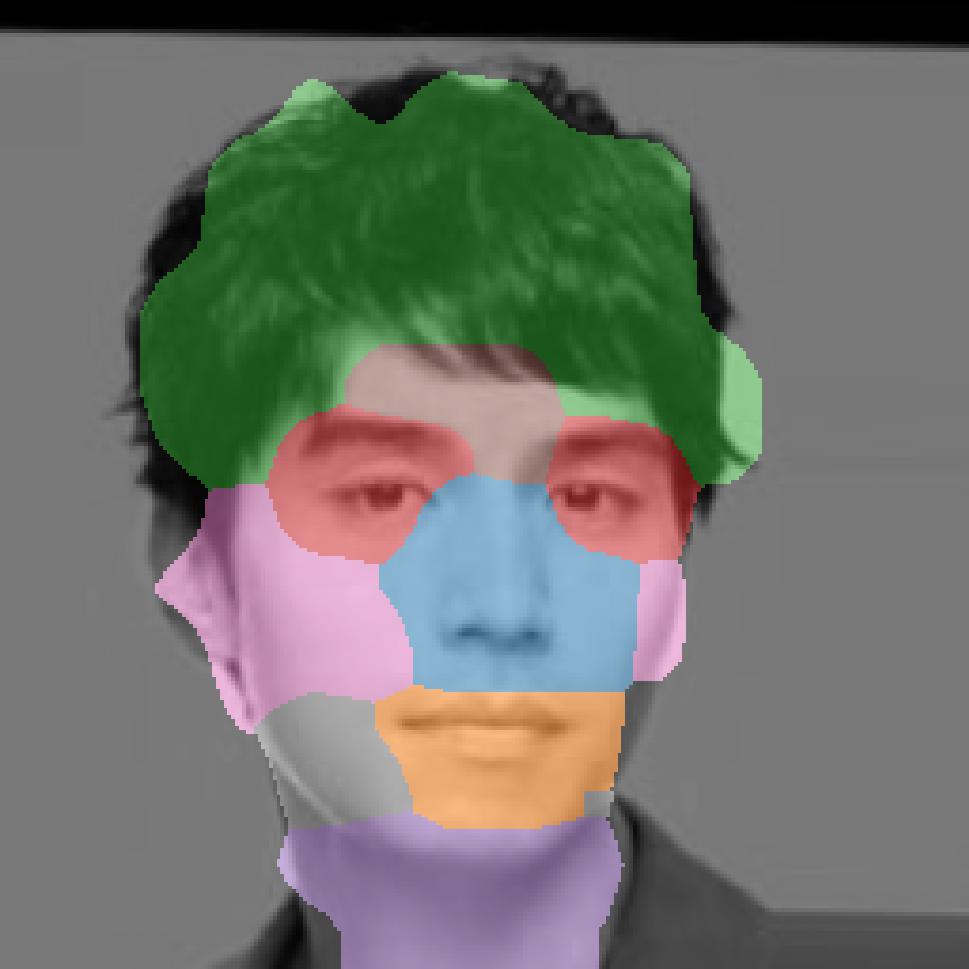} & \includegraphics[width=0.088\linewidth]{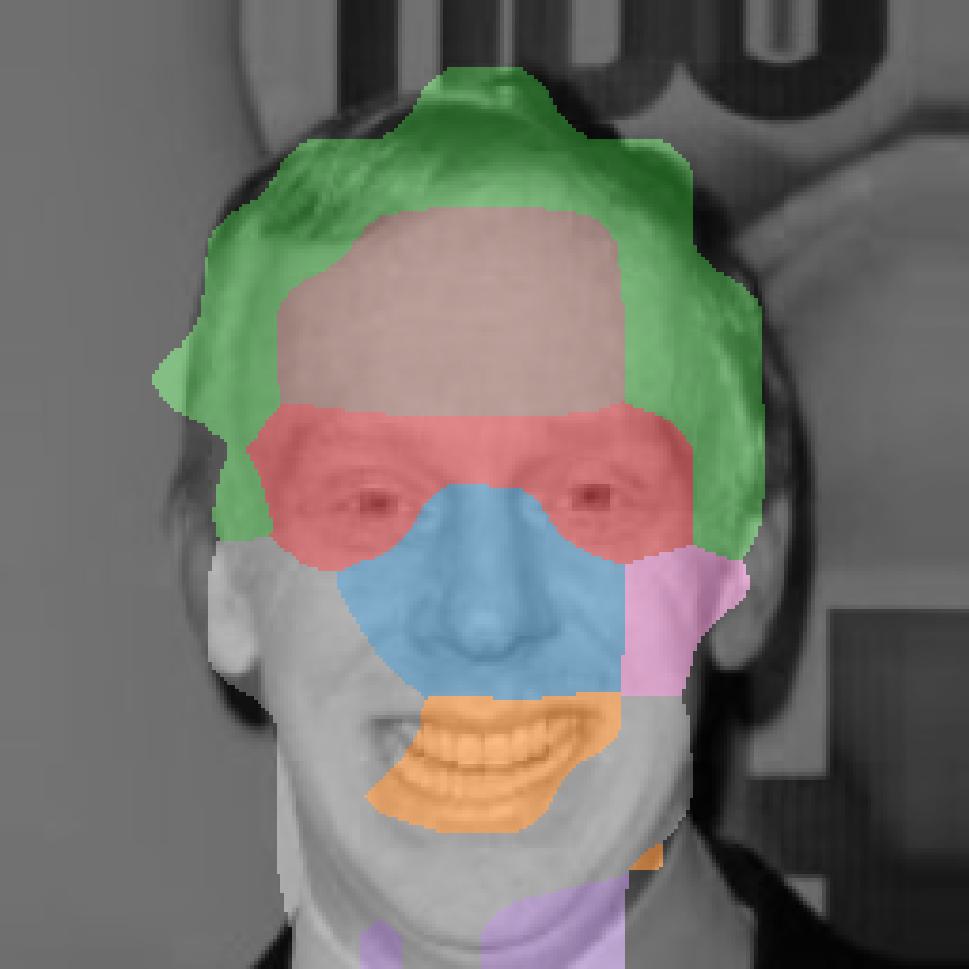} & \includegraphics[width=0.088\linewidth]{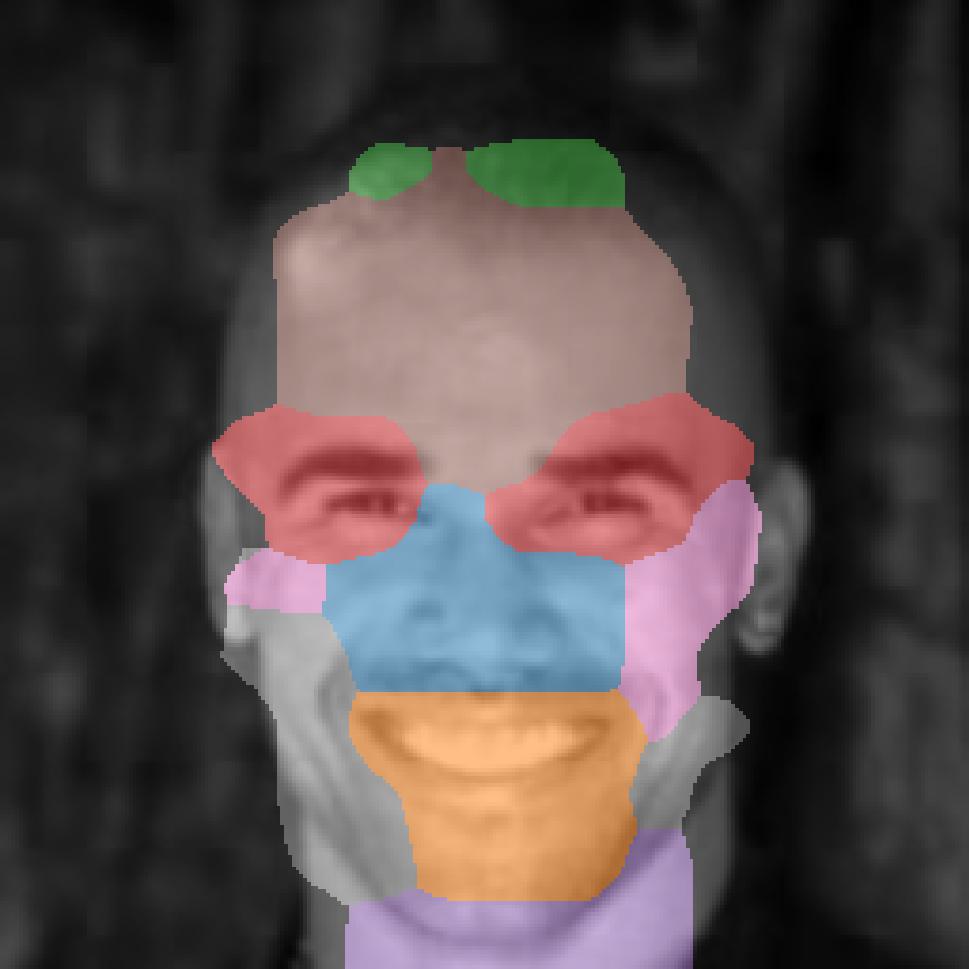} & \includegraphics[width=0.088\linewidth]{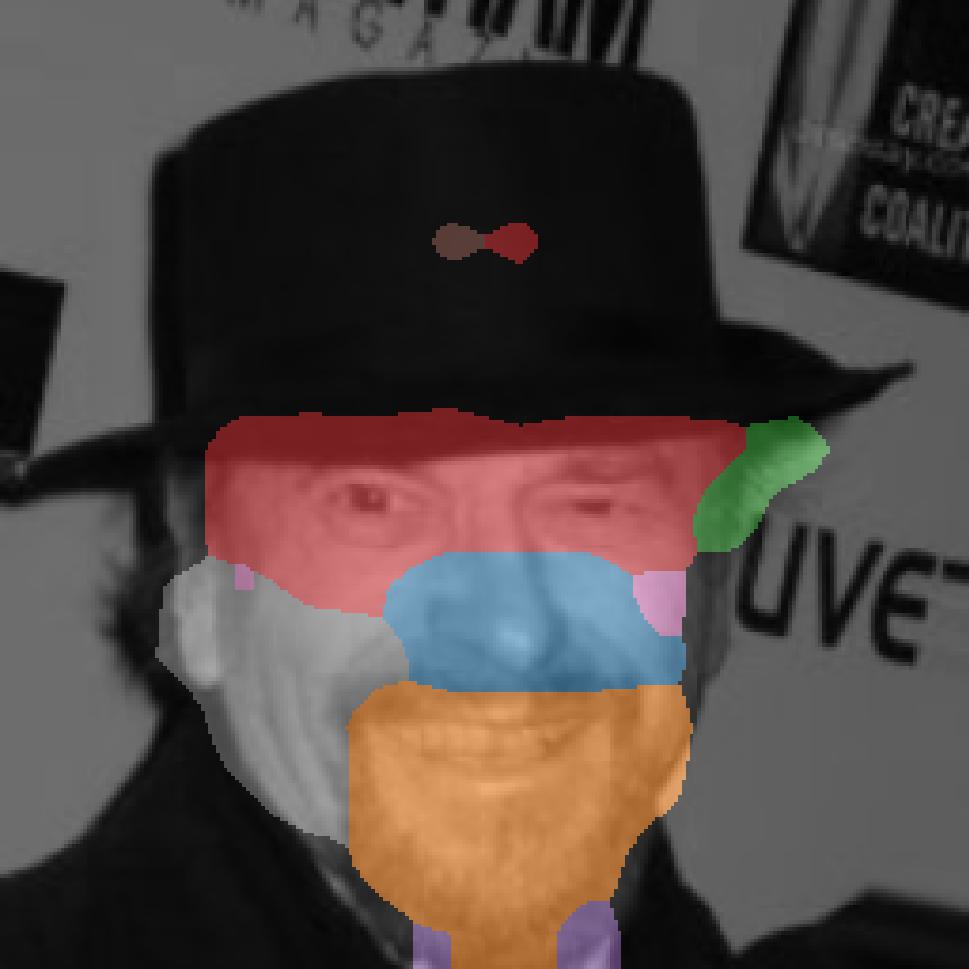} & \includegraphics[width=0.088\linewidth]{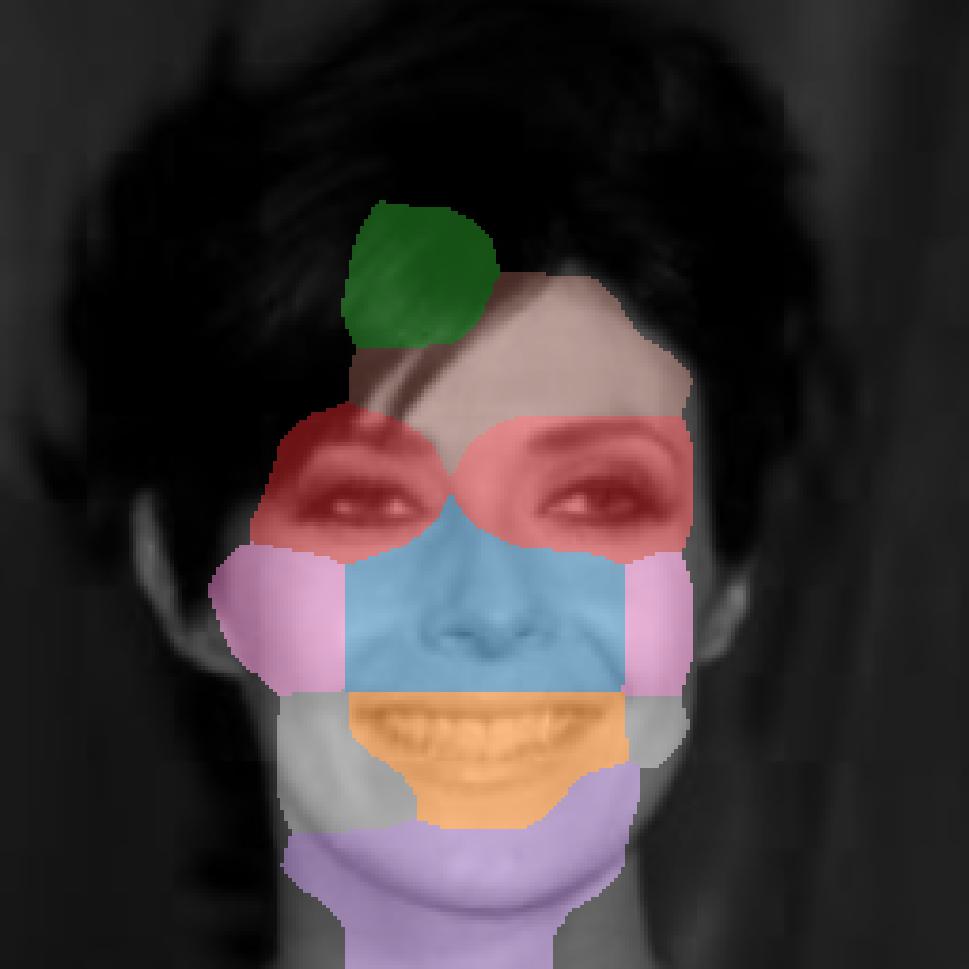} & \includegraphics[width=0.088\linewidth]{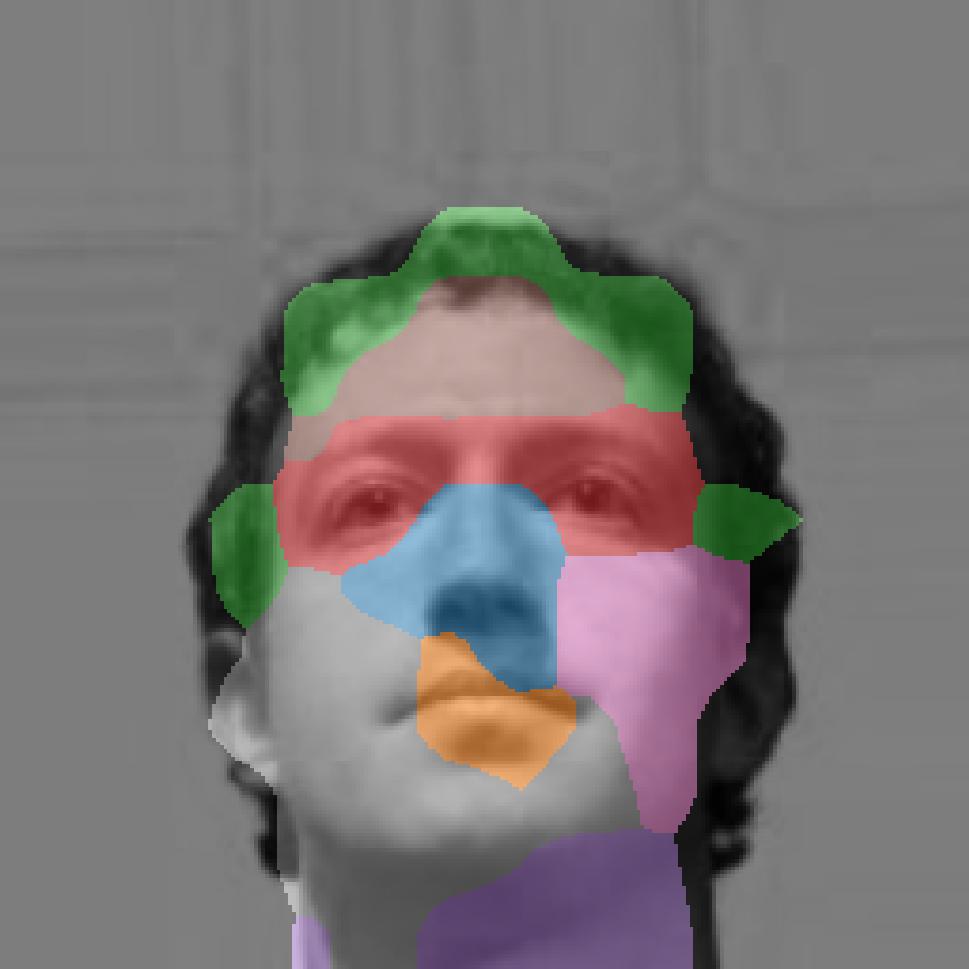} & \includegraphics[width=0.088\linewidth]{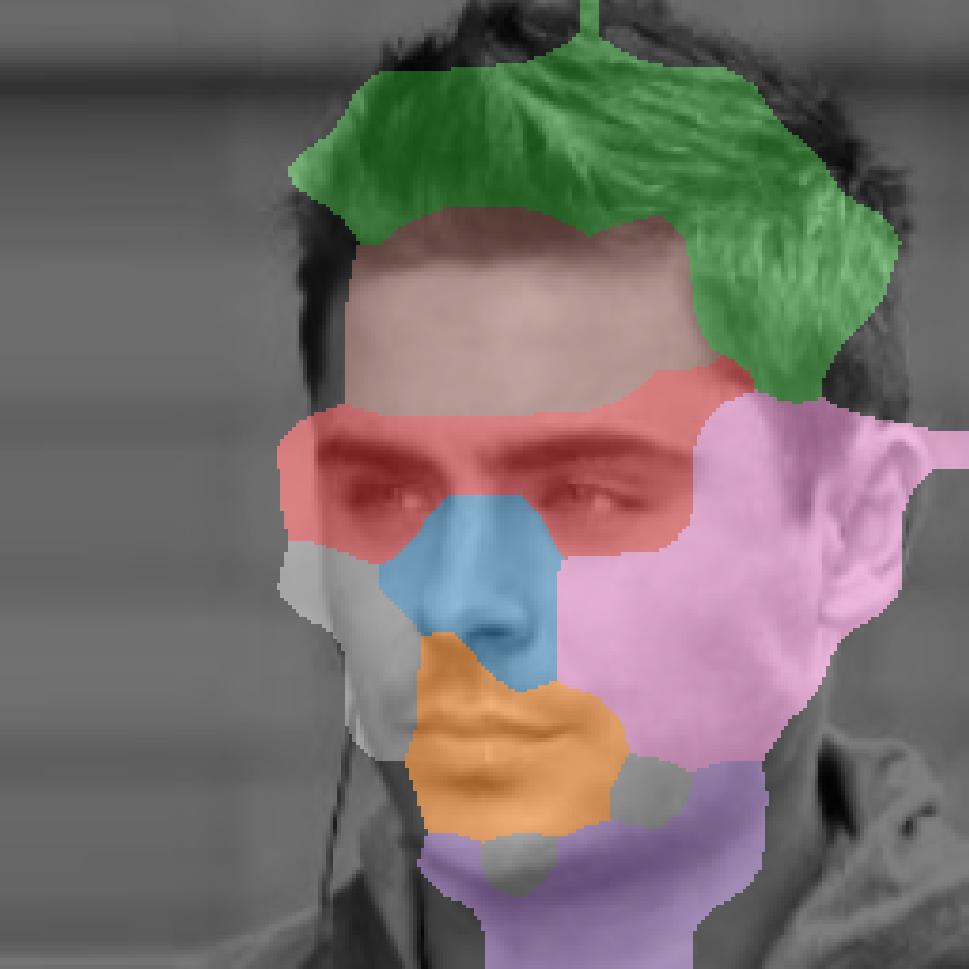} \\
  \bottomrule
  \end{tabular}
  \caption{Additional qualitative results on CelebA.}
  \label{tab:celeba}
\end{figure}

\begin{figure}[t]
  \centering
  \setlength{\tabcolsep}{0pt}
  \renewcommand{\arraystretch}{0.5}
  \begin{tabular}{@{}r@{\hspace{2pt}}c@{\hspace{1pt}}c@{\hspace{1pt}}c@{\hspace{1pt}}c@{\hspace{1pt}}c@{\hspace{1pt}}c@{\hspace{1pt}}c@{\hspace{1pt}}c@{\hspace{1pt}}c@{\hspace{1pt}}c@{\hspace{1pt}}@{}}
  \toprule
    \rotatebox{90}{\makebox[1cm][c]{\scriptsize Images}} & \includegraphics[width=0.088\linewidth]{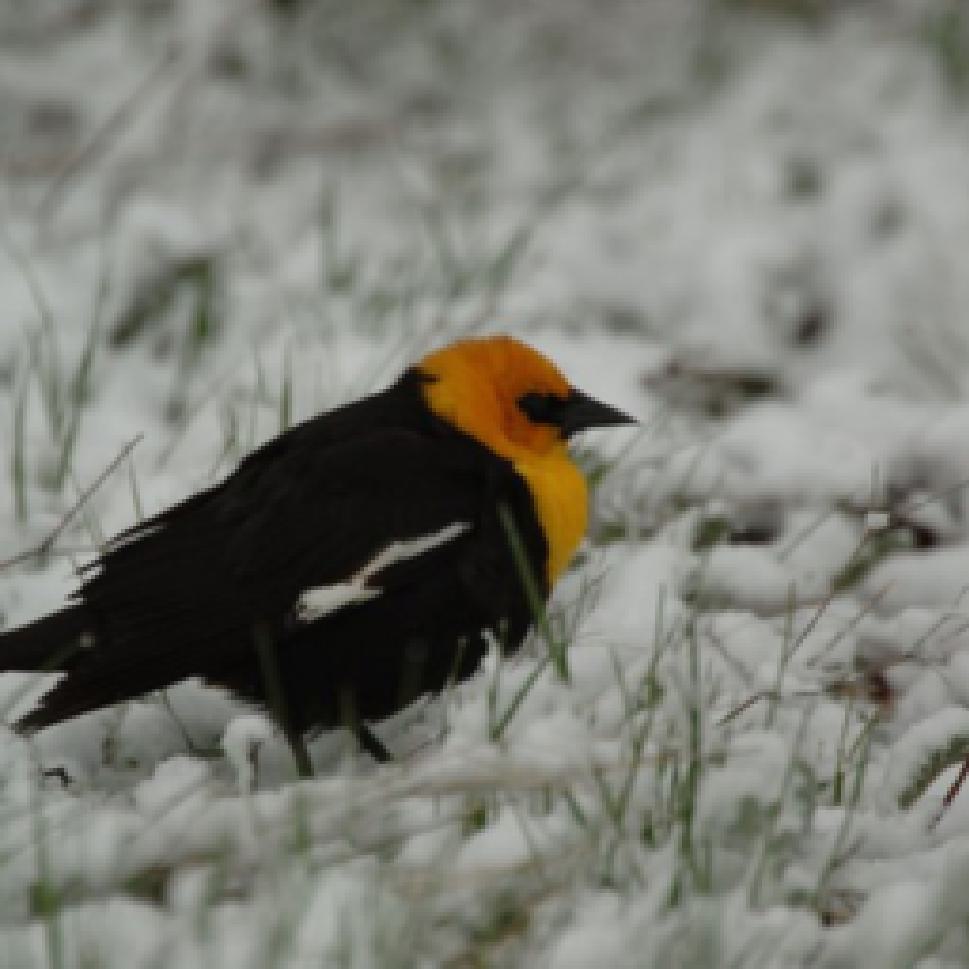} & \includegraphics[width=0.088\linewidth]{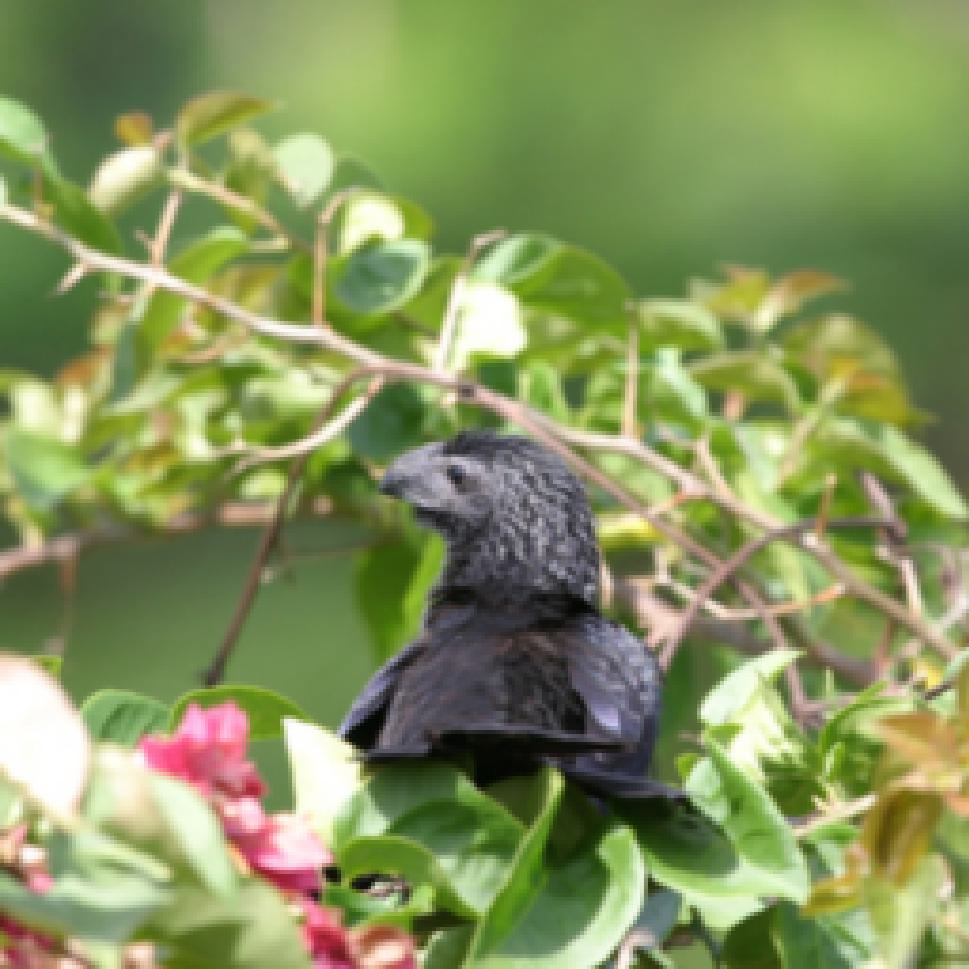} & \includegraphics[width=0.088\linewidth]{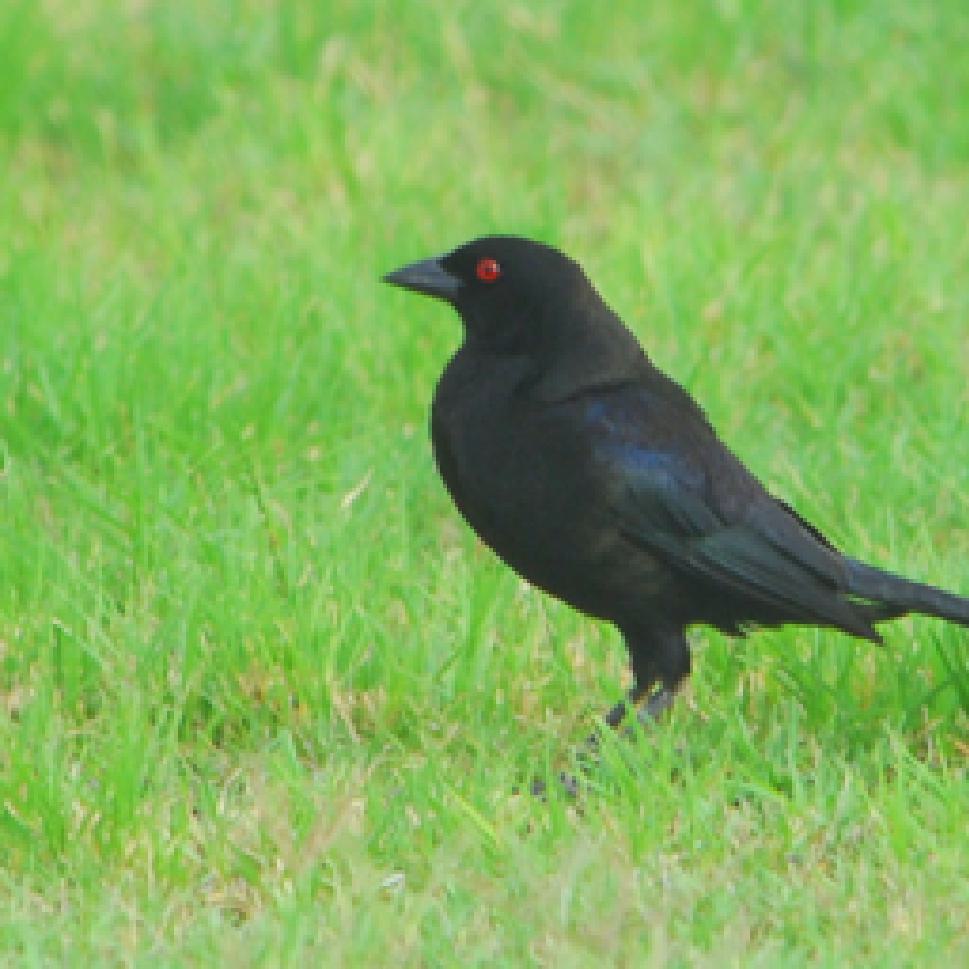} & \includegraphics[width=0.088\linewidth]{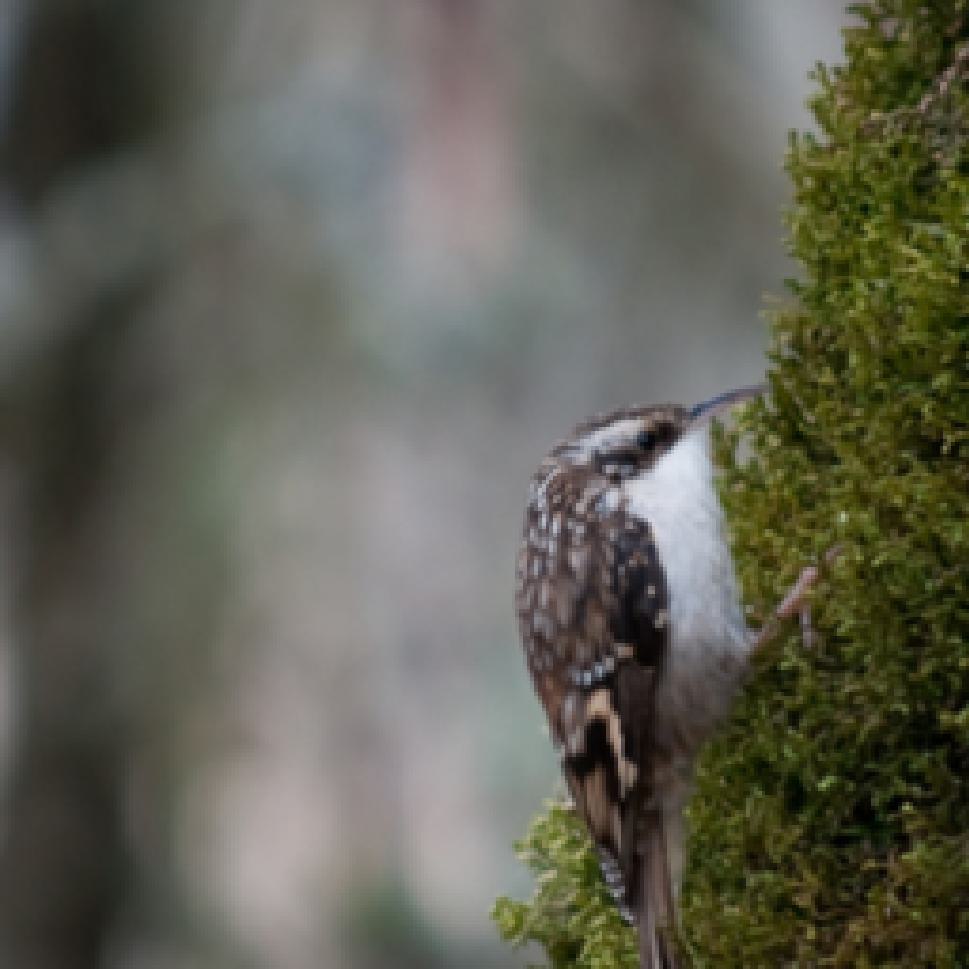} & \includegraphics[width=0.088\linewidth]{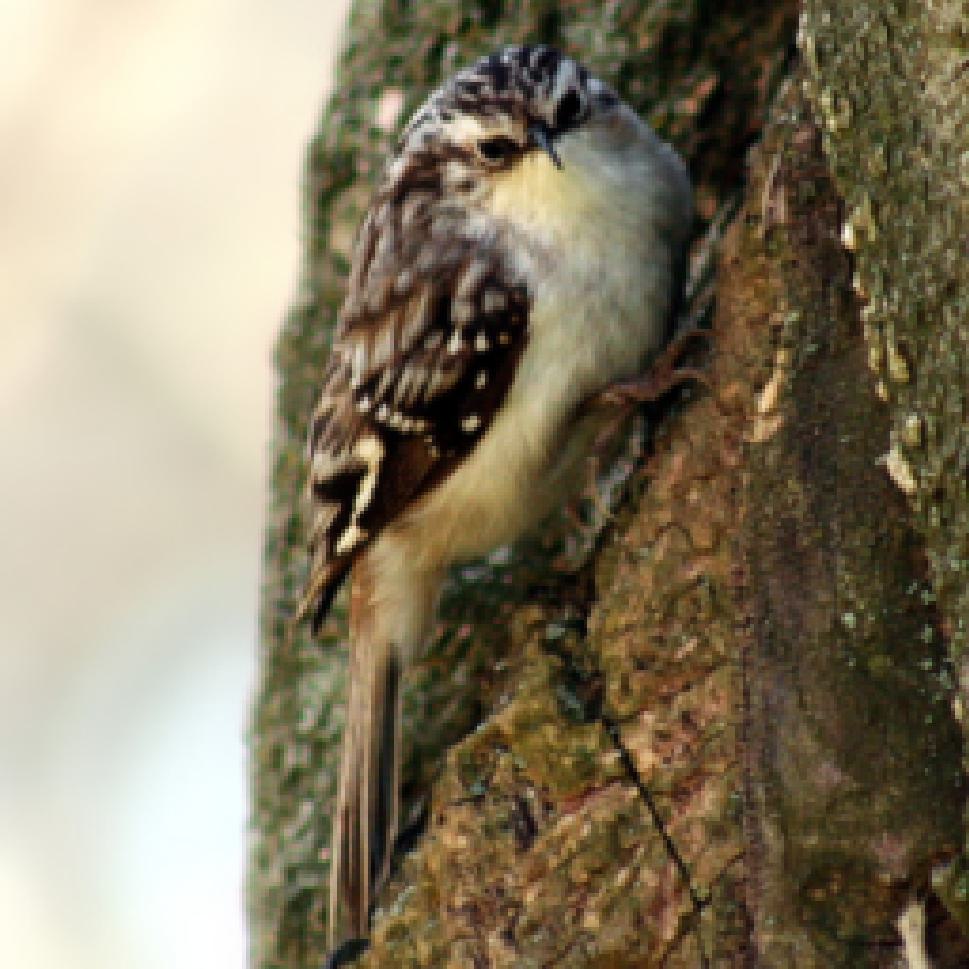} & \includegraphics[width=0.088\linewidth]{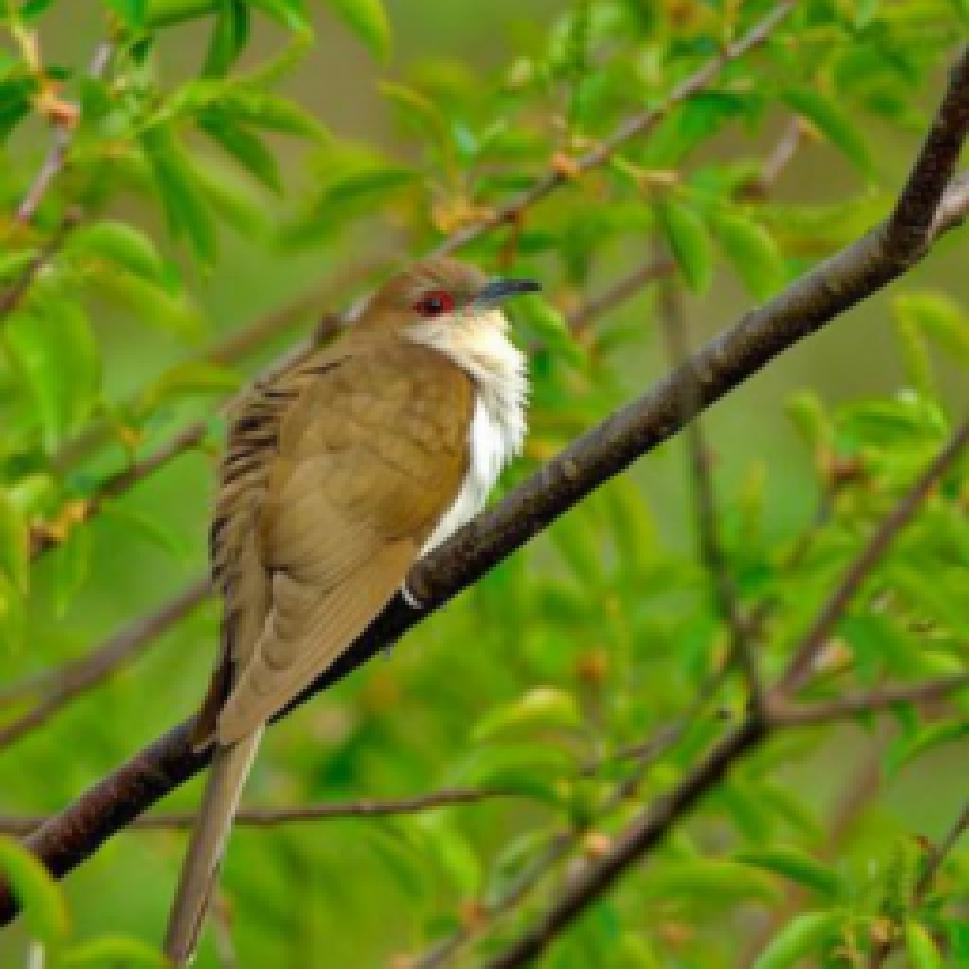} & \includegraphics[width=0.088\linewidth]{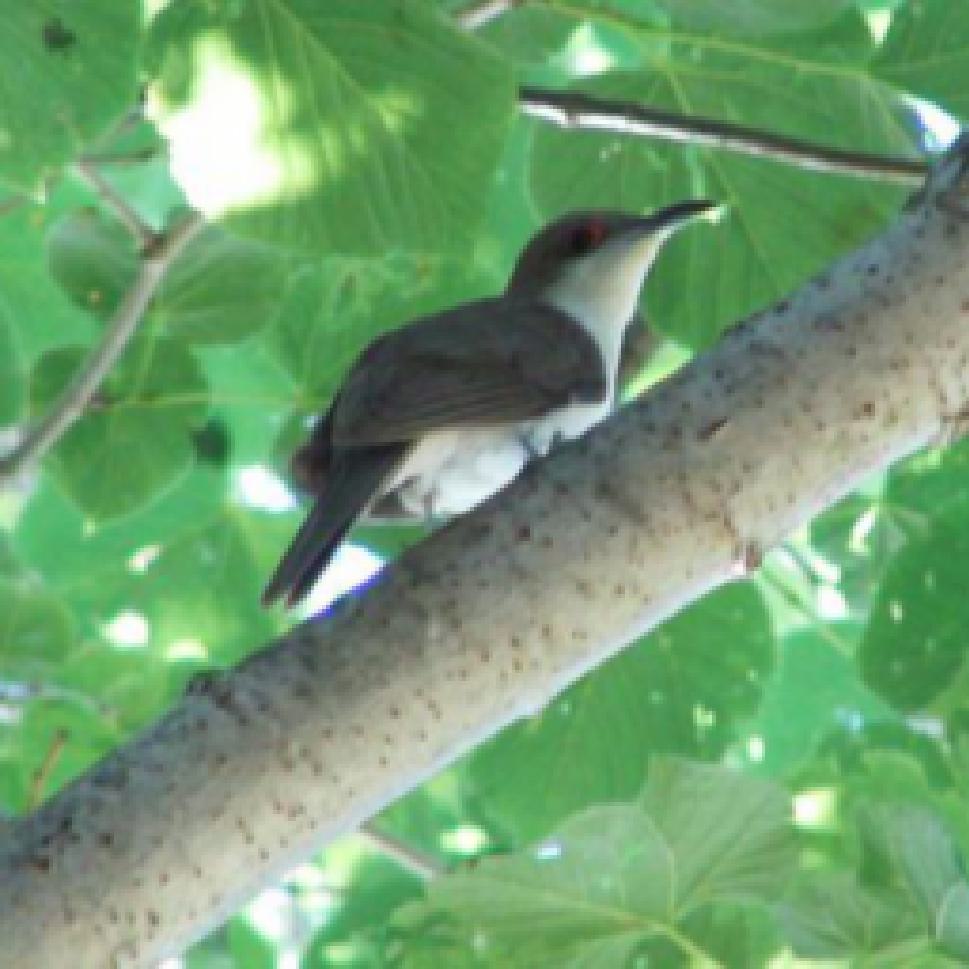} & \includegraphics[width=0.088\linewidth]{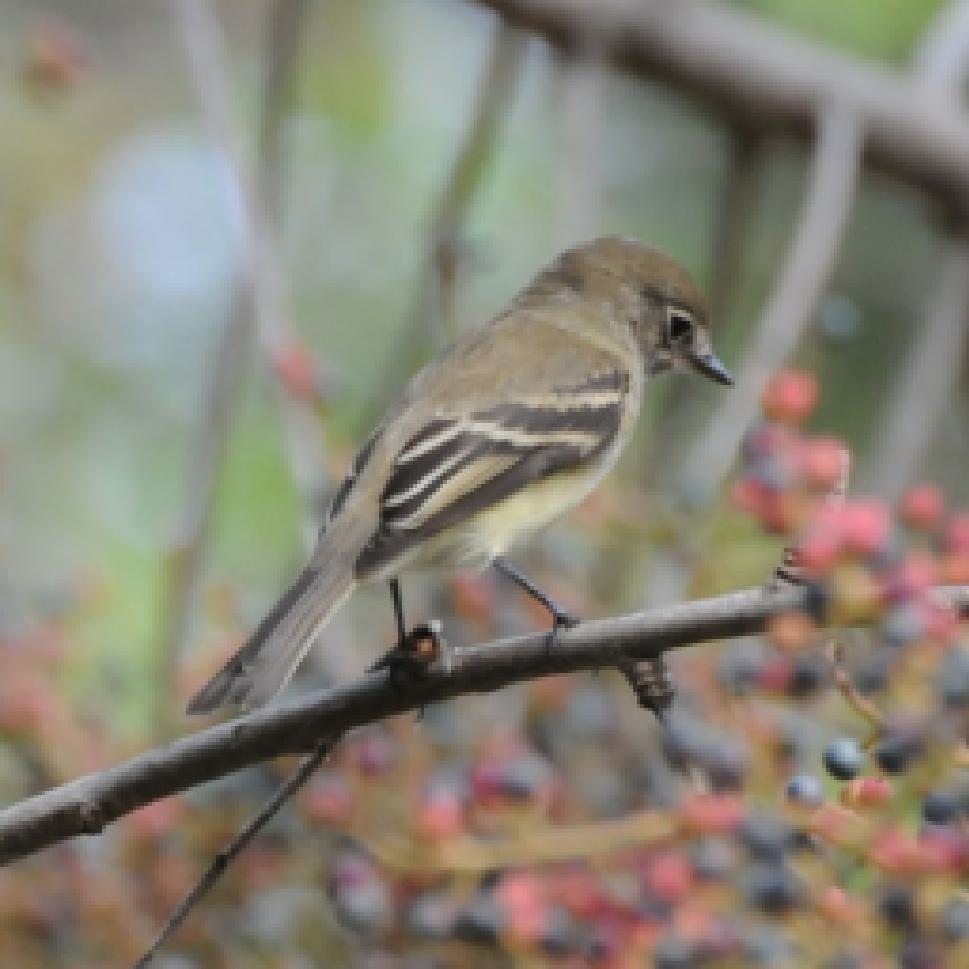} & \includegraphics[width=0.088\linewidth]{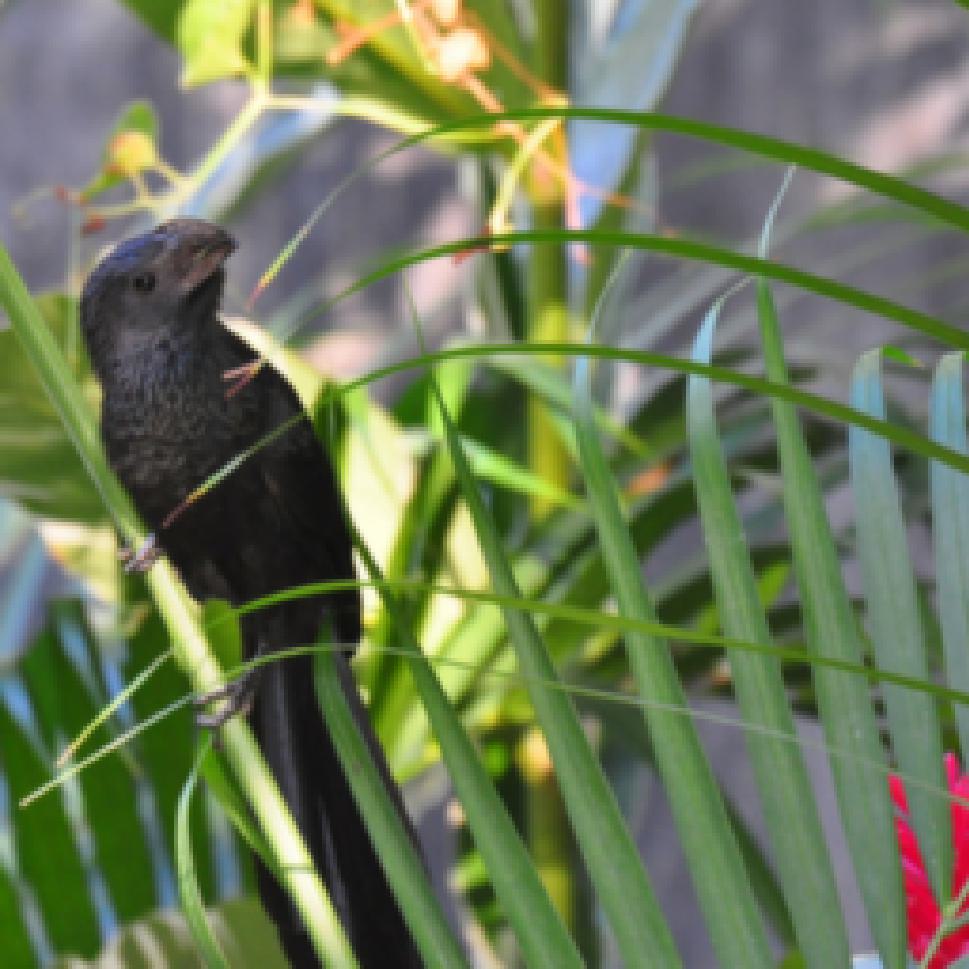} & \includegraphics[width=0.088\linewidth]{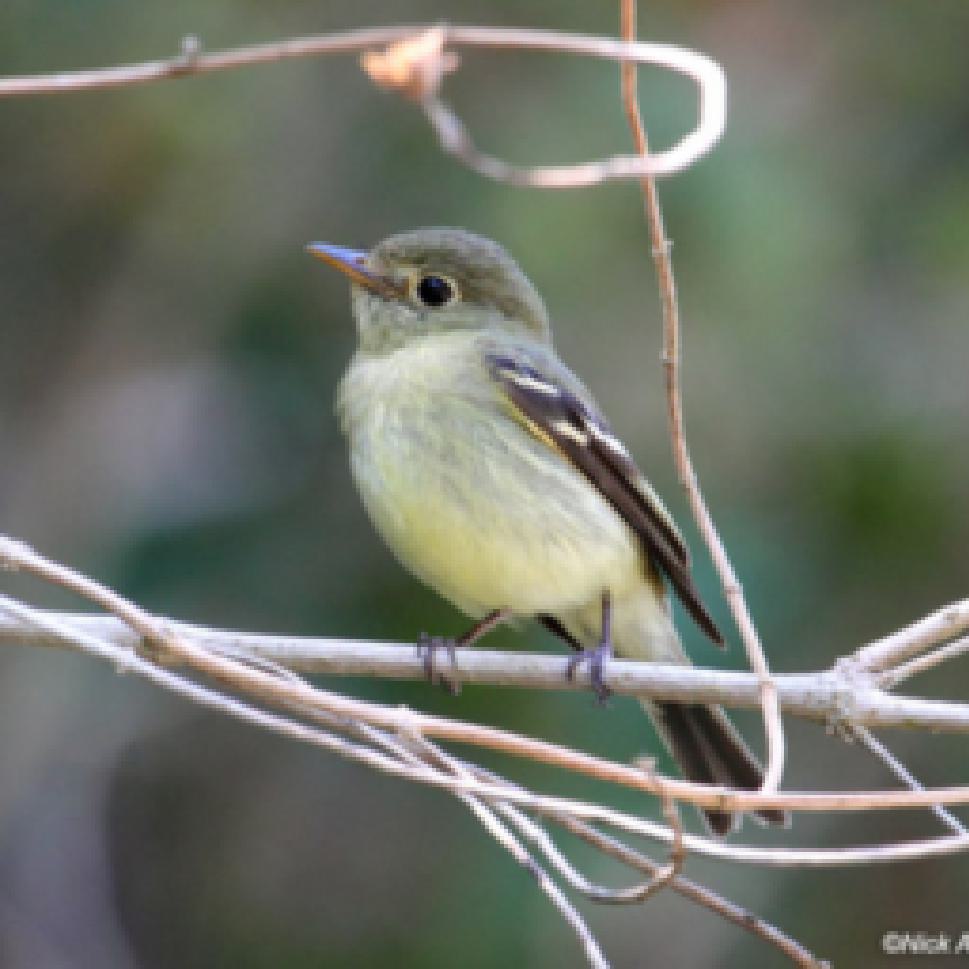} \\
    \rotatebox{90}{\makebox[1cm][c]{\scriptsize 1-stage}} & \includegraphics[width=0.088\linewidth]{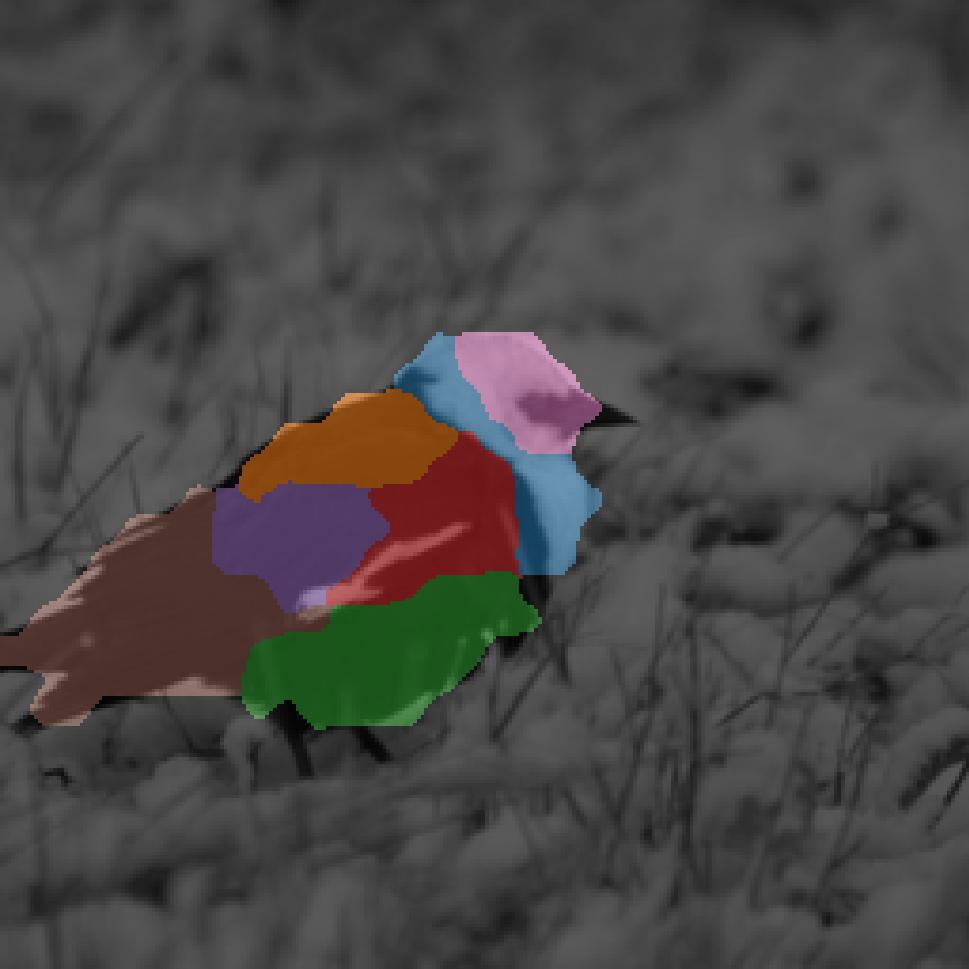} & \includegraphics[width=0.088\linewidth]{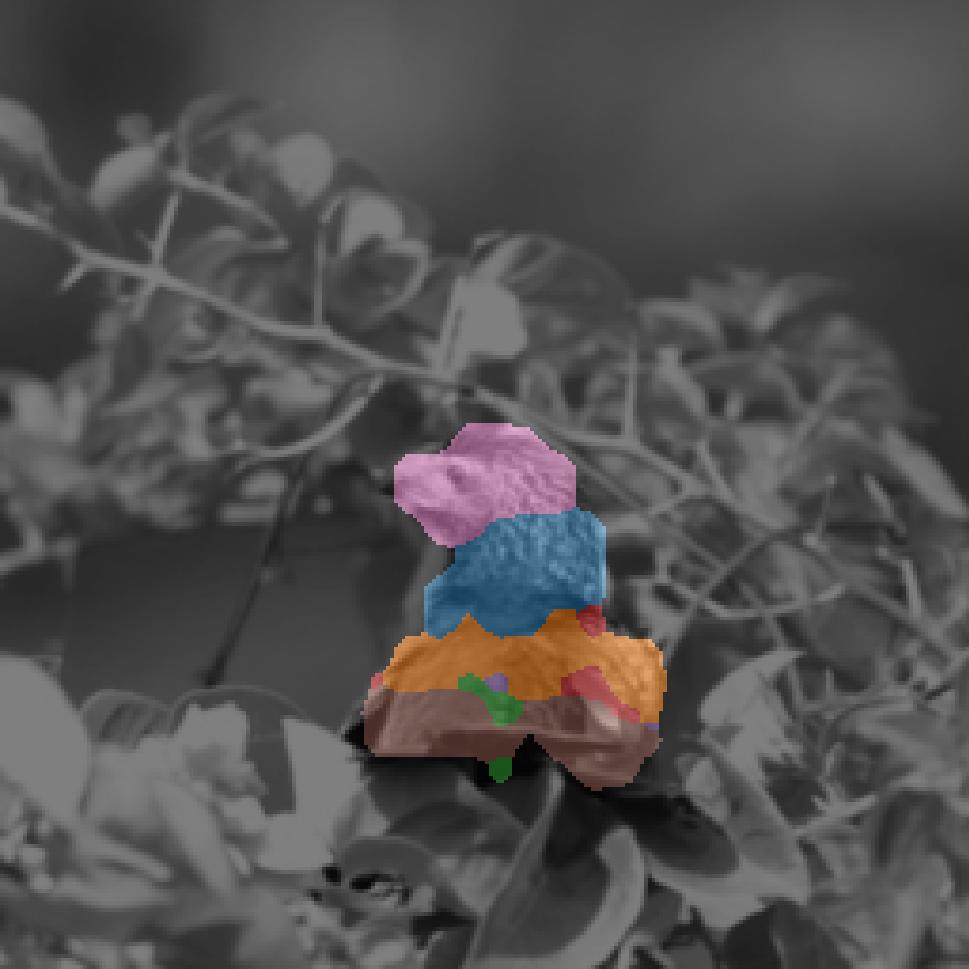} & \includegraphics[width=0.088\linewidth]{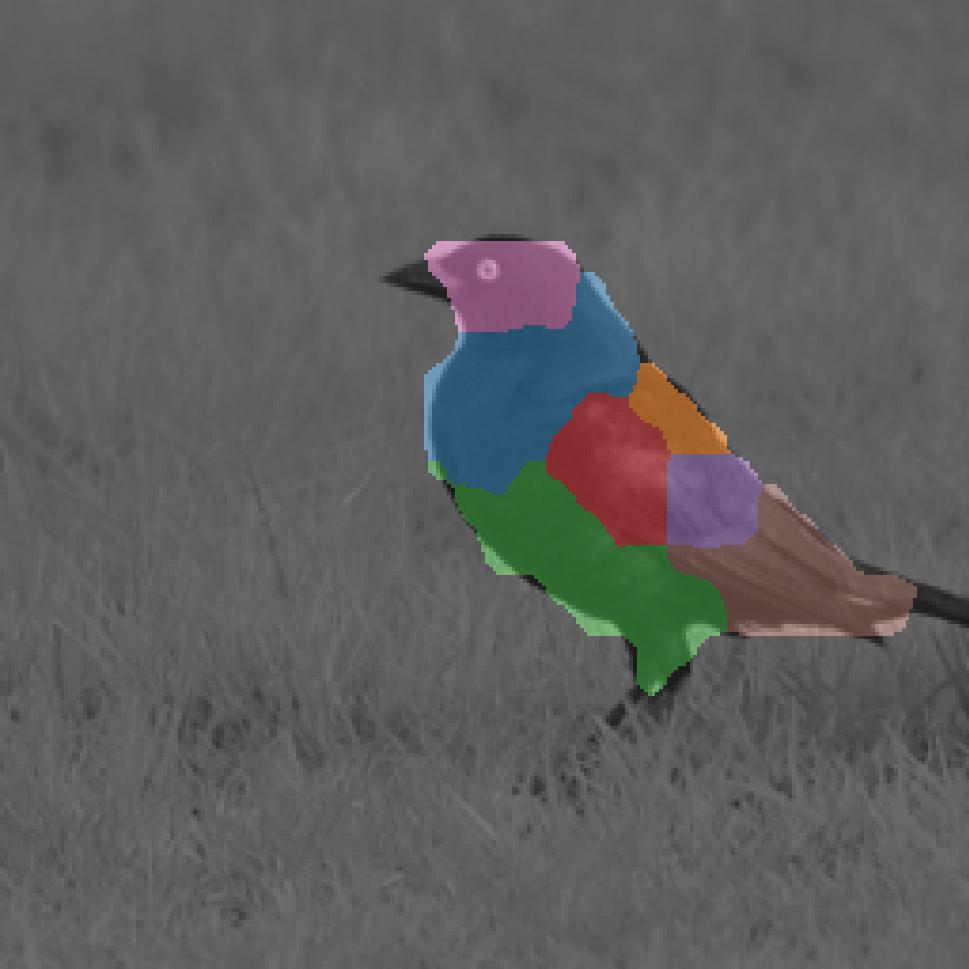} & \includegraphics[width=0.088\linewidth]{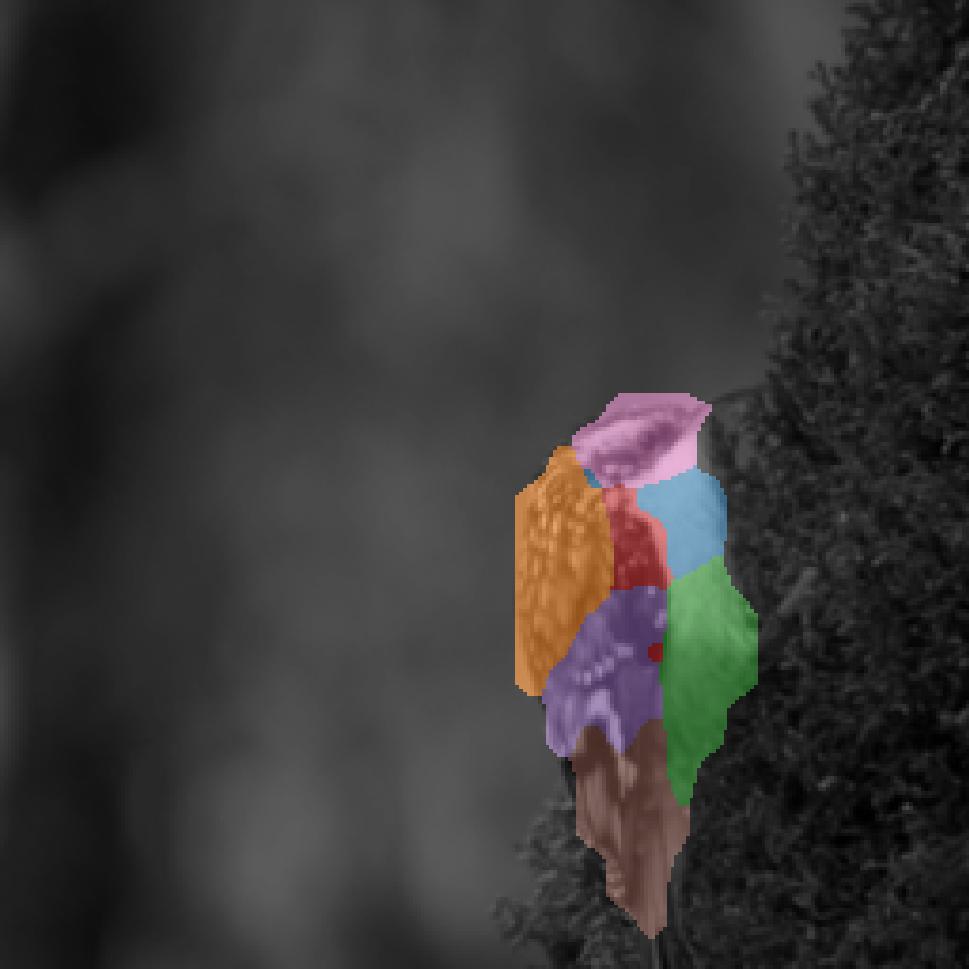} & \includegraphics[width=0.088\linewidth]{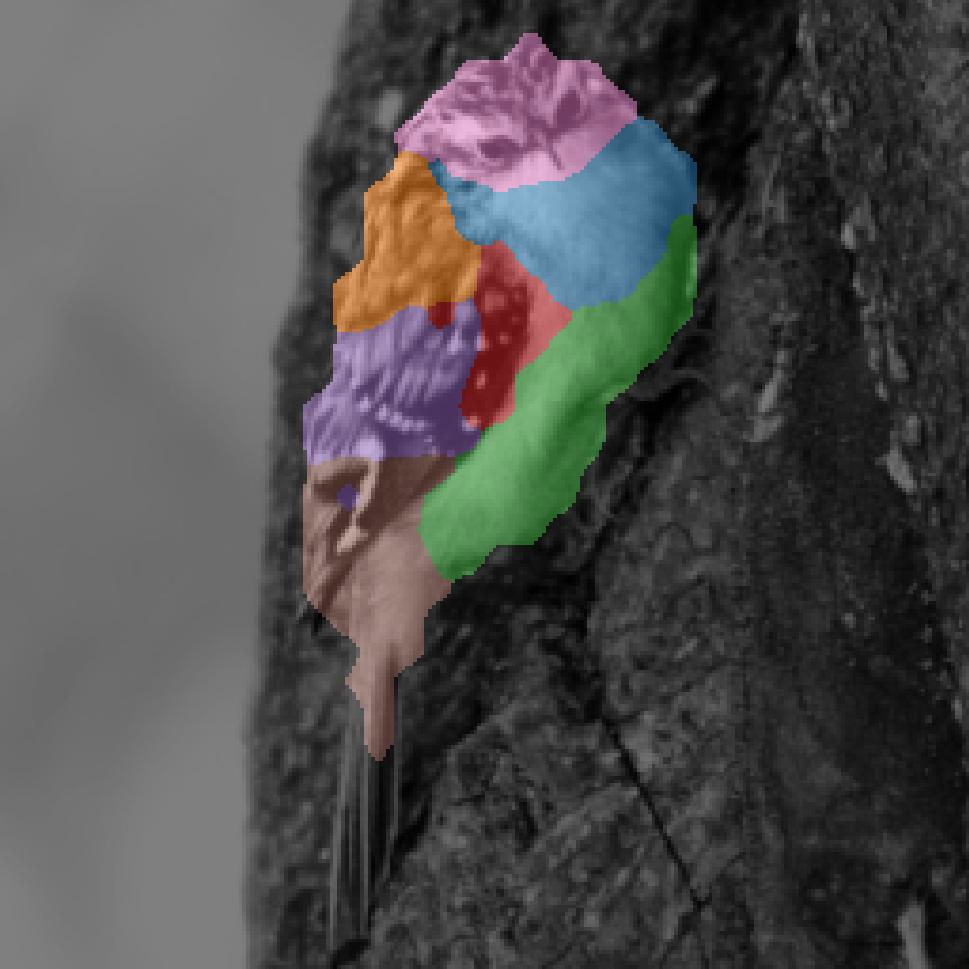} & \includegraphics[width=0.088\linewidth]{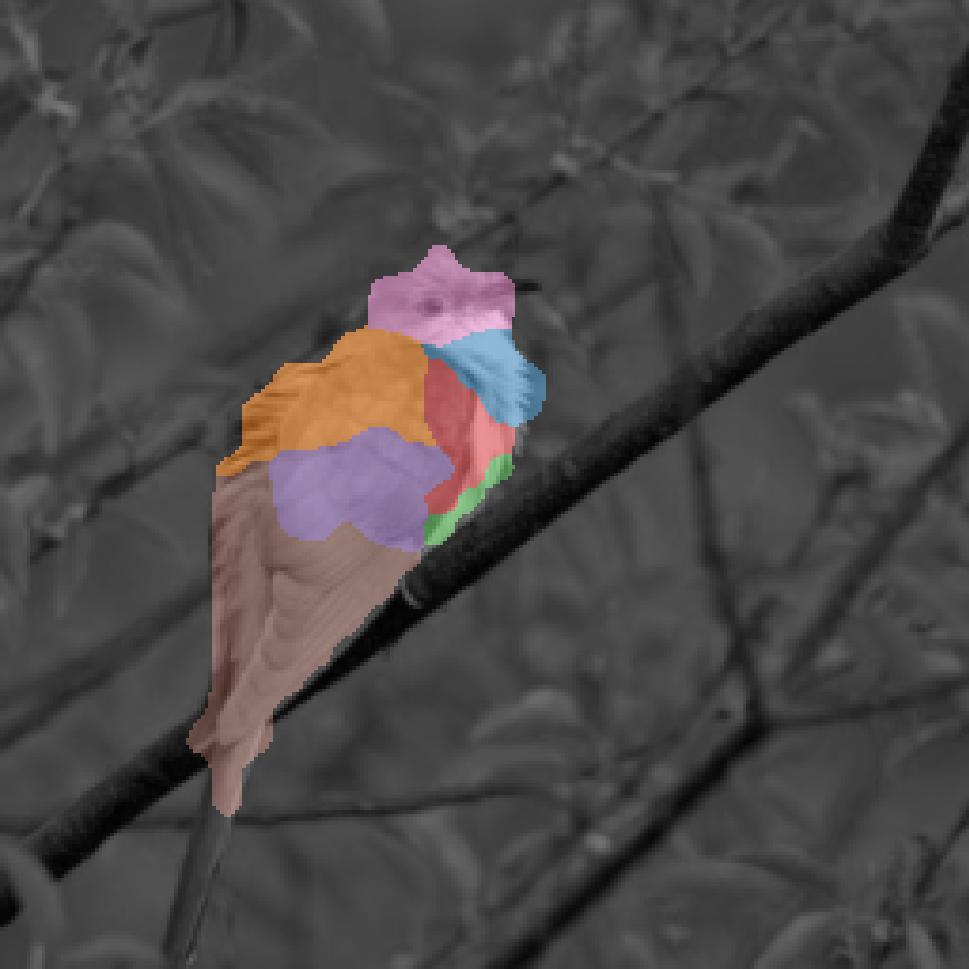} & \includegraphics[width=0.088\linewidth]{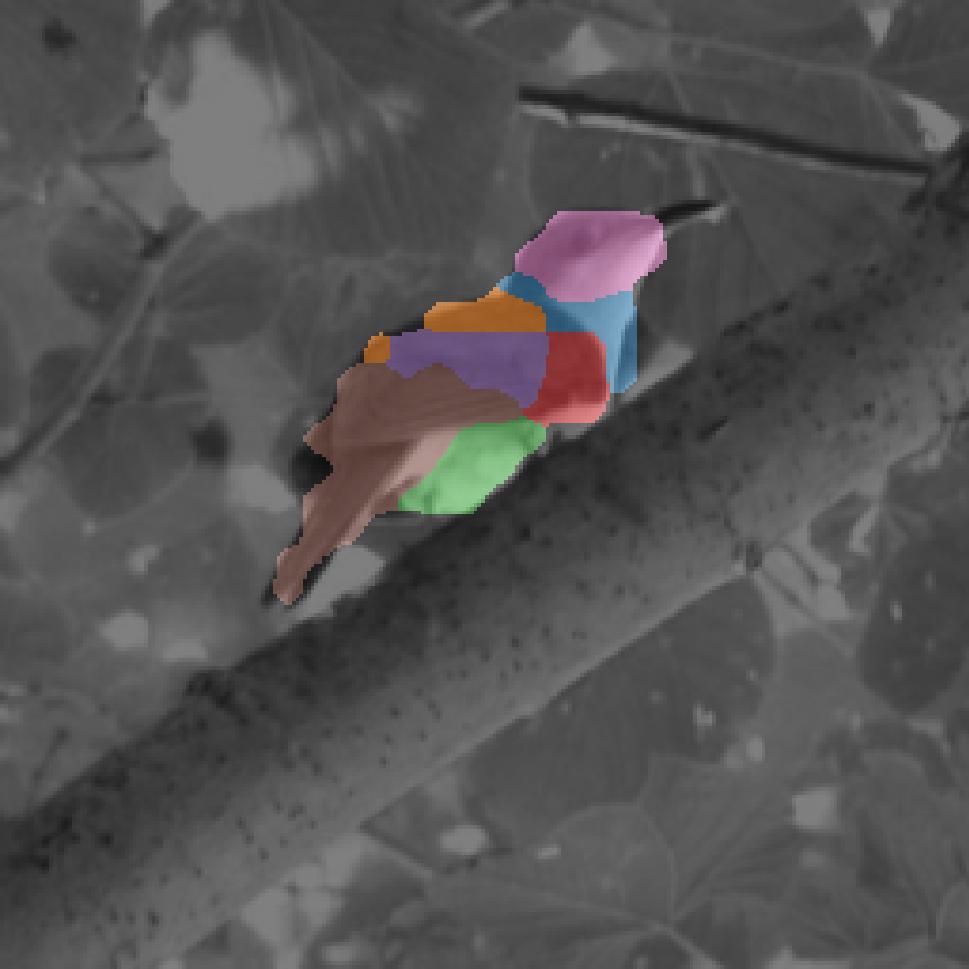} & \includegraphics[width=0.088\linewidth]{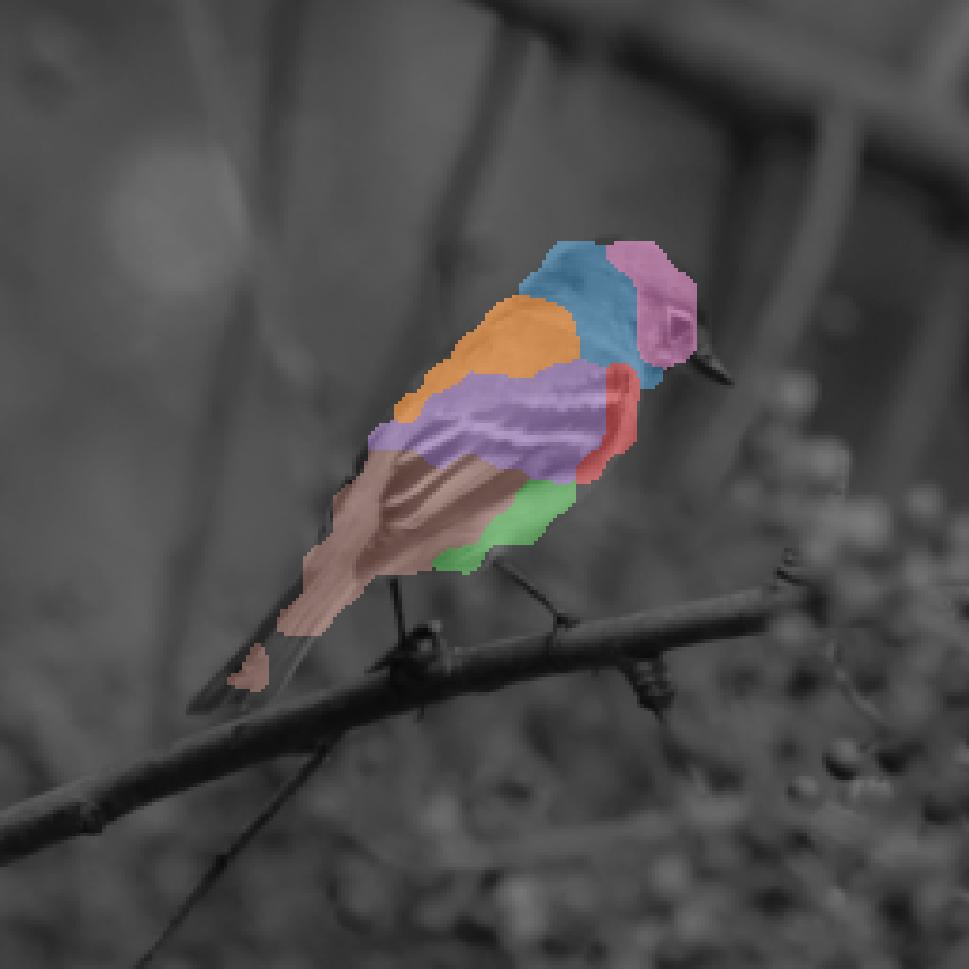} & \includegraphics[width=0.088\linewidth]{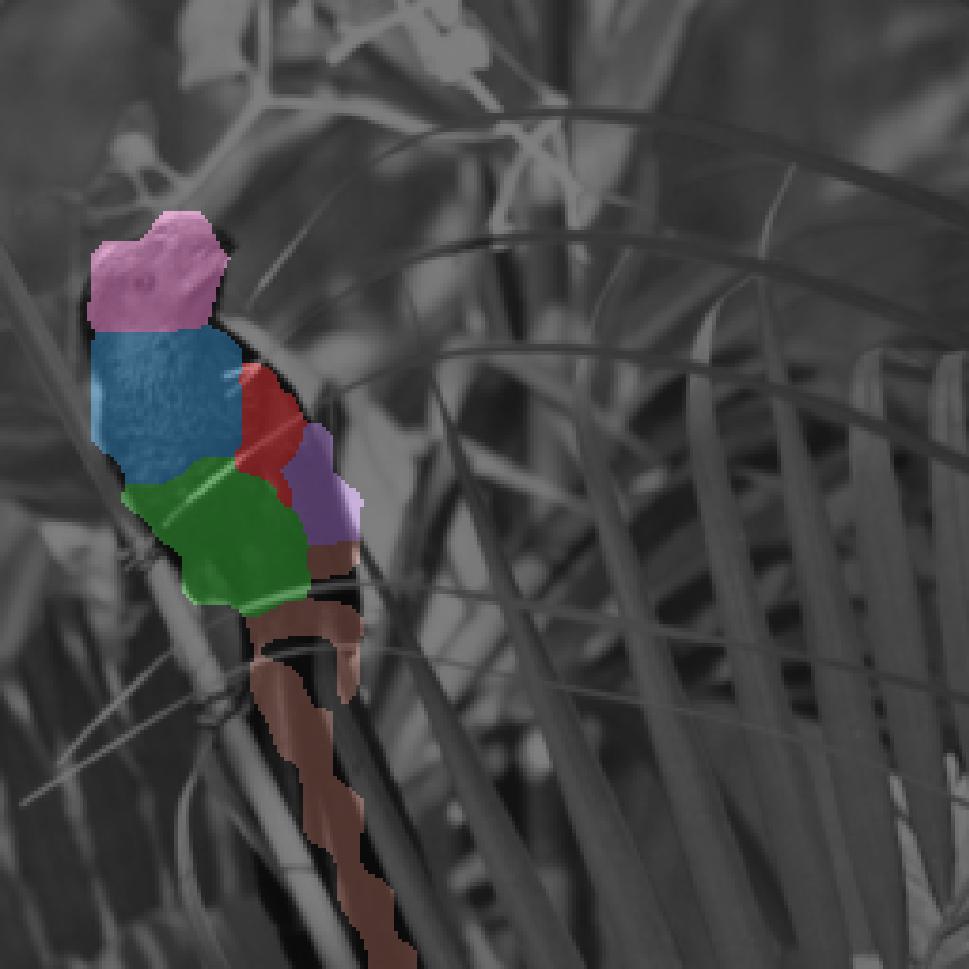} & \includegraphics[width=0.088\linewidth]{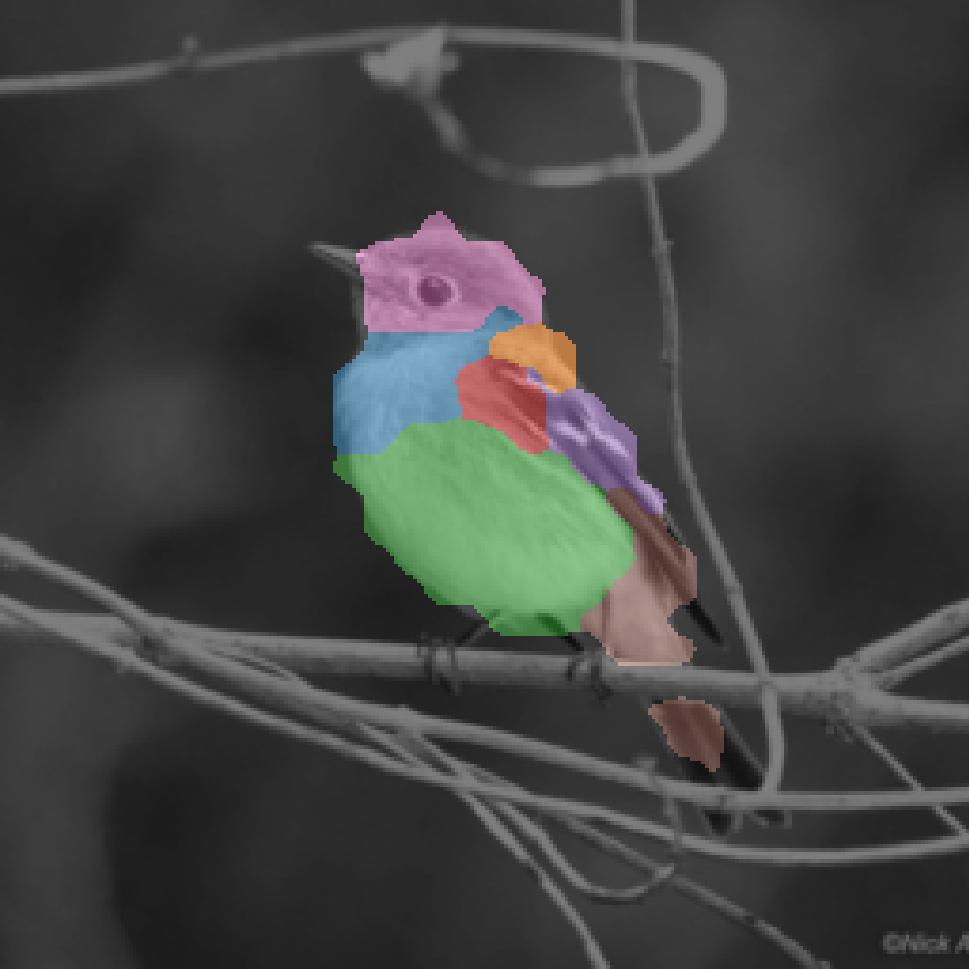} \\

    \rotatebox{90}{\makebox[1cm][c]{\scriptsize Soft}} & \includegraphics[width=0.088\linewidth]{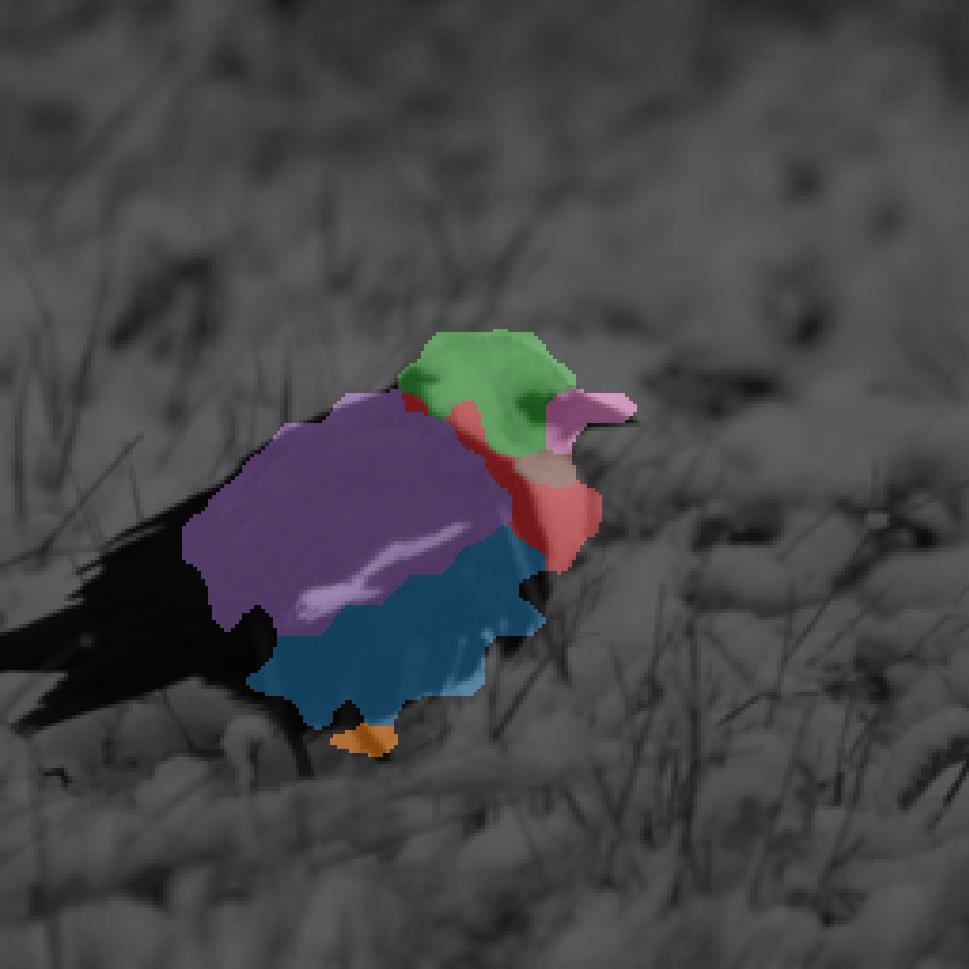} & \includegraphics[width=0.088\linewidth]{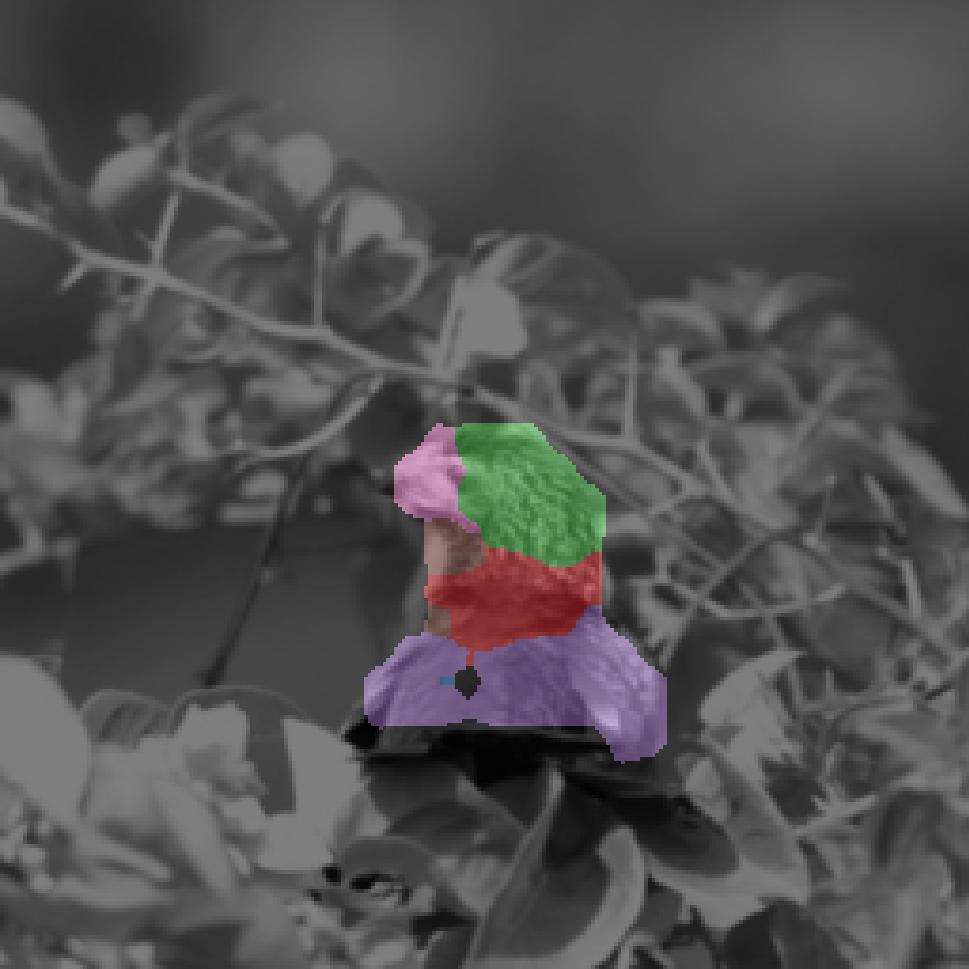} & \includegraphics[width=0.088\linewidth]{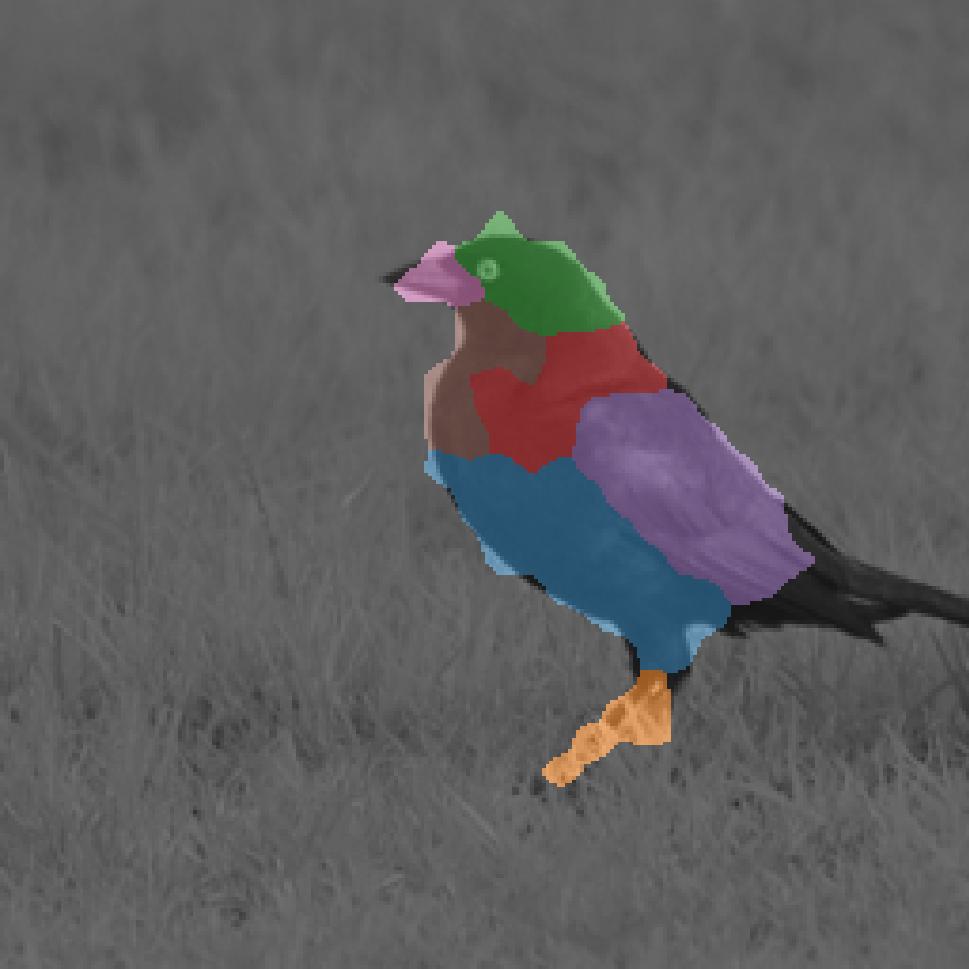} & \includegraphics[width=0.088\linewidth]{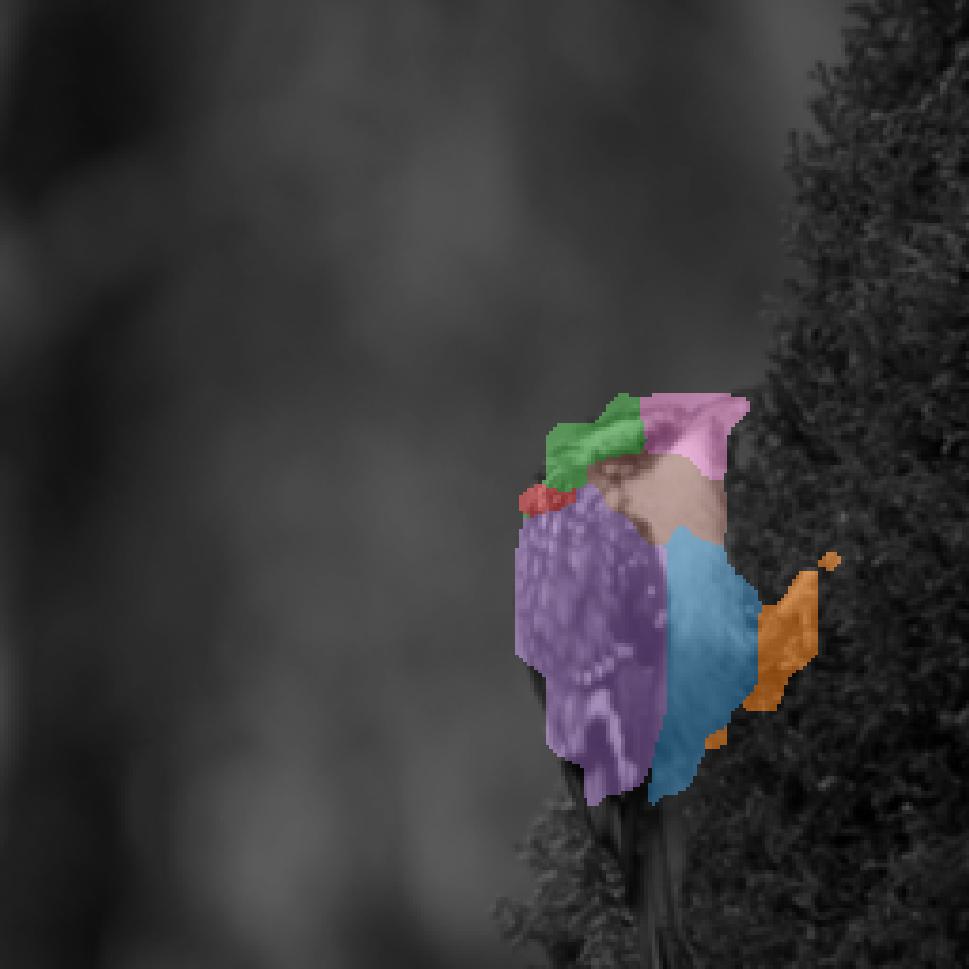} & \includegraphics[width=0.088\linewidth]{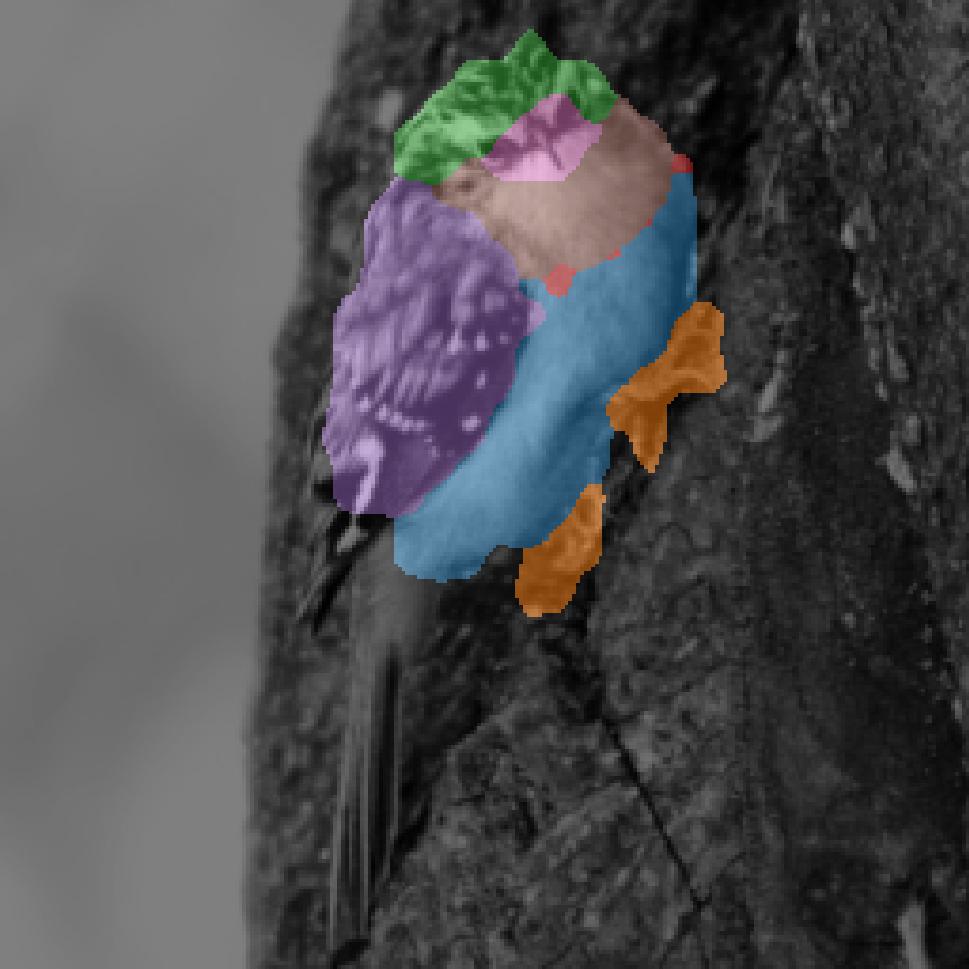} & \includegraphics[width=0.088\linewidth]{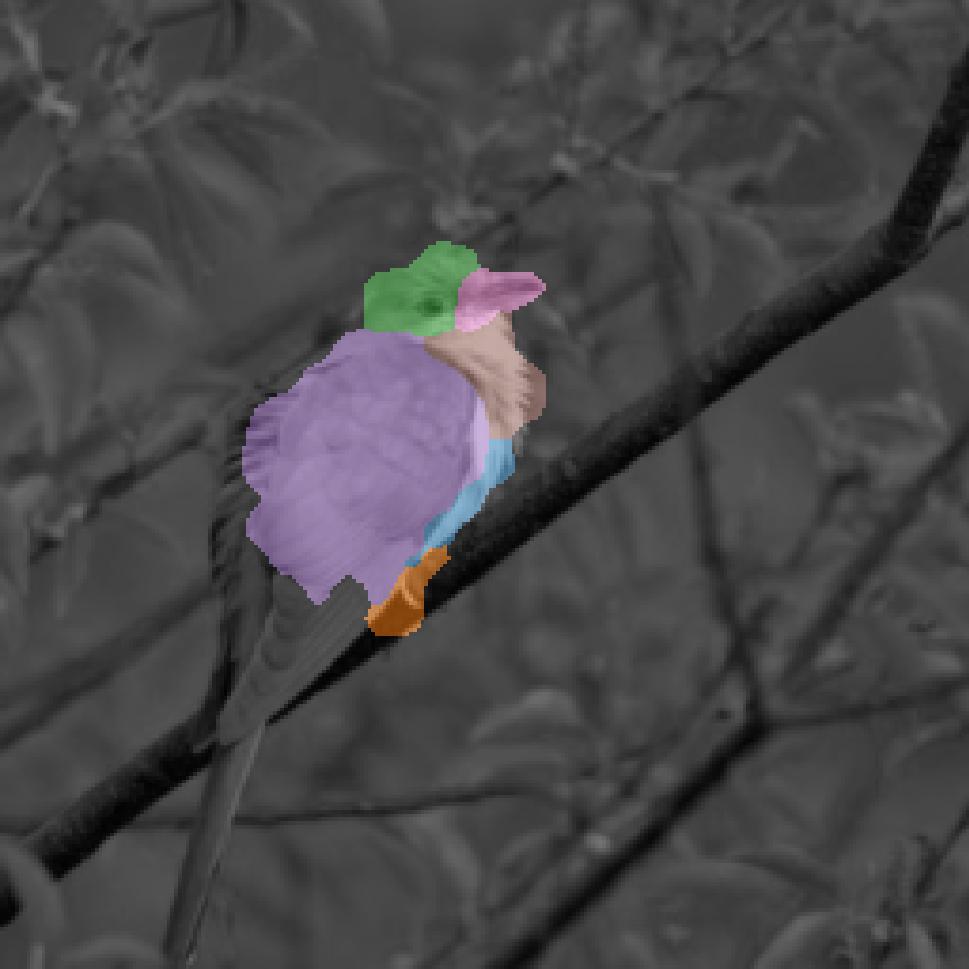} & \includegraphics[width=0.088\linewidth]{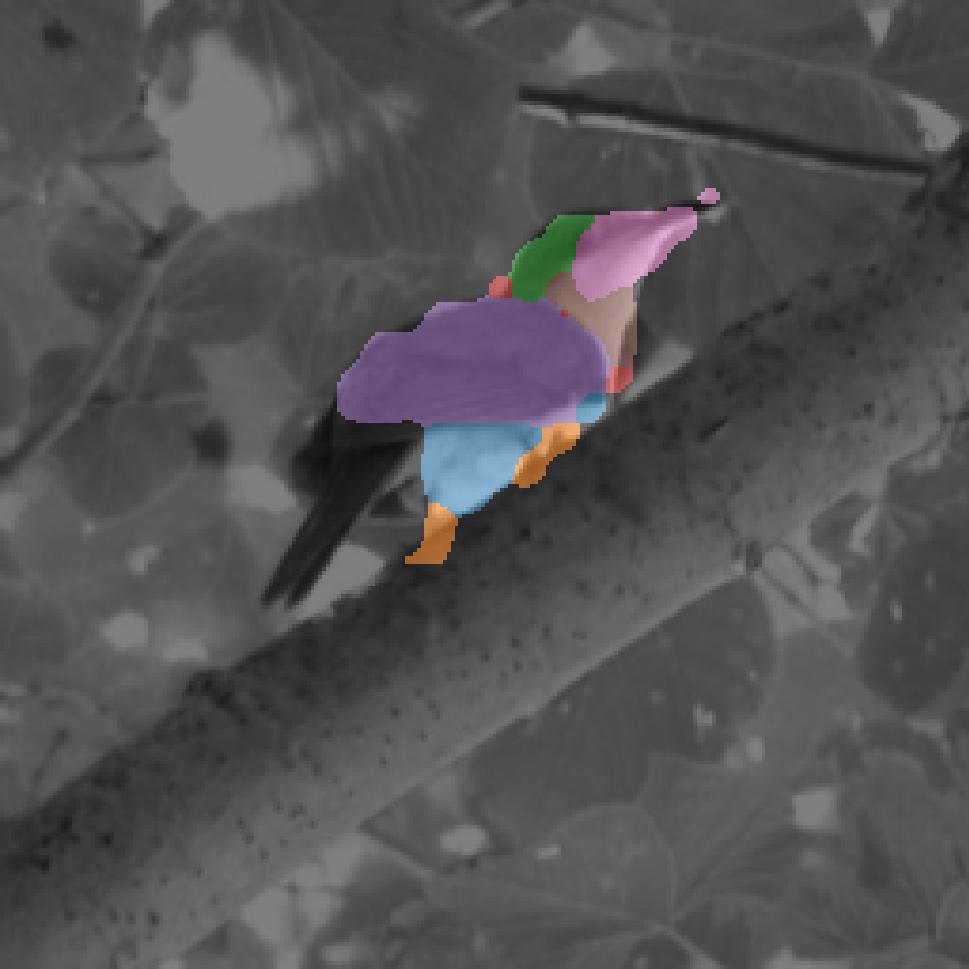} & \includegraphics[width=0.088\linewidth]{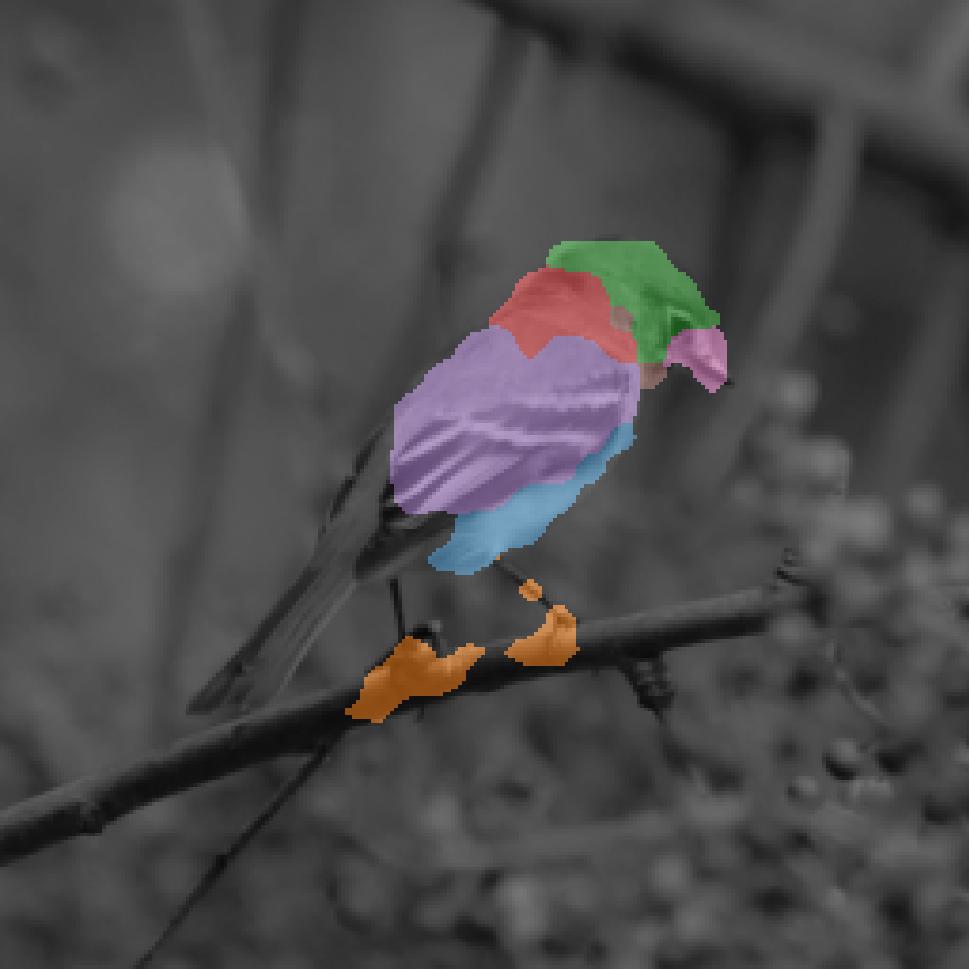} & \includegraphics[width=0.088\linewidth]{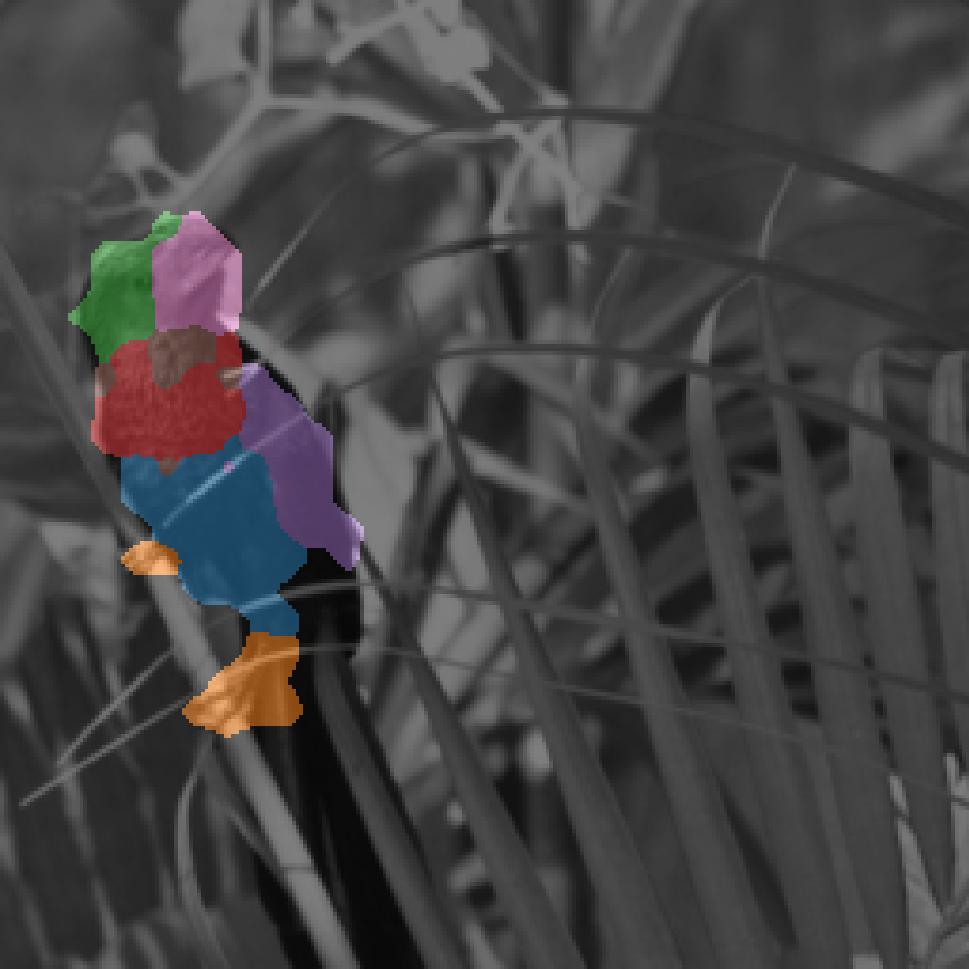} & \includegraphics[width=0.088\linewidth]{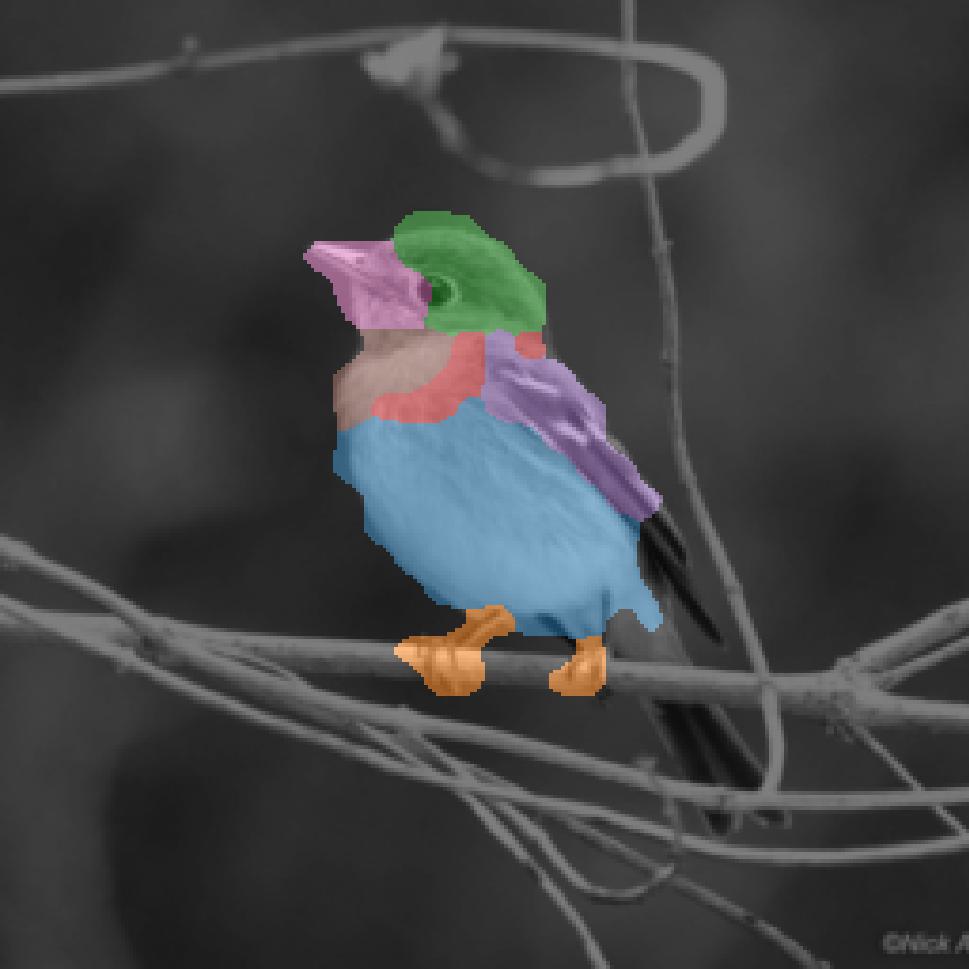} \\
    
    \rotatebox{90}{\makebox[1cm][c]{\scriptsize Hard}} & \includegraphics[width=0.088\linewidth]{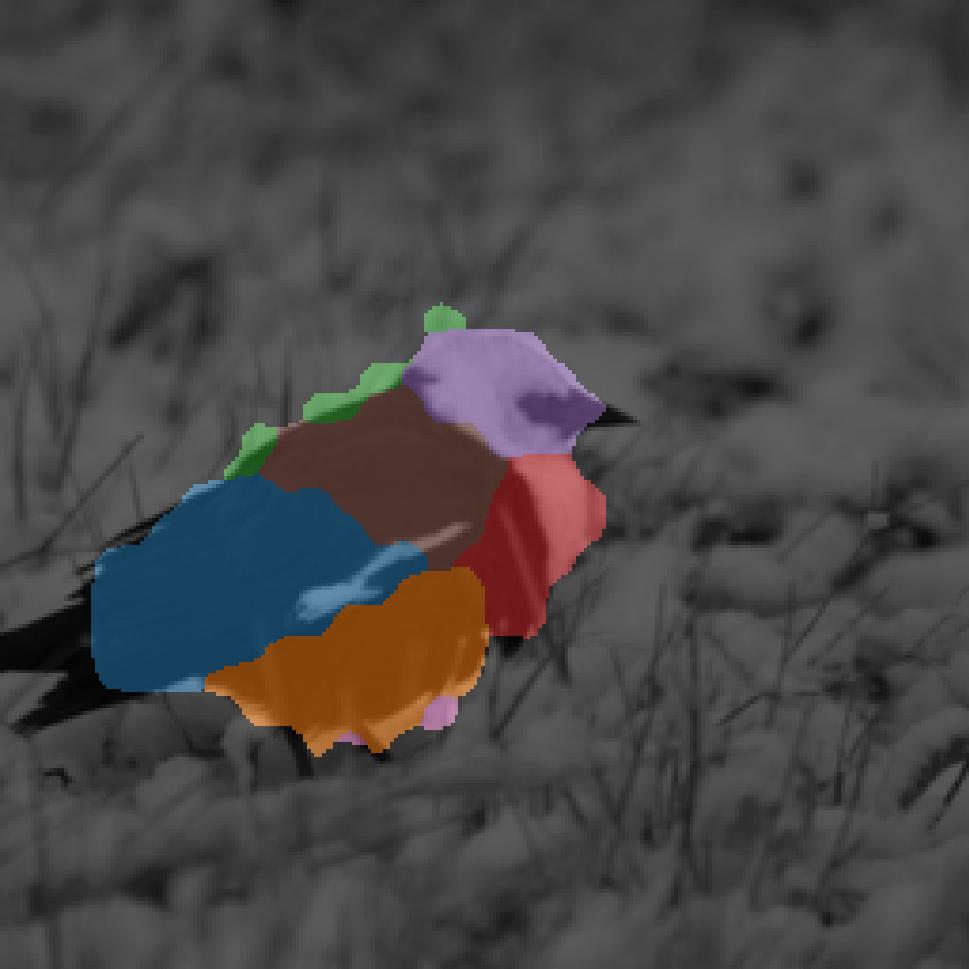} & \includegraphics[width=0.088\linewidth]{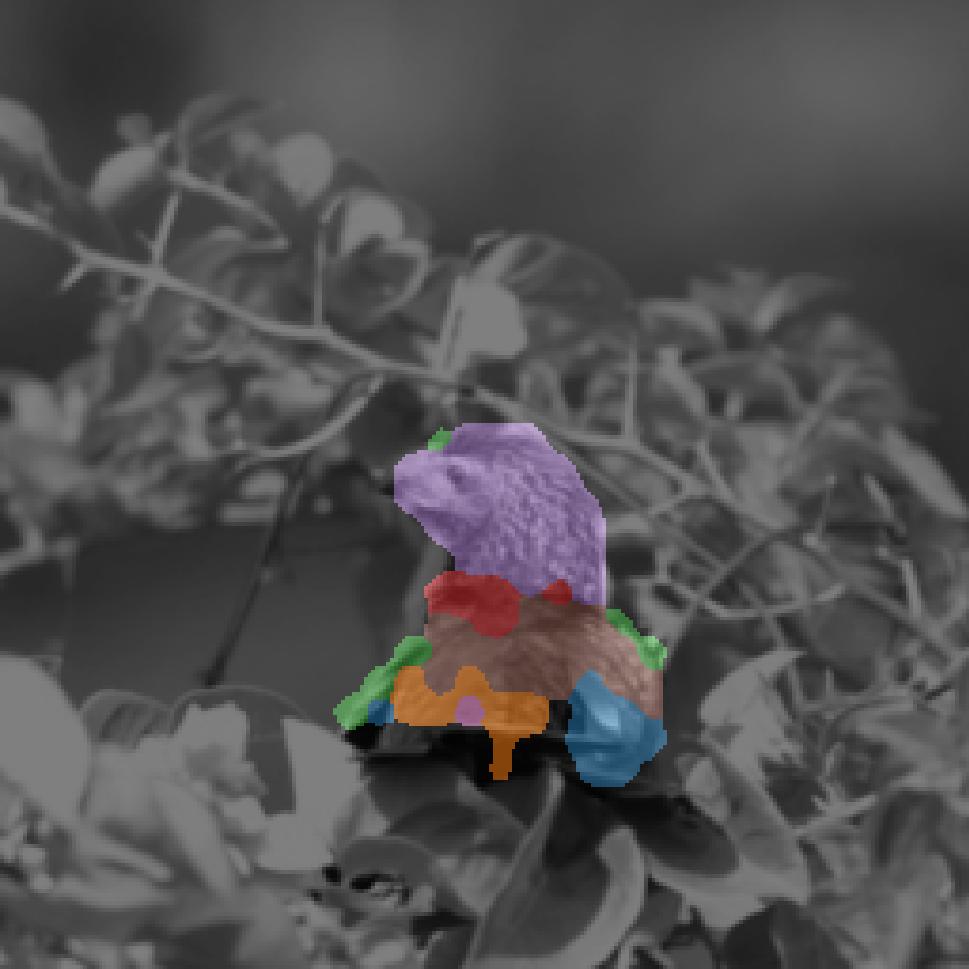} & \includegraphics[width=0.088\linewidth]{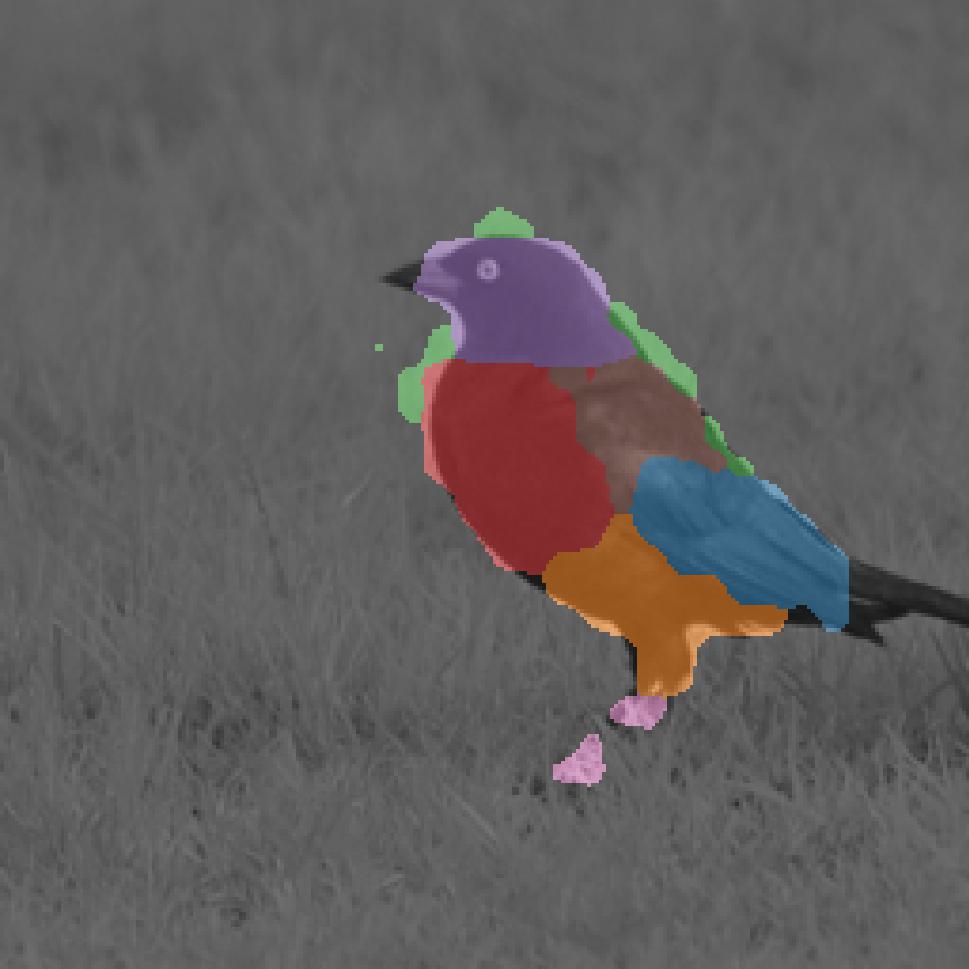} & \includegraphics[width=0.088\linewidth]{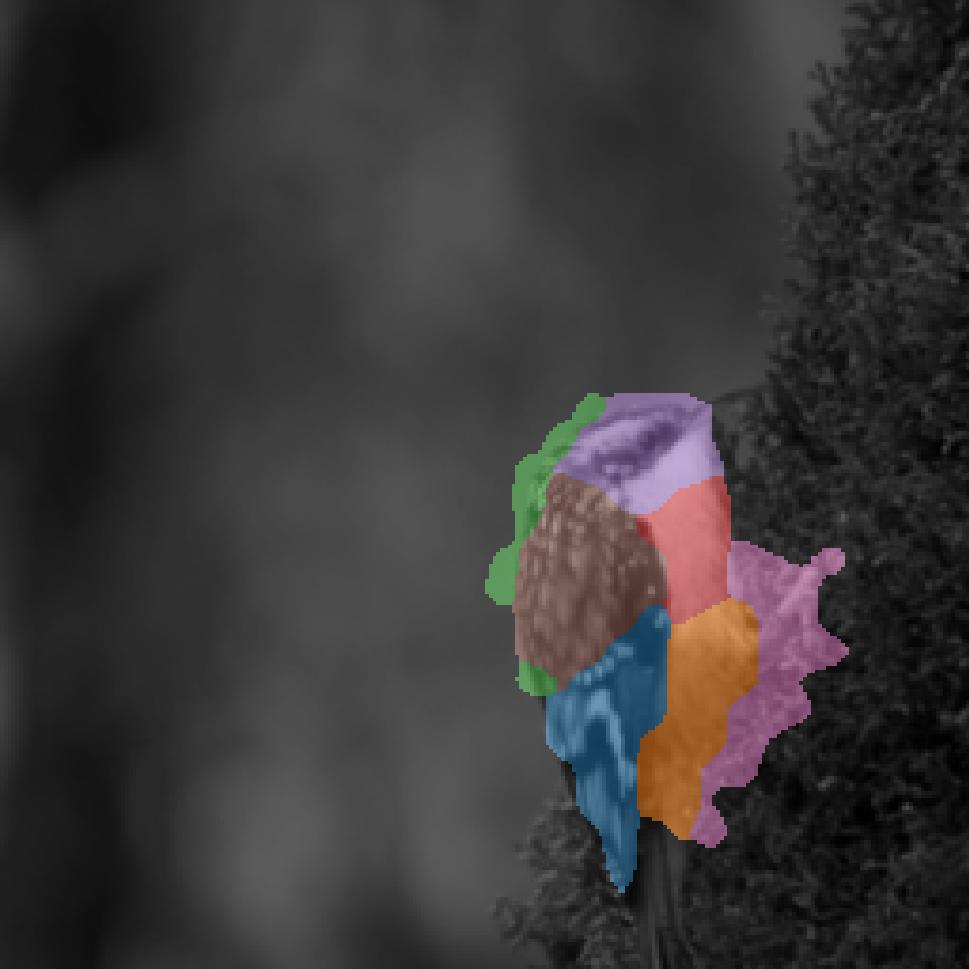} & \includegraphics[width=0.088\linewidth]{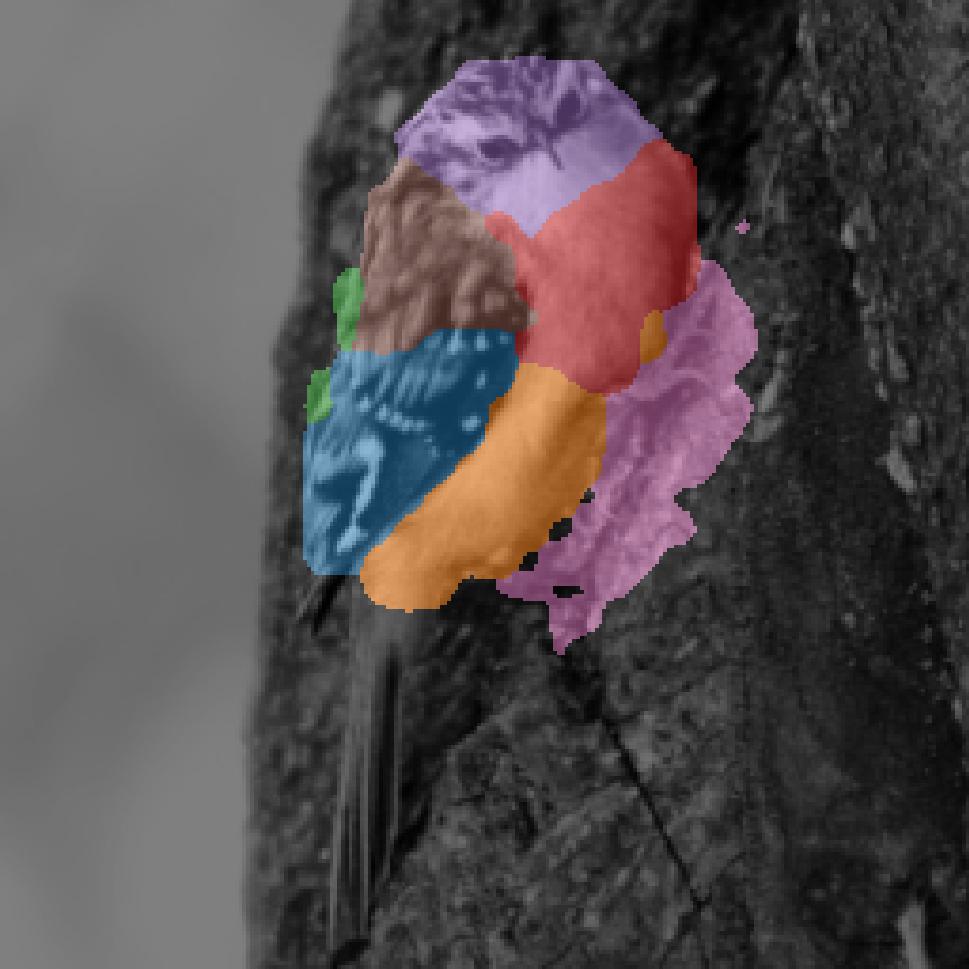} & \includegraphics[width=0.088\linewidth]{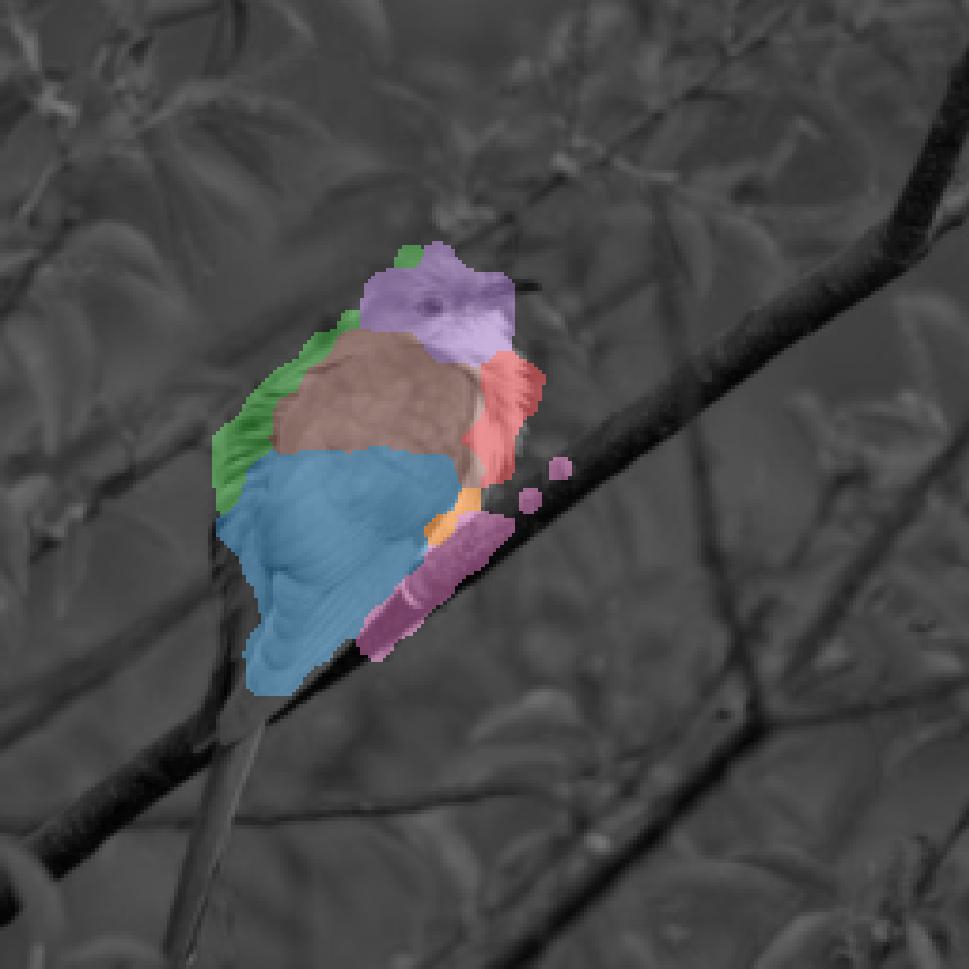} & \includegraphics[width=0.088\linewidth]{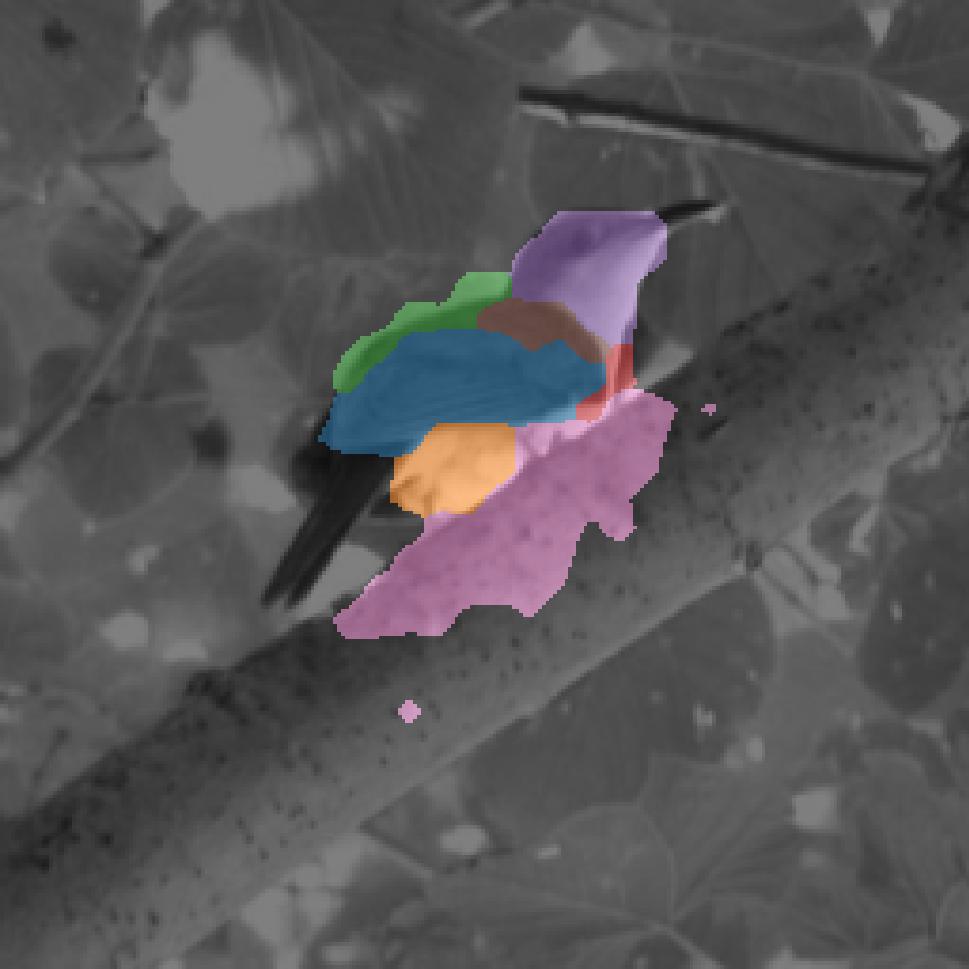} & \includegraphics[width=0.088\linewidth]{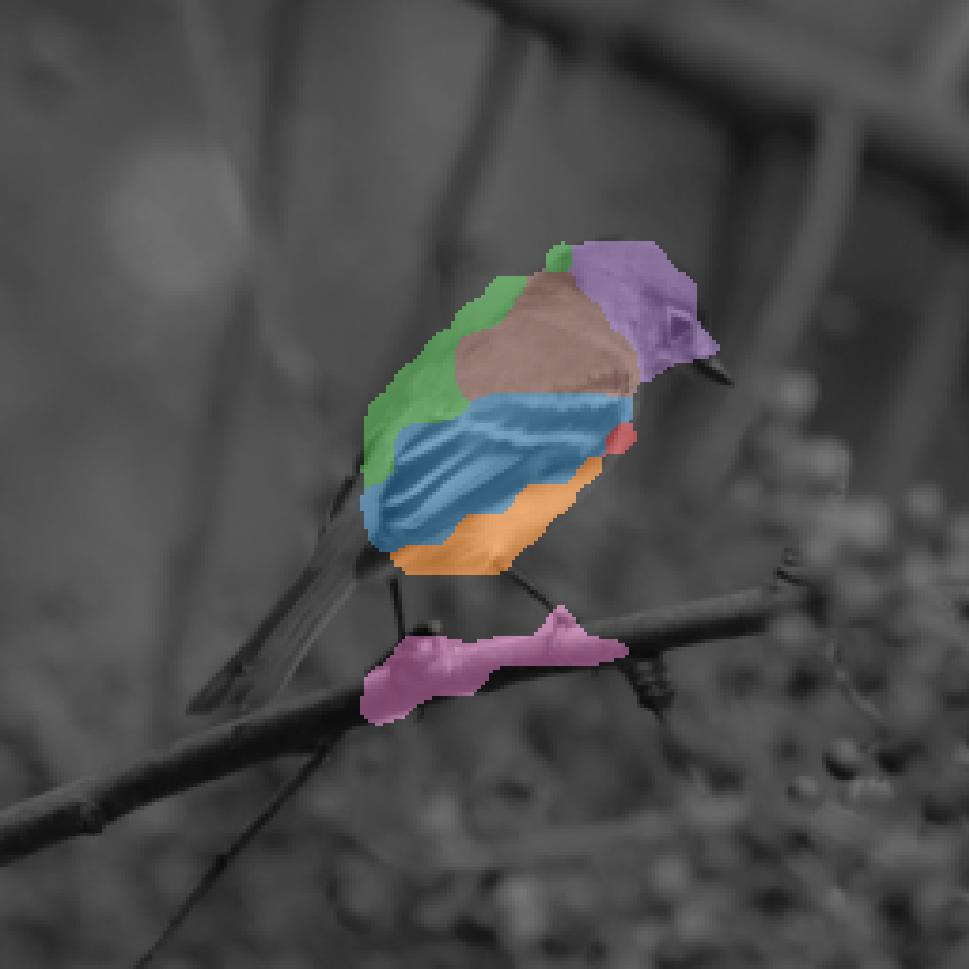} & \includegraphics[width=0.088\linewidth]{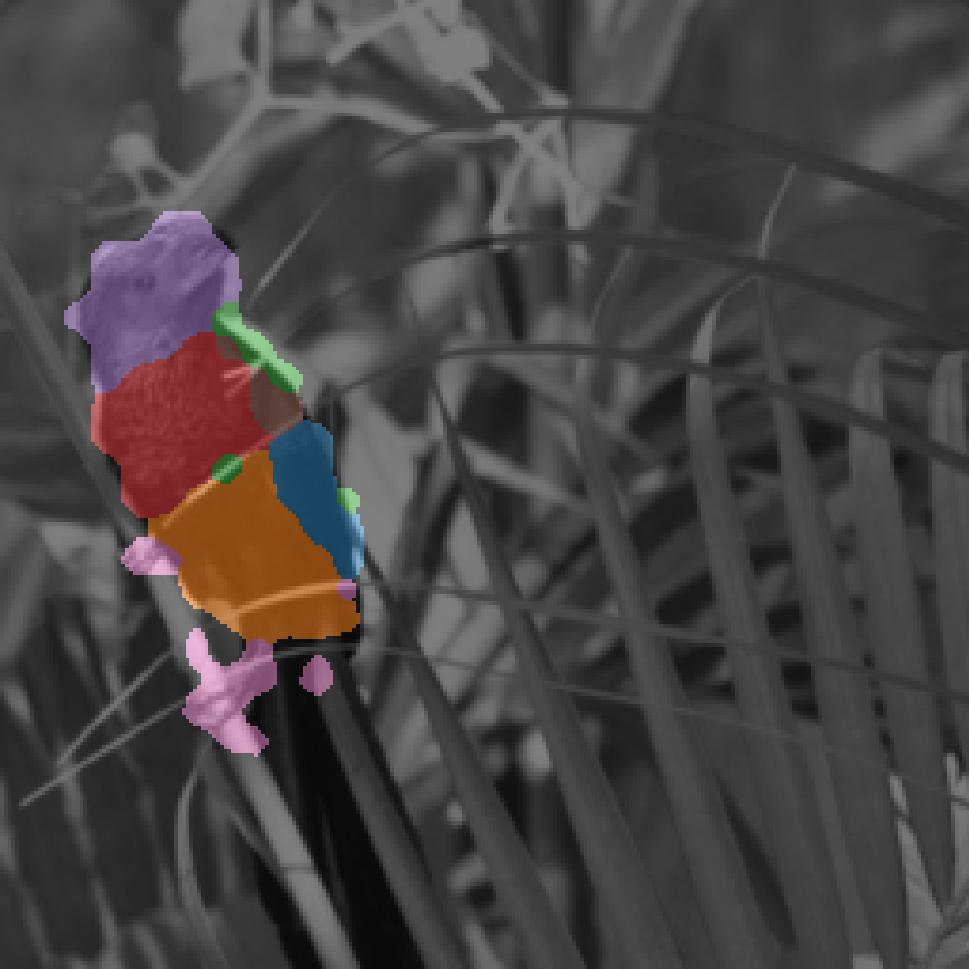} & \includegraphics[width=0.088\linewidth]{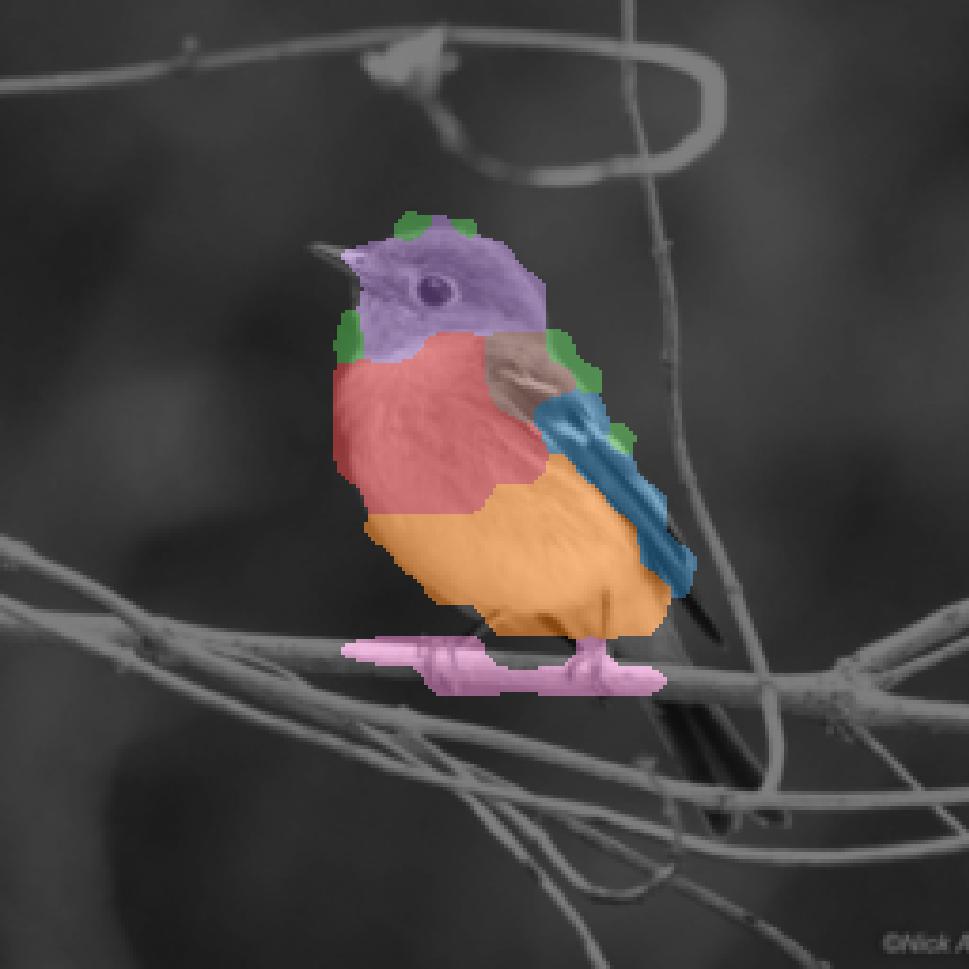} \\
    
    \rotatebox{90}{\makebox[1cm][c]{\scriptsize ST}} & \includegraphics[width=0.088\linewidth]{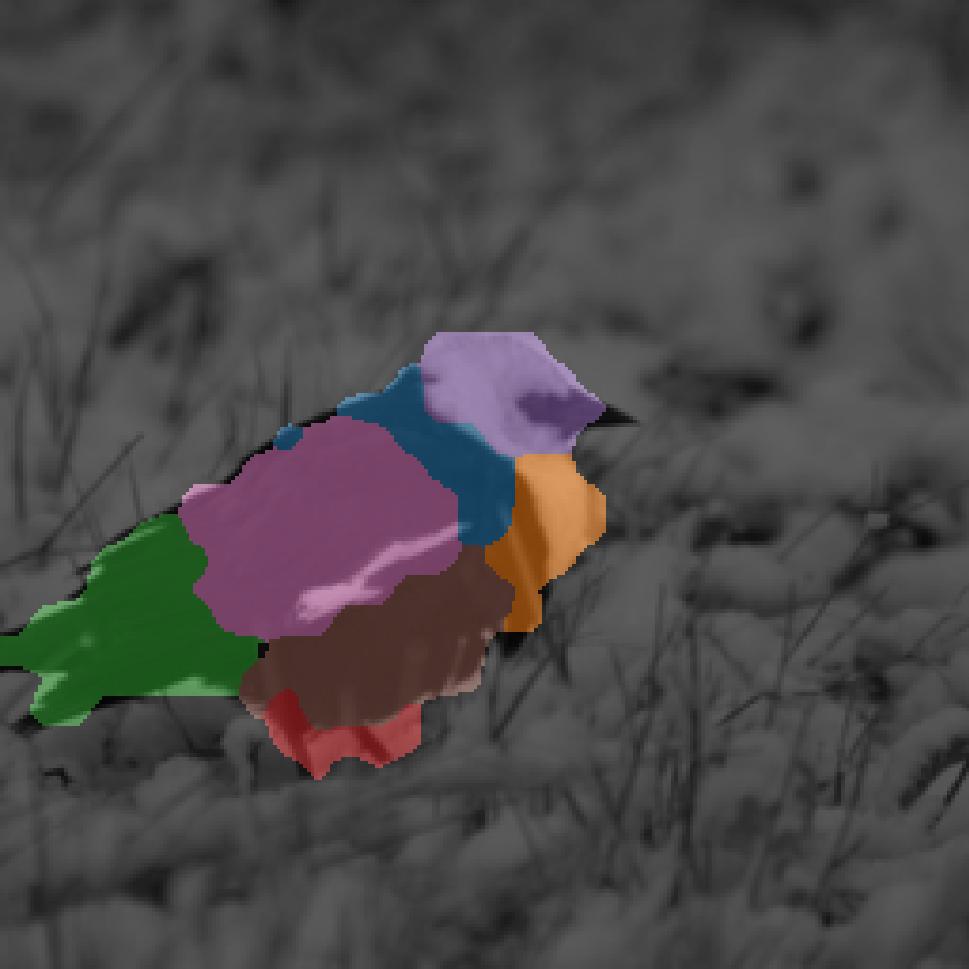} & \includegraphics[width=0.088\linewidth]{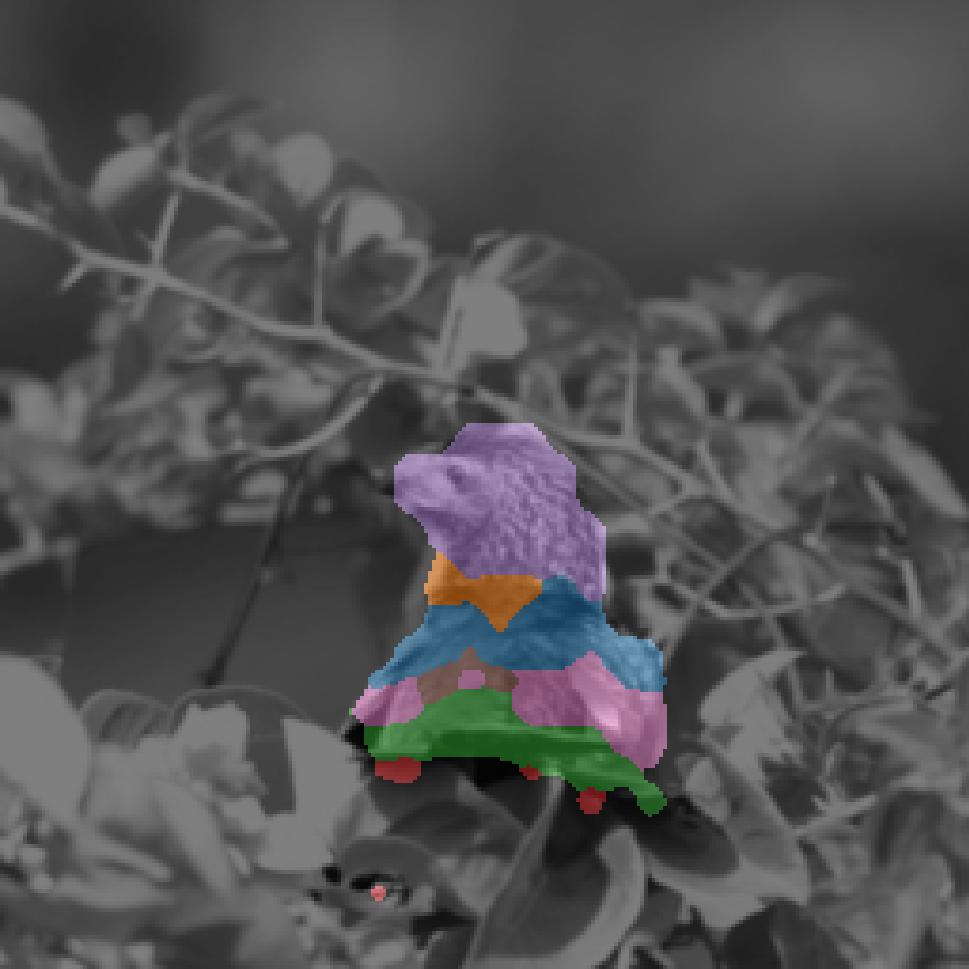} & \includegraphics[width=0.088\linewidth]{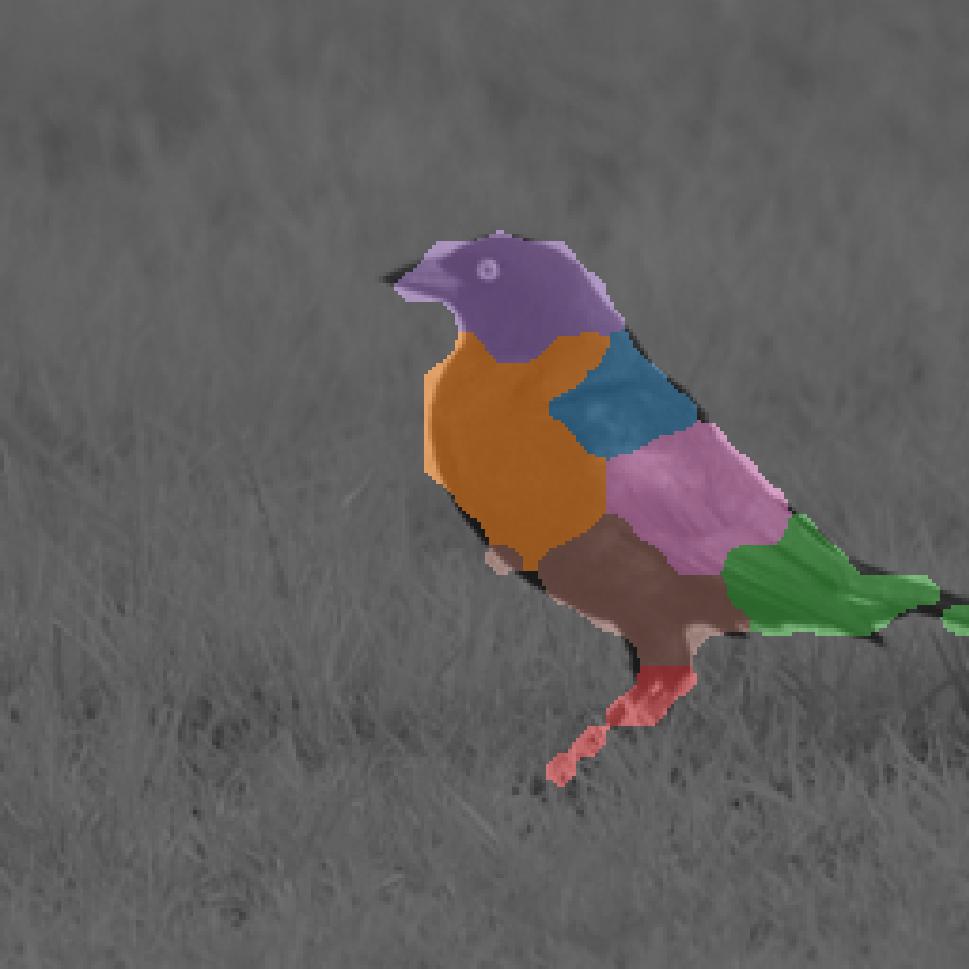} & \includegraphics[width=0.088\linewidth]{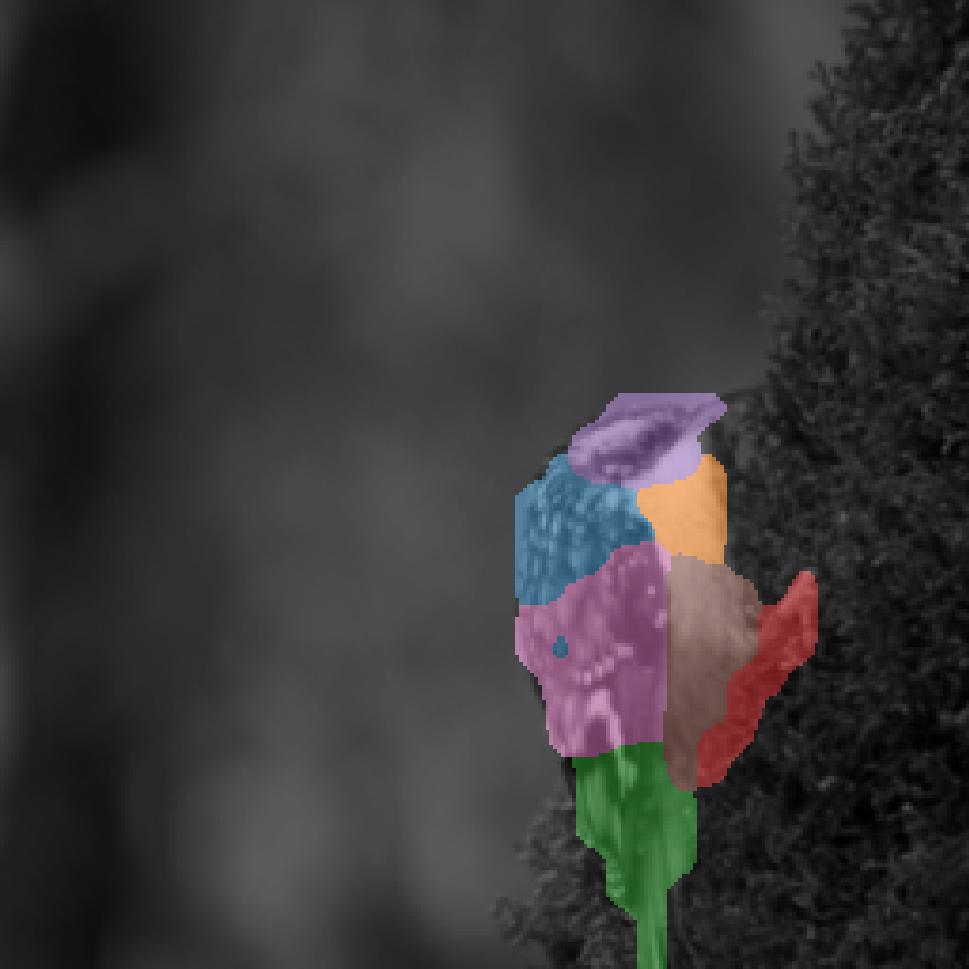} & \includegraphics[width=0.088\linewidth]{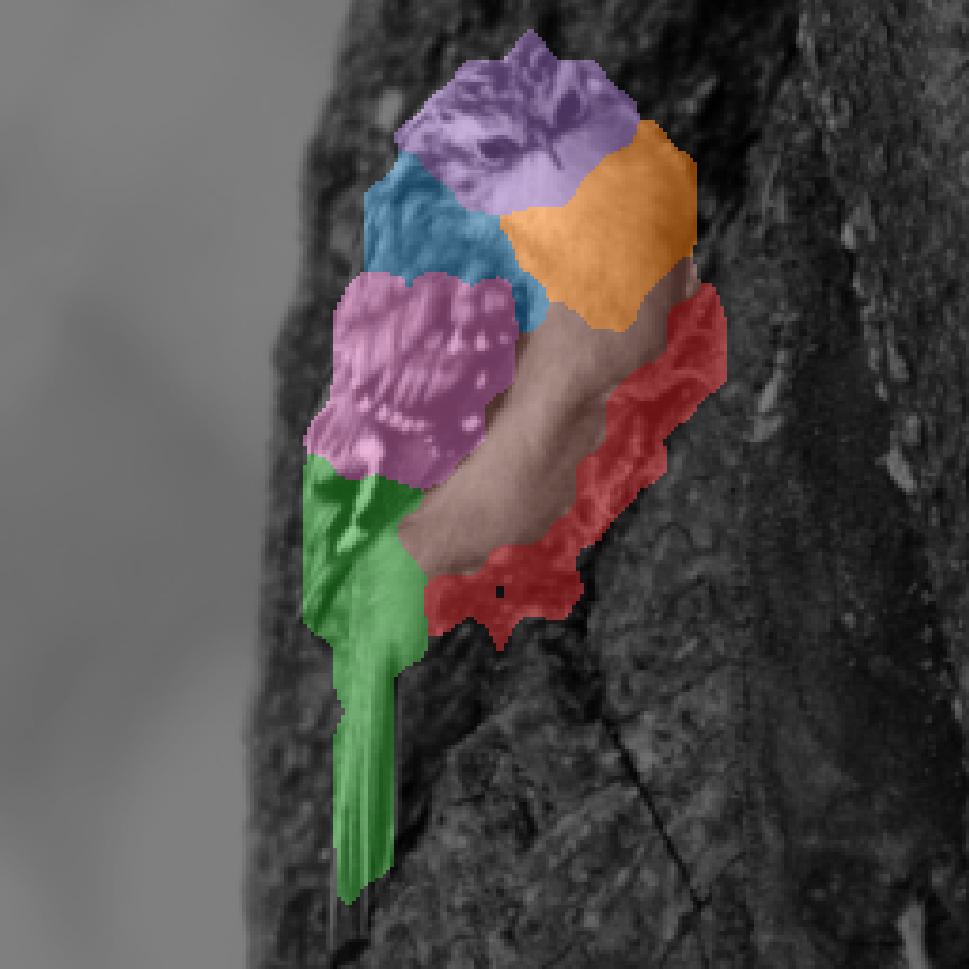} & \includegraphics[width=0.088\linewidth]{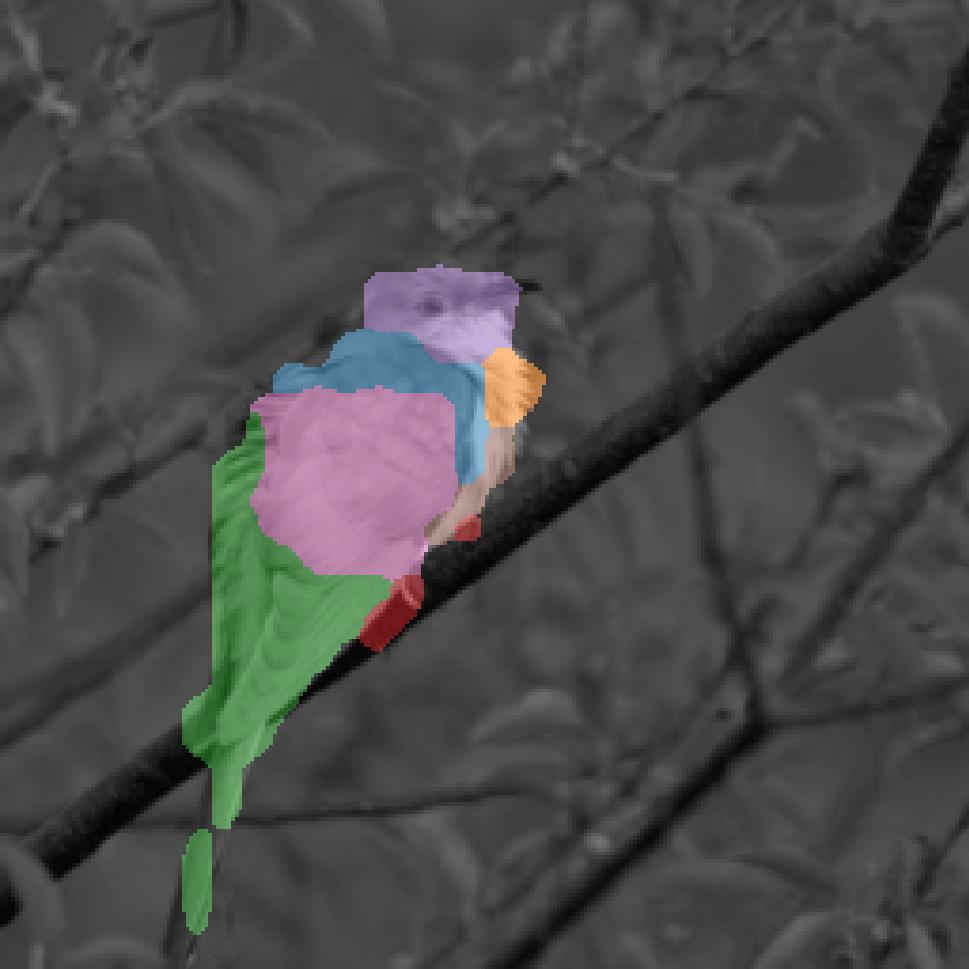} & \includegraphics[width=0.088\linewidth]{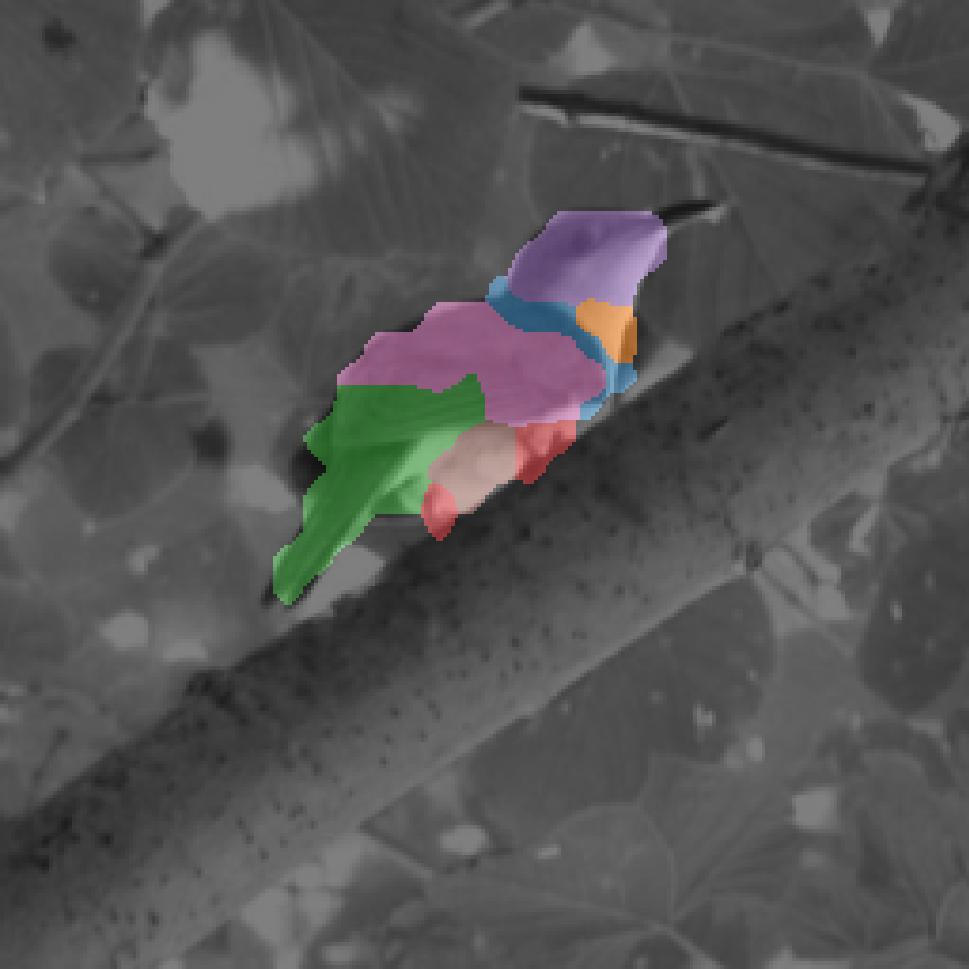} & \includegraphics[width=0.088\linewidth]{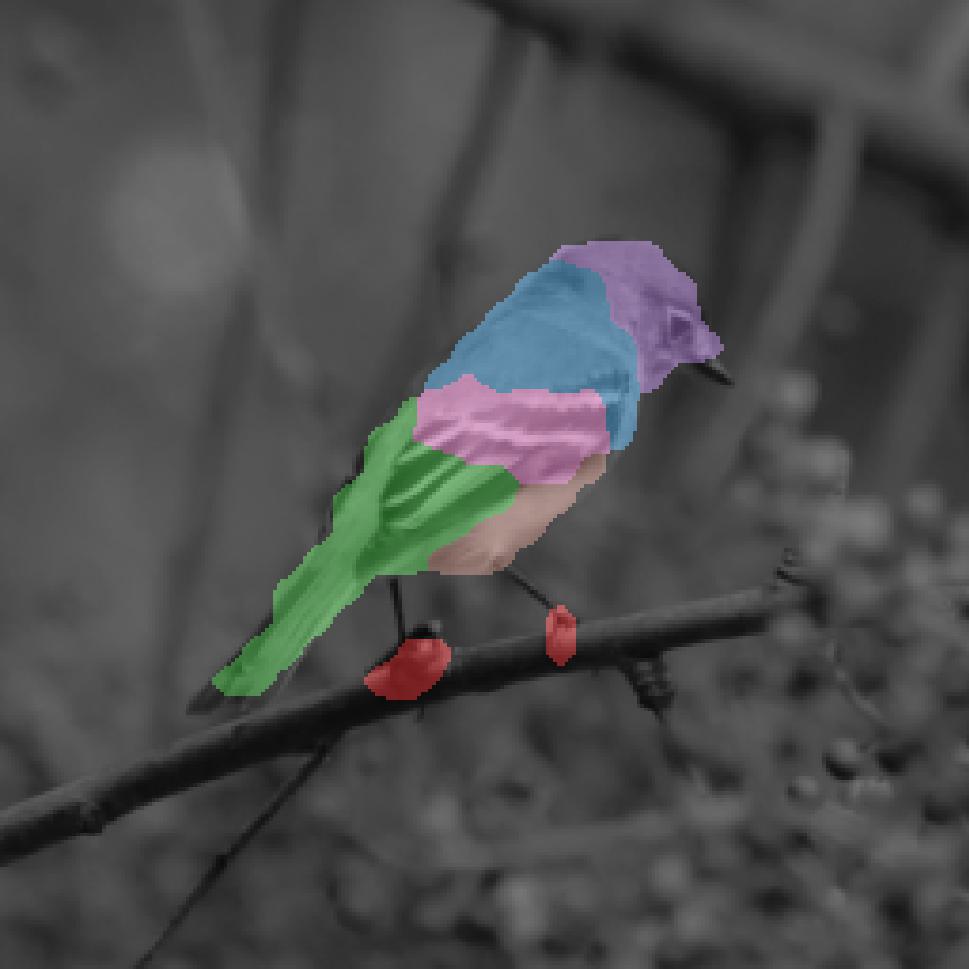} & \includegraphics[width=0.088\linewidth]{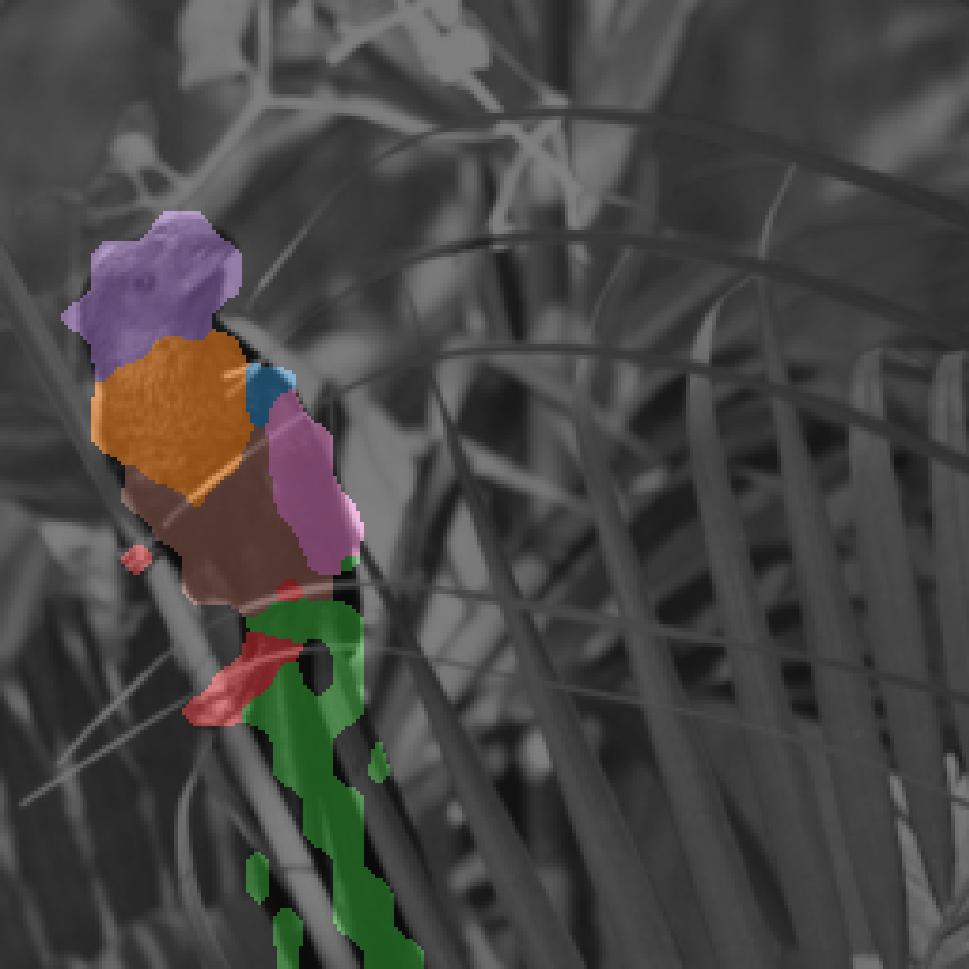} & \includegraphics[width=0.088\linewidth]{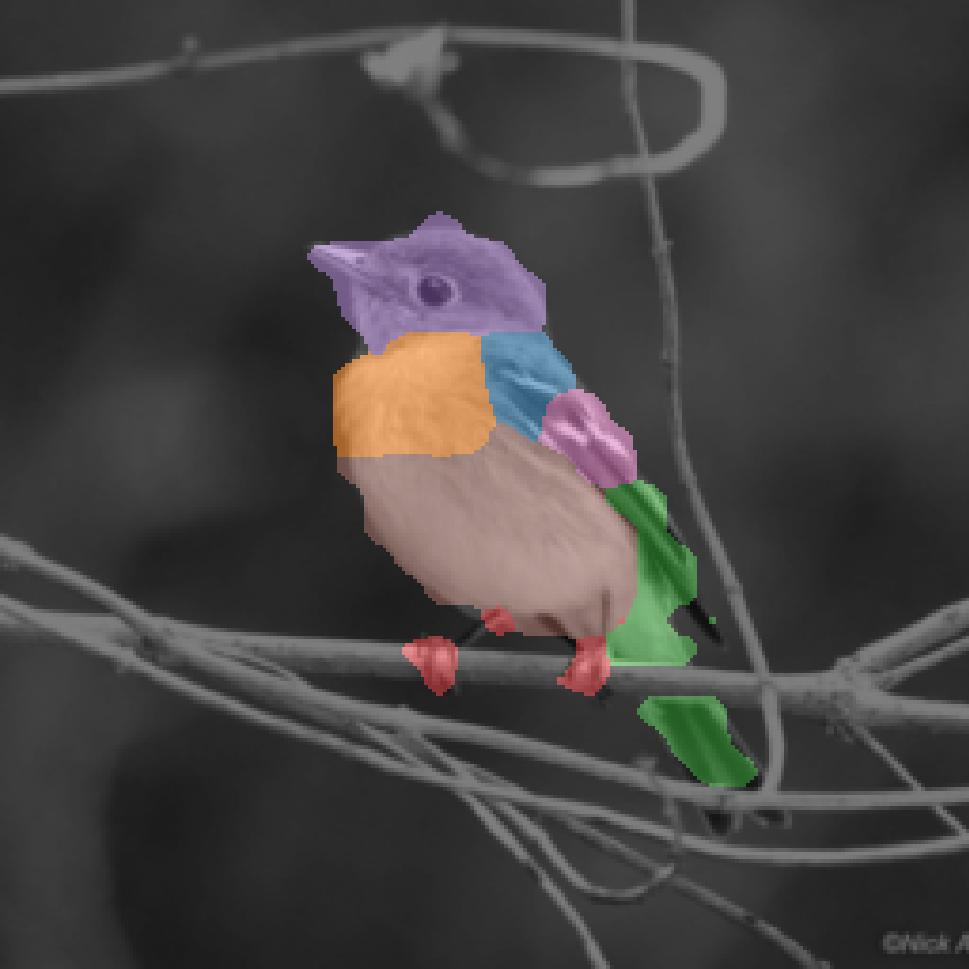} \\
  \bottomrule
  \end{tabular}
  \caption{Additional qualitative results on CUB.}
  \label{tab:class_level_cub}
\end{figure}

\begin{figure}[t]
  \centering
  \setlength{\tabcolsep}{0pt}
  \renewcommand{\arraystretch}{0.5}
  \begin{tabular}{@{}r@{\hspace{2pt}}c@{\hspace{1pt}}c@{\hspace{1pt}}c@{\hspace{1pt}}c@{\hspace{1pt}}c@{\hspace{1pt}}c@{\hspace{1pt}}c@{\hspace{1pt}}c@{\hspace{1pt}}c@{\hspace{1pt}}c@{\hspace{1pt}}@{}}
  \toprule
    \rotatebox{90}{\makebox[1cm][c]{\scriptsize Images}}& \includegraphics[width=0.088\linewidth]{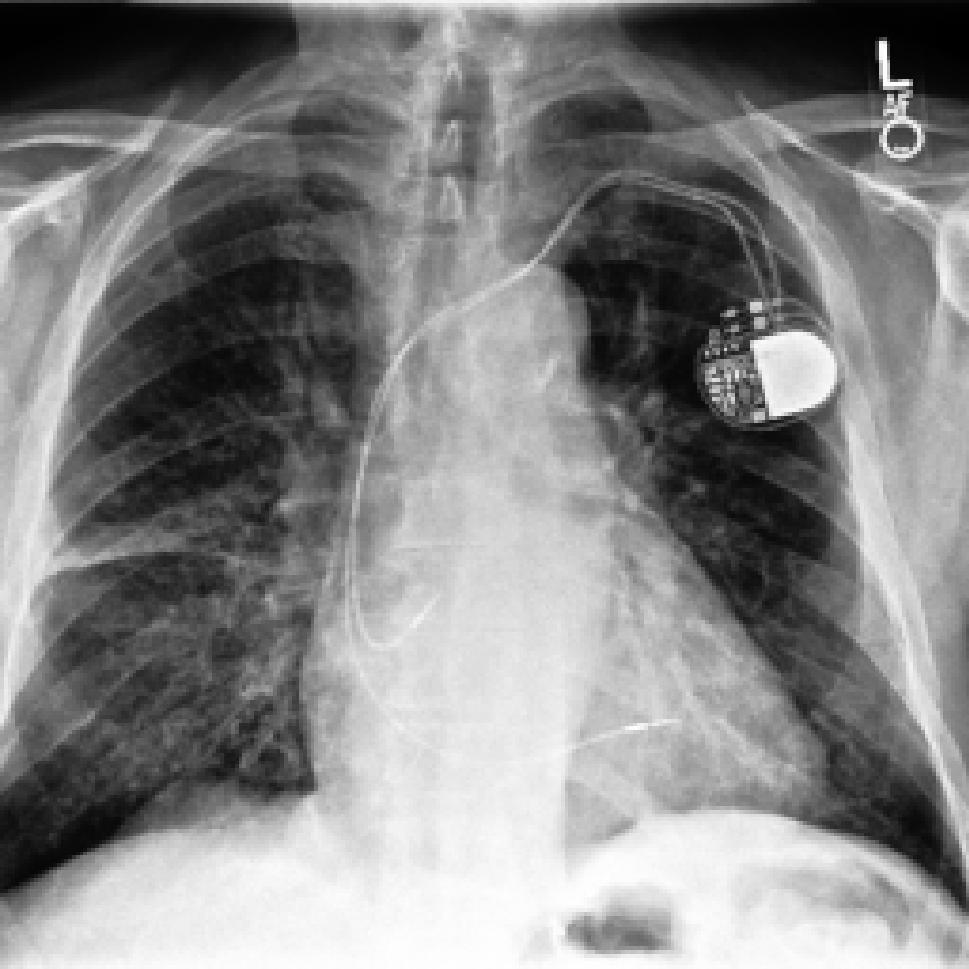} & \includegraphics[width=0.088\linewidth]{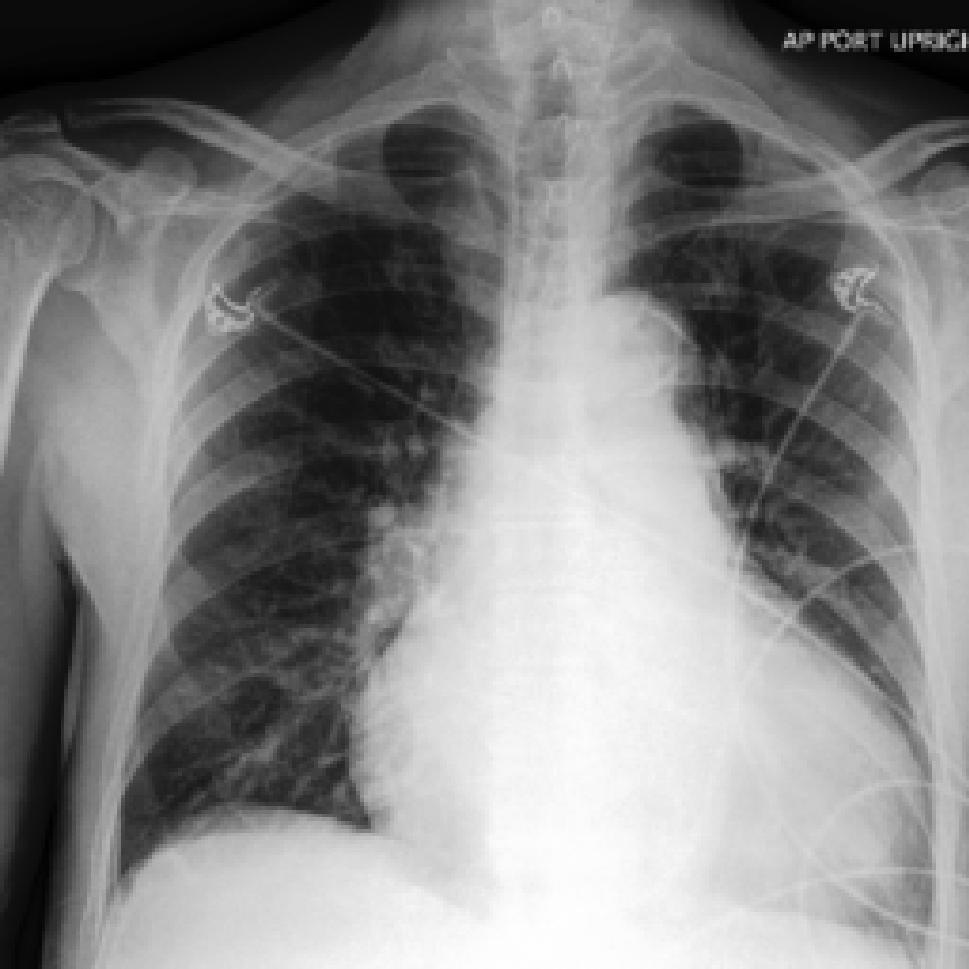} & \includegraphics[width=0.088\linewidth]{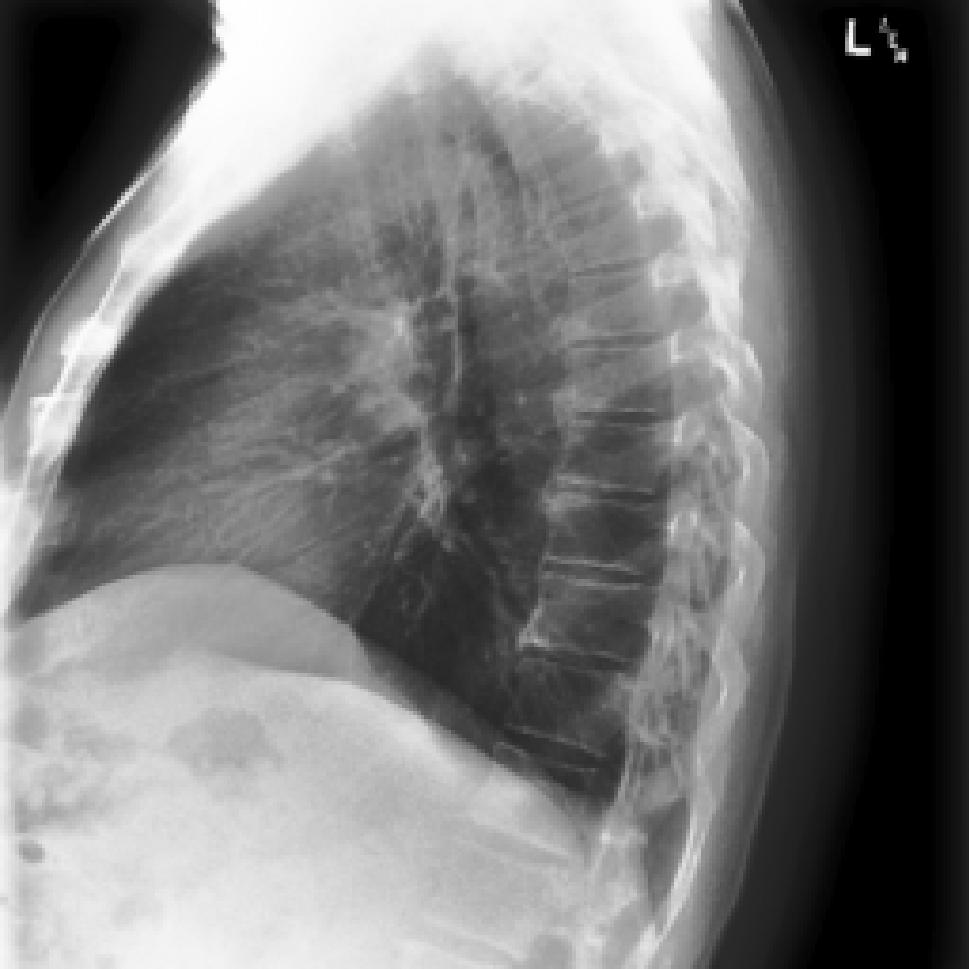} & \includegraphics[width=0.088\linewidth]{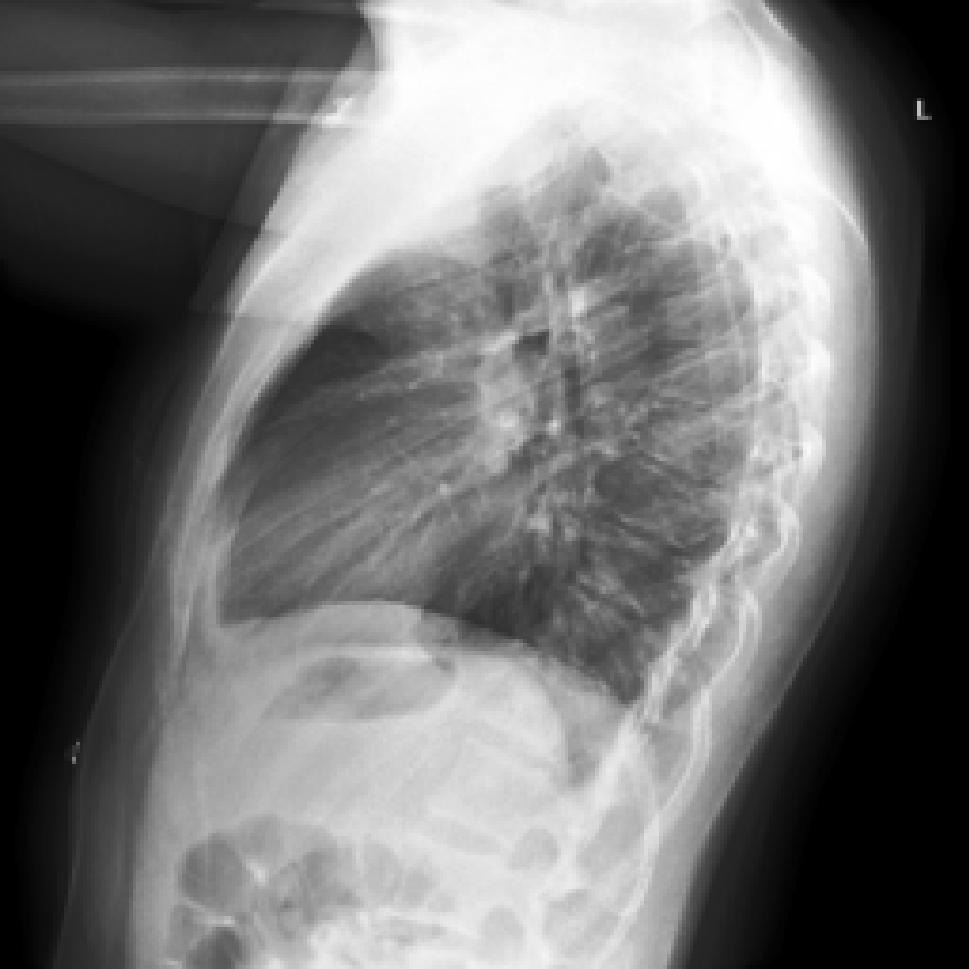} & \includegraphics[width=0.088\linewidth]{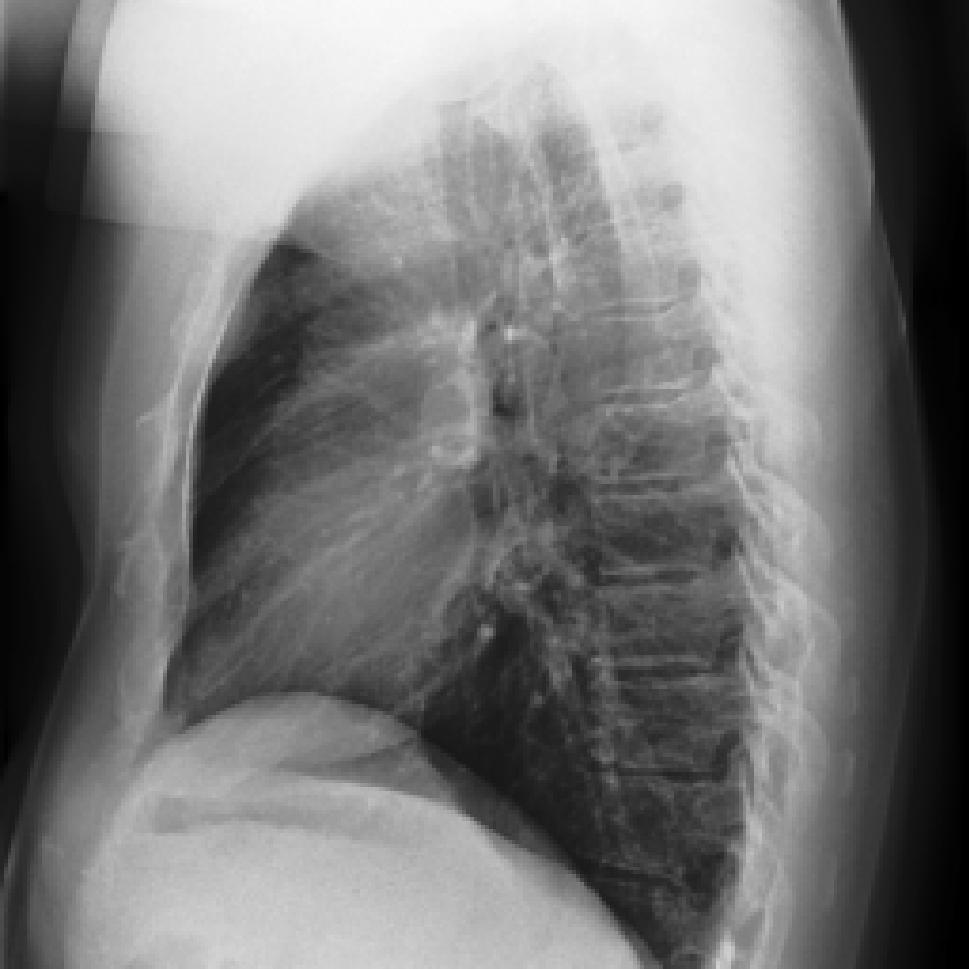} & \includegraphics[width=0.088\linewidth]{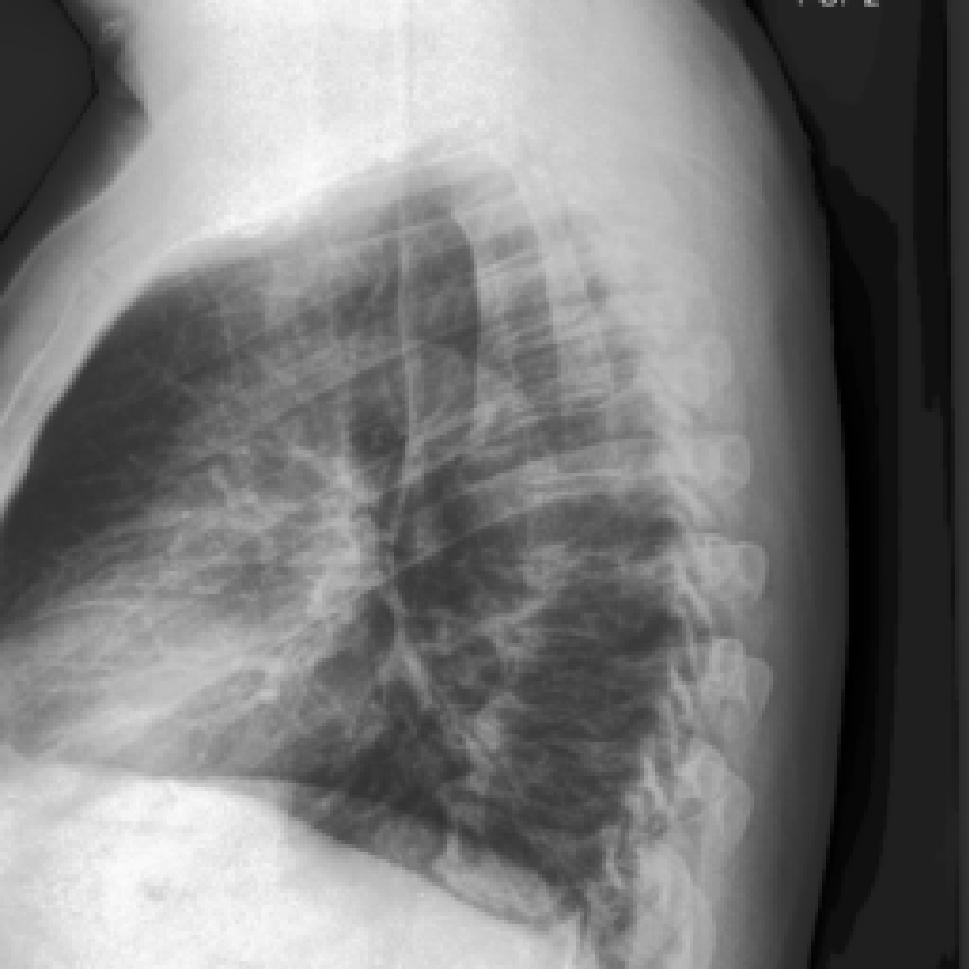} & \includegraphics[width=0.088\linewidth]{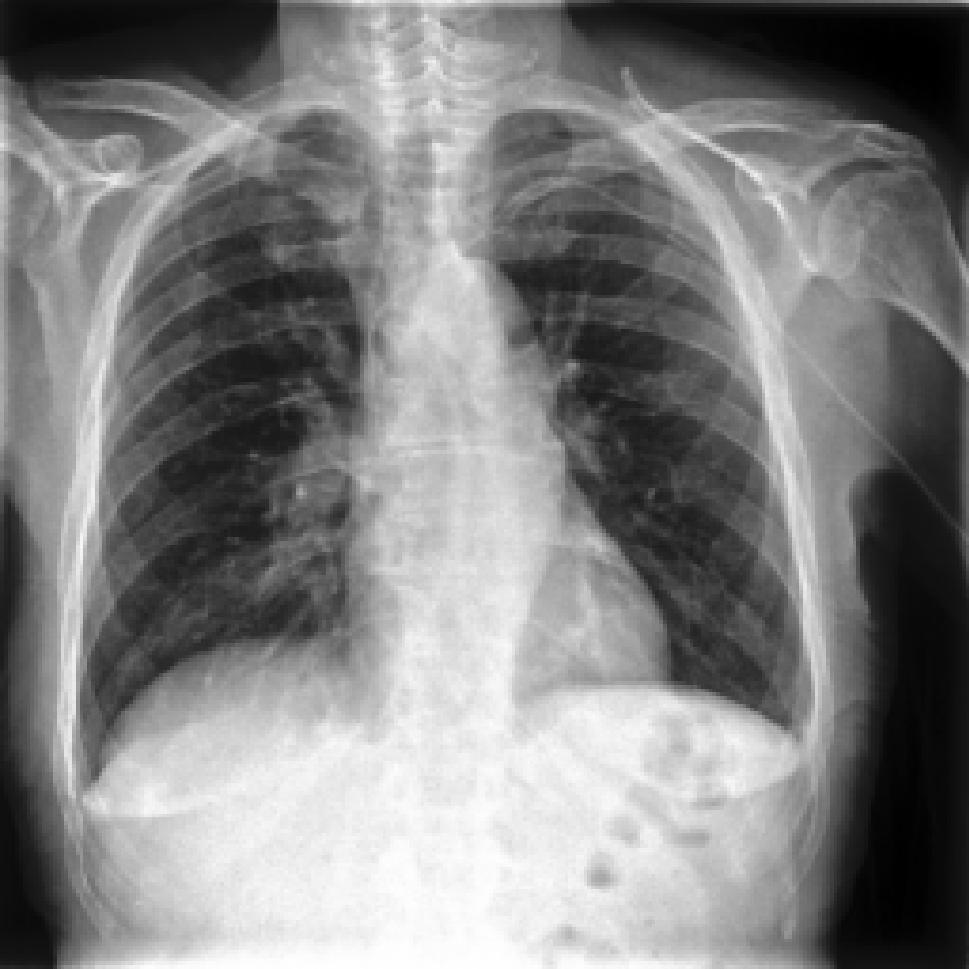} & \includegraphics[width=0.088\linewidth]{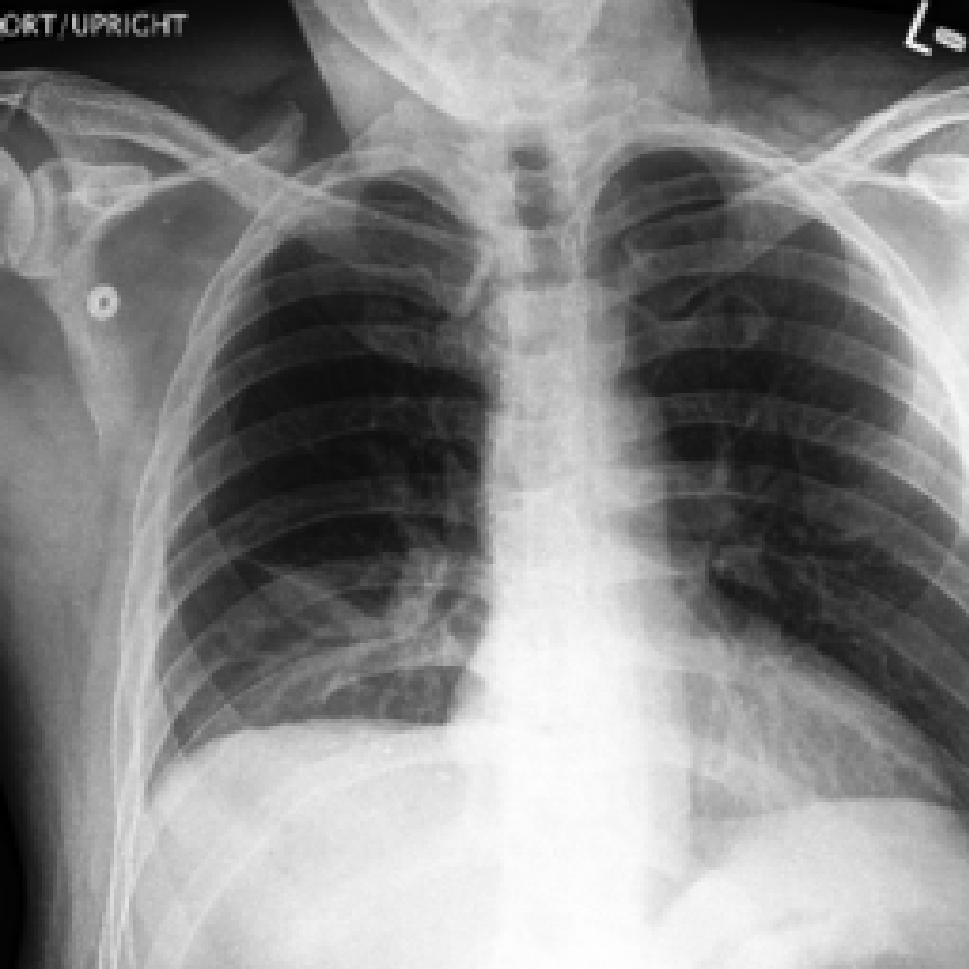} & \includegraphics[width=0.088\linewidth]{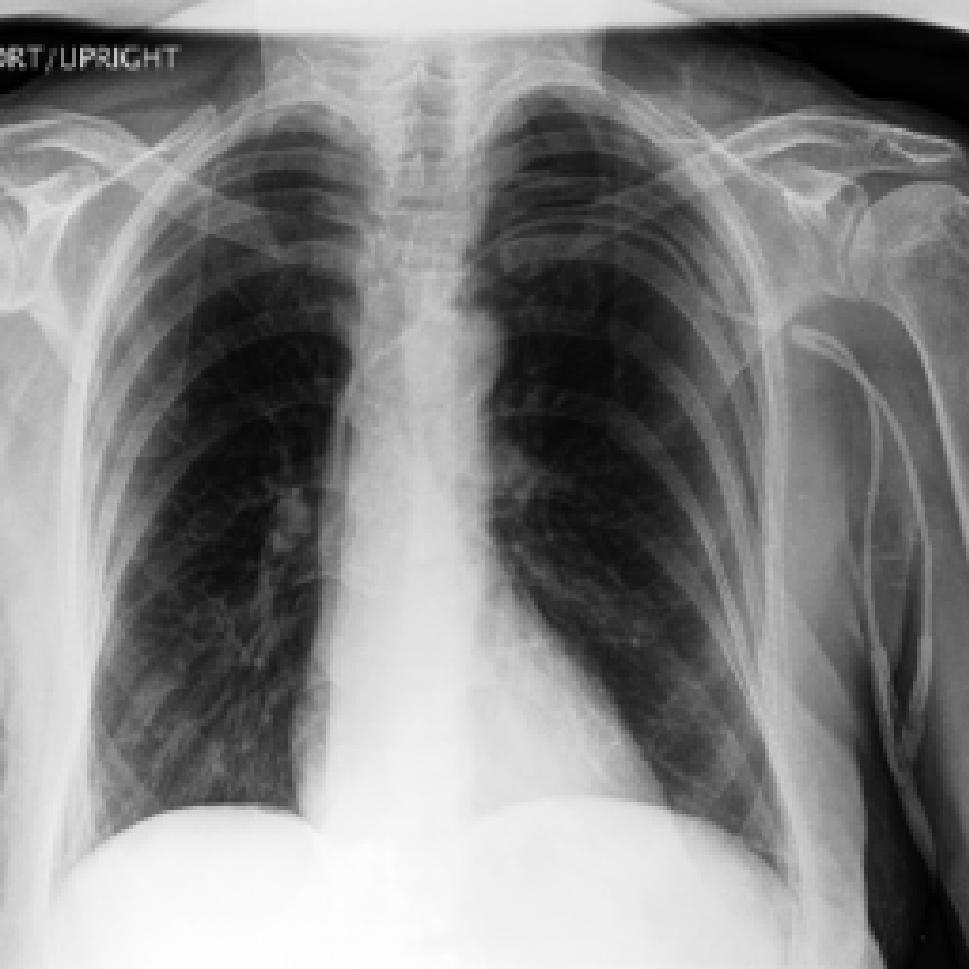} & \includegraphics[width=0.088\linewidth]{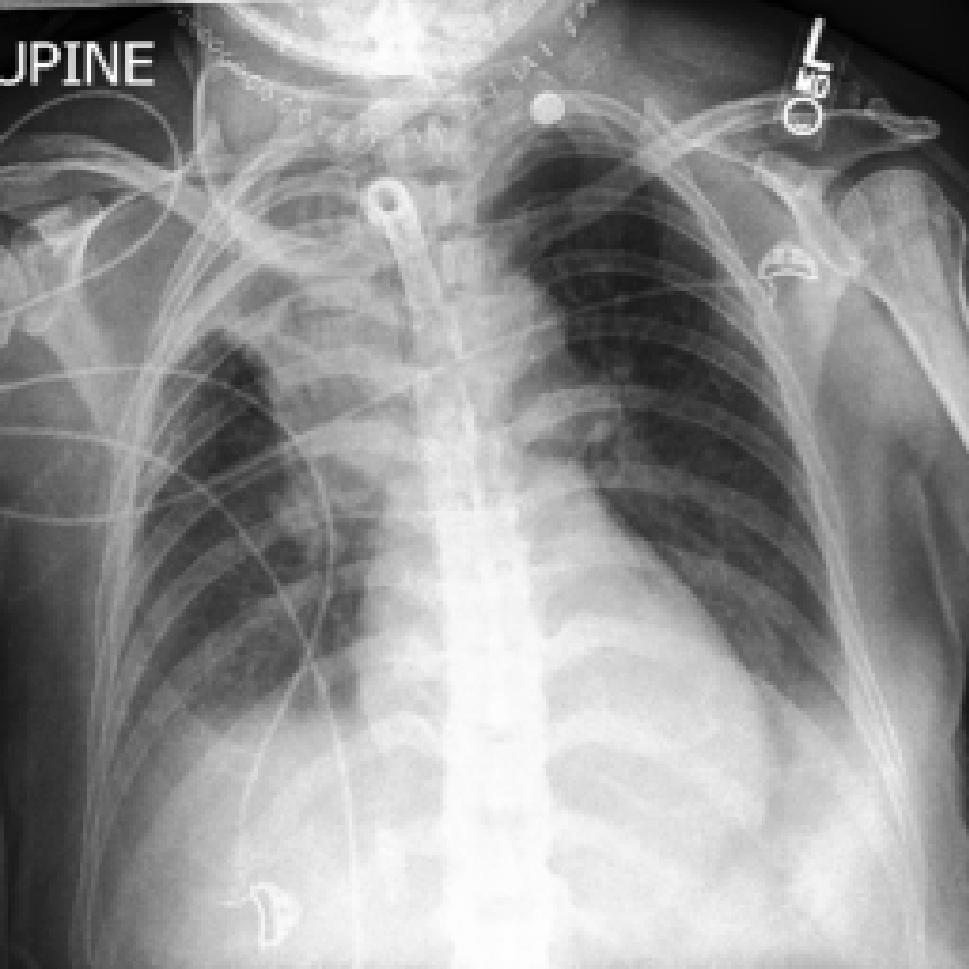} \\

    \rotatebox{90}{\makebox[1cm][c]{\scriptsize 1-stage}}  & \includegraphics[width=0.088\linewidth]{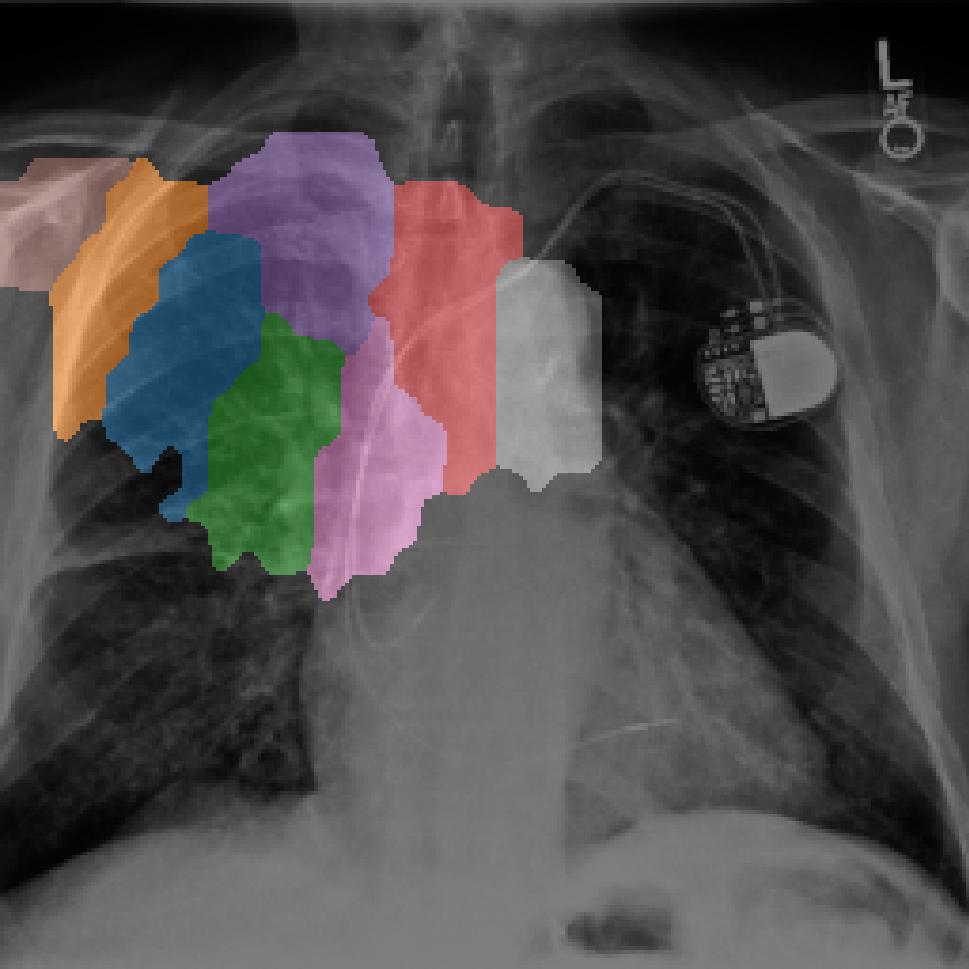} & \includegraphics[width=0.088\linewidth]{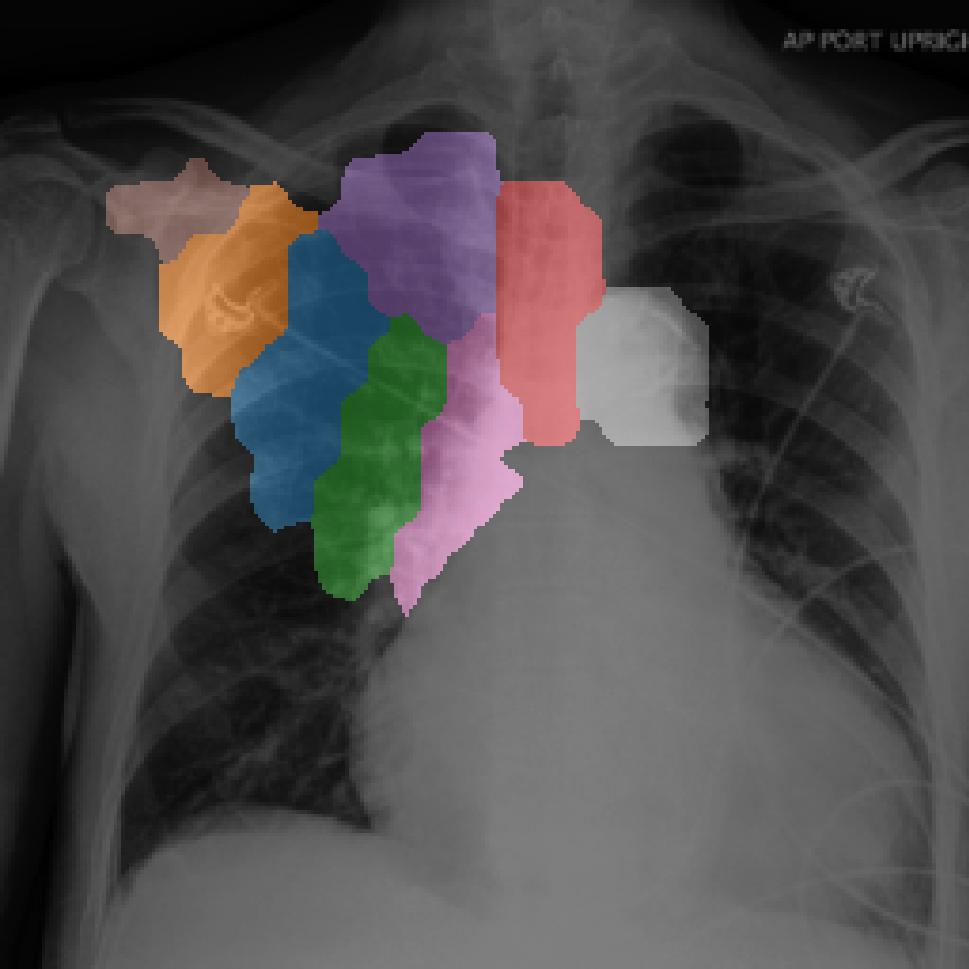} & \includegraphics[width=0.088\linewidth]{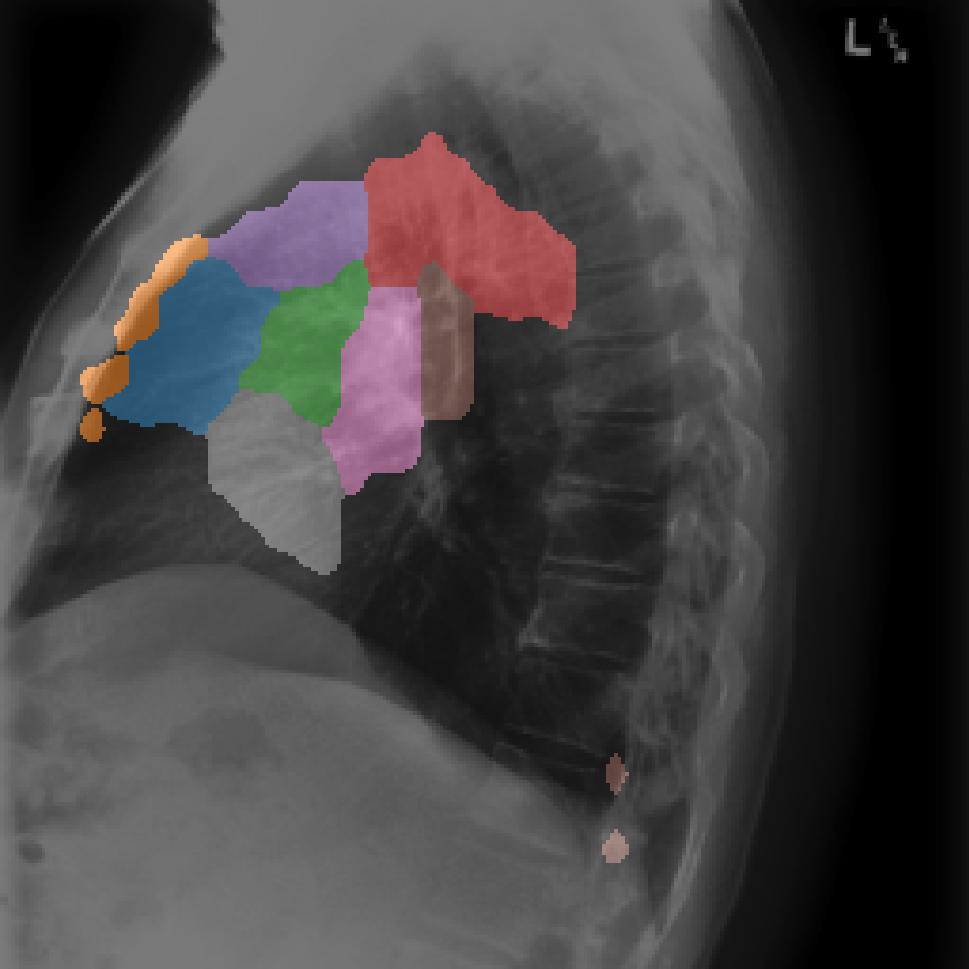} & \includegraphics[width=0.088\linewidth]{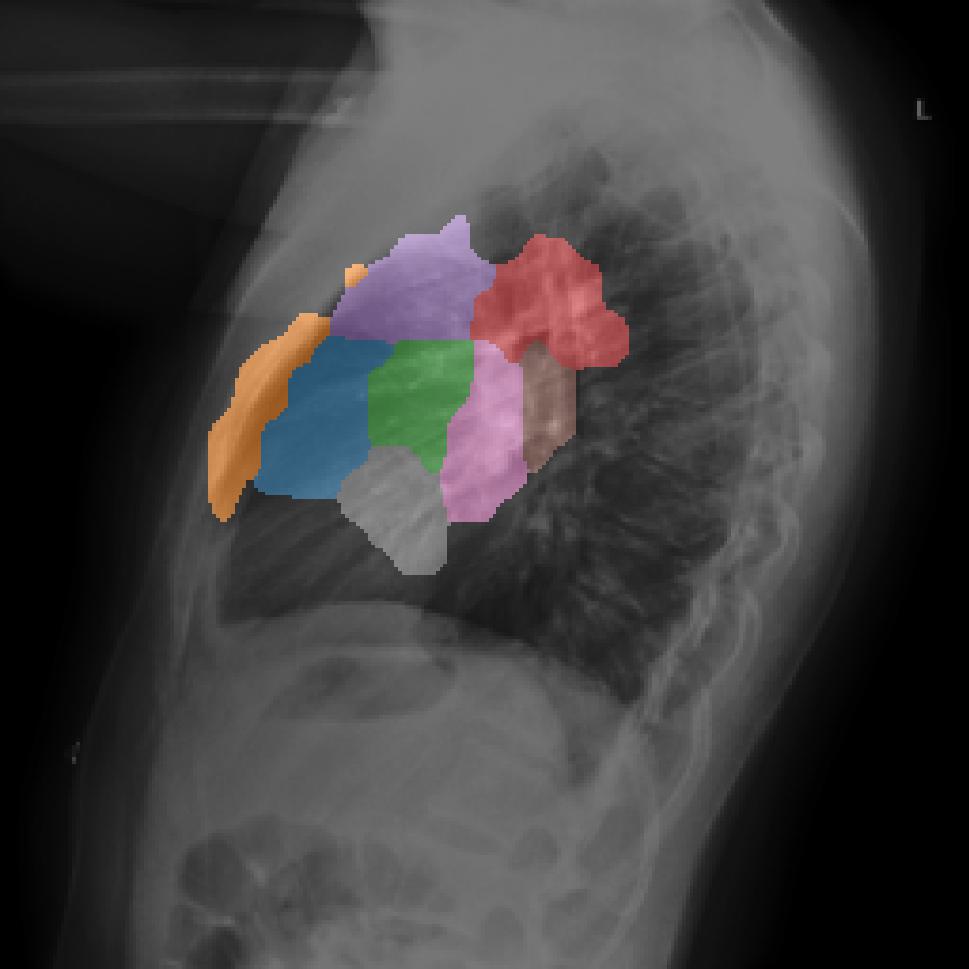} & \includegraphics[width=0.088\linewidth]{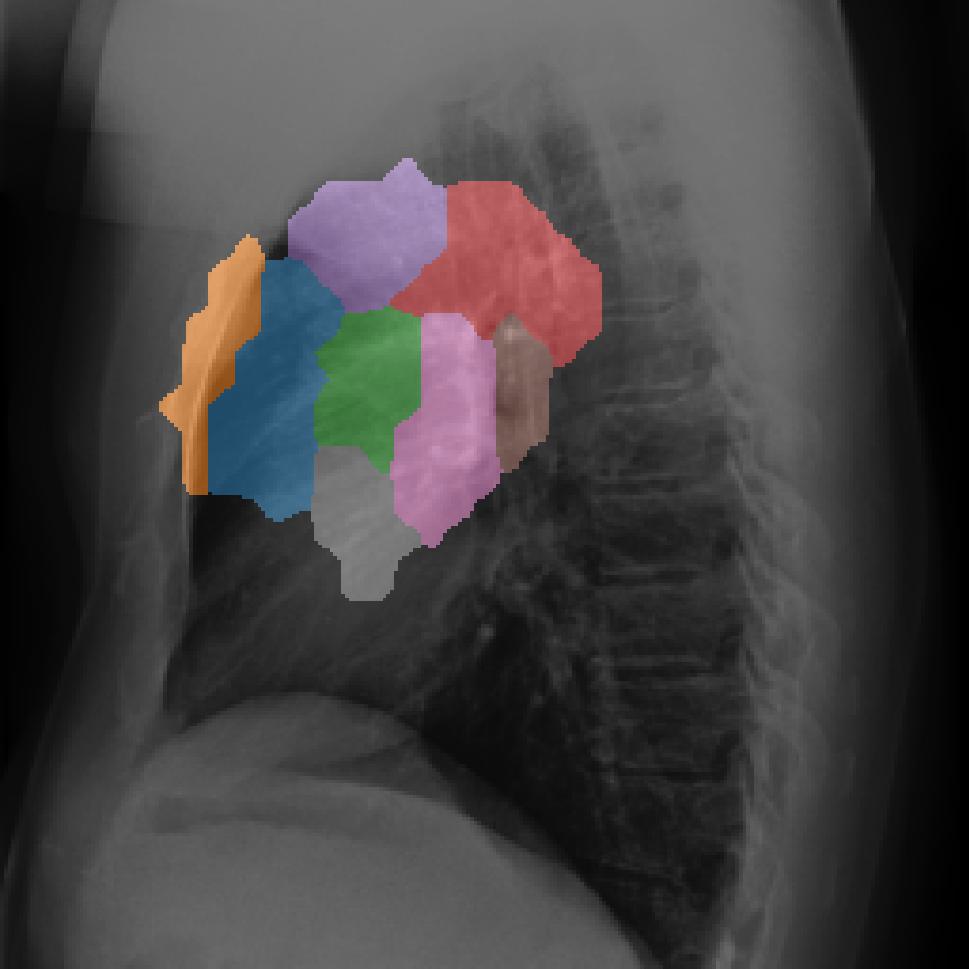} & \includegraphics[width=0.088\linewidth]{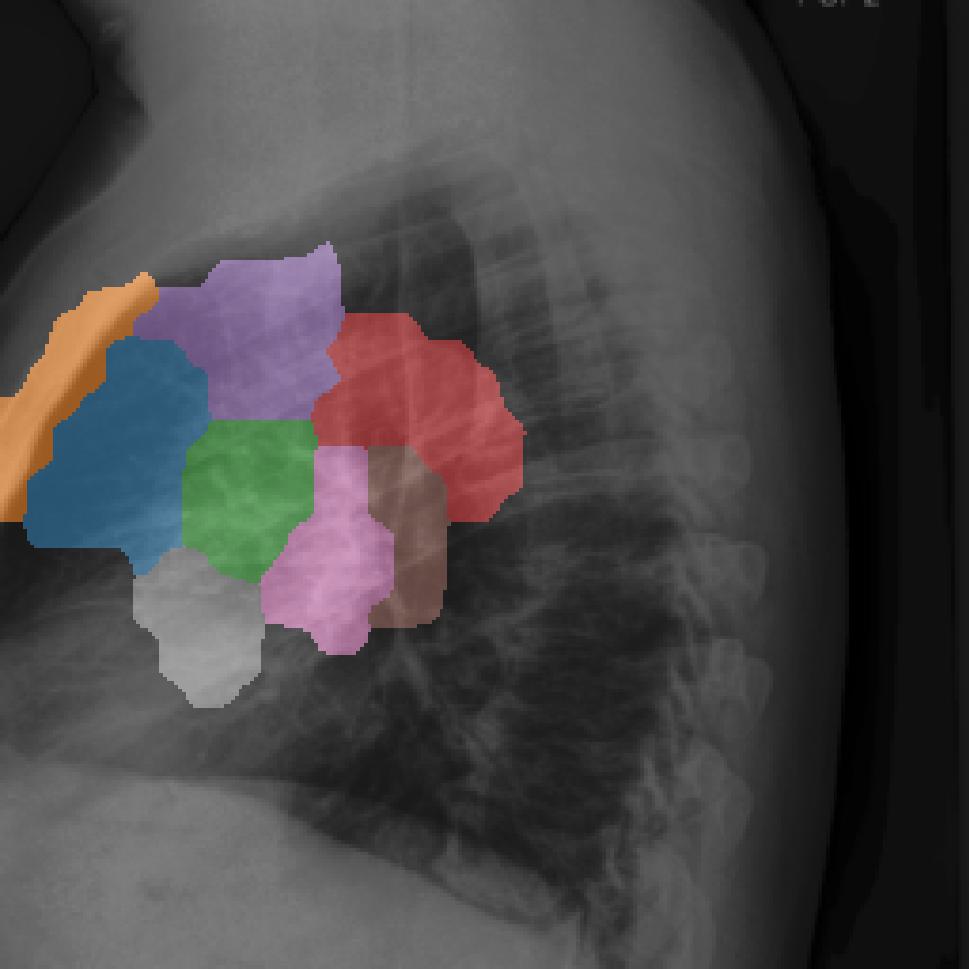} & \includegraphics[width=0.088\linewidth]{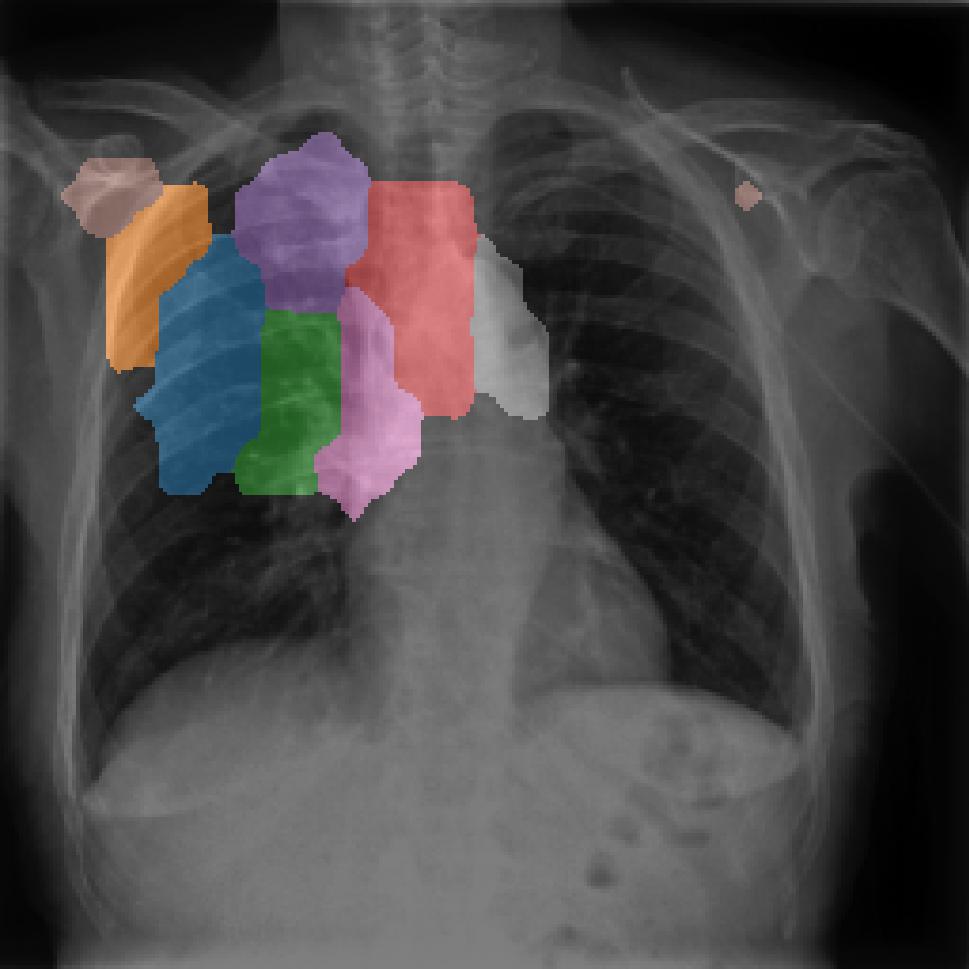} & \includegraphics[width=0.088\linewidth]{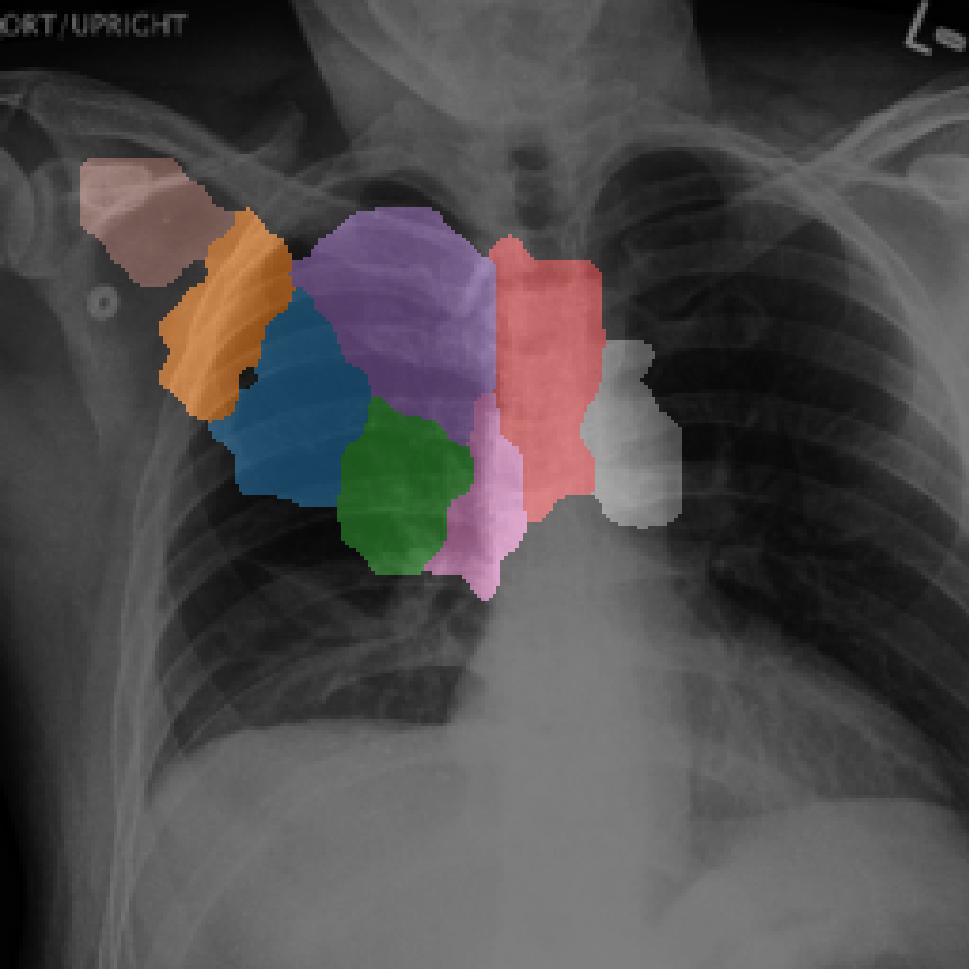} & \includegraphics[width=0.088\linewidth]{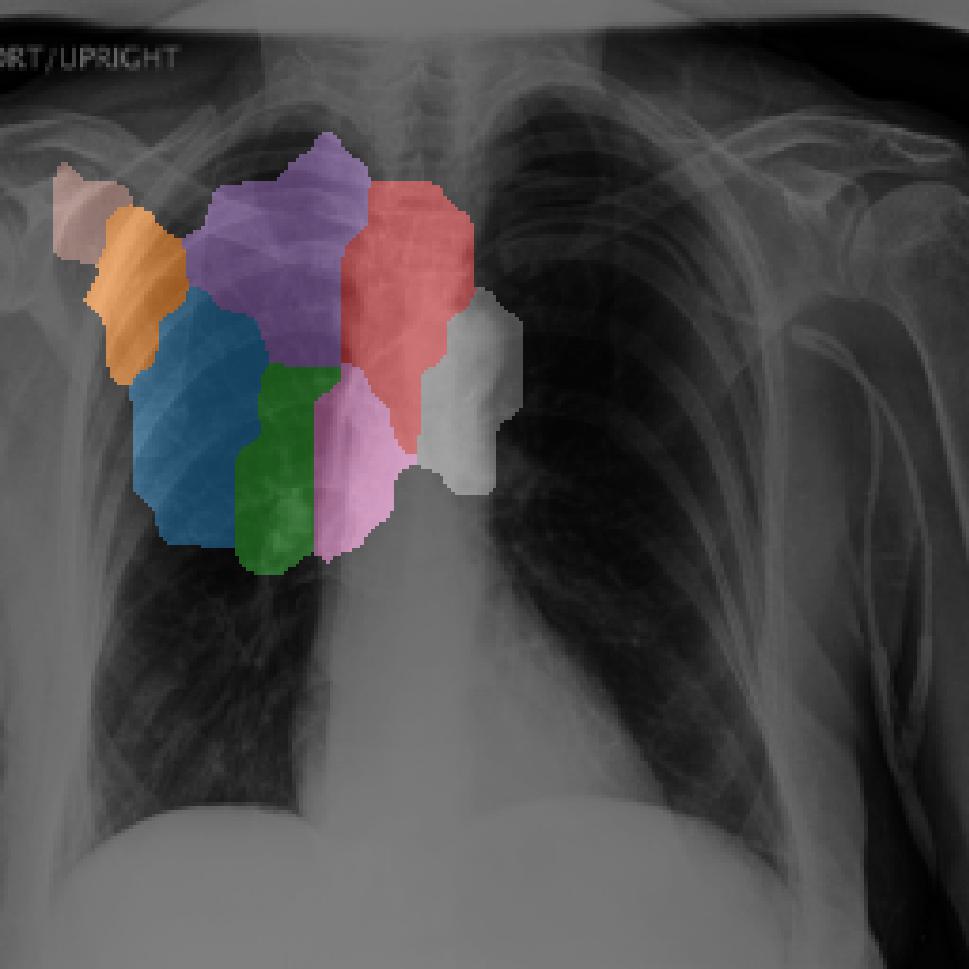} & \includegraphics[width=0.088\linewidth]{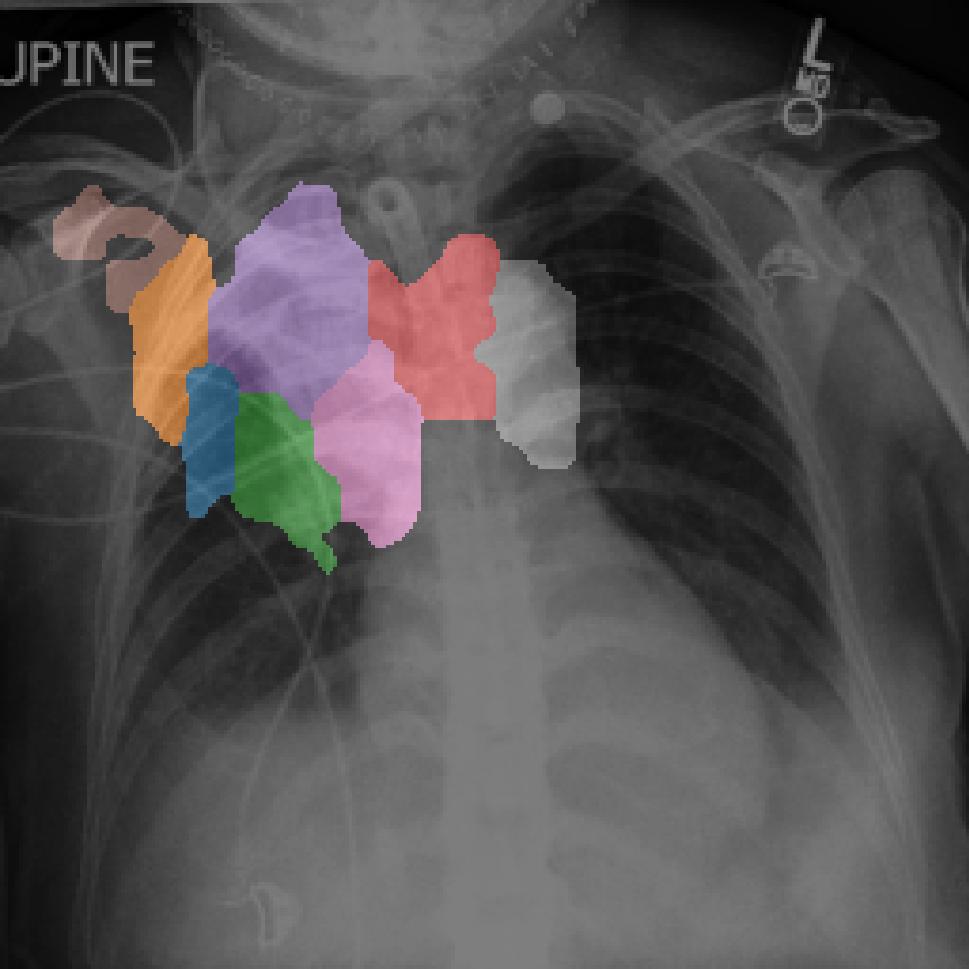} \\

    \rotatebox{90}{\makebox[1cm][c]{\scriptsize Soft}} & \includegraphics[width=0.088\linewidth]{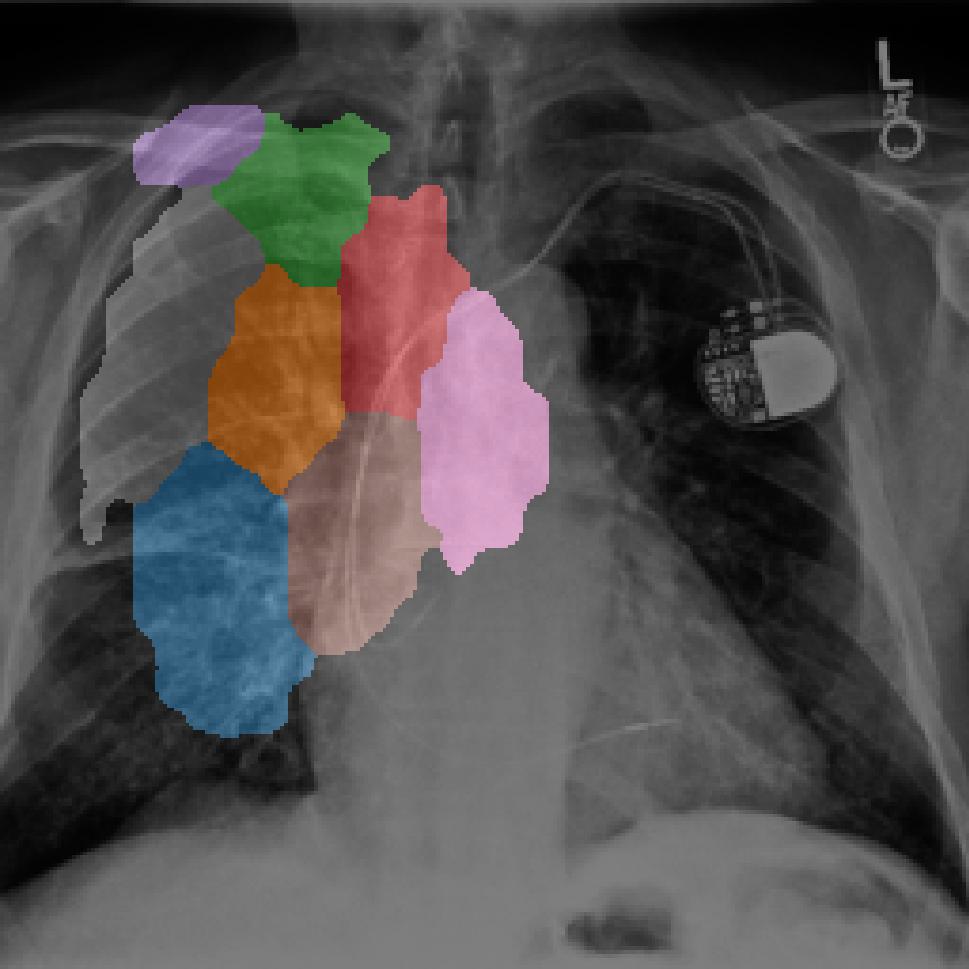} & \includegraphics[width=0.088\linewidth]{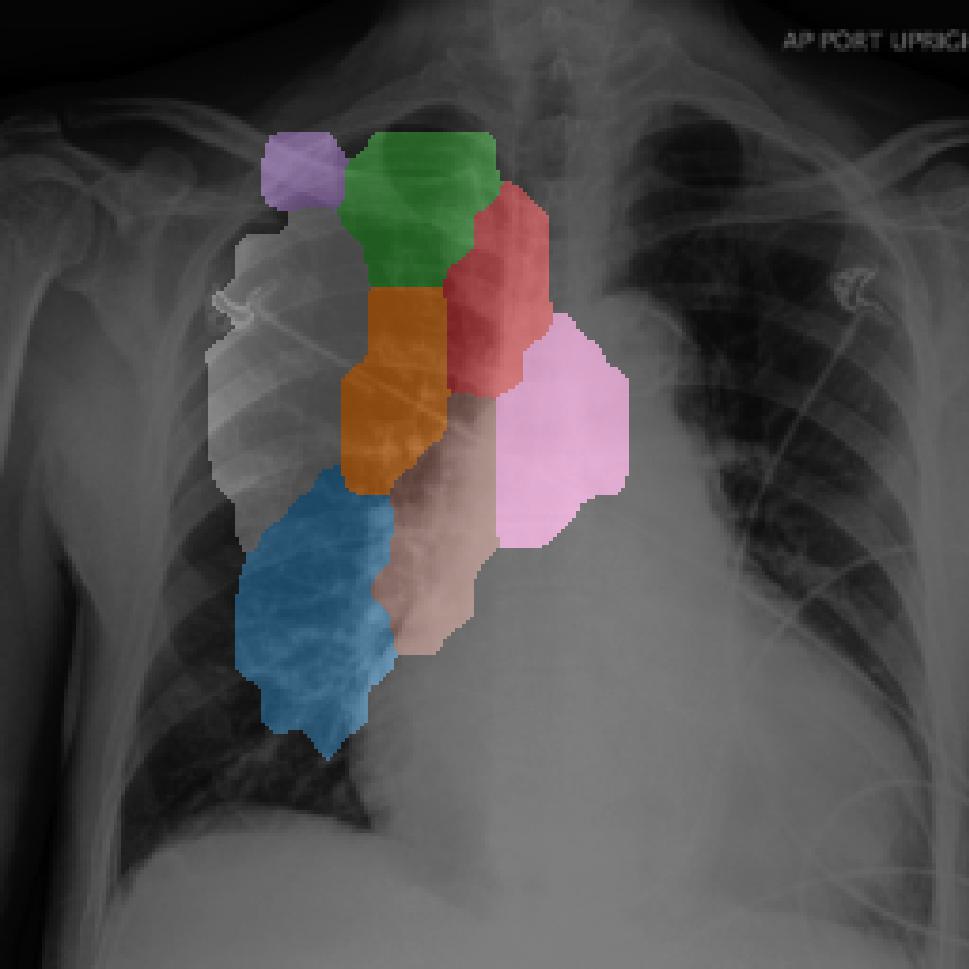} & \includegraphics[width=0.088\linewidth]{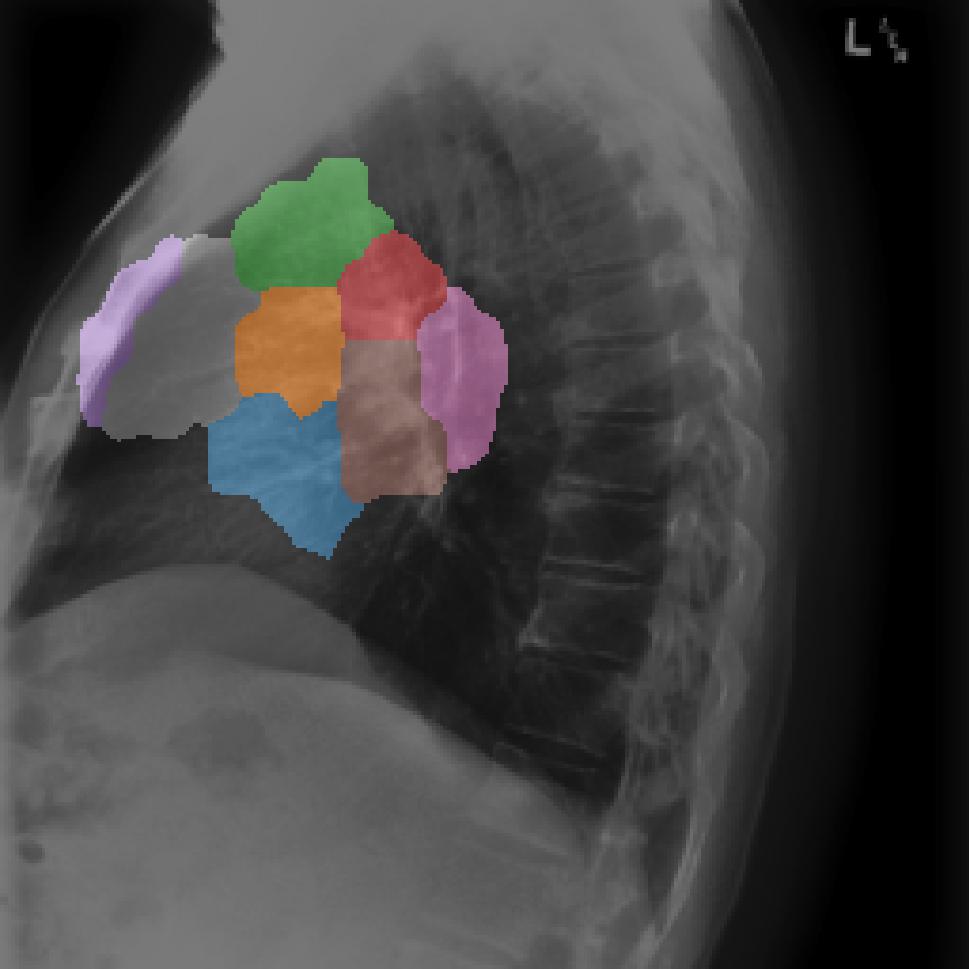} & \includegraphics[width=0.088\linewidth]{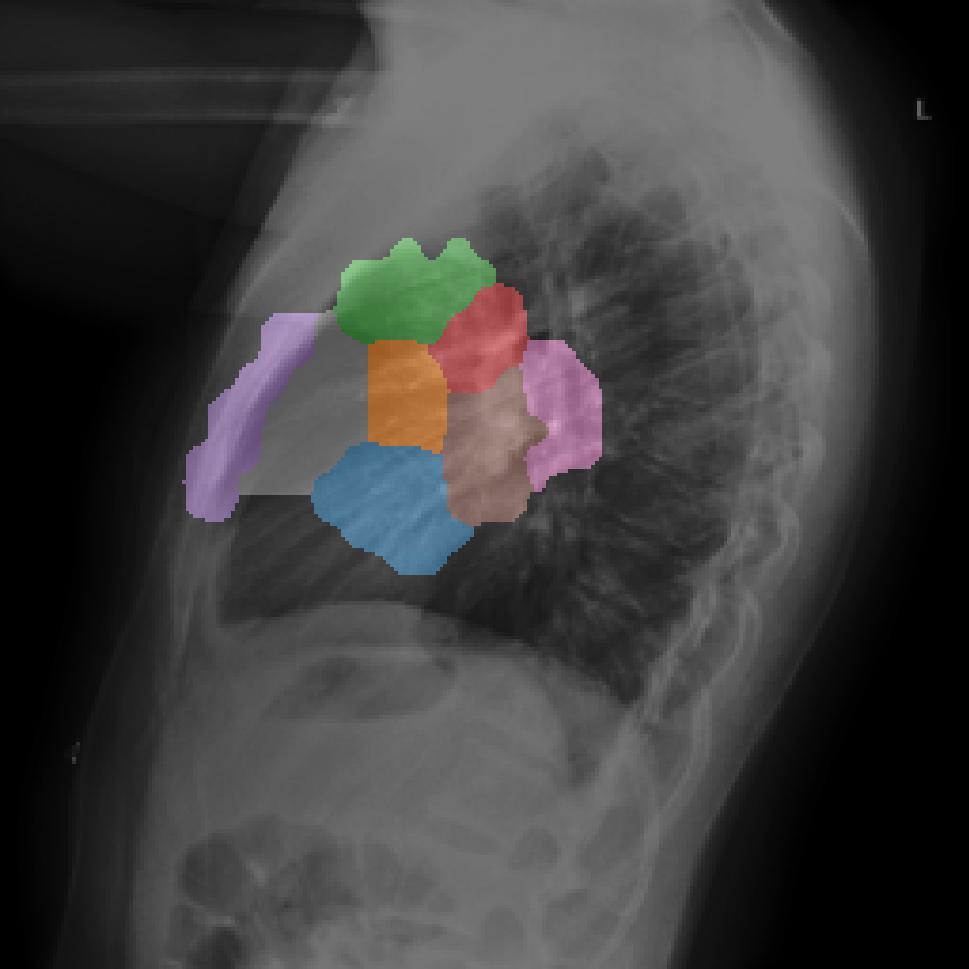} & \includegraphics[width=0.088\linewidth]{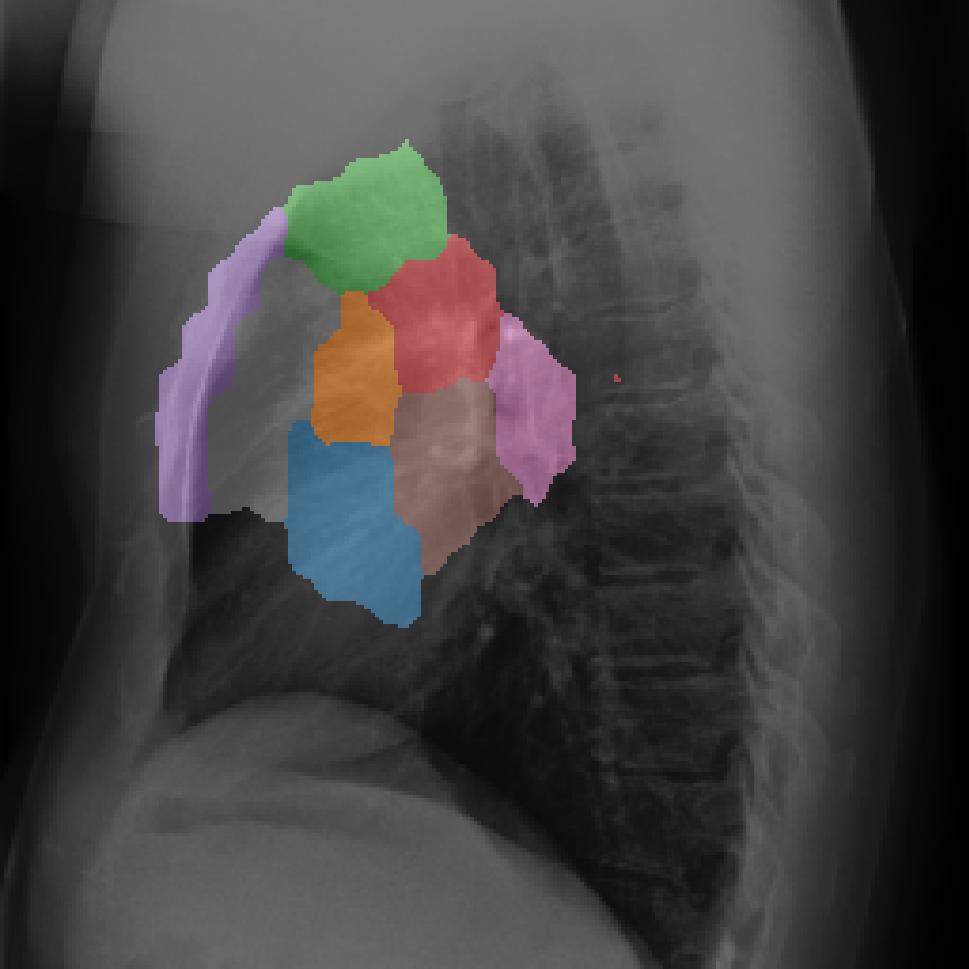} & \includegraphics[width=0.088\linewidth]{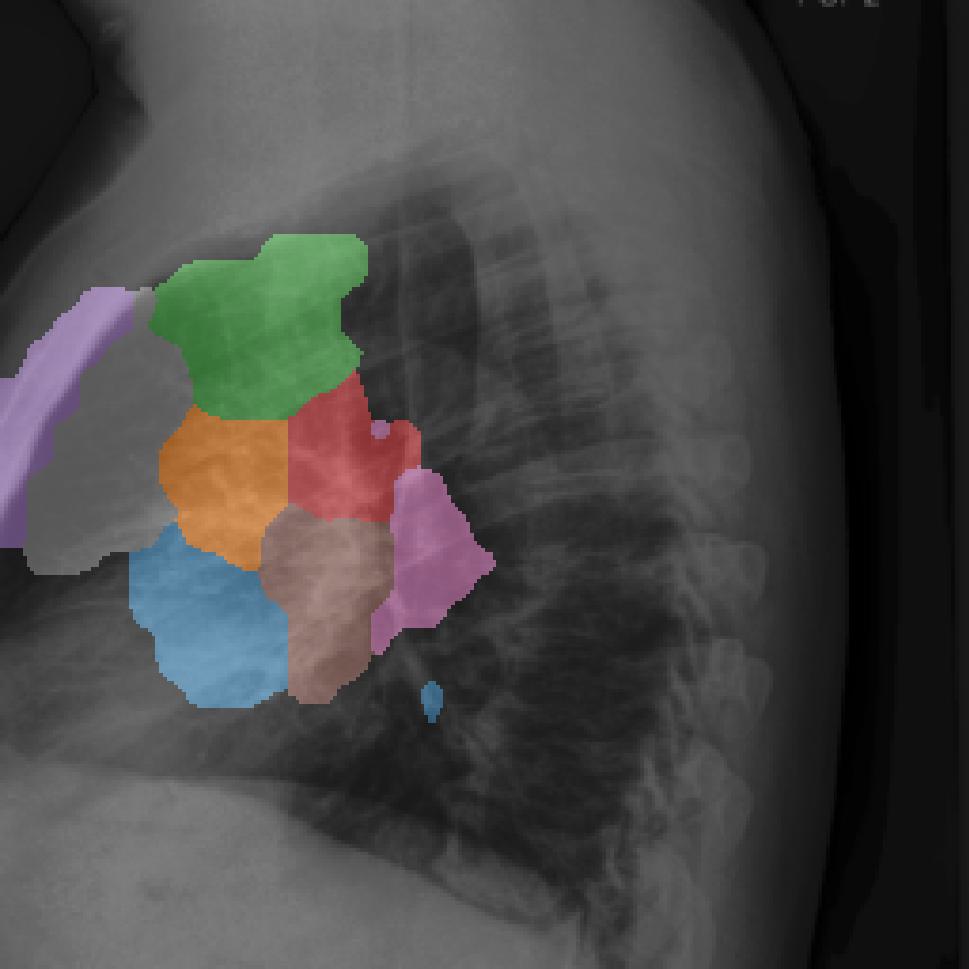} & \includegraphics[width=0.088\linewidth]{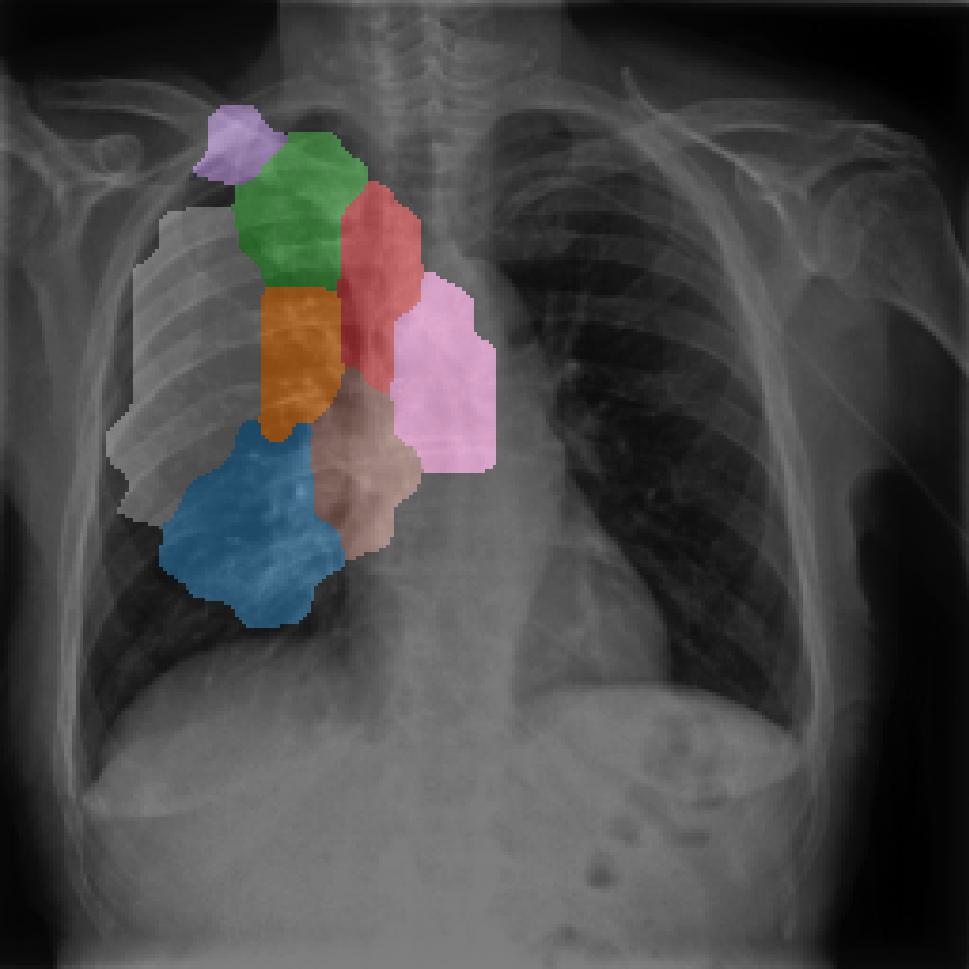} & \includegraphics[width=0.088\linewidth]{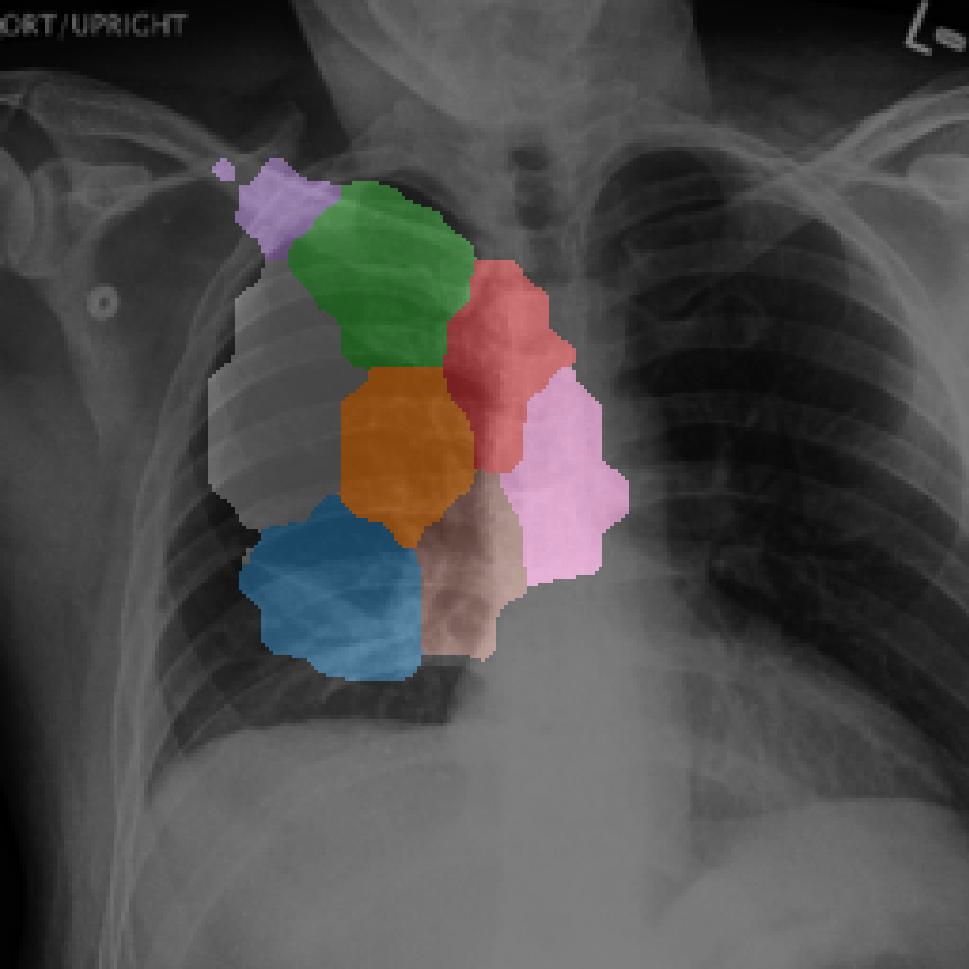} & \includegraphics[width=0.088\linewidth]{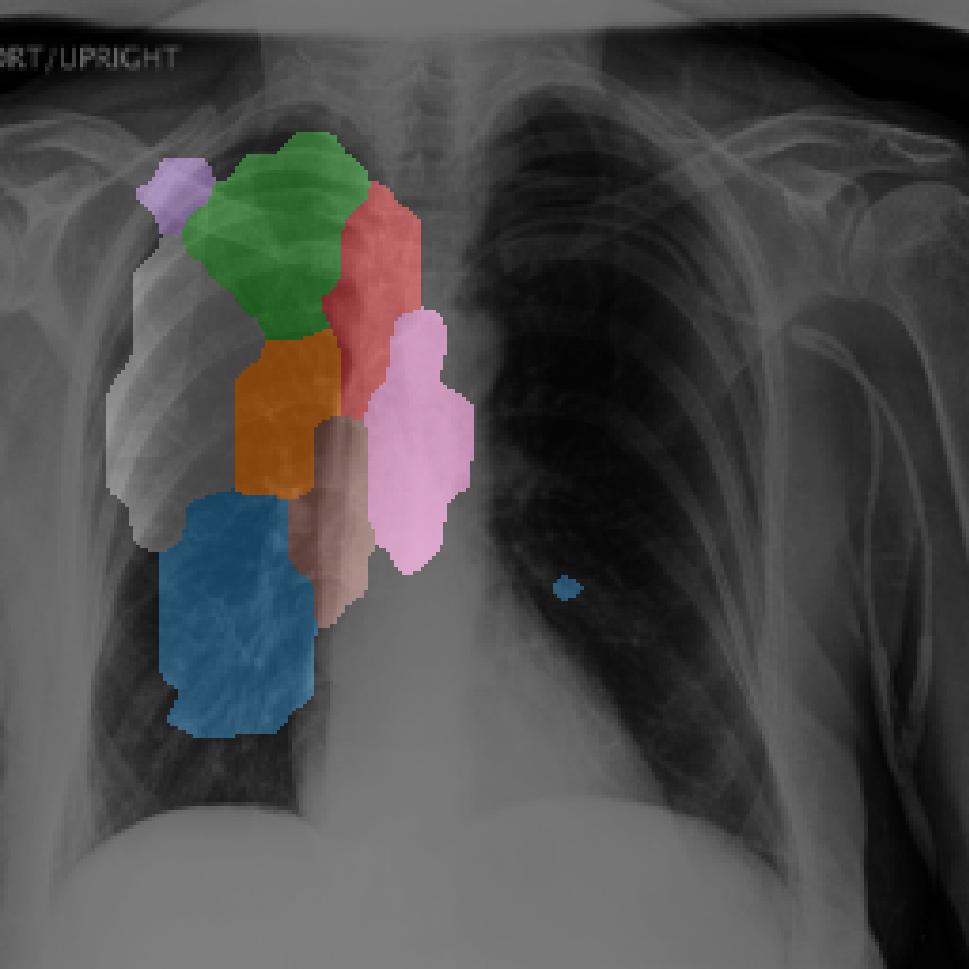} & \includegraphics[width=0.088\linewidth]{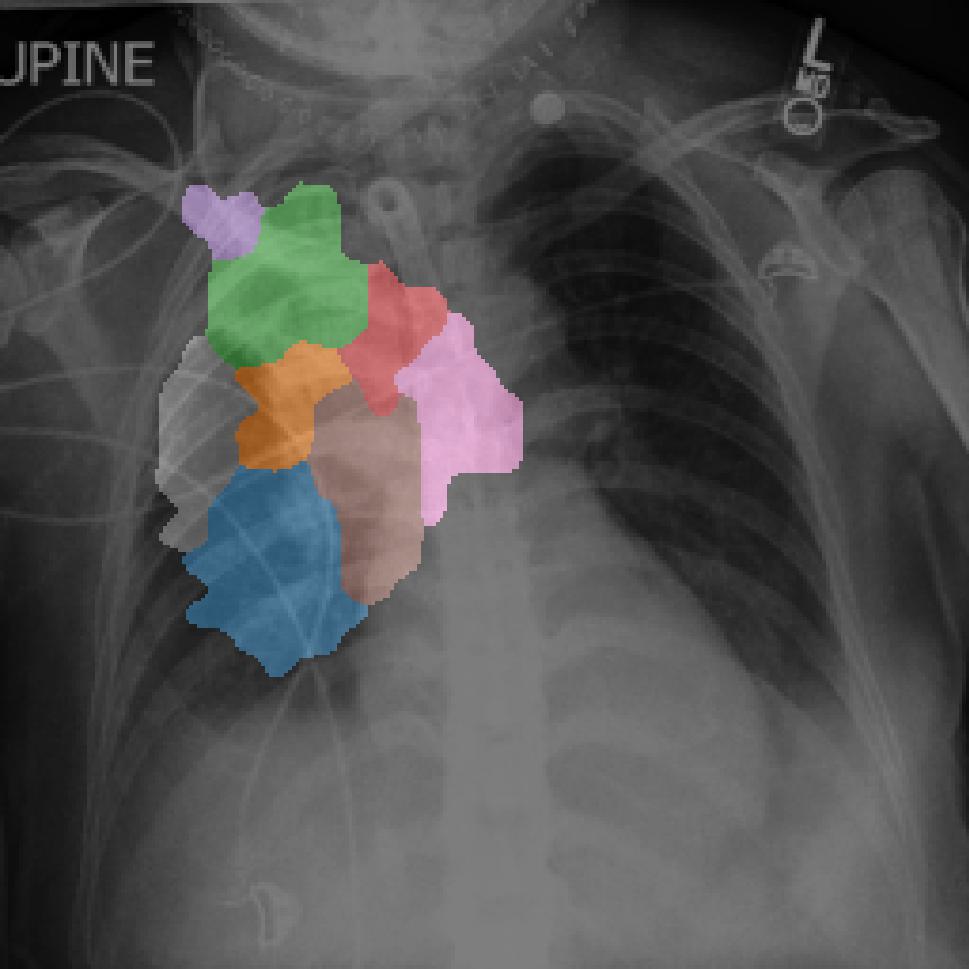} \\
    
    \rotatebox{90}{\makebox[1cm][c]{\scriptsize Hard}} & \includegraphics[width=0.088\linewidth]{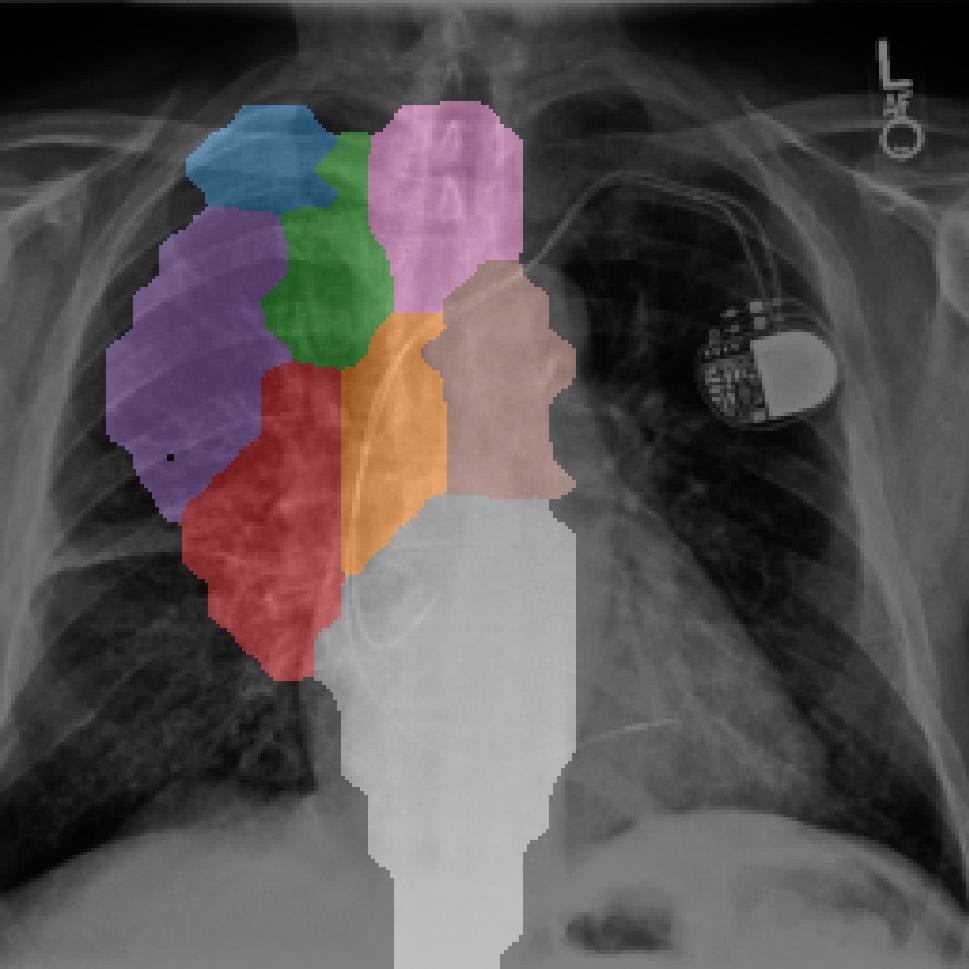} & \includegraphics[width=0.088\linewidth]{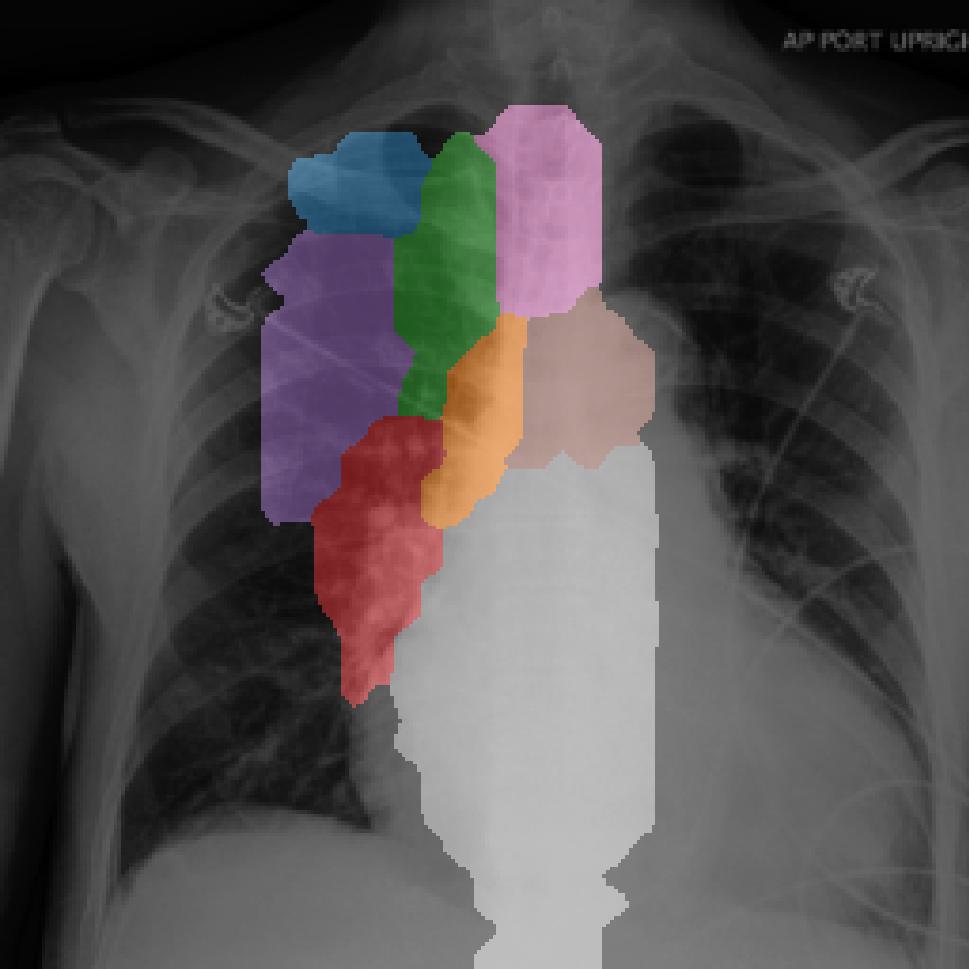} & \includegraphics[width=0.088\linewidth]{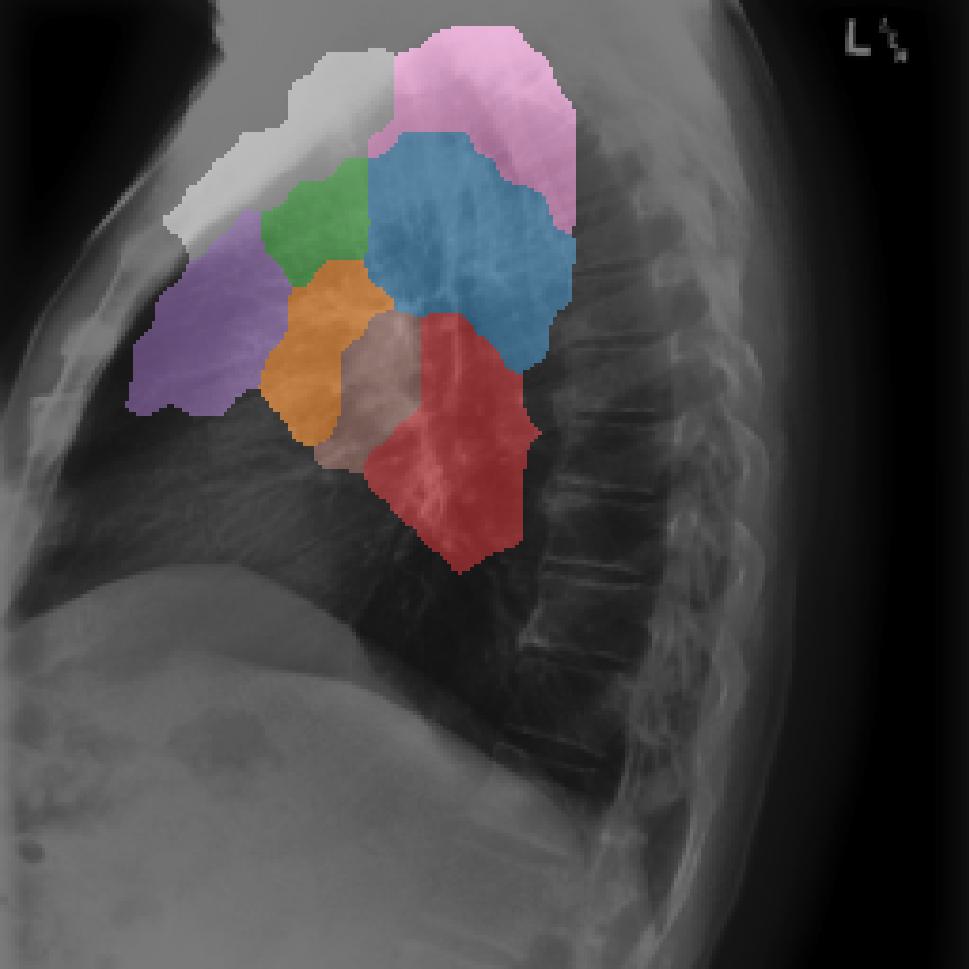} & \includegraphics[width=0.088\linewidth]{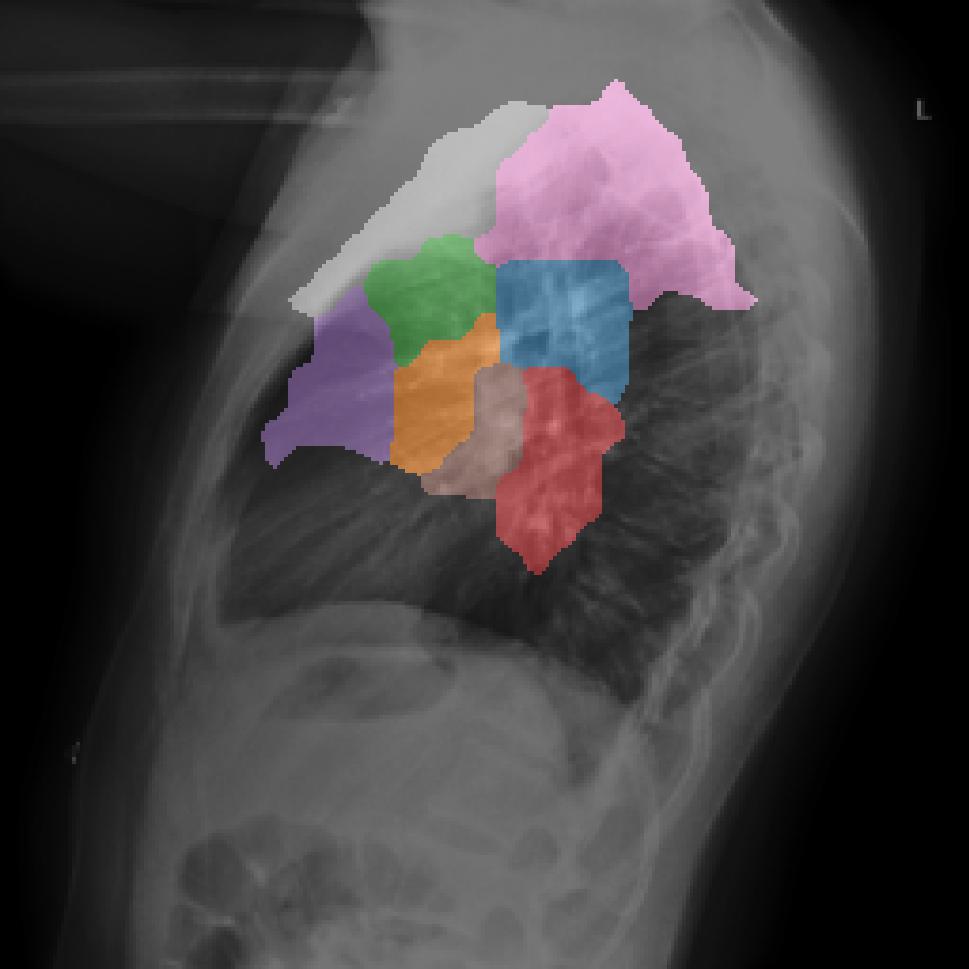} & \includegraphics[width=0.088\linewidth]{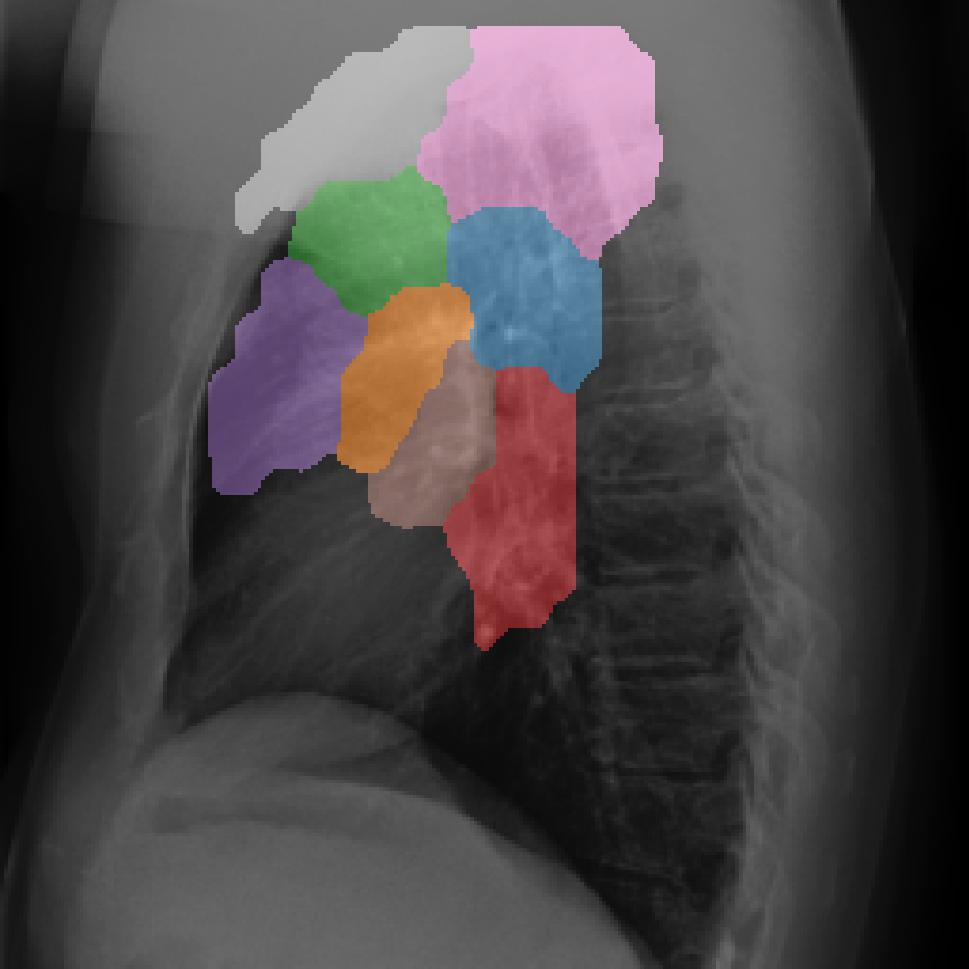} & \includegraphics[width=0.088\linewidth]{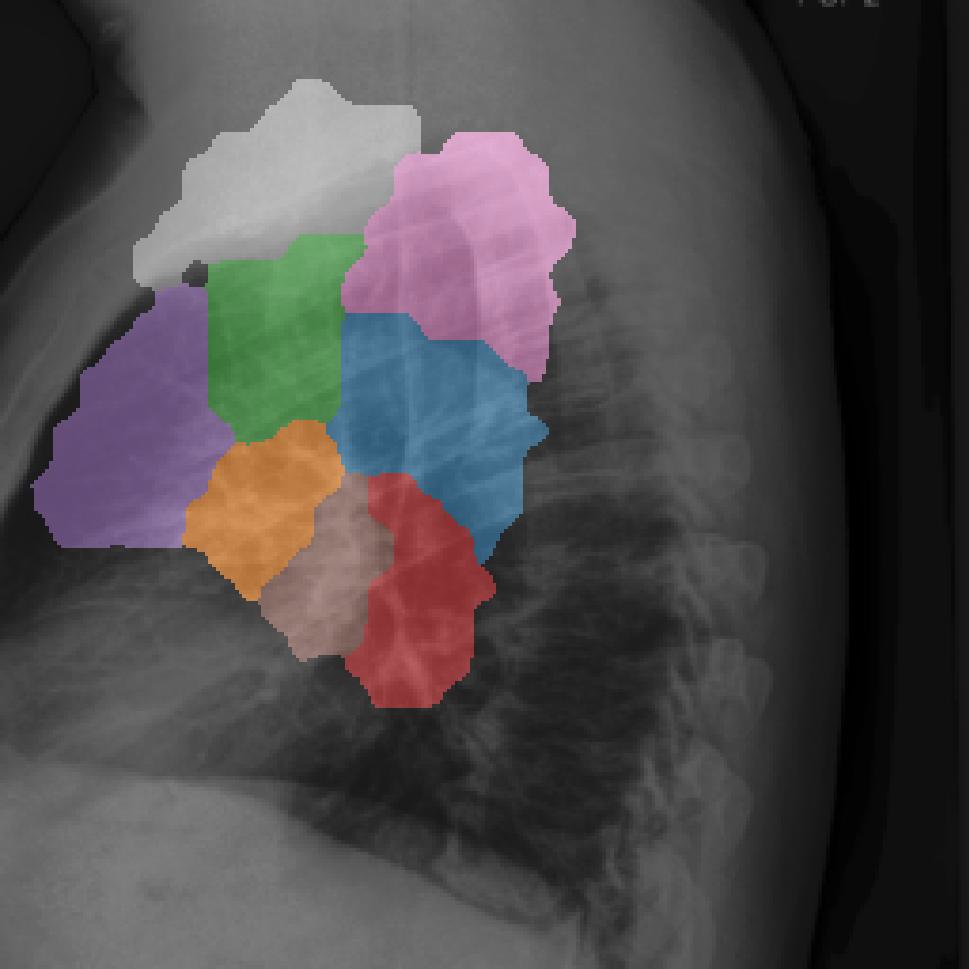} & \includegraphics[width=0.088\linewidth]{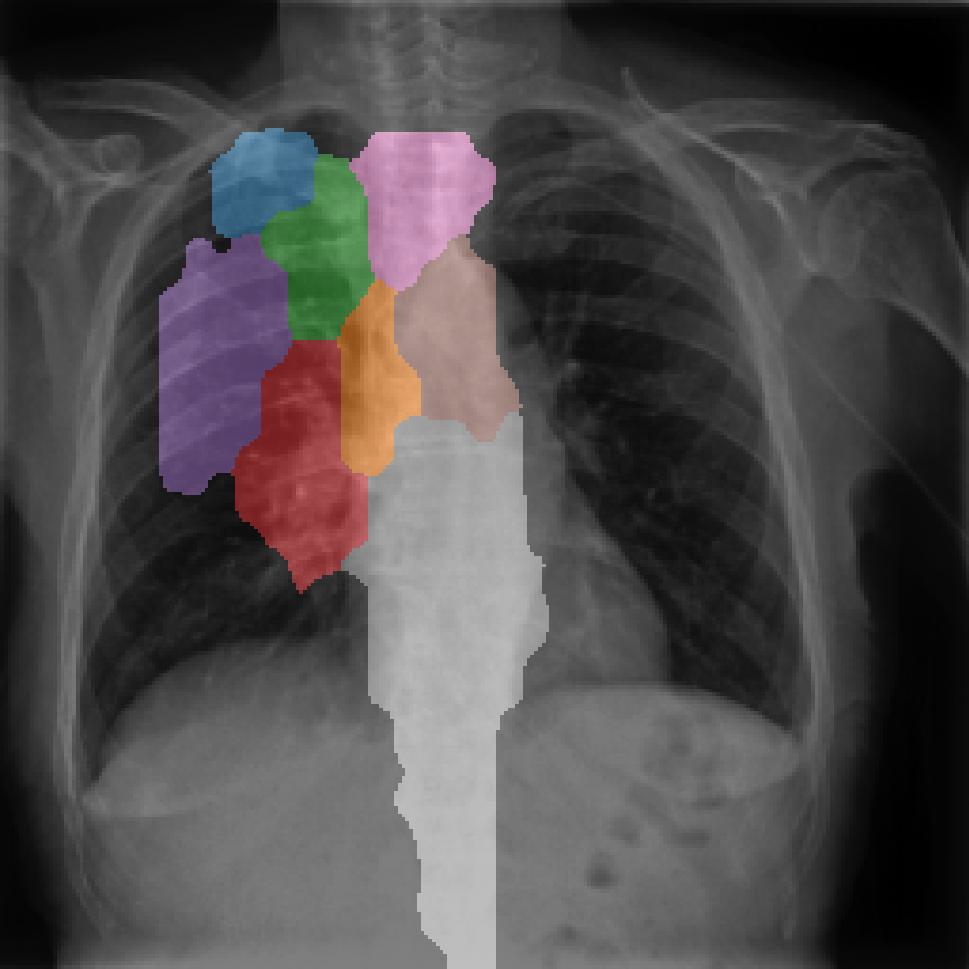} & \includegraphics[width=0.088\linewidth]{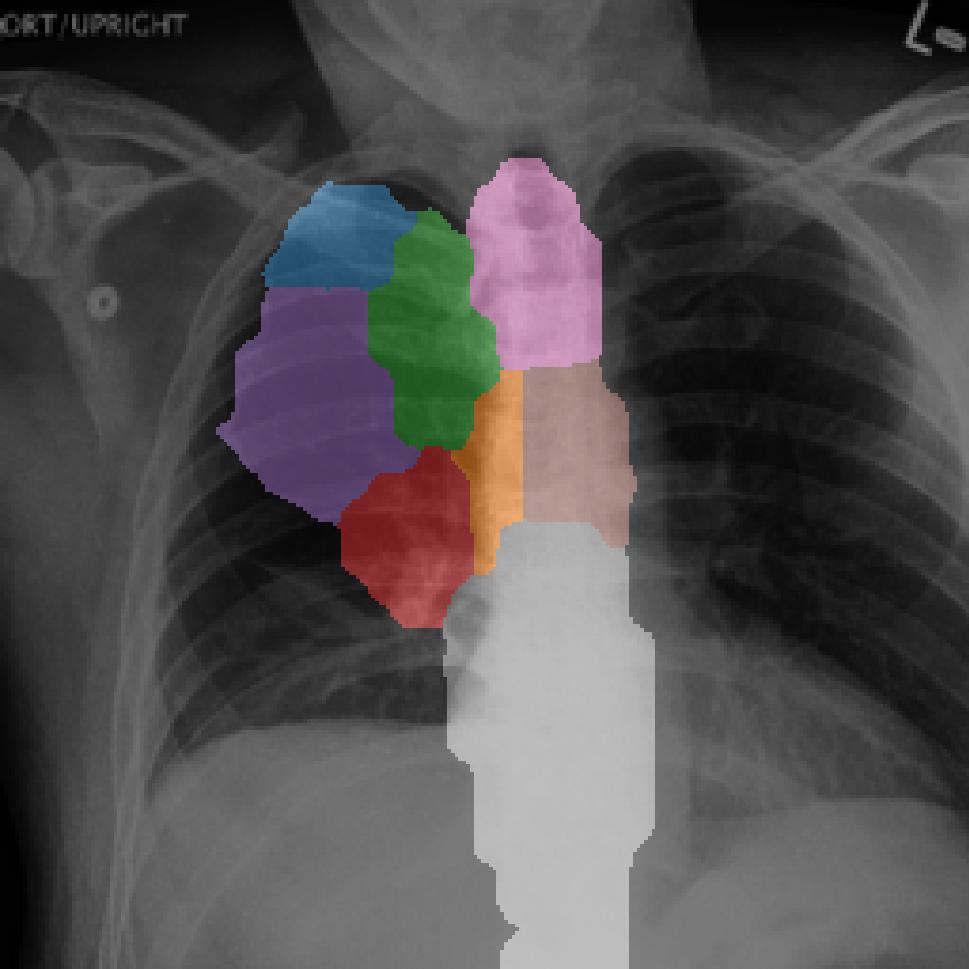} & \includegraphics[width=0.088\linewidth]{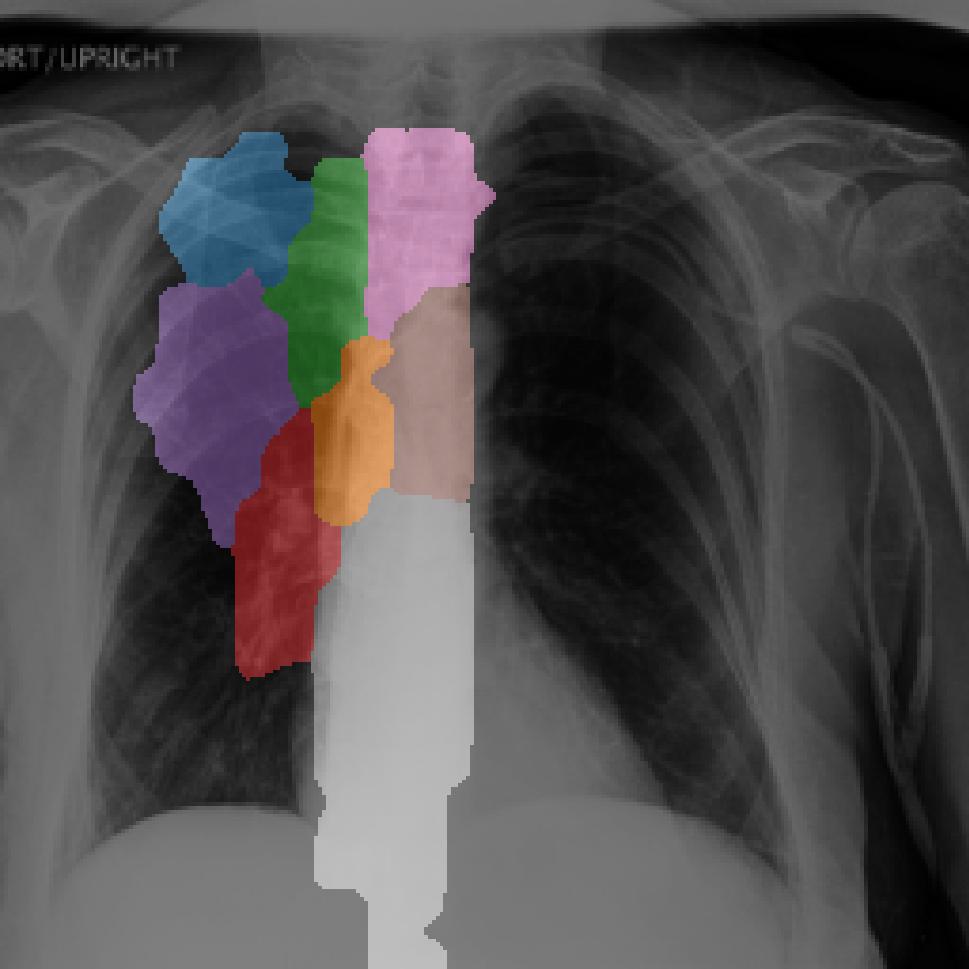} & \includegraphics[width=0.088\linewidth]{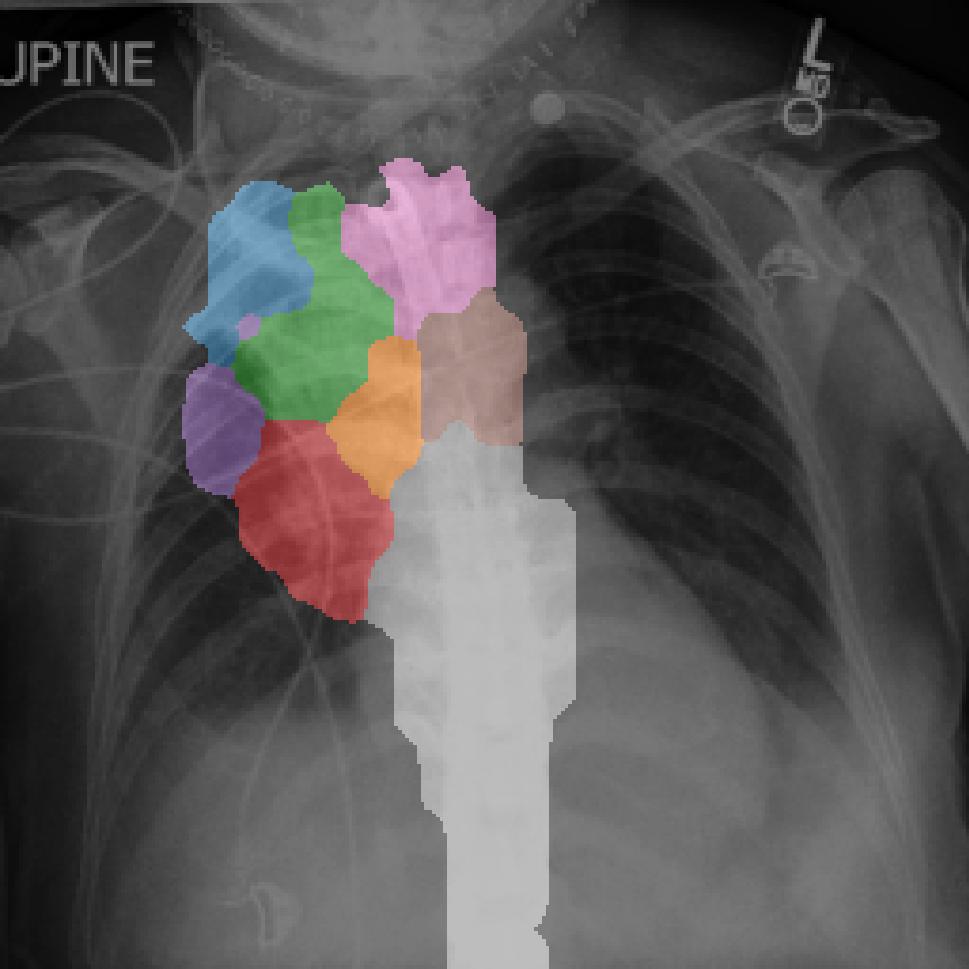} \\

    \rotatebox{90}{\makebox[1cm][c]{\scriptsize ST}} & \includegraphics[width=0.088\linewidth]{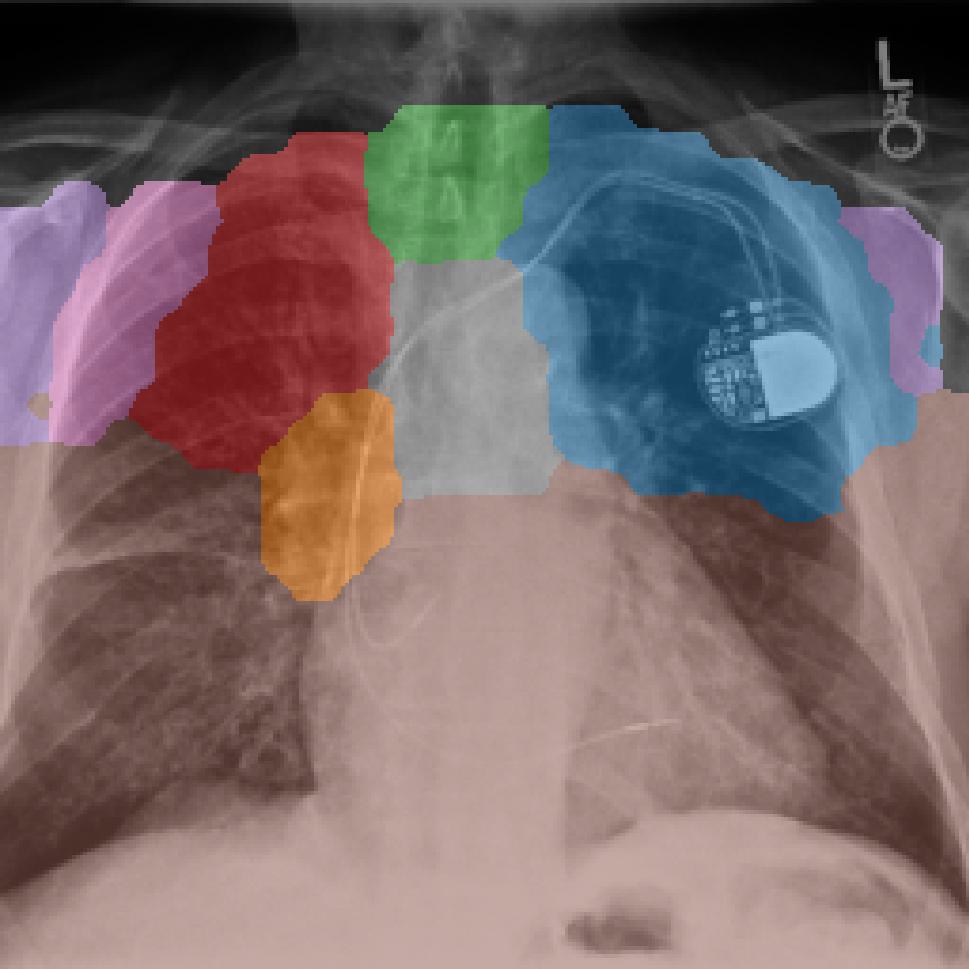} & \includegraphics[width=0.088\linewidth]{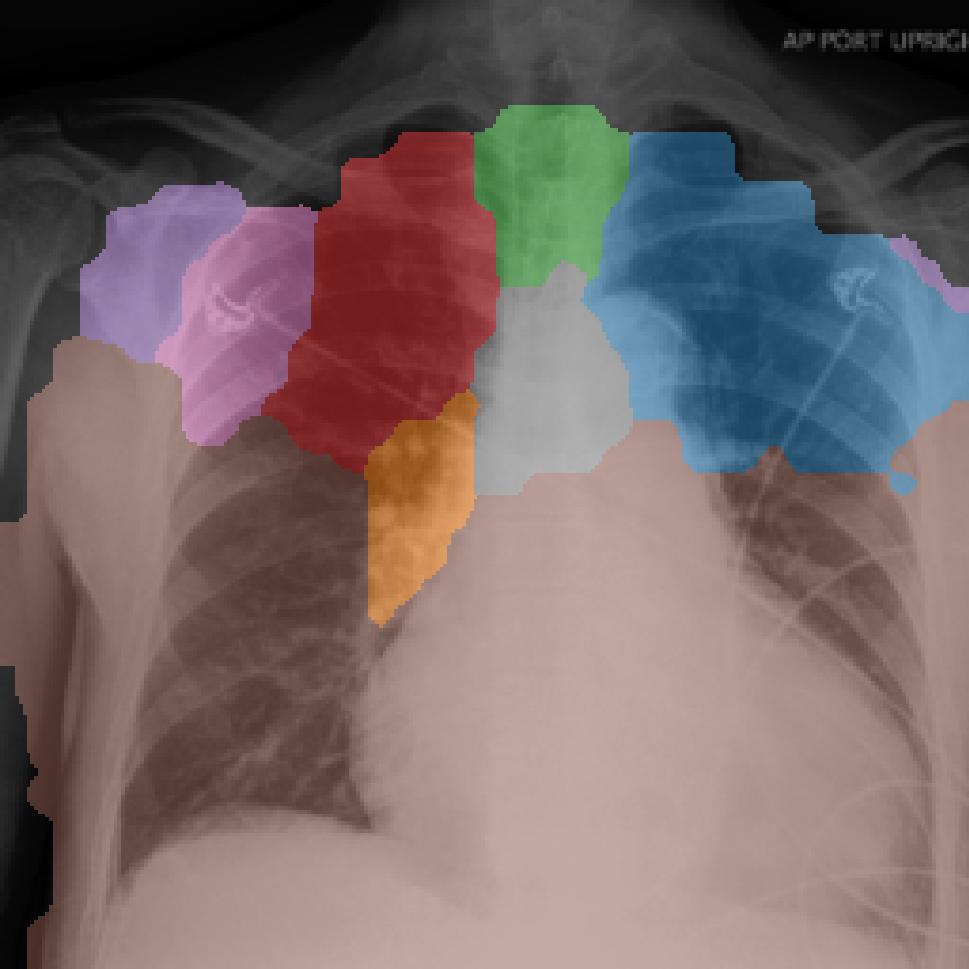} & \includegraphics[width=0.088\linewidth]{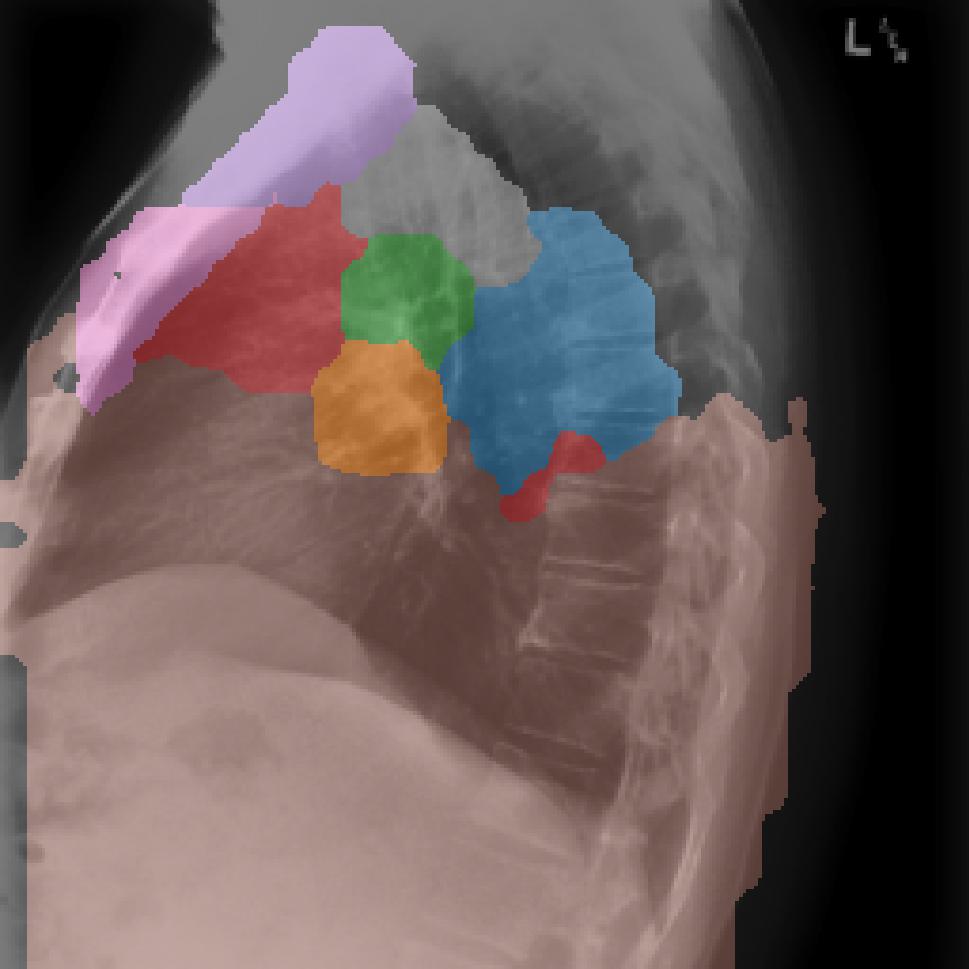} & \includegraphics[width=0.088\linewidth]{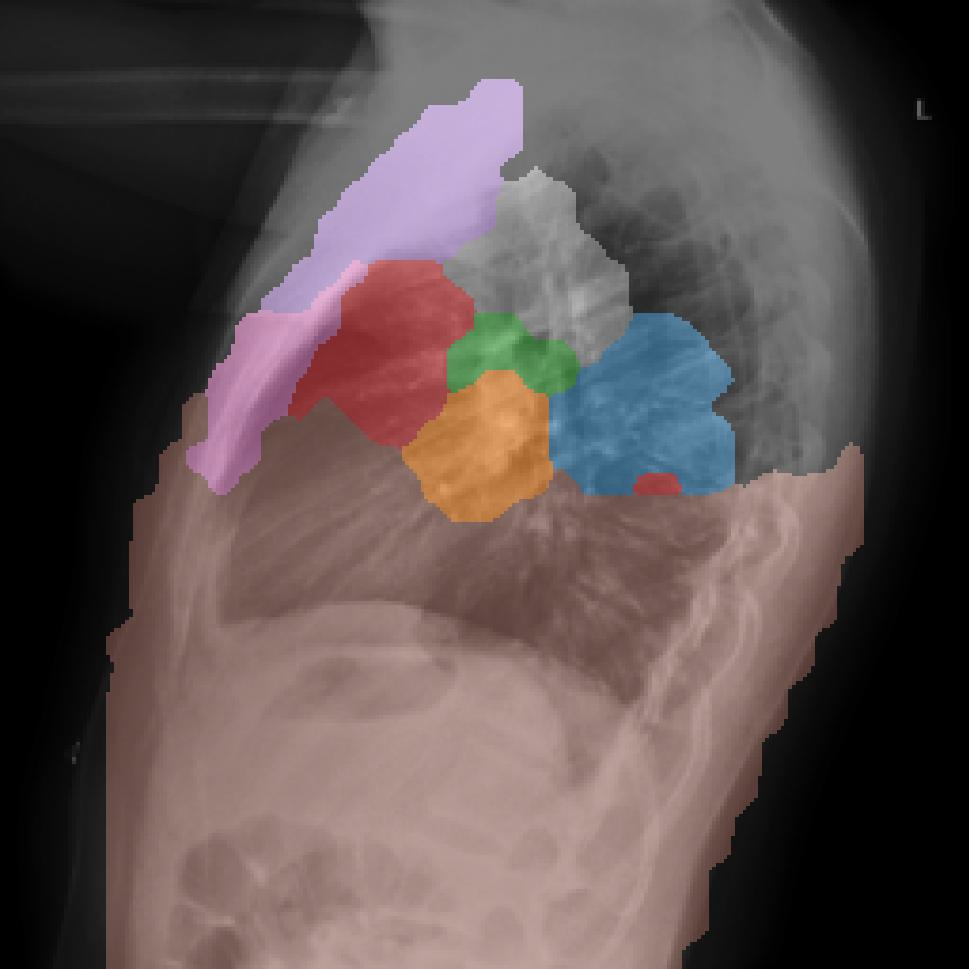} & \includegraphics[width=0.088\linewidth]{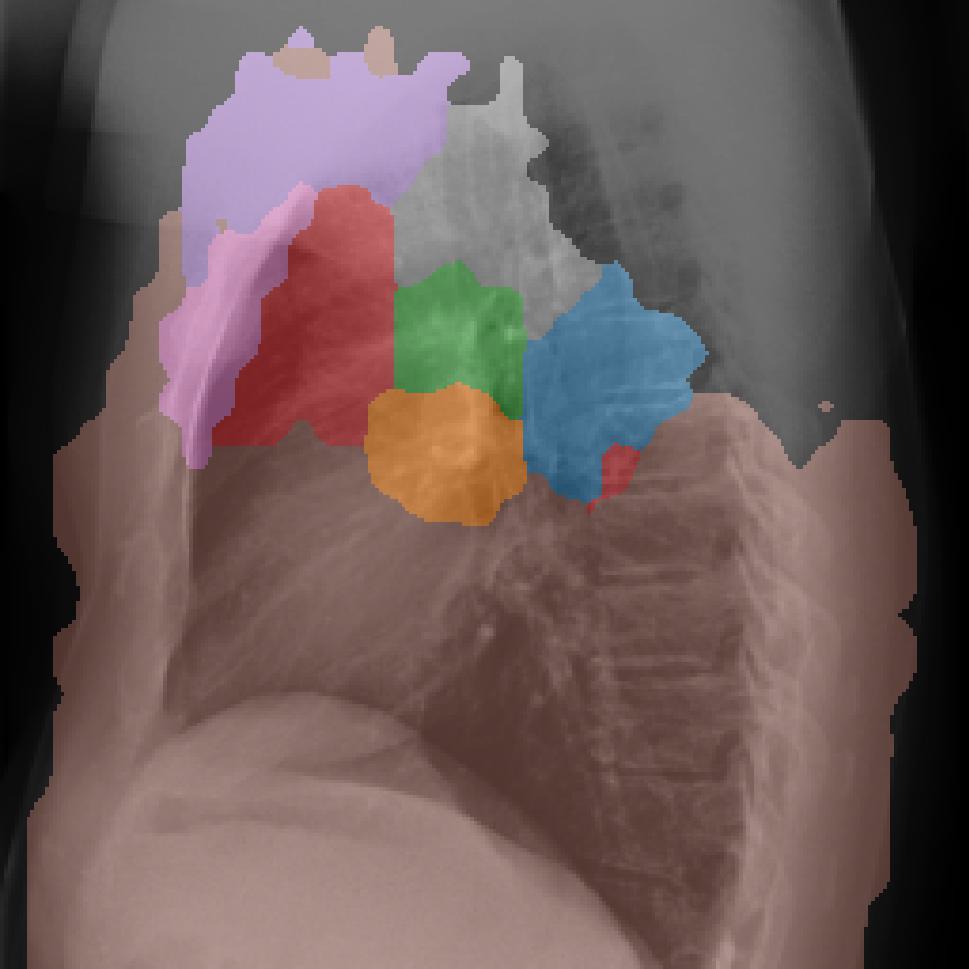} & \includegraphics[width=0.088\linewidth]{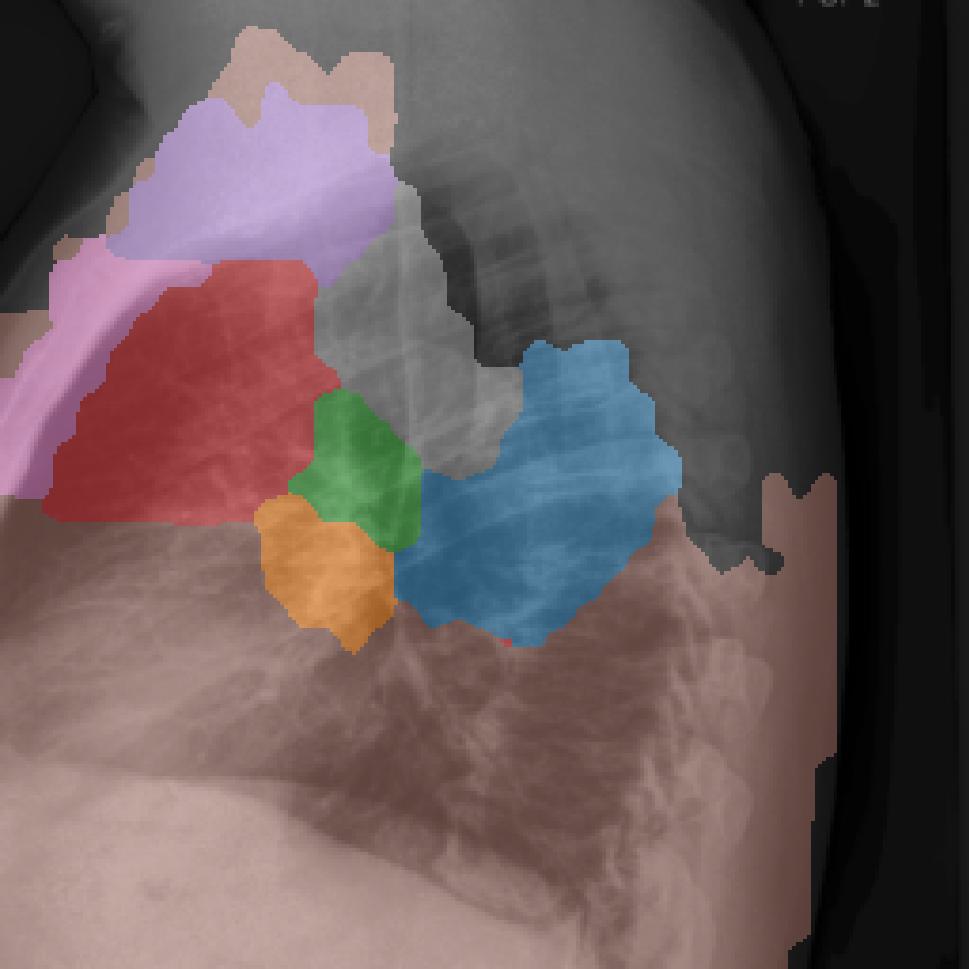} & \includegraphics[width=0.088\linewidth]{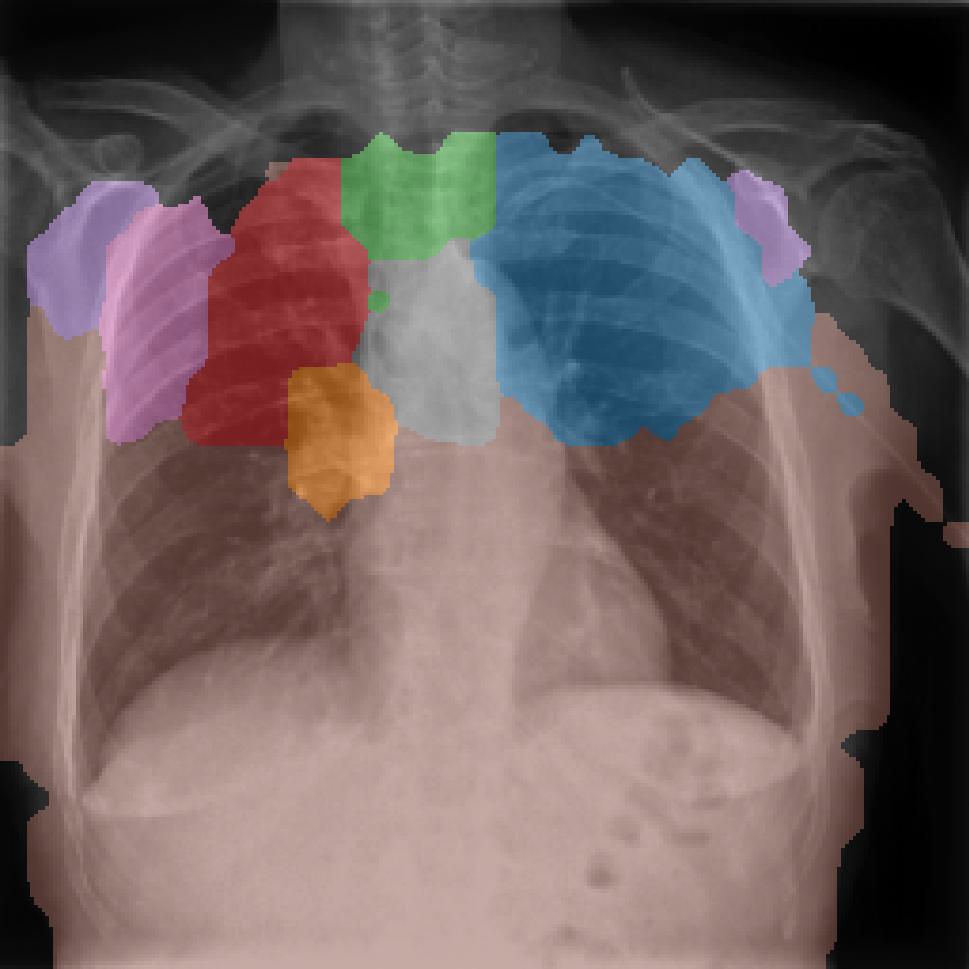} & \includegraphics[width=0.088\linewidth]{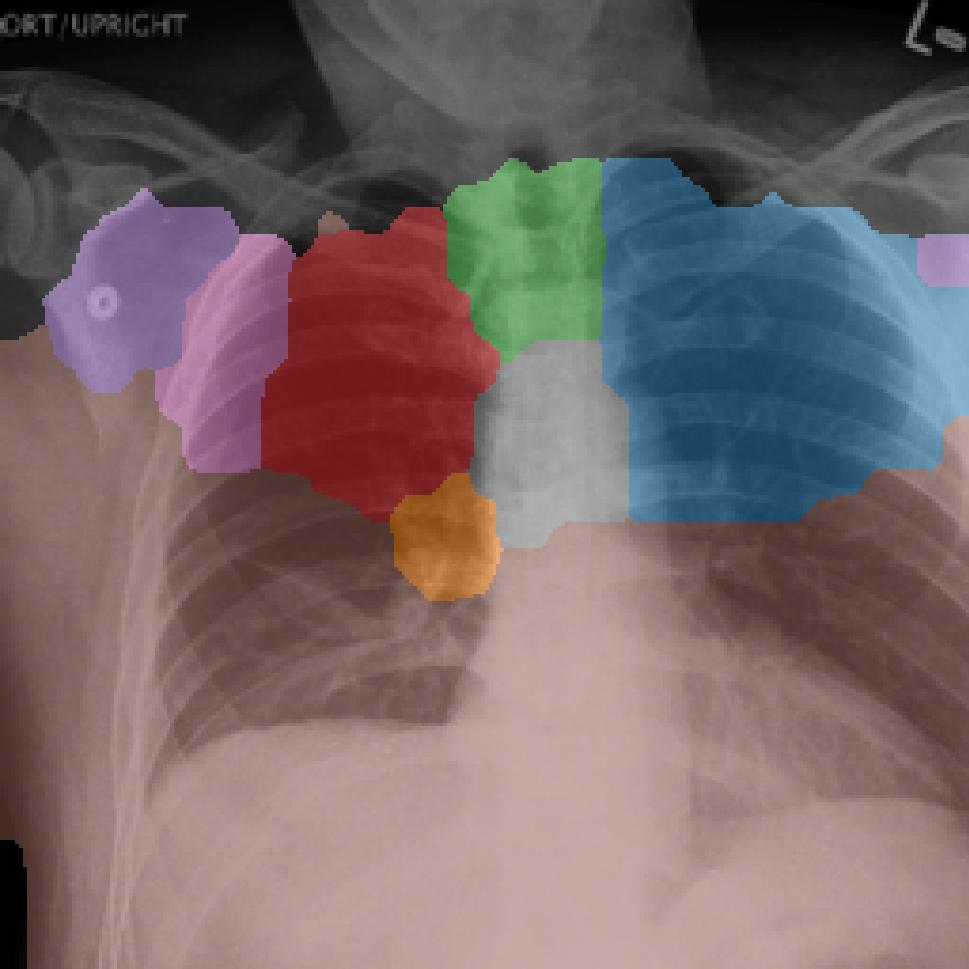} & \includegraphics[width=0.088\linewidth]{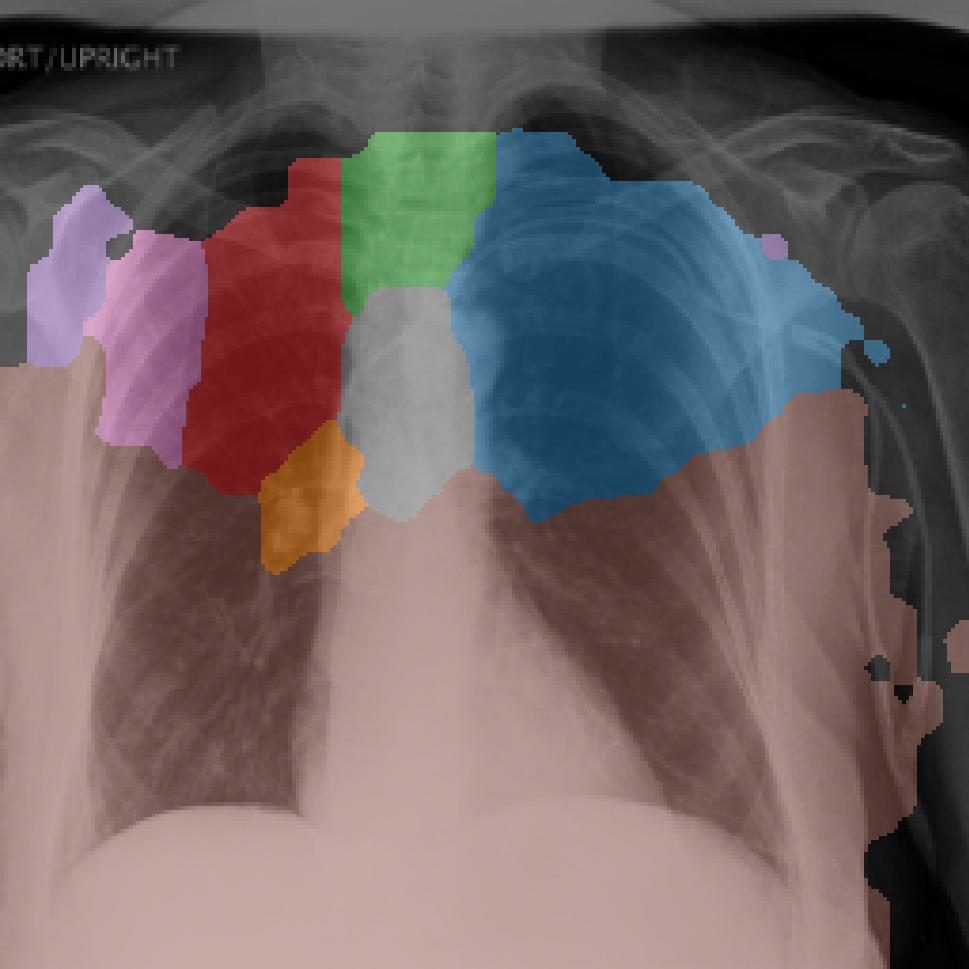} & \includegraphics[width=0.088\linewidth]{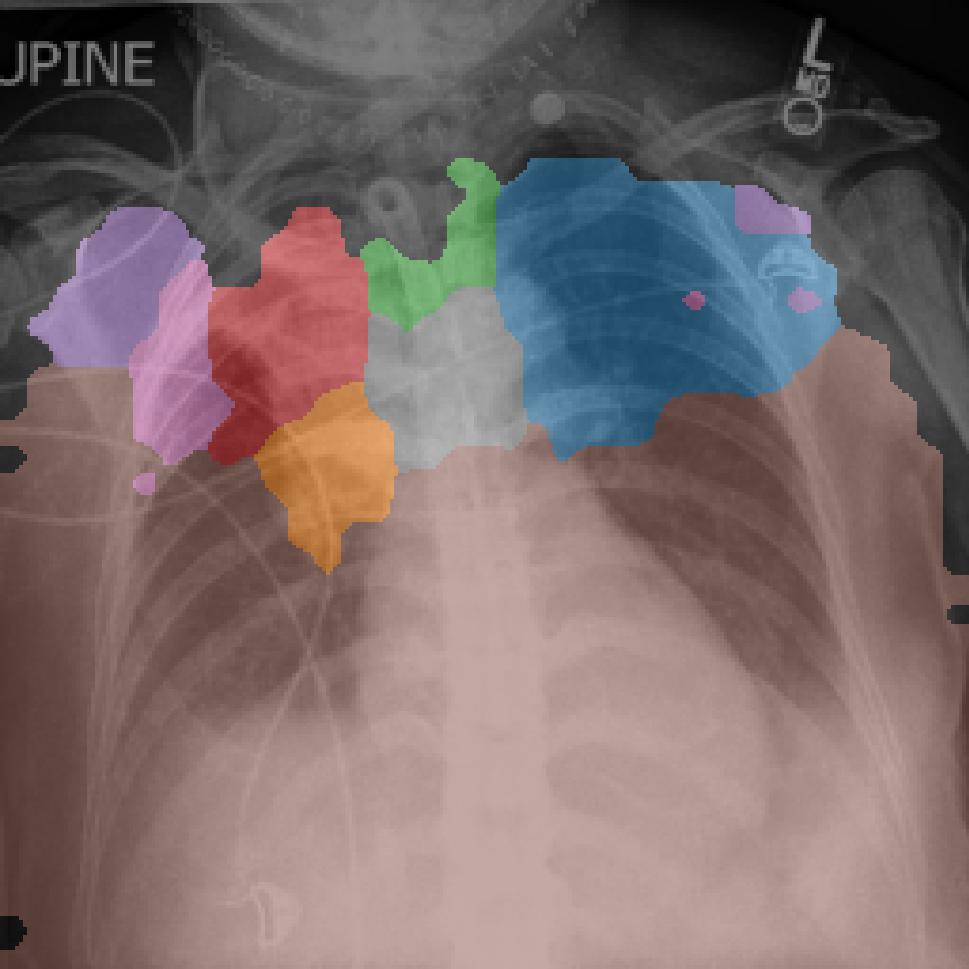} \\

  \bottomrule
  \end{tabular}
  \caption{Additional qualitative results on CheXpert.}
  \label{tab:chexpert}
\end{figure}

We provide additional randomly-selected qualitative results on CelebA, CUB and CheXpert in Figures \ref{tab:celeba}, \ref{tab:class_level_cub} and \ref{tab:chexpert} respectively, in order to allow a qualitative assessment of the part discovery consistency. Note that colour coding of parts is only consistent within each model and not across models.

\section{Extended discussion on mask expansion and leakage}
\label{sec_supp:mask_expansion}

In the main text, we note that our two-stage Straight-Through Estimator (STE) variant produces more spatially expanded masks that better align with semantic ground-truth. Here, we provide a deeper qualitative analysis of why mitigating intra-object leakage directly prevents ``mask collapse''.

\textbf{Insufficient mask expansion in leaky models.} In standard single-stage models, or two-stage models with soft masking, the receptive fields allow information to bleed across spatial locations. Consequently, the part discovery module becomes ``lazy''. To accurately predict an attribute, the model does not need to segment the entire relevant anatomical region; it only needs to highlight a small, highly discriminative patch, relying on feature leakage to supply the rest of the context.

\textbf{Insufficient mask expansion with hard masks.} On the other hand, pure hard masks in a 2-stage setting result in limited learning signal flowing to the part discovery module: gradients only flow through the currently selected tokens, meaning that the second stage can give a signal to shrink the current part, but not to expand it, potentially resulting in missed to too small parts.

\textbf{The CheXpert Left/Right Lung Phenomenon.} This lazy masking behavior is starkly visible in our CheXpert experiments (Fig.~\ref{tab:chexpert}). We observed that the single-stage baseline, as well as the two-stage hard and soft variants, frequently collapsed their part discovery to focus entirely on the left lung, effectively ignoring the right lung. Because features were either leaking (single-stage/soft) or gradients were blocked (hard), the masks never expanded. In contrast, the STE variant successfully discovered parts covering both lungs. By enforcing strict isolation in the forward pass while maintaining continuous gradients in the backward pass, STE allows the mask to expand over all relevant predictive tokens to minimize the attribute loss.

\textbf{Observations on CUB and CelebA.} We observe the exact same dynamic on CelebA (\cref{tab:celeba}) and CUB (\cref{tab:class_level_cub}). The leaky 1-stage baseline produces smaller attention masks that fail to cover the full anatomy, often leaving out peripheral parts like the hair in CelebA and the legs in CUB. STE expands these masks into more coherent and complete partitions.

\textbf{Minor Limitations in Peripheral Discovery.} While STE provides superior part discovery overall, manual inspection reveals a minor limitation regarding peripheral parts. On the CelebA dataset (\cref{tab:celeba}), parts located at the extreme periphery of the image (specifically the ears and outer edges of hair) are sometimes imperfectly segmented. We hypothesize this is due to frequent truncation at the image borders along with a trade-off to satisfy the background loss. However, because our architecture isolates whatever mask is discovered, this minor geometric truncation does not compromise the faithfulness of the downstream attribute prediction: if the model fails to discover one part that is relevant to the attribute prediction task, we will be able to notice it due to the drop in attribute prediction performance.

\section{Trade-offs, limitations, and ensemble performance}
\label{sec_supp:tradeoffs}

\begin{table}[t]
\centering
\caption{\textbf{Computational Overhead.} Comparison of inference throughput, memory footprint, and computational operations (GFLOPS) for a single image (batch size 1, input resolution $224 \times 224$) on an Nvidia RTX 3090 GPU. The two-stage model requires more computation to process parts in isolated attention streams, resulting in a roughly $2.1\times$ decrease in throughput.}
\label{tab_supp:compute_overhead}
\begin{adjustbox}{max width=\linewidth}
\begin{tabular}{@{}l ccc@{}}
\toprule
\textbf{Variant} & \textbf{Speed} & \textbf{Memory} & \textbf{GFLOPS} \\
 & im/s & MiB & \\
\midrule
Single-Stage & 310 & 23.25 & 17.21 \\
Two-Stage & 143 & 28.89 & 37.82 \\
\bottomrule
\end{tabular}
\end{adjustbox}
\end{table}

\begin{table}[t]
\centering
\caption{\textbf{Absolute attribute prediction performance.} While enforcing strict spatial locality (Stage 2 Only) removes unfaithful spatial shortcuts and can lead to a drop in raw accuracy (e.g., on CelebA and CheXpert), ensembling the global context (Stage 1) with the strictly local evidence (Stage 2) yields the highest overall performance across all datasets.}
\label{tab_supp:ensemble_performance}
\begin{adjustbox}{max width=\linewidth}
\begin{tabular}{@{}l ccc@{}}
\toprule
\textbf{Variant} & \textbf{CUB} & \textbf{CelebA} & \textbf{CheXpert} \\
 & mAP & mAP & AUROC \\
\midrule
Single-Stage & 34.7 & 72.0 & 87.6 \\
\midrule
Two-Stage (Stage 2 Only) & 34.2 & 63.5 & 81.0 \\
Two-Stage (Ensemble) & \textbf{36.3} & \textbf{73.9} & \textbf{88.4} \\
\bottomrule
\end{tabular}
\end{adjustbox}
\end{table}

In the main paper, we demonstrate that enforcing strict spatial locality via our two-stage architecture successfully mitigates intra-object leakage. While this architectural guarantee introduces an inherent computational overhead, we find that it ultimately offers a highly flexible trade-off between strict faithfulness and absolute predictive performance.

\textbf{Compute overhead.} In terms of inference speed, our two-stage method (incorporating Stage 1 part discovery and Stage 2 strict parallel masking) is roughly $2.1\times$ slower than the standard single-stage baseline (Table~\ref{tab_supp:compute_overhead}). For instance, evaluating models trained on CelebA at a batch size of 1 with an input resolution of $224 \times 224$, throughput drops from 310 images/second for the single-stage baseline down to 143 images/second for the two-stage model. This increase in computational cost (17.21 to 37.82 GFLOPS) is a direct consequence of the architectural design required to prevent leakage. In the standard baseline, the image is passed through the vision transformer once, and features are aggregated at the end. In our approach, to guarantee that the feature representation of the ``head'' is not contaminated by the ``wing'', the Stage 2 ViT must process the prefix tokens for all $K$ discovered parts (plus the background) in parallel, isolated attention streams.

\textbf{The cost of strict locality.} For CelebA and CheXpert, relying \textit{solely} on the strictly isolated Stage 2 representations results in a lower absolute attribute prediction accuracy compared to the unconstrained single-stage baseline (Table~\ref{tab_supp:ensemble_performance}). This is a consequence of mitigating metonymy for attributes that can be better predicted using context beyond their corresponding part. Standard backbones achieve artificially high predictive performance precisely because they exploit intra-object leakage: they utilize highly correlated visual cues from the entire object to make guesses. By severing these attention links, our Stage 2 module forces the classifier to rely only on local evidence, leading to a natural drop in raw accuracy that directly reflects the baseline's reliance on unfaithful spatial shortcuts. Remarkably, attribute prediction on CUB suffers only a minimal performance degradation (a drop of 0.5 mAP) when using early masking, which suggests that the fine-grained attributes in this dataset are already described well enough by the purely local representation.

\textbf{Performance recovery via ensembling.} However, our architecture does not force a strict compromise between interpretability and accuracy in practice. Because our model natively computes attribute predictions at both the global level (Stage 1, prior to masking) and the strictly local level (Stage 2, post-masking), practitioners can ensemble these signals. We empirically find that summing the logits from both stages allows the two-stage model to cleanly surpass the pure single-stage baseline in absolute predictive performance across all datasets (Table~\ref{tab_supp:ensemble_performance}). This demonstrates that the unconstrained global context (Stage 1) and the strictly faithful local evidence (Stage 2) provide highly complementary predictive signals, yielding a model that is both highly accurate and capable of strictly localized reasoning.

\textbf{Discussion.} While a roughly $2.1\times$ inference slowdown introduces a computational cost, our ensembling results demonstrate that this architecture ultimately does not force a compromise between raw performance and interpretability. For instance, in our CheXpert~\cite{irvin2019chexpert} medical imaging experiments, trading inference speed not only yields a guarantee that the model is diagnosing a pathology based on the correct anatomical region (evidenced by the MPPO spatial grounding score jumping from 0.58 to 0.76), but also improves the overall predictive AUROC from 87.6 to 88.4. This proves that mitigating metonymy and enforcing constrained, local reasoning is not just a safety-critical necessity, but also a pathway to superior overall model performance.

\section{Part-Attribute Groupings}
\label{sec_supp:attribute_groupings}

\begin{table}[t]
\centering
\caption{\textbf{Semantic Part to Keypoint Groupings.} The exact mapping of the original dataset keypoints into the merged semantic macro-regions. These groupings align the spatial keypoints with the attribute vocabularies and are used to evaluate part discovery quality (NMI, ARI) and spatial contingency.}
\label{tab_supp:keypoint_groupings}

\vspace{0.5em}
\begin{minipage}[t]{0.5\textwidth}
\centering
\textbf{CelebA Keypoints} \\
\vspace{0.2em}
\begin{adjustbox}{max width=\linewidth}
\begin{tabular}{@{}ll@{}}
\toprule
\textbf{Merged} & \multicolumn{1}{c}{\textbf{Original}} \\
\midrule
\textbf{Eye} & left eye, right eye \\
\textbf{Nose} & nose \\
\textbf{Mouth} & left mouth, right mouth (averaged) \\
\bottomrule
\end{tabular}
\end{adjustbox}
\end{minipage}\hfill
\begin{minipage}[t]{0.45\textwidth}
\centering
\textbf{CUB Keypoints} \\
\vspace{0.2em}
\begin{adjustbox}{max width=\linewidth}
\begin{tabular}{@{}lp{0.6\linewidth}@{}}
\toprule
\textbf{Merged} & \multicolumn{1}{c}{\textbf{Original}} \\
\midrule
\textbf{Back} & back \\
\textbf{Belly} & belly \\
\textbf{Breast} & breast \\
\textbf{Head} & beak, crown, eye, forehead, nape, throat \\
\textbf{Leg} & left leg, right leg \\
\textbf{Tail} & tail \\
\textbf{Wing} & left wing, right wing \\
\bottomrule
\end{tabular}
\end{adjustbox}
\end{minipage}
\end{table}

\begin{table}[t]
\centering
\caption{\textbf{CelebA Part-Attribute Groupings.} The exact mapping of the 32 localized facial attributes to their corresponding 8 semantic parts.}
\label{tab_supp:celeba_attributes}
\begin{adjustbox}{max width=\linewidth}
\begin{tabular}{@{}lp{0.8\textwidth}@{}}
\toprule
\textbf{Semantic Part} & \multicolumn{1}{c}{\textbf{Associated Attributes}} \\
\midrule
\textbf{Hair} & Bald, Bangs, Black Hair, Blond Hair, Brown Hair, Gray Hair, Receding Hairline, Straight Hair, Wavy Hair \\
\textbf{Hat} & Wearing Hat \\
\textbf{Eye} & Arched Eyebrows, Bags Under Eyes, Bushy Eyebrows, Narrow Eyes \\
\textbf{Eyeglasses} & Eyeglasses \\
\textbf{Nose} & Big Nose, Pointy Nose \\
\textbf{Mouth / Jaw} & Big Lips, Mouth Slightly Open, Smiling, Wearing Lipstick, 5 o' Clock Shadow, Double Chin, Goatee, Mustache, No Beard, Sideburns, High Cheekbones, Rosy Cheeks \\
\textbf{Ear} & Wearing Earrings \\
\textbf{Neck} & Wearing Necklace, Wearing Necktie \\
\bottomrule
\end{tabular}
\end{adjustbox}
\end{table}

\begin{table}[t]
\centering
\caption{\textbf{CUB Part-Attribute Groupings.} The exact mapping of the 248 localized fine-grained attributes to the 7 semantic bird parts.}
\label{tab_supp:cub_attributes}
\begin{adjustbox}{max width=\linewidth}
\begin{tabular}{@{}lp{0.8\textwidth}@{}}
\toprule
\textbf{Semantic Part} & \multicolumn{1}{c}{\textbf{Associated Attributes}} \\
\midrule
\textbf{Back} 
& \textbf{has\_back\_color:} black, blue, brown, buff, green, grey, iridescent, olive, orange, pink, purple, red, rufous, white, yellow \newline
\textbf{has\_back\_pattern:} multi-colored, solid, spotted, striped \\
\midrule
\textbf{Belly} 
& \textbf{has\_belly\_color:} black, blue, brown, buff, green, grey, iridescent, olive, orange, pink, purple, red, rufous, white, yellow \newline
\textbf{has\_belly\_pattern:} multi-colored, solid, spotted, striped \\
\midrule
\textbf{Breast} 
& \textbf{has\_breast\_color:} black, blue, brown, buff, green, grey, iridescent, olive, orange, pink, purple, red, rufous, white, yellow \newline
\textbf{has\_breast\_pattern:} multi-colored, solid, spotted, striped \\
\midrule
\textbf{Head} 
& \textbf{has\_bill\_color:} black, blue, brown, buff, green, grey, iridescent, olive, orange, pink, purple, red, rufous, white, yellow \newline
\textbf{has\_bill\_length:} about\_the\_same\_as\_head, longer\_than\_head, shorter\_than\_head \newline
\textbf{has\_bill\_shape:} all-purpose, cone, curved\_(up\_or\_down), dagger, hooked, hooked\_seabird, needle, spatulate, specialized \newline
\textbf{has\_crown\_color:} black, blue, brown, buff, green, grey, iridescent, olive, orange, pink, purple, red, rufous, white, yellow \newline
\textbf{has\_eye\_color:} black, blue, brown, buff, green, grey, olive, orange, pink, purple, red, rufous, white, yellow \newline
\textbf{has\_forehead\_color:} black, blue, brown, buff, green, grey, iridescent, olive, orange, pink, purple, red, rufous, white, yellow \newline
\textbf{has\_head\_pattern:} capped, crested, eyebrow, eyeline, eyering, malar, masked, plain, spotted, striped, unique\_pattern \newline
\textbf{has\_nape\_color:} black, blue, brown, buff, green, grey, iridescent, olive, orange, pink, purple, red, rufous, white, yellow \newline
\textbf{has\_throat\_color:} black, blue, brown, buff, green, grey, iridescent, olive, orange, pink, purple, red, rufous, white, yellow \\
\midrule
\textbf{Leg} 
& \textbf{has\_leg\_color:} black, blue, brown, buff, green, grey, iridescent, olive, orange, pink, purple, red, rufous, white, yellow \\
\midrule
\textbf{Tail} 
& \textbf{has\_tail\_pattern:} multi-colored, solid, spotted, striped \newline
\textbf{has\_tail\_shape:} fan-shaped\_tail, forked\_tail, notched\_tail, pointed\_tail, rounded\_tail, squared\_tail \newline
\textbf{has\_under\_tail\_color:} black, blue, brown, buff, green, grey, iridescent, olive, orange, pink, purple, red, rufous, white, yellow \newline
\textbf{has\_upper\_tail\_color:} black, blue, brown, buff, green, grey, iridescent, olive, orange, pink, purple, red, rufous, white, yellow \\
\midrule
\textbf{Wing} 
& \textbf{has\_wing\_color:} black, blue, brown, buff, green, grey, iridescent, olive, orange, pink, purple, red, rufous, white, yellow \newline
\textbf{has\_wing\_pattern:} multi-colored, solid, spotted, striped \newline
\textbf{has\_wing\_shape:} broad-wings, long-wings, pointed-wings, rounded-wings, tapered-wings \\
\bottomrule
\end{tabular}
\end{adjustbox}
\end{table}

To evaluate part-specific attribute prediction, we mapped the localized attributes from both CelebA and CUB to their corresponding semantic parts. Holistic attributes that describe the entire object (e.g., ``Attractive'' in CelebA or general size in CUB) were excluded from this part-level analysis. 

For completeness and reproducibility, \cref{tab_supp:keypoint_groupings} details how the raw dataset keypoints were grouped into these corresponding semantic parts for spatial evaluation. Specifically for CelebA, we compute the center of the mouth by averaging the spatial coordinates of the left and right mouth keypoints.

The exact attribute groupings are provided below. For CelebA (\cref{tab_supp:celeba_attributes}), we mapped 32 localized facial attributes to 8 semantic facial parts. For CUB (\cref{tab_supp:cub_attributes}), we mapped 248 fine-grained attributes to 7 semantic bird parts. To ensure the CUB table remains compact and readable, attributes sharing the same prefix (e.g., \texttt{has\_bill\_color}) are grouped together alongside their possible values.

\end{document}